%% file: main.tex
\documentclass{article}

\RequirePackage{etex} %

\usepackage[script]{changdefs}
\usepackage[numbers,compress]{natbib}
\usepackage{booktabs}

\usepackage{graphicx}
\graphicspath{{./figures/}}
\usepackage{wrapfig}
\usepackage{subcaption}
\usepackage{xspace}
\usepackage{multirow}
\usepackage{comment}
\usepackage{xcolor}
\usepackage{float}
\usepackage{algorithm}
\usepackage{algpseudocode}
\usepackage{chemformula}
\usepackage{titlesec}
\usepackage{tocloft}

\usepackage[normalem]{ulem}

\makeatletter
\newcommand{\subalign}[1]{%
  \vcenter{%
    \Let@ \restore@math@cr \default@tag
    \baselineskip\fontdimen10 \scriptfont\tw@
    \advance\baselineskip\fontdimen12 \scriptfont\tw@
    \lineskip\thr@@\fontdimen8 \scriptfont\thr@@
    \lineskiplimit\lineskip
    \ialign{\hfil$\m@th\scriptstyle##$&$\m@th\scriptstyle{}##$\hfil\crcr
      #1\crcr
    }%
  }%
}
\makeatother

\newcommand{\appxref}[1]{Supplementary Section~\ref{appx:#1}}
\renewcommand{\appxfigref}[1]{Supplementary Figure~\ref{fig:#1}}
\renewcommand{\appxtblref}[1]{Supplementary Table~\ref{tbl:#1}}

\newcommand{\bfppt}{\tilde{\bfpp}}

\newcommand{\ext}{\textnormal{ext}}
\newcommand{\XC}{\textnormal{XC}}

\newcommand{\eff}{\textnormal{eff}}
\newcommand{\ee}{\textnormal{ee}}
\newcommand{\TXC}{\textnormal{TXC}}
\newcommand{\res}{\textnormal{S,res}}
\newcommand{\TF}{\textnormal{TF}}
\newcommand{\vW}{\textnormal{vW}}
\newcommand{\APBE}{\textnormal{APBE}}
\newcommand{\base}{\textnormal{S,base}}
\newcommand{\grid}{\textnormal{grid}}
\newcommand{\angstrom}{\text{\r{A}}}
\newcommand{\ourmethod}{M-OFDFT\xspace}
\newcommand{\mlff}{M-NNP\xspace}
\newcommand{\mlffden}{M-NNP-Den\xspace}
\renewcommand{\grad}{\textnormal{grad}}
\newcommand{\coeff}{\textnormal{coeff}}
\newcommand{\std}{\mathrm{std}}

\begin{document}

\title{Overcoming the Barrier of Orbital-Free Density Functional Theory for Molecular Systems Using \\ Deep Learning}
\date{}

\author{
He Zhang$^{1,2\dagger\sharp}$,
Siyuan Liu$^{2\dagger\sharp}$,
Jiacheng You$^{2\sharp}$,
Chang Liu$^{2*}$,
Shuxin Zheng$^{2*}$, \\
Ziheng Lu$^{2}$,
Tong Wang$^{2}$,
Nanning Zheng$^{1}$,
Bin Shao$^{2*}$ \\[0.3cm]
\normalsize{$^1$ National Key Laboratory of Human-Machine Hybrid Augmented Intelligence, National Engineering Research Center for Visual Information and Applications, and Institute of Artificial Intelligence and Robotics, Xi’an Jiaotong University, Xi'an, China. \\
$^2$ Microsoft Research AI4Science, Beijing, China.} \\[0.4cm]
$^*$ Corresponding authors.
E-mails: \texttt{\{changliu, shuz, binshao\}@microsoft.com} \\
$^\dagger$ These authors contributed equally. \\
$^\sharp$ These authors did this work during an internship at Microsoft Research AI4Science.
}

\maketitle
\begin{abstract}
  Orbital-free density functional theory (OFDFT) is a quantum chemistry formulation that has a lower cost scaling than the prevailing Kohn-Sham DFT, which is increasingly desired for contemporary molecular research. However, its accuracy is limited by the kinetic energy density functional, which is notoriously hard to approximate for non-periodic molecular systems. Here we propose \ourmethod, an OFDFT approach capable of solving molecular systems using a deep learning functional model.
  We build the essential non-locality into the model, which is made affordable by the concise density representation as expansion coefficients under an atomic basis.
  With techniques to address unconventional learning challenges therein,
  \ourmethod achieves a comparable accuracy with Kohn-Sham DFT on a wide range of molecules untouched by OFDFT before. More attractively, \ourmethod extrapolates well to molecules much larger than those seen in training, %
  which unleashes the appealing scaling of OFDFT for studying large molecules including proteins, representing an advancement of the accuracy-efficiency trade-off frontier in quantum chemistry.
\end{abstract}

\input{intro}
\input{result}
\input{conclusion}
\input{method}

\section*{Inclusion \& Ethics}
All collaborators who have fulfilled all the criteria of authorship defined by Nature Portfolio journals have been listed as authors, and other collaborators have been listed in the Acknowledgements.
The research is not on a location-specific topic, but has included local researchers related to the topic throughout the research process.
Roles, responsibilities and workload plan were agreed among collaborators ahead of the research.
This research was not severely restricted or prohibited in the setting of the researchers, and does not result in stigmatization, incrimination, discrimination or personal risk to participants.

\section*{Data Availability}
All molecular structures used in this study can be freely accessed from public sources:
ethanol structures are from the MD17 dataset~\citep{chmiela2017machine} at \url{http://www.sgdml.org/#datasets}, QM9~\citep{ramakrishnan2014quantum} molecular structures are from \url{http://dx.doi.org/10.6084/m9.figshare.978904}, QMugs~\citep{isert2022qmugs} molecular structures are from \url{https://doi.org/10.3929/ethz-b-000482129},
and structures of Chignolin, the BBL-H142W system (PDB ID: \texttt{2WXC}), and the protein B system (PDB ID: 
\texttt{1PRB}) are from ref.~\citep{lindorff2011fast} at \url{https://www.deshawresearch.com/downloads/download_trajectory_science2011.cgi}. Example evaluation data for reproducing the analyses in this work are available at \url{https://doi.org/10.6084/m9.figshare.c.6877432} (ref.~\citep{zhang_liu_you_liu_zheng_lu_wang_zheng_shao_2024}). Source data are provided with this paper.

\section*{Code Availability}
The code for implementing the proposed methodology is available at \url{https://doi.org/10.5281/zenodo.10616893} (ref.~\citep{zhang_2024_10616893}).
Trained neural network model checkpoints are available at \url{https://doi.org/10.6084/m9.figshare.c.6877432} (ref.~\citep{zhang_liu_you_liu_zheng_lu_wang_zheng_shao_2024}).

\bibliographystyle{unsrtnat}
\bibliography{ofdft}

\section*{Acknowledgements}
We thank Paola Gori Giorgi, William Chuck Witt, Sebastian Ehlert, Zun Wang, Lixue Cheng, Jan Hermann and Ziteng Liu for insightful discussions and constructive feedback;
Xingheng He and Yaosen Min for suggestions on protein preprocessing;
Han Yang for help with trying other OFDFT software;
Yu Shi for suggestions and feedback on model design and optimization;
and Jingyun Bai for help with figure design.
We received no specific funding for this work. 
He Zhang, Siyuan Liu and Jiacheng You did this work during an internship at Microsoft Research AI4Science.

\section*{Author Contributions}
C.L. led the research under the support from N.Z. and B.S. C.L. is the lead contact.
C.L., S.Z. and B.S. conceived the project.
S.L., C.L., H.Z. and J.Y. deduced and designed data generation methods, enhancement modules, training pipeline, and density optimization.
H.Z., S.Z. and J.Y. designed and implemented the deep learning model.
H.Z. and S.L. conducted the experiments.
Z.L. and T.W. contributed to the experiment design and evaluation protocol.
C.L., H.Z., S.L. and S.Z. wrote the paper with inputs from all authors.

\section*{Competing Interests}
C.L., S.Z. and B.S. have filed a patent on \ourmethod (application number: PCT/CN2023/112628).
The other authors declare no competing interests.

\newpage
\input{appendix}

\end{document}

%% file: intro.tex
\section{Introduction} \label{sec:intro}

\begin{figure}[t]
  \centering
  \includegraphics[width=0.98\textwidth]{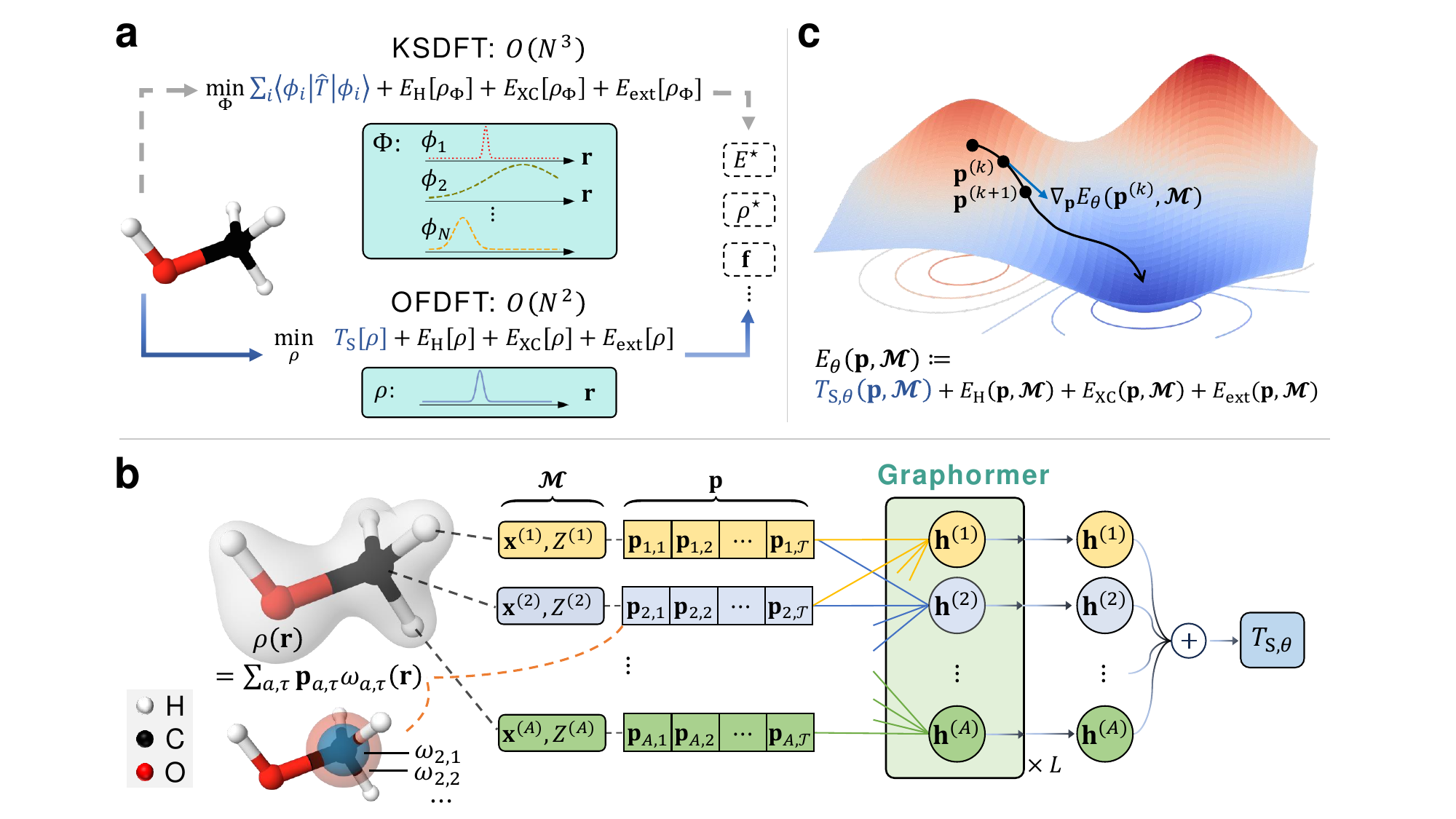}
  \caption{\textbf{Overview of \ourmethod.}
  \textbf{(a)}~KSDFT solves the properties (\eg, the ground-state electron density $\rho^{\star}$, the energy $E^{\star}$, and the force $\mathbf{f}$) of a molecular structure $\clM := \{(\bfxx^{(a)}, Z^{(a)})\}_{a=1}^A$ with $N$ electrons, where $\bfxx^{(a)}$ and $Z^{(a)}$ denote the coordinates and atomic number of the $a$-th atom out of a total of $A$ atoms in the molecule, by optimizing $N$ orbital functions $\Phi := \{\phi_i(\bfrr)\}_{i=1}^N$, where $\phi_i$ denotes the $i$-th orbital which is a function of the coordinates $\bfrr$ of an electron, so that the kinetic energy can be evaluated directly ($\Th$ is the kinetic-energy operator). In contrast, OFDFT only needs to optimize one density function $\rho(\bfrr)$ if the kinetic energy density functional (KEDF) $T_\tnS[\rho]$ is available, which reduces the complexity by an order of $N$.
  \textbf{(b)}~The proposed \ourmethod uses a deep learning model $T_{\tnS,\theta}(\bfpp,\clM)$ ($\theta$ denotes learnable parameters) to approximate KEDF, which is learned from data. The model incorporates non-local interaction of density over the space, which is made affordable by inputting a concise representation of the density (gray shaded region around the molecule): the expansion coefficients $\bfpp$ on an atomic basis $\{\omega_\mu(\bfrr)\}_\mu$, where $\omega_\mu(\bfrr)$ is the $\mu$-th basis function and the index $\mu = (a, \tau)$ is composed of the center atom index $a$ and the pattern index $\tau$ (\eg, the blue and red spheres located bottom-left illustrate two basis functions of two patterns centered at atom~2 (the carbon)).
  The coefficients are correspondingly distributed over the atoms. %
  Non-locality is captured by the attention mechanism which updates features on one atom by calculation with features on all other atoms, including distant ones (\eg, the solid blue lines represent the update of features $\bfhh^{(2)}$ of atom~2 incorporates features on all other atoms).
  After updates by $L$ layers, the final scalar features over atoms are summed up to produce the kinetic energy value.
  \textbf{(c)}~\ourmethod solves a molecular structure $\clM$ by optimizing the density coefficients $\bfpp$ to minimize the electronic energy $E_\theta(\bfpp,\clM)$, which is constructed by the learned KEDF model and three other terms that can be directly evaluated. The red and blue hues represent values of electronic energy.
  }
  \label{fig:whole-framework}
  \vspace{-0.1in}
\end{figure}

Density functional theory (DFT) is a powerful quantum chemistry method for solving electronic states, and hence the energy and properties of molecular systems. It is among the most popular choices owing to its appropriate accuracy-efficiency trade-off, and has fostered many scientific discoveries~\citep{seminario1996recent,jain2013materials}. %
For solving a system with $N$ electrons,
the prevailing Kohn-Sham formulation (KSDFT)~\citep{kohn1965self} minimizes the electronic energy as a functional of $N$ orbital functions $\{\phi_i(\bfrr)\}_{i=1}^N$, where $\phi_i$ denotes the $i$-th orbital, which is a function of the coordinates $\bfrr$ of an electron.
Although the orbitals allow explicitly calculating the non-interacting part of kinetic energy, optimizing $N$ functions deviates from the original idea of DFT~\citep{thomas1927calculation,fermi1928statistische,slater1951simplification,hohenberg1964inhomogeneous}
to optimize one function, the (one-body reduced) electron density $\rho(\bfrr)$, and hence immediately increases the cost scaling by an order of $N$ (\figref{whole-framework}(a)).
This higher complexity is increasingly undesired for the current research stage where large-scale system simulations for practical applications are in high demand.
For this reason, there is a growing interest in methods following the original spirit of DFT, now called \emph{orbital-free DFT} (OFDFT)~\citep{wang2000orbital,karasiev2014progress,huang2023central}.

The central task in OFDFT is to approximate the non-interacting part of kinetic energy as a density functional (KEDF), which is denoted as $T_\tnS[\rho]$.
Classical approximations are developed based on the uniform electron gas theory~\citep{hodges1973quantum,brack1976extended,wang1992kinetic,wang1999orbital,huang2010nonlocal}, %
and have achieved many successes for periodic material systems~\citep{hung2009accurate,witt2018orbital}. %
But the accuracy is still limited for molecules%
~\citep{garcia2007kinetic,xia2012can,teale2022dft}, %
mainly because the electron density in molecules is far from uniform. %

For approximating a complicated functional, recent triumphant progress in machine learning creates new opportunities. %
By leveraging labeled data, the theoretical mismatch can be compensated.
Pioneering works~\citep{snyder2012finding,li2016pure,brockherde2017bypassing} use kernel ridge regression to approximate KEDF and have shown success on 1-dimensional systems. %
Deep learning models have been recently explored for a broader applicability. %
Some works learn to output point-wise kinetic energy density from electron density features~\citep{seino2018semi,seino2019semi,imoto2021order} and hence are (semi-)local. %
Others account for the non-local interaction of density at different points~\citep{yao2016kinetic,meyer2020machine,remme2023kineticnet}.
They enable 3-dimensional calculations, %
yet are still limited to dozen atoms.
\appxref{relw} gives more details.
Moreover, few studies have shown accuracy on molecules much larger than those in training data.
However, such an extrapolation study is imperative to demonstrate the dominating value of the scaling advantage of an OFDFT method, since molecules in a similar scale as training molecules are already affordable by the data-generating methods. 

In this work, we develop an OFDFT method called \textbf{M}-OFDFT that can handle common \textbf{M}olecules using a deep learning KEDF model.
We attribute the limited applicability of previous works on molecular systems to the grid-based representation of density as the model input, which is not sufficiently efficient to represent the uneven density in molecules.
Even an irregular grid requires unaffordably many points ($\sim 10^4 N$)
for a non-local calculation, while the non-locality has been found indispensable for approximating KEDF~\citep{garcia1996nonlocal,wang1999orbital,wang2000orbital,mi2018nonlocal} (\appxref{ablat-nonlocal}); hence, a stringent accuracy-efficiency trade-off is raised. %
To afford non-local calculation for approximating KEDF, we %
adopt an atomic basis set $\{\omega_\mu(\bfrr)\}_{\mu=1}^M$, where $\omega_\mu(\bfrr)$ represents the $\mu$-th basis function and $M$ is the number of basis functions, %
to expand the density as $\rho(\bfrr) = \sum_\mu \bfpp_\mu \omega_\mu(\bfrr)$, and take the coefficients $\bfpp$ as the model input (\figref{whole-framework}(b)).
Each basis function $\omega_{\mu = (a,\tau)}(\bfrr)$ depicts a function pattern $\tau$ around an atom $a$, %
which aligns with the pattern that the electron density in the molecule distributes around atoms. %
They are even designed to mimic the nuclear cusp condition~\citep{kato1957eigenfunctions} for sculpting the sharp density change near a nucleus. %
Such an alignment makes an efficient representation of density: typically $\sim 10 N$ basis functions are sufficient, leading to thousands times fewer dimensions than a grid-based representation, which is especially desired in a non-local calculation.
Moreover, atomic basis functions form a shell structure, %
facilitating an OFDFT method to overcome the challenge of recovering the shell structure of density in molecules. %

Under this representation, the KEDF model follows the form $T_{\tnS,\theta}(\bfpp, \clM)$, where $\theta$ denotes learnable parameters, and $\clM := \{(\bfxx^{(a)},Z^{(a)})\}_{a=1}^A$ denotes the molecular structure, comprising the coordinates and atomic numbers of all atoms, of the target system, which is required in the input for specifying the locations and types of basis functions.
As the coefficients $\bfpp$ can be distributed over atoms according to the center of the corresponding basis function,
the input is a set of %
nodes each with a location, a type, and coefficient features, representing the electron density in its locality (\figref{whole-framework}(b)).
To process such input, we build a deep learning model based on Graphormer~\citep{ying2021transformers,shi2022benchmarking}, a variant of the Transformer model~\citep{vaswani2017attention}. %
It iteratively processes features on all nodes, and adds up the final features %
over the nodes as the kinetic energy output.
Non-locality is covered by the attention mechanism, which updates features on a node by first calculating a weight (``attention'') for the interaction with every other node using features on the two nodes and their distance, then adding the features on every other node, each with the above calculated weight, to the features on this node (\figref{whole-framework}(b)).
This process accounts for the interaction of density features in distant localities, hence non-local effect is captured.
Details on the model architecture are provided in \appxref{model}.
After the KEDF model is learned, \ourmethod solves a given molecular system by optimizing the density coefficients to minimize the electronic energy as the objective, where the KEDF model is used to construct the energy (\figref{whole-framework}(c)).
The optimization result gives ground-state properties such as energy and electron density.
We note that our formulation of KEDF model resembles neural network potentials (NNPs)~\citep{smith2017ani,schutt2018schnet,tholke2021equivariant,liao2023equiformer}, %
which predicts the ground-state energy from $\clM$ end-to-end. %
Without the need to optimize density, they predict energy faster, but do not describe electronic state. %
The \ourmethod formulation also exhibits better extrapolation %
(Results~\ref{sec:res-larger-scale}).

Perhaps unexpectedly, learning a KEDF model %
is more challenging beyond conventional machine learning.
Since the model is used as an optimization objective, %
it needs to capture the energy landscape over the coefficient space for each molecular structure,
for which only one datapoint per molecular structure is far from sufficient.
We %
hence design methods to produce \emph{multiple} coefficient datapoints, each also with a \emph{gradient} (with respect to the coefficients) label, for each molecular structure (Methods~\ref{sec:stage1}).
Moreover, the input %
coefficients are tensors equivariant to the rotation of the molecule, but the output energy is invariant.
We employ \emph{local frames} to guarantee this geometric invariance
(Methods~\ref{sec:local-frame}).
Finally, the model needs to fit a physical mechanism which may vary steeply with the input. %
For expressing large gradients, we introduce a series of \emph{enhancement modules} that balances the sensitivity over coefficient dimensions, rescales the gradient dimension-wise, and offsets the gradient with a reference (Methods~\ref{sec:learn-grad}).

We demonstrate the practical utility and advantage in the following aspects.
\itemone~\ourmethod achieves \emph{chemical accuracy} compared to KSDFT on a range of molecular systems in similar scales as those in training.
This is hundreds times more accurate than classical OFDFT.
The optimized density shows a clear shell structure, which is regarded as challenging for an orbital-free approach. %
\itemtwo~\ourmethod achieves an attractive \emph{extrapolation capability} that its per-atom error stays constant or even decreases on increasingly larger molecules all the way to 10 times (224 atoms) beyond those in training.
The absolute error is still much smaller than classical OFDFT.
In contrast, the per-atom error keeps increasing by NNP variants. %
\ourmethod also improves more efficiently on limited data at large scale. %
\itemthr~With the accuracy and extrapolation capability, \ourmethod unleashes the scaling advantage of OFDFT to large-scale molecular systems.
We find that its empirical time complexity is $O(N^{1.46})$, %
indeed lower by order-$N$ than $O(N^{2.49})$ of KSDFT.
The absolute time is always shorter, achieving %
a 27.4-fold speedup on the protein B system (738 atoms). %
In all, \ourmethod improves the accuracy-efficiency trade-off frontier in quantum chemistry, pushing the applicability of OFDFT for solving large-scale molecular science problems. %

%% file: result.tex
\section{Results} \label{sec:res}

\subsection{Workflow of \ourmethod} \label{sec:res-framework}

OFDFT solves the electronic state of a molecular structure $\clM$ by minimizing the electronic energy as a functional of the electron density $\rho$, which is decomposed in the same way as KSDFT:
$E[\rho] = T_\tnS[\rho] + E_\tnH[\rho] + E_\XC[\rho] + E_\ext[\rho]$ (\appxref{dft-functional}),
where the classical internal (Hartree) and external potential energies, $E_\tnH[\rho]$ and $E_\ext[\rho]$, have exact expressions, and the exchange-correlation (XC) functional $E_\XC[\rho]$ already has accurate approximations. In \ourmethod, the kinetic energy density functional (KEDF) $T_\tnS[\rho]$ is approximated by the $T_{\tnS, \theta}(\bfpp, \clM)$ model under the atomic-basis representation introduced above.
After the $T_{\tnS, \theta}(\bfpp, \clM)$ model is learned (Methods~\ref{sec:stage1}), \ourmethod solves the electronic state of a given molecular structure $\clM$ through the density optimization procedure (\figref{whole-framework}(c)), which minimizes the electronic energy $E_\theta(\bfpp, \clM)$:
\begin{align+}
  \min_{\bfpp: \bfpp\trs \bfww = N} E_\theta(\bfpp, \clM) := T_{\tnS, \theta}(\bfpp, \clM) + E_\tnH(\bfpp, \clM) + E_\XC(\bfpp, \clM) + E_\ext(\bfpp, \clM),
  \label{eqn:den-min}
\end{align+}
where $E_\tnH$, $E_\ext$ and $E_\XC$ can be computed from $(\bfpp, \clM)$ in a conventional way (\appxref{dft-mat-of}).
The constraint on $\bfpp$ fulfills a normalized density, where $\bfww_\mu := \int \omega_\mu(\bfrr) \dd \bfrr$ is the basis normalization vector.
The optimization is solved by gradient descent: %
\begin{align}
  \bfpp^{(k+1)} := \bfpp^{(k)} - \veps \lrparen{\bfI - \frac{\bfww \bfww\trs}{\bfww\trs \bfww}} \nabla_\bfpp E_\theta(\bfpp^{(k)}, \clM),
  \label{eqn:den-gd}
\end{align}
where $\veps$ is a step size, and the gradient is projected onto the admissible plane in respect to the linear constraint.
Notably, due to directly operating on density, the complexity of \ourmethod in each iteration is $O(N^2)$ (\appxref{dft-mat-of}), %
which is order-$N$ less than that $O(N^3)$ (with density fitting; \appxref{dft-mat-ks}) of KSDFT.

\subsection{Performance of \ourmethod on Molecular Systems} \label{sec:res-in-scale}

We first evaluate the performance of \ourmethod on molecules in similar scales but unseen in training.
We generate datasets based on two settings: %
ethanol structures from the MD17 dataset~\citep{chmiela2017machine,chmiela2019sgdml} for studying conformational space generalization, and %
molecular structures from the QM9 dataset~\citep{ruddigkeit2012enumeration,ramakrishnan2014quantum} for studying chemical space generalization.
Each dataset is split into three parts for the training and validation of the KEDF model, and the test of \ourmethod. %
For ease of training, we use the asymptotic PBE-like (APBE) functional~\citep{constantin2011semiclassical} as a base KEDF and let the deep learning model learn the residual (\appxref{func-var-tsres}).

We evaluate \ourmethod in terms of the mean absolute error (MAE) from KSDFT results in energy, as well as in the Hellmann-Feynman (HF) force (\appxref{hfforce}).
The results are 0.18$\,\mathrm{kcal/mol}$ and~1.18$\, \mathrm{kcal/mol/\angstrom}$ on ethanols, and 0.93$\, \mathrm{kcal/mol}$ and~2.91$\, \mathrm{kcal/mol/\angstrom}$ on QM9 (\appxref{res-in-scale} shows more results).
We see \ourmethod achieves chemical accuracy (1$\, \mathrm{kcal/mol}$ energy MAE) in both cases.

\begin{figure}[!ht]
  \centering
  \includegraphics[width=0.98\textwidth]{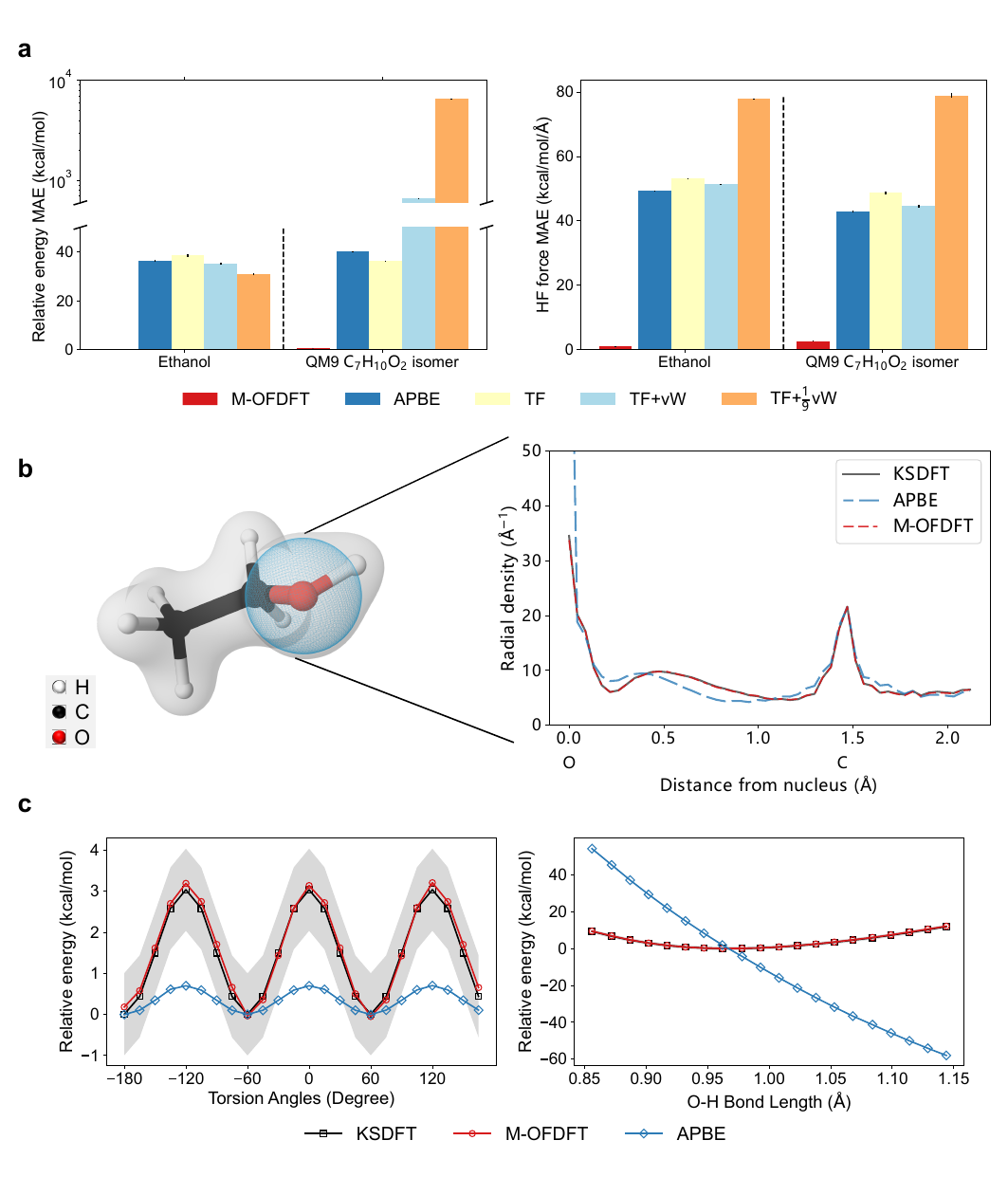}
  \caption{\textbf{Results of \ourmethod compared with classical OFDFT on molecular systems.}
  \textbf{(a)}~Relative energy \textbf{(left)} and Hellmann-Feynman (HF) force \textbf{(right)} results in terms of the MAE from KSDFT, with error bars showing~95\% confidence intervals. Results for ethanol are statistics over 10,000 test structures, and results for QM9 $\mathrm{C_7 H_{10} O_2}$ isomer are statistics over 619 test isomers.
  \textbf{(b)}~Visualization of optimized density. Each curve plots the integrated density on spheres with varying radii centered at the oxygen atom in an ethanol structure.
  \textbf{(c)}~PES study on ethanol. \textbf{(left)} PES over various torsion angles along the \ch{H-C-C-O} bond; \textbf{(right)} PES over various \ch{O-H} bond lengths. The shaded region denotes the range within chemical accuracy (1${\, \mathrm{kcal/mol}}$) with respect to KSDFT.
  }
  \label{fig:res-in-scale}
\end{figure}

To show the advantage of this result, we compare \ourmethod with classical OFDFT using well-established KEDFs, including the Thomas-Fermi (TF) KEDF~\citep{thomas1927calculation,fermi1928statistische} which is exact in the uniform electron gas limit, its corrections TF+$\frac19$vW (ref.~\citep{brack1976extended}) and TF+vW (ref.~\citep{karasiev2012issues}) with the von Weizs\"acker (vW) KEDF~\citep{weizsacker1935theorie}, and the base KEDF APBE (Methods~\ref{sec:classic-kedf}).
We note that different KEDFs may have different absolute energy biases, so for the energy error we compare the MAE in relative energy.
On ethanol structures, the relative energy is taken with respect to the energy on the equilibrium conformation. %
On QM9, as each molecule only has one conformation, we evaluate the relative energy between every pair from the~6,095 isomers of $\mathrm{C_7 H_{10} O_2}$ in the QM9 dataset.
These isomers can be seen as different conformations of the same set of atoms~\citep{smith2017ani}.
As shown in \figref{res-in-scale}(a), \ourmethod still achieves chemical accuracy on relative energy, and is two orders more accurate than classical OFDFT.

As a qualitative investigation of \ourmethod, we visualize the density on a test ethanol structure optimized by these methods in \figref{res-in-scale}(b) (\appxref{visual-den} shows more).
Radial density by spherical integral around the oxygen atom is plotted.
We find that the \ourmethod curves coincide with the KSDFT curve precisely. %
Particularly, the two major peaks around 0$\, \mathrm{\angstrom}$ and 1.4$\, \mathrm{\angstrom}$ correspond to the density of core electrons of the oxygen atom and the bonded carbon atom, while the minor peak in between reflects the density of electrons in the covalent bonds with the hydrogen atom and the carbon atom.
\ourmethod successfully recovers this shell structure, which is deemed difficult for OFDFT.
In comparison, the classical OFDFT using the APBE KEDF does not align well with the true density around the covalent bonds.
We further assess the density numerically by Hirshfeld partial charges~\citep{hirshfeld1977} (\appxref{visual-den} presents a visualization) and dipole moment,
on which the MAEs of \ourmethod results on test ethanol structures are 1.92$\times 10^{-3}\, \mathrm{e}$ and 0.0180$\,\mathrm{D}$, which are substantially better than the results 0.155$\, \mathrm{e}$ and 0.985$\,\mathrm{D}$ of the classical TF+$\frac19$vW OFDFT.
These results suggest that \ourmethod is a working OFDFT for molecular systems.

To further demonstrate the utility of \ourmethod, we investigate its potential energy surface (PES). %
\figref{res-in-scale}(c) shows the PES on ethanol over two coordinates: the torsion angle along the \ch{H-C-C-O} bond and the \ch{O-H} bond length.
We see both curves %
are sufficiently smooth, and stay closely (within chemical accuracy) with KSDFT results.
In comparison, the classical APBE OFDFT fails to maintain chemical accuracy, and does not produce the correct energy barrier and equilibrium bond length.
\appxref{res-in-scale} presents more details.
We also verify the effectiveness of \ourmethod for geometry optimization in \appxref{res-in-scale}.

\subsection{Extrapolation of \ourmethod to Larger-Scale Molecules} \label{sec:res-larger-scale}

\begin{figure}[!t]
  \centering
  \includegraphics[width=0.98\textwidth]{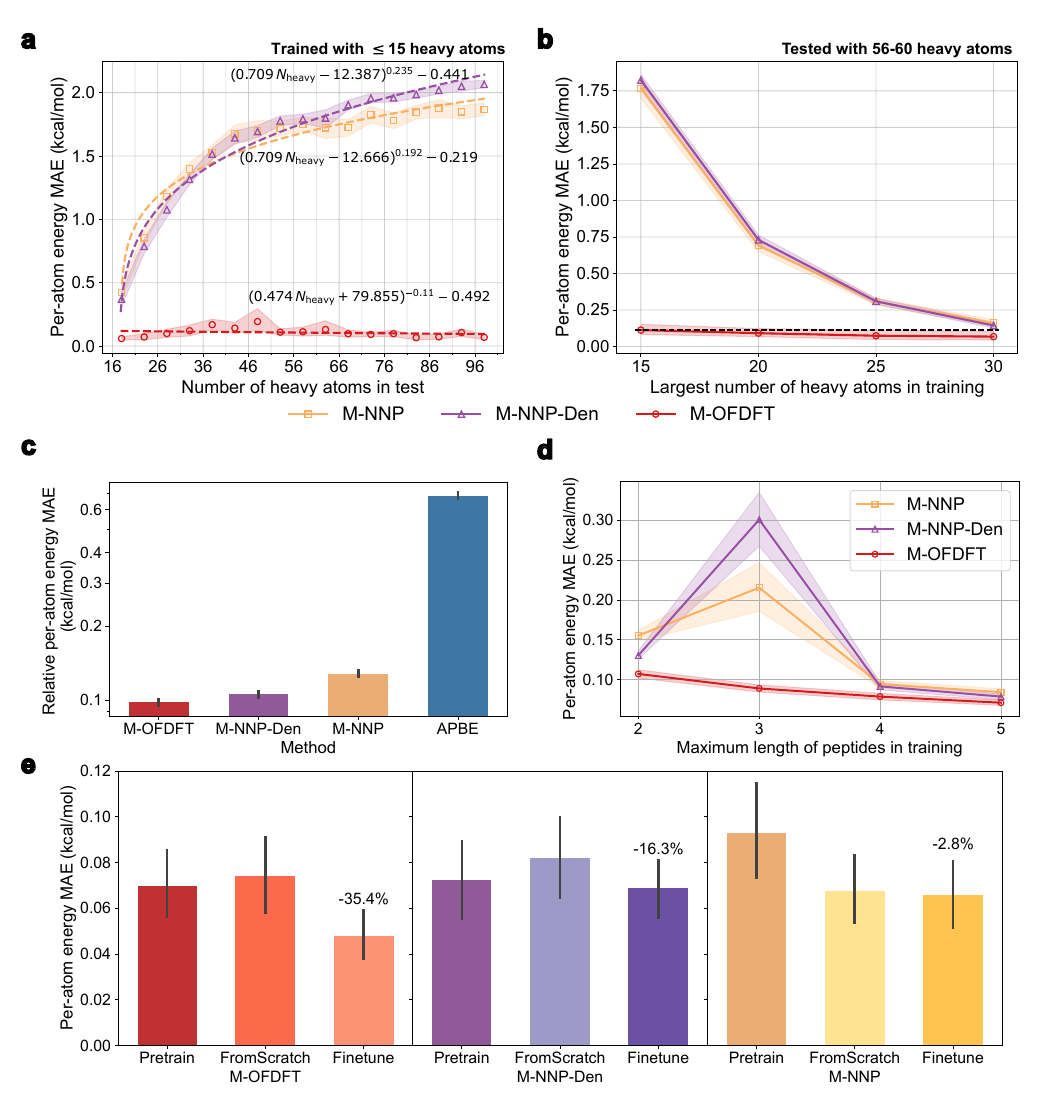}
  \caption{\textbf{Extrapolation performance of \ourmethod compared with other deep learning methods.}
    Considered are \mlff and \mlffden, which use deep learning models to predict the ground-state energy end-to-end.
    The shades and error bars show~95\% confidence intervals.
    \textbf{(a)}~MAE of per-atom energy on increasingly larger molecules from the QMugs dataset, using models trained on molecules with no more than 15 heavy atoms from QM9 and QMugs datasets. Each value is calculated on 50 QMugs molecules.
    \textbf{(b)}~Energy error on QMugs test molecules ($n$=50) with~56-60 heavy atoms, using models trained on a series of datasets containing increasingly larger QMugs molecules up to~30 heavy atoms. The horizontal dashed black line marks the performance of \ourmethod trained on the first dataset.
    \textbf{(c)}~Relative energy error on chignolin structures ($n$=50), using models trained on all peptides (lengths~2-5).
    Also shown is the result of the classical OFDFT using APBE.
    \textbf{(d)}~Energy error on chignolin structures ($n$=1,000), using models trained on a series of datasets including increasingly longer peptides.
    \textbf{(e)}~Energy error on chignolin structures ($n$=50), using models trained on all peptides without (`Pretrain') and with (`Finetune') finetuning on~500 chignolin structures. Also marked are error reduction ratios by the finetuned models over models trained from scratch (`FromScratch') on the~500 chignolin structures only.
  }
  \vspace{-0.1cm}
  \label{fig:res-outscale}
\end{figure}

To wield the advantage of the lower cost scaling of \ourmethod for a more meaningful impact, we evaluate its accuracy on molecular systems with a scale beyond affordable for generating abundant training data.
For running on large molecules, we train the deep learning model targeting the sum of the kinetic and XC energy to get rid of the demanding calculation on grid (Supplementary Sections~\ref{appx:func-var-tsexc} and~\ref{appx:scala-exp}).
This modification does not lead to obvious accuracy lost (\appxref{res-in-scale}). %

To evaluate the extrapolation performance, we compare \ourmethod with a neural-network-potential counterpart, dubbed \mlff. It is a natural deep learning variant, which directly predicts the ground-state energy from %
$\clM$.
We also consider a variant dubbed \mlffden that additionally takes the MINimal Atomic Orbitals (MINAO)~\citep{sun2018pyscf} initialized density into input for investigating the effect of density feature on extrapolation.
Both variants use the same non-local architecture and training settings as \ourmethod for fair comparison (Methods~\ref{sec:mlff-baseline}).
\appxref{res-out-scale-main} presents comparisons with more recent NNP architectures, ET~\citep{tholke2021equivariant} and Equiformer~\citep{liao2023equiformer}, which suggest the same conclusion.

\paragraph{QMugs}
We first study the extrapolation on the QMugs dataset~\citep{isert2022qmugs}, containing much larger molecules than those in QM9 which have no more than~9 heavy atoms.
We train the models on QM9 together with QMugs molecules with no more than 15 heavy atoms, %
and test the methods on larger QMugs molecules up to~101 heavy atoms, which are grouped according to the number of heavy atoms into bins of width~5,
and are randomly subsampled to ensure the same number~(50) of molecular structures in each bin to eliminate statistical effects.

The result is shown in \figref{res-outscale}(a).
We see that the per-atom MAE of \ourmethod is always orders smaller than \mlff and \mlffden in absolute value, even though \mlff and \mlffden achieve a lower validation error (\appxtblref{e2e-out-scale}).
More attractively, the error of \ourmethod keeps constant and even decreases (note the negative exponent) when the molecule scale increases, while the errors of \mlff and \mlffden keep increasing, %
even though they use the same non-local architecture capable of capturing long-range effects, and \mlffden also has a density input.
We attribute the qualitatively better extrapolation to appropriately formulating the machine-learning task. %
The ground-state energy of a molecular structure is the \emph{result} of an intricate, many-body interaction among electrons and nuclei, leading to a highly challenging %
function to extrapolate from one region to another.
\ourmethod converts the task into learning the objective function for the target output.
The objective needs to capture only the \emph{mechanism} through which the particles interact, which has a reduced level of complexity, while transferring a large portion of complexity to the optimization process, which optimization tools can handle effectively without an extrapolation issue.
Similar phenomena have also been observed recently in machine learning that learning an objective shows better extrapolation than learning an end-to-end map~\citep{du2019implicit,du2022learning}.

To further substantiate the extrapolation capability of \ourmethod, we investigate
the magnitude by which the training molecule scale must be increased
for \mlff and \mlffden to achieve the same level of performance as \ourmethod on a given workload of large-scale molecules.
We take~50 QMugs molecules with~50-60 heavy atoms as the extrapolation benchmark, and train the models on a series of equal-sized datasets that include increasingly larger molecules up to~30 heavy atoms. %
As shown in \figref{res-outscale}(b), \mlff and \mlffden require at least twice as large molecules in the training dataset (30\,\emph{vs.}~15 heavy atoms) to achieve a commensurate accuracy (0.068${\, \mathrm{kcal/mol}}$) as \ourmethod provides.
These extrapolation results suggest that \ourmethod can be applied to systems much larger than training to exploit the scaling advantage, and is more affordable to develop for solving large-scale molecular systems. %

\paragraph{Chignolin}
An increasingly important portion of the demand for large-scale quantum chemistry calculation comes from biomolecular systems, particularly proteins, which are not touched by OFDFT previously.
We assess the capability of \ourmethod for protein systems on the chignolin protein (10 residues, 168 atoms after neutralization). %
We consider the common set-up where it is unaffordable to generate abundant data for the large target system and hence requires extrapolation.
We generate training data on smaller-scale systems of short peptide structures containing~2 to~5 residues, cropped %
from~1,000 chignolin structures selected from ref.~\citep{lindorff2011fast}. %
To account for non-covalent effects, non-consecutive fragments are also used in training, including systems of two dipeptides and systems of one dipeptide and one tripeptide. %
See more details in Methods~\ref{sec:data-chig}.
For this task, we let the model target the total energy for a learning stability consideration (\appxref{func-var-others}). %

We first train the model on all available peptides, and compare the relative energy error on chignolin with other methods in \figref{res-outscale}(c).
Notably, \ourmethod achieves a substantially lower per-atom error than the classical OFDFT using the APBE KEDF (0.098${\, \mathrm{kcal/mol}}$ \emph{vs.}~0.684${\, \mathrm{kcal/mol}}$), providing an effective OFDFT method for biomolecular systems.
\ourmethod also outperforms deep learning variants \mlff and \mlffden, indicating a better extrapolation capability.
To investigate extrapolation in more detail, we train the deep learning models on peptides with increasingly larger scale and plot the error on chignolin in \figref{res-outscale}(d) (similar to the setting of \figref{res-outscale}(b)).
Remarkably, \ourmethod consistently outperforms end-to-end energy prediction methods \mlff and \mlffden across all lengths of training peptides, and halves the required length for the same level of accuracy. %
We note the spikes of \mlff and \mlffden at peptide length~3 despite extensive hyperparameter tuning, possibly due to that their harder extrapolation difficulty magnifies %
the gap between in-scale validation and larger-scale performance in this case.

After being trained on data in accessible scale, which is called ``pretraining'' in the following context, a deep learning model for a larger-scale workload can be further improved if a few larger-scale data are available for finetuning.
In this situation, a method capable of good extrapolation could be roughly aligned with the larger-scale task in advance using accessible data, more efficiently leveraging the limited larger-scale data, and outperforming the model trained from scratch on these limited data only.
To investigate the benefit of \ourmethod in this scenario, we build a finetuning dataset on~500 chignolin structures. %
Results in \figref{res-outscale}(e) show that \ourmethod achieves the most gain from pretraining, reducing the energy error by~35.4\% over training from scratch, %
showing the appeal of extracting a more generalizable rule from accessible-scale data.
With finetuning, \ourmethod still gives the best absolute accuracy.
These results suggest that \ourmethod could effectively handle as large a molecular system as a protein, even without abundant training data on the same large scale.

\subsection{Empirical Time Complexity of \ourmethod} \label{sec:res-time}

\begin{figure}[t]
  \centering
  \includegraphics[width=0.8\textwidth]{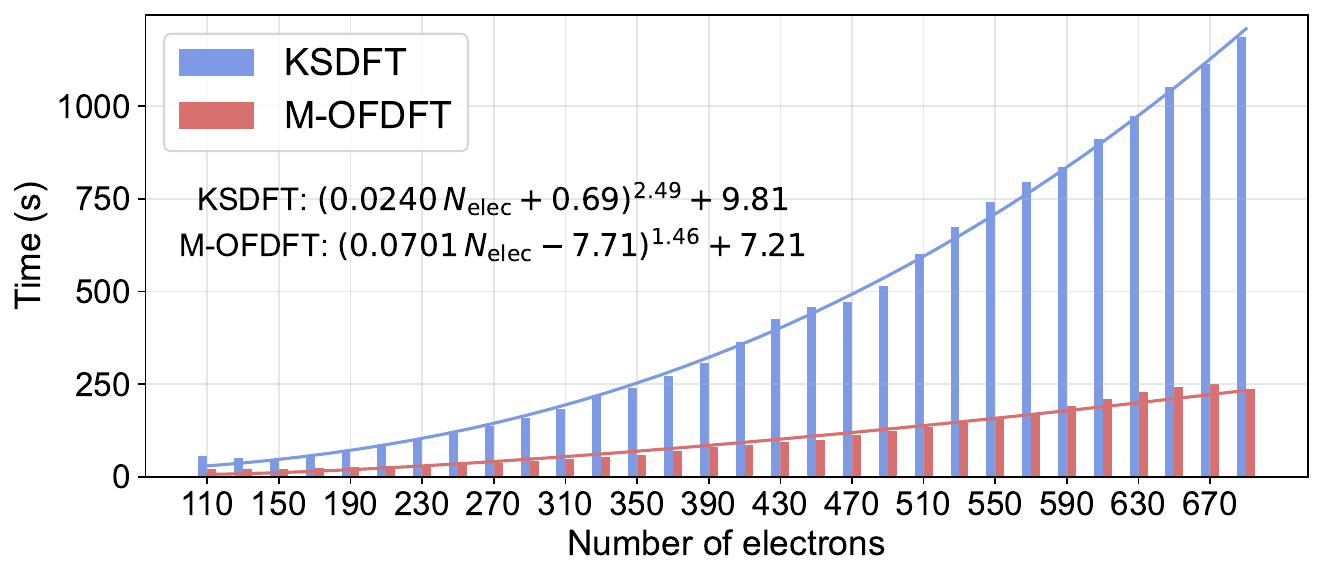}
  \caption{\textbf{Empirical time cost of \ourmethod compared with KSDFT on molecules ($n$=808) at various scales.}
  Each plotted value is the average of running times on molecules whose number of electrons falls in the corresponding bin of width~20.
  }
  \label{fig:res-time}
\end{figure}

After validating the accuracy and extrapolation capability, %
we now demonstrate the scaling advantage of \ourmethod empirically.
The time cost for running both methods on increasingly larger molecules from the QMugs dataset~\citep{isert2022qmugs} is plotted in \figref{res-time}. %
\ourmethod calculations are run on a 32-core CPU server with 216 GiB memory and one Nvidia A100 GPU with 80 GiB memory, and KSDFT calculations on servers %
with 256 GiB memory and 32 Intel Xeon Platinum 8272CL cores with hyperthreading disabled.
We see the absolute running time of \ourmethod is always shorter than that of KSDFT, achieving up to 6.7-fold speedup.
The empirical complexity of \ourmethod is $O(N^{1.46})$, which is %
indeed at least order-$N$ less than the empirical complexity $O(N^{2.49})$ of KSDFT.
\appxref{scala-exp} %
details the running setup and per-component cost.

To further wield the advantage, we run \ourmethod on two molecular systems as large as proteins:
\itemone the peripheral subunit-binding domain BBL-H142W (PDB ID: 
\texttt{2WXC})~\citep{neuweiler2009folding} containing~2,676 electrons (709 atoms), and %
\itemtwo the K5I/K39V double mutant of the Albumin binding domain of protein B (PDB ID: \texttt{1PRB})~\citep{wang2004folding} containing~2,750 electrons (738 atoms). %
Such a scale exceeds the typical workload of KSDFT~\citep{rayson2008rapid}. %
\ourmethod costs~0.41~h and~0.45~h on the two systems, while using KSDFT costs~10.5~h and~12.3~h; hence, a 25.6-fold and 27.4-fold speedup is achieved.
\appxref{scala-exp} provides more details. %

%% file: conclusion.tex
\section{Discussion} \label{sec:conclusion}

We have developed \ourmethod, a deep learning implementation of orbital-free density functional theory that works accurately on molecules while maintaining the appealing low cost scaling, hence providing a powerful tool for exploring complex molecular systems with a higher level of detail and scale.
\ourmethod represents an attempt to leverage deep learning to overcome longstanding challenges and initiate resurgence of alternative quantum chemistry formulations, demonstrating the potential to improve the accuracy-efficiency trade-off.

The presented study is focused on neutral molecules without spin polarization, but the methodology is not limited to such systems. For example,  since the formulation only requires the description of electron density, it can also support charged and open-shell systems. \appxref{res-out-scale-qm9} shows a preliminary demonstration for charged molecules, and more investigation could be conducted in the future.
The formulation can also be applied to material systems, for which atomic basis is also an effective choice for representing electronic states~\citep{vandevondele2005quickstep,ye2021fast}.
Demonstration for a broader interest in the method includes running molecular dynamics simulation and constructing quantum embedding~\citep{cortona1991self,govind1998accurate}, which could be elaborated in the future.

Even though \ourmethod has demonstrated improved extrapolation by choosing an appropriate formulation to leverage deep learning models for molecular science, extrapolation remains an obstacle to the universal application of \ourmethod.
Future exploration to improve extrapolation may include leveraging analytical properties of the KEDF. \appxref{dft-scaling-property} discusses the possibility to leverage the scaling property, and \appxref{func-var-tsres} for leveraging a lower bound.
From the deep learning perspective, it is possible to improve extrapolation using more data and larger model with proper architecture, as suggested by recent progress in large language model~\citep{brown2020language,zhao2023survey} and an application of the Graphormer architecture~\citep{zheng2023towards}.
Considering the higher cost of obtaining data compared with conventional deep learning, active learning could be leveraged to collect more informative data, which can be identified by, \eg, a large disagreement among an ensemble of models~\citep{zhang2019active}, or a large (relative) variance estimation for the model prediction~\citep{krondorfer2023symmetry}.
These possibilities give room to further increase the applicability of \ourmethod.

%% file: method.tex
\section{Methods} \label{sec:method}
As designing and learning the KEDF model are more challenging than conventional (deep) machine learning tasks, we describe methodological details for KEDF model training (Methods~\ref{sec:stage1}), additional design for geometric invariance (Methods~\ref{sec:local-frame}) and large gradient capacity (Methods~\ref{sec:learn-grad}), and density optimization strategies (Methods~\ref{sec:stage2}) of \ourmethod.
For the empirical evaluation, we provide details for dataset preparation (Methods~\ref{sec:exp-dataset}) and generation (Methods~\ref{sec:basis-setup}), implementation of classical OFDFT methods (Methods~\ref{sec:classic-kedf}) and \mlff/\mlffden variants (Methods~\ref{sec:mlff-baseline}), and method for fitting the curves in \figref{res-outscale}(a) and \figref{res-time} (Methods~\ref{sec:curve-fit}).

We first summarize and discuss our major technological innovations here. %
Instead of a grid-based representation, we used coefficients on atomic basis as input density feature, whose much lower dimensionality allows a non-local architecture for accuracy and extrapolation.
We also introduce local frames (Methods~\ref{sec:local-frame}) to guarantee the invariance with respect to the rotation of the molecule.
Some works (\eg, refs.~\citep{dick2020machine,chen2021deepks}) on learning the XC functional also adopt the coefficient input, but without the molecular structure input, and hence cannot properly capture inter-atomic density feature interaction.
Regarding the additional challenge for learning an objective, %
we generated \emph{multiple} datapoints each also with a \emph{gradient} label for each molecular structure to train the model (Methods~\ref{sec:stage1}).
Although the possibility has been noted by previous works (\eg, refs.~\citep{snyder2012finding,snyder2013orbital}),
none has %
fully leveraged such abundant data for training (some only incorporated gradient~\citep{meyer2020machine,fujinami2020orbital,imoto2021order,del2023variational}; ref.~\citep{remme2023kineticnet} also produced multiple datapoints but by perturbing the external potential). %
There are other ways to regularize the optimization behavior of a functional model~\citep{nagai2020completing,kirkpatrick2021pushing,li2021kohn,chen2021deepks}, %
but our trials in \appxref{other-train-strategy} show that they are not as effective.
To express intrinsically large gradient, we introduce enhancement modules (Methods~\ref{sec:learn-grad}) in addition to a conventional neural network.
With these techniques, \ourmethod %
achieves a stable density optimization process, which is regarded as challenging using a deep learning KEDF model. %
Some previous deep learning KEDFs~\citep{yao2016kinetic,seino2018semi,seino2019semi}
do not support density optimization, and some of the others require projection onto the training-data manifold in each step~\citep{snyder2012finding,snyder2013orbital,brockherde2017bypassing,meyer2020machine}. %
\ourmethod achieves stable density optimization using an on-manifold initialization, which is a weaker requirement (Methods~\ref{sec:stage2}).
We note that some previous studies (\eg, ref.~\citep{imoto2021order}) have achieved stable density optimization using a self-consistent field (SCF) scheme. The applicability of the scheme to \ourmethod will be investigated in the future.

\subsection{Training the KEDF Model} \label{sec:stage1}

Although learning the KEDF model $T_{\tnS,\theta}(\bfpp,\clM)$ can be converted to a supervised machine learning task, it is more challenging than the conventional form.
The essential difference is rooted in the way that the model is used: instead of as an end-to-end mapping to predict the kinetic energy of $(\bfpp,\clM)$ queries, the model is used as the objective to optimize the density coefficients $\bfpp$ for a given molecular structure $\clM$ (\figref{whole-framework}(c)).
To eliminate instability and achieve accurate optimization result, the model is required to capture how to vary with $\bfpp$ for a fixed $\clM$, \ie, the optimization landscape on the coefficient space.
The conventional data format $\{\clM^{(d)}, \bfpp^{(d)}, T_\tnS^{(d)}\}_d$ ($d$ indexes training molecular structures) does not effectively convey such information, since only one labeled $\bfpp$ datapoint is seen for each $\clM$.
Hence, the first requirement on training data is multiple coefficient datapoints per structure, following the format $\{ \clM^{(d)}, \{\bfpp^{(d,k)}, T_\tnS^{(d,k)}\}_k \}_d$ ($k$ indexes the multiple coefficient datapoints). %
On such data, the model is trained by minimizing:
\begin{align}
  \sum_d \sum_k \lrvert{ T_{\tnS, \theta}(\bfpp^{(d,k)}, \clM^{(d)}) - T_\tnS^{(d,k)} }.
  \label{eqn:loss-eng}
\end{align}
After some trials, we found this is still not sufficient.
The trained model, although accurately predicts the kinetic energy value, still decreases the electronic energy in density optimization (\eqnref{den-gd}) even starting from the ground-state density. %
This indicates the gradient $\nabla_\bfpp T_{\tnS,\theta}(\bfpp,\clM)$ w.r.t the coefficients is still not accurately recovered.
We hence also desire a \emph{gradient} label for each datapoint, which constitutes data in the format $\{ \clM^{(d)}, \{\bfpp^{(d,k)}, T_\tnS^{(d,k)}, \nabla_\bfpp T_\tnS^{(d,k)}\}_k \}_d$.
As only the projected gradient matters for density optimization following \eqnref{den-gd}, the gradient data is used for training the model by minimizing:
\begin{align}
  \sum_d \sum_k \lrVert*[\bigg]{
    \bigg( \bfI - \frac{\bfww^{(d)} {\bfww^{(d)}}\trs}{{\bfww^{(d)}}\trs \bfww^{(d)}} \bigg) \Big( \nabla_\bfpp T_{\tnS, \theta}(\bfpp^{(d,k)}, \clM^{(d)}) - \nabla_\bfpp T_\tnS^{(d,k)} \Big)
  },
  \label{eqn:loss-grad}
\end{align}
where $\mathbf{I}$ is the identity matrix with matching dimension.
The gradient label provides additional information on the local landscape near each coefficient datapoint.
As the model is used in density optimization only through its gradient, the gradient data directly stabilizes and regularizes density optimization, and enforces stationary-point condition for correct convergence.
\appxref{abl-mult-grad} verifies the improvement empirically through an ablation study. %

To generate such multiple-coefficient and gradient-labeled data, we note that it is tractable from running the conventional KSDFT on each molecular structure $\clM^{(d)}$, which conducts a self-consistent field (SCF) iteration.
The rationale is that, the task in each SCF step $k$ is to solve a non-interacting fermion system in an effective one-body potential constructed from previous steps.
The ground-state wavefunction solution is a Slater determinant specified by the $N$ orbital solutions in that step, by which the non-interacting kinetic energy $T_\tnS^{(d,k)}$ can be directly calculated. %
The corresponding density coefficients $\bfpp^{(d,k)}$ can be calculated from these orbitals by density fitting~\citep{dunlap2000robust}; see \appxref{dft-label-dfit} for details.
For the gradient label, since $\bfpp^{(d,k)}$ represents the ground-state density of the non-interacting system, it minimizes the energy of the non-interacting system as a function of density coefficient, $T_\tnS(\bfpp, \clM^{(d)}) + \bfpp\trs \bfvv_\eff^{(d,k)}$, where $\bfvv_\eff^{(d,k)}$ is the effective potential in SCF step $k$ in vector form under the atomic basis.
This indicates $\nabla_\bfpp T_\tnS(\bfpp^{(d,k)}, \clM^{(d)}) = -\bfvv_\eff^{(d,k)}$ up to the normalization projection.
\appxref{dft-variation} elaborates more on the reasoning, and \appxref{dft-label} provides calculation details, including an efficient implementation to generate the gradient label.

In our implementation of \ourmethod, the atomic basis for representing density is taken as the even-tempered basis set~\citep{bardo1974even} with tempering ratio $\beta = 2.5$.
For generating data, restricted-spin KSDFT is conducted at the PBE/6-31G(2df,p) level, which is sufficient for the considered systems which are uncharged, in near-equilibrium conformation, and only involve light atoms (up to fluorine).
Here, the basis sets for expanding electron density and orbitals are different, since the density corresponding to an orbital state is effectively expanded on the paired orbital basis whose number of basis functions is squared (see Supplementary \eqnref{den-denmat}), so the basis set to expand density needs to be larger than the orbital basis set.
Using a different basis set for density is also the common practice in density fitting, in which context the basis is called an auxiliary basis. The even-tempered basis set is a common choice in density fitting, which is finer than other auxiliary basis choices. It achieves a lower density fitting error in our trials, and could facilitate calculation under other basis by projection onto this finer basis.

\subsection{Geometric Invariance} \label{sec:local-frame}

Another challenge beyond conventional machine learning is that the target physical functional exhibits symmetry w.r.t transformations on the input $(\bfpp, \clM = \{\bfX, \bfZ\})$ arising from the translation and rotation of the molecule, where $\bfX$ and $\bfZ$, respectively, denote the coordinates and atomic numbers of the atoms in the molecule.
This is formally referred to as $\mathrm{SE}(3)$-invariance, following ``3-dimensional special Euclidean group'' that comprises these transformations.
This is because the non-interacting kinetic energy of electrons does not change with the translation and rotation of the molecule,
but the input atomic coordinates $\bfX$ do, and the input density coefficients $\bfpp$ also change with the rotation. %
The change of $\bfpp$ is due to that the electron density rotates with the molecule, but the atomic basis functions do not, since their orientations are aligned with the (global) coordinate system, a.k.a frame.
Formally, such input features are geometric vectors and tensors that change equivariantly with the translation and/or rotation of the molecule.
Subsequently, the model is expected to have this $\mathrm{SE}(3)$-invariance built-in.
This allows the model to learn the essential dependency of the energy on the density irrespective of geometric variability, reducing the problem space, and facilitating data efficiency and effective training.
The invariance also enhances generalization and extrapolation performance, as an important physical property is always guaranteed. %

For the invariance w.r.t atomic coordinates $\bfX$, the neural network model of Graphormer is naturally $\mathrm{SE}(3)$-invariant, since the model only uses relative distances of atom pairs for later processing, which are inherently invariant w.r.t the translation and rotation of the molecule.
To ensure the invariance of the model w.r.t the density coefficients $\bfpp$, we introduce a transformation on $\bfpp$ under \emph{local frames} to make invariant coefficient features.
Each local frame is associated to an atom, and specifies the orientation of atomic basis functions on that atom.
It is determined by the relative positions among nearby atoms, hence the basis function orientations rotate with the molecule and the density, making the density coefficients under the local frame invariant.
Specifically, the local frame on the atom located at $\bfxx_a^{(0)}$ is determined following previous works (\eg, \citep{han2018deep,li2022deep}):
the x-axis unit vector $\hat{\bfxx} := \mathrm{Normalize}(\bfxx_a^{(1)} - \bfxx_a^{(0)})$ is pointed to its nearest heavy atom located at $\bfxx_a^{(1)}$, then the z-axis is pointed to $\hat{\bfzz} := \mathrm{Normalize} \big( \hat{\bfxx} \times (\bfxx_a^{(2)} - \bfxx_a^{(0)}) \big)$, where $\bfxx_a^{(2)}$ is the coordinates of the second-nearest heavy atom not collinear with the nearest one, and finally the y-axis is pointed to $\hat{\bfyy} := \hat{\bfzz} \times \hat{\bfxx}$ following a right-handed system.
See \appxref{local-frame} for more details.

Moreover, the local frame approach offers an additional benefit that the coefficient features are stabilized for local molecular substructures, %
\eg, bond or functional group, of the same type. %
Such substructures on one molecule may have different orientations relative to the whole molecule, but the electron density on them are naturally close, up to a rotation.
Other invariant implementations, \eg, using an equivariant global frame~\citep{li2021closer,puny2022frame} or processing tensorial input invariantly~\citep{schutt2021equivariant,batzner20223}, bind the basis orientations on different atoms together, so the resulting coefficients on the substructures appear vastly different. %
In contrast, using local frames, basis orientations on different atoms are decoupled, and since they are determined only by nearby atoms, the basis functions rotate from one substructure to another accordingly. %
Hence, the resulting density coefficients on the same type of substructures are aligned together, whose difference only indicates the minor density fluctuation on the same type of substructure but not the different orientations of the copies.
This makes it much easier for the model to identify that such local density components follow the same pattern and contribute similarly to the energy.
\appxref{local-frame} provides an illustrative explanation. %
We numerically demonstrate the benefit in Supplementary Figures~\ref{fig:coeff-local-frame}-\ref{fig:grad-local-frame} that using local frame instead of equivariant global frame substantially reduces the variance of both density coefficients and gradients on atoms of each type.
Especially, on most basis functions of hydrogen, the coefficient and gradient scales are reduced by over 60\%.
This substantially stabilizes the training process and immediately reduces training error, resulting in a considerable improvement of overall performance as empirically verified in \appxref{ablat-den-prep}.

\subsection{Enhancement Modules for Vast Gradient Range} %
\label{sec:learn-grad}

After reducing the geometric variability of data using local frame, the raw gradient values still show a vast range, which conventional neural networks are not designed for (\eg, ref.~\citep{fazlyab2019efficient}) and indeed causes training difficulties in our trials.
This is an intrinsic challenge for learning a physical functional since we require non-ground-state density in the data, which would increase the energy steeply. %
The large gradient range cannot be trivially reduced by conventional data normalization techniques, since its scale is associated with the scale of energy and coefficient, %
hence downscaling the gradient would either proportionally downscale the energy values which requires a higher prediction resolution, or inverse-proportionally upscale the coefficients which is also numerically unfriendly to process.
To handle this challenge, we introduce a series of enhancement modules to allow expressing a vast gradient range, including dimension-wise rescaling, a reparameterization of the density coefficients, and an atomic reference module to offset the large mean of gradient.

\paragraph{Dimension-wise Rescaling}
We first upgrade data normalization more flexibly to trade-off coefficient-gradient scales dimension-wise.
Considering the number of coefficient dimensions vary from different molecules, %
we propose to center and rescale the coefficients using biases $\bar{\bfpp}_{Z,\tau}$ and factors $\lambda_{Z,\tau}$ each specific to one coefficient/gradient dimension $\tau$ associated with one \emph{atom type} (\ie, chemical element) $Z$ (instead of one atom).
The bias $\bar{\bfpp}_{Z,\tau} := \mathrm{mean} \{\bfpp_{a,\tau}^{(d,k)}\}_{a: Z^{(a)} = Z, \; k, \, d}$ for $(Z,\tau)$ is the average over coefficient values in dimension $\tau$ on all atoms of type $Z$ in all molecular structures in the training dataset.
After centering the coefficients using the bias (which does not affect gradients), the scaling factor $\lambda_{Z,\tau}$ is determined by upscaling the centered coefficient and simultaneously inverse-proportionally downscaling the gradient, until the gradient achieves a chosen target scale $s_\grad$ or the coefficient exceeds a chosen maximal scale $s_\coeff$. In equation:
\begin{align}
  \lambda_{Z,\tau} := \begin{dcases}
    \min \bigg\{ \frac{\mathrm{mean\_abs\_grad}_{Z,\tau}}{s_\grad}, \, \frac{s_\coeff}{\mathrm{std\_coeff}_{Z,\tau}} \bigg\}, & \text{if } \mathrm{mean\_abs\_grad}_{Z,\tau} > s_\grad, \\
    1, & \text{otherwise},
  \end{dcases}
  \label{eqn:dim-rescale-def}
\end{align}
where the scales of gradient and coefficient in the $(Z,\tau)$-th component are measured by the mean of the absolute value of the gradient component, $\mathrm{mean\_abs\_grad}_{Z,\tau} := \mathrm{mean} \big\{\big\lvert \nabla_{\bfpp_{a,\tau}} T_\tnS^{(d,k)} \big\rvert\big\}_{a: Z^{(a)} = Z, \; k, \, d}$,
and the standard derivation of the coefficient component, $\mathrm{std\_coeff}_{Z,\tau} := \std\{\bfpp_{a,\tau}^{(d,k)}\}_{a: Z^{(a)} = Z, \; k, \, d}$, both collected on the training dataset. %
Using the rescaling factors, each centered coefficient is rescaled by:
\begin{align}
  \bfpp'_{a,\tau} := \lambda_{Z^{(a)},\tau} \, \bfpp_{a,\tau},
  \label{eqn:dim-rescale}
\end{align}
and gradient by $\nabla_{\bfpp_{a,\tau}} T_\tnS' := \nabla_{\bfpp_{a,\tau}} T_\tnS \, / \, \lambda_{Z^{(a)},\tau}$ ($\lambda_{Z,\tau} > 1$ in most cases).

\paragraph{Natural Reparameterization}
On quite a few dimensions, both the coefficient and gradient scales are large, making dimension-wise rescaling ineffective.
We hence introduce \emph{natural reparameterization} applied before rescaling to balance the rescaling difficulties across dimensions hence reduce the worst-case difficulty. %
The unbalanced scales come from the different sensitivities of the density function on different coefficient dimensions: %
the change of density function from a coefficients change $\Delta \bfpp$ is measured by the L2-metric in the function space, $\int \lrvert{\Delta \rho(\bfrr)}^2 \dd \bfrr$, which turns out to be $\Delta \bfpp\trs \bfW \Delta \bfpp$,
in which different dimensions indeed contribute with different weights since the overlap matrix $\bfW_{\mu\nu} := \int \omega_\mu(\bfrr) \omega_\nu(\bfrr) \dd \bfrr$ ($\mu$ and $\nu$ are indexes of basis functions) therein is generally anisotropic.
The reparameterized coefficients $\bfppt$ are expected to contribute equally across the dimensions:
$\int \lrvert{\Delta \rho(\bfrr)}^2 \dd \bfrr = \Delta \bfppt\trs \Delta \bfppt$.
We hence take:
\begin{align}
  \bfppt := \bfM\trs \bfpp,
  \label{eqn:nat-reparam}
\end{align}
where $\bfM$ is a square matrix satisfying $\bfM \bfM\trs = \bfW$.
See \appxref{nat-reparam} for more details.
This reparameterization also leads to natural gradient descent~\citep{amari1998natural} in density optimization, which is known to converge faster than vanilla gradient descent.

\paragraph{Atomic Reference Module}
Recall that in dimension-wise rescaling, the large bias of coefficients can be offset by the mean on a dataset, but this does not reduce the bias scale of gradient labels. %
To further improve the coefficient-gradient scale trade-off, we introduce an \emph{atomic reference module}:
\begin{align}
  T_\mathrm{AtomRef}(\bfpp,\clM) := \bar{\bfgg}_\clM\trs \bfpp + \Tb_\clM,
  \label{eqn:atom-ref}
\end{align}
which is linear in the coefficients $\bfpp$ and whose output is added to the neural network output as the kinetic energy value.
By this design, the gradient of the atomic reference model $\nabla_\bfpp T_\mathrm{AtomRef}(\bfpp,\clM) = \bar{\bfgg}_\clM$ is a constant, which offsets the target gradient for the neural network to capture, effectively reducing the scale of gradient labels %
and facilitating neural network training.
The weights $\bar{\bfgg}_\clM := \mathrm{concat} \{ \bar{\bfgg}_{Z^{(a)},\tau} \}_{\tau,\, a \in \clM}$ and bias $\Tb_\clM := \sum_{a \in \clM} \Tb_{Z^{(a)}} + \Tb_\mathrm{global}$ of the linear model are constructed by tiling and summing the per-type statistics, which are derived over all atoms of each type in a dataset.
The per-type gradient statistics is defined by $\bar{\bfgg}_{Z,\tau} := \mathrm{mean} \{ \nabla_{\bfpp_{a,\tau}} T_\tnS^{(d,k)} \}_{a: Z^{(a)} = Z, \; k, \, d}$, which represents the average response to $T_\tnS$ from the change of coefficients on an atom of type $Z$.
Per-type bias statistics $\{\Tb_Z\}_Z$ and $\Tb_\mathrm{global}$ are fit by least squares.
See \appxref{atom-ref} for more details.

The final KEDF model is constructed from these enhancement modules and the neural network model in the following way (\appxfigref{model-arch}(a)):
the density coefficients are first transformed under local frame and processed by natural reparameterization; the processed coefficients, through one branch, are fed to the atomic reference module to calculate the reference part of output energy, and through another branch, are processed by dimension-wise rescaling and then input to the neural network model which produces the rest part of output energy.
Comparative results in Supplementary Sections~\ref{appx:tech-effi-train} and~\ref{appx:ablat-den-prep} highlight the empirical benefits of each module.

\subsection{Density Optimization} \label{sec:stage2}

In the deployment stage, \ourmethod solves the ground state of a given molecular structure $\clM$ by minimizing the electronic energy as a function of density coefficients $\bfpp$, where the learned KEDF model $T_{\tnS, \theta}(\bfpp, \clM)$ is used to construct the energy function (\figref{whole-framework}(a)).
As described in Results~\ref{sec:res-framework}, we use gradient descent to optimize $\bfpp$ (\eqnref{den-gd}), since it is unnatural to formulate the optimization problem into a self-consistent iteration. %
Gradient descent has also been used in KSDFT, which bears the merit of being more stable~\citep{yoshikawa2022automatic}.

A subtlety in density optimization using a learned functional model is that the model may be confronted with densities far from the training-data manifold (or ``out of distribution'' in machine-learning term), which may lead to unstable optimization.
Such an issue has been observed in previous machine-learning OFDFT~\citep{snyder2012finding,brockherde2017bypassing}, %
which mitigates the problem by projecting the density onto the training-data manifold in each optimization step.
A similar phenomenon is also observed in \ourmethod. %
For example, we find that, when starting from the MINAO initialization~\citep{sun2018pyscf} which is common for KSDFT, the density optimization process %
leads to an obvious gap from the target KSDFT energy.
\appxfigref{stage2-init} provides an illustrative case.
We note that the initial density by MINAO already lies off the manifold inherently:
each density entry in the training data comes from the eigensolution to an effective one-electron Hamiltonian matrix, which exactly solves an effective non-interacting fermion system (\appxref{dft-variation}), %
while the MINAO density comes from the superposition of orbitals of each atom in isolation, which is a different mechanism.

We hence propose using two other initialization methods to resolve the mismatch.
The first approach is to use an established initialization that solves an eigenvalue problem, for which we choose the H\"uckel initialization~\citep{hoffman1963an}. %
In our observation, although the H\"uckel density shows a much larger energy error than MINAO density at initialization, it %
ultimately indeed leads the optimization process to converge closely to the target energy. \appxfigref{stage2-init} provides an illustrative case.

The second choice is to project the MINAO density onto the training-data manifold, which we call ProjMINAO.
In contrast to previous methods, \ourmethod conducts optimization on the coefficient space which varies with molecular structure, so the training-data manifold of coefficient is unknown for an unseen molecular structure.
We hence use another deep learning model $\Delta\bfpp_\theta(\bfpp,\clM)$ %
to predict the required correction to project the input coefficient $\bfpp$ towards the ground-state coefficient $\bfpp^\star$ of the input molecular structure $\clM$, which is always on the manifold. %
See \appxref{den-proj} for details. %
In our observation, we see ProjMINAO initialization indeed converges the optimization curve close to the target energy, even better than H\"uckel initialization. See \appxfigref{stage2-init} for an illustrative case.
We also note from a few examples that even though ProjMINAO already closely approximates the ground state density, density optimization still continues to %
improve the accuracy. This suggests a potential advantage over end-to-end ground-state density prediction followed by energy prediction from ground-state density, %
which may also encounter extrapolation challenges similar to \mlff and \mlffden.

Remarkably, \ourmethod only requires an on-manifold initialization but does not need projection in each optimization step, suggesting better robustness than previous methods; see \appxref{den-res} for details.
\ourmethod in Results~\ref{sec:res} is conducted using ProjMINAO, although using H\"uckel still achieves a reasonable accuracy; see \appxref{res-in-scale}.
\appxref{den-res} provides curves in density error and comparison with classical KEDFs.

\subsection{Dataset Preparation}
\label{sec:exp-dataset}
To substantiate the efficacy of the proposed method, we conduct evaluations on two distinct molecular datasets: ethanol and QM9. These datasets are specifically chosen to evaluate the generalization performance of \ourmethod within both conformation space and chemical space. Furthermore, we seek to assess the extrapolation capability of \ourmethod with larger molecular systems, for which two additional datasets are prepared: QMugs and chignolin. Comprehensive supplementary information and generation details for each dataset are provided below.

\subsubsection{Ethanol} 
To investigate the generalization ability of \ourmethod in conformation space, we created a set of 100,000 non-equilibrium ethanol structures, randomly drawn from the MD17 dataset~\citep{chmiela2017machine, chmiela2019sgdml}. These geometries were then randomly partitioned into training, validation and test sets using an 8:1:1 ratio.

\subsubsection{QM9} 
The QM9 dataset~\citep{ramakrishnan2014quantum} serves as a popular benchmark for predicting quantum-chemical properties using deep learning methods. It comprises equilibrium geometries of approximately 134k small organic molecules, composed of $\rmH$, $\rmC$, $\rmO$, $\rmN$, and $\rmF$ atoms. These molecules represent a subset of all species with up to nine heavy atoms from the GDB-17 chemical universe dataset~\citep{ruddigkeit2012enumeration}. As the dataset contains only equilibrium geometries, we employ it as a benchmark for evaluating the generalization ability of \ourmethod in chemical space. Furthermore, to compare \ourmethod with classical OFDFT methods, we include the 6905 isomers of $\mathrm{C_7 H_{10} O_2}$ present in the QM9 dataset as a benchmark for assessing \ourmethod in conformation space. We continue to randomly split the molecules into training, validation, and test sets using an 8:1:1 ratio.

\subsubsection{QMugs}
The QMugs dataset, proposed in ref.~\citep{isert2022qmugs}, comprises over 665k biologically and pharmacologically relevant molecules extracted from the ChEMBL database, totaling approximately 2M conformers. Notably, QMugs provides molecular samples that are considerably larger than those in the QM9 and MD17 datasets, with an average of 30.6 and a maximum of 100 heavy atoms per compound. This allows us to study the extrapolation capabilities of \ourmethod.

We group QMugs molecules based on the number of heavy atoms into bins with a width of 5, except for the first bin, which contains molecules with 10-15 heavy atoms. To construct an extrapolation task, we divide the union of QM9 and the first QMugs bin into training and validation sets with a 9:1 ratio, and test \ourmethod on 50 molecular structures from each of the other bins with increasing scales. The results are shown in \figref{res-outscale}(a).

For the extrapolation setting in \figref{res-outscale}(b) for investigating the amount of additional larger-scale molecular data required by end-to-end counterparts \mlff and \mlffden to achieve a comparable extrapolation performance as \ourmethod, the target workload of large molecules is fixed, while the affordable training datasets contain increasingly larger molecules.
Specifically, 
taking the QM9 dataset as the base data source, the molecular structures from the first four QMugs bins are progressively incorporated to build four training datasets. All training datasets are prepared with the same size to eliminate statistical effects, while the composition ratio of each data source is designed to follow their relative composition ratios in the original joint dataset (QM9 $\cup$ QMugs). %
The detailed statistics of data source composition ratios are summarized in~\appxfigref{qmugs-comp-ratio}.

\subsubsection{Chignolin} \label{sec:data-chig}

\paragraph{Data Selection} To further evaluate the extrapolation ability of \ourmethod on large-scale biomolecular molecules, such as proteins, we sampled 1,000 chignolin structures from an extensive molecular dynamic (MD) simulation~\citep{lindorff2011fast}. Due to its fast-folding property and short length (TYR-TYR-ASP-PRO-GLU-THR-GLY-THR-TRP-TYR), chignolin has been widely used in molecular dynamics studies.
Following \citet{wang2019machine}, we first featurized the backbone conformations into all pairwise $\alpha$-carbon-atom distances, excluding pairs of nearest neighbor residues. Then, a time-lagged independent component analysis (TICA) was conducted with a lag time of $\tau_{\rm lag}=20$ ns, projecting the conformations onto a low-dimensional subspace. The first eight TICA components were clustered into 1,000 groups using the $k$-means algorithm. The centroid structure of each cluster was extracted using MDTraj~\citep{McGibbon2015MDTraj} and taken as the representative structure. %

\paragraph{Conformation Neutralization}
In this study, we focus on isolated molecular systems where all atoms or functional groups stay neutral and all electrons are paired.
However, protein molecules extracted from MD trajectories are simulated in aqueous solution and some of its functional groups are ionized, such as the amine group in the N-terminal and the carboxyl group in the C-terminal.
In our setting, we neutralize protein conformations by editing (adding/deleting) hydrogen atoms in ionizable groups using OpenMM~\citep{eastman2017openmm}, which is expected to have a minimal impact on the original geometry. The hydrogen definition template was modified to handle hydrogen atoms not defined in standard amino acids, such as the missing hydrogen atom in the carboxyl group in the C-terminal amino acid. We then performed local energy optimization on the neutralized conformations using Amber~\citep{case2008amber}, with the Amber-ff14SB force field and a maximum of 100 optimization cycles. To prevent substantial geometry changes, we only allowed hydrogen atoms associated with heavy atoms involved in the hydrogen editing process to be optimized. For neutralized amino acids without standard force field templates, we built force field parameters using the antechamber and parmchk2 tools in Amber.

\paragraph{Protein Fragmentation} 
To investigate the extrapolation utility of \ourmethod from ``local'' fragments to ``global'' proteins, we created our datasets by cutting each of the 1,000 chignolin structures into protein fragments (\ie, polypeptides) of varying lengths. To construct these fragments, we enumerated all possible polypeptides with sequence lengths up to five in the chignolin sequence. 
Specifically, we allow fragments of length 5 to be composed of two dipeptides or a dipeptide and a tripeptide, \ie, dipeptide-[GAP]-dipeptide and tripeptide-[GAP]-dipeptide. The extracted fragment molecules were capped with hydrogen atoms instead of larger capping groups to minimize the introduction of substantial geometry changes.
These fragments were then used as training and validation sets with a 9:1 ratio. It should be noted that when benchmarking the performance of deep learning methods trained with different molecule scales, the training set with a maximum peptide length of $L_{\rm pep}$ is comprised of all polypeptides with sequence lengths less than $L_{\rm pep}$.

For the comparison of \ourmethod with classical KEDFs in \figref{res-outscale}(c), we only used 50 chignolin structures as the test set, since classical KEDFs require grid-based quadrature which is very costly for running on all the 1,000 chignolin structures.

\paragraph{Data Filtration} After preliminary analysis, we found that the range of energy and gradient labels in fragments datasets are too vast to fit even if equipped with several efficient training techniques (see \appxref{tech-effi-train}). Moreover, most of the vast datapoints come from the first several steps of the SCF iterations that are far from convergent. To mitigate this issue, we filter out datapoints if the residual between its energy and the ground-state energy is larger than 500$\, \mathrm{kcal/mol}$. This strategy can empirically improve the optimization robustness and prediction performance.

\subsection{Data Generation Configurations}
\label{sec:basis-setup}

\paragraph{KSDFT Calculations}
All KSDFT calculations were carried out using the PySCF~\citep{sun2018pyscf} software package, which is open-sourced and has a convenient Python interface that allows flexible customization, \eg, for implementing the data generation process detailed in \appxref{dft-label}, and implementing \ourmethod calculation described in Results~\ref{sec:res-framework} using automatic differentiation packages in Python.
A well-acknowledged pure XC functional (\ie, only depends on density but not orbitals), \eg, PBE, %
is used to make sure the data pairs demonstrate a density functional. %
Restricted-spin calculation and the 6-31G(2df,p) basis set were adopted, which is sufficient for the considered molecules which are uncharged, in near-stable conformations, and only involve light atoms (up to fluorine).
To accelerate the calculations, density fitting with the def2-universal-jfit basis set was enabled in the calculations for molecules with more than 30 atoms.  Grid level was set to 2.  Convergence tolerence was set to 1 $\mathrm{meV}$.  %
Slight modifications to the PySCF code were made to log the required information (\eg, the molecular orbitals) for SCF intermediate steps.  Direct Inversion in Iterative Subspace (DIIS) was enabled, following the defaults.  Per-molecule statistics such as running time and number of atoms were collected alongside the calculation for benchmarking purposes.

\paragraph{KSDFT Initialization}
The MINAO initialization method is employed to conduct KSDFT calculation for data generation, adhering to the default settings.
We would like to mention that as explained and empirically shown in Methods~\ref{sec:stage2}, the H\"uckel initialization achieves a good convergence behavior for density optimization of \ourmethod, since it comes from an eigenvalue-problem solution and hence aligns with the way how the training data are generated (\appxref{dft-label}).
It is then motivated to generate data also using the H\"uckel initialization for running KSDFT, in hope to further align the training data with the use cases (\ie, density optimization of \ourmethod using H\"uckel initialization), making the use cases lie more in-distribution with the training data and expecting better density optimization results.
We tried generating and leveraging such data, but found the generated data entries from different SCF iterations on the same molecular structure show much larger variance, especially the gradient labels.
This brings more challenges on training stability and effectiveness, which are hard to handle effectively even with the techniques in Methods~\ref{sec:learn-grad} and \appxref{tech-effi-train}.
Although using H\"uckel-initialized data improves the accuracy of H\"uckel-initialized \ourmethod on ethanol, the improvement is not obvious on QM9, and using H\"uckel-initialized data does not improve the accuracy of ProjMINAO-initialized \ourmethod on either dataset,
due to that the training challenges outweigh the benefits.

\paragraph{Hardware Details}
All KSDFT calculations and preprocessing (\eg, density fitting) procedures were carried out on a cluster of 700 capable CPU servers. Each server in the cluster has 256 GiB of memory and 32 Intel Xeon Platinum 8272CL cores with hyperthreading disabled.

\paragraph{Dataset Preprocessing}
After the SCF runs, extra procedures were done to transform the SCF results into datasets for training and evaluation.  The density coefficients were obtained using the density fitting procedure detailed in \appxref{dft-label-dfit}, with the auxiliary basis set being an even-tempered basis set~\citep{bardo1974even} generated on-the-fly with $\beta=2.5$ (see Supplementary \eqnref{even-tempered-basis}).  After that, gradient and force labels were calculated using the obtained density coefficients following \appxref{dft-label-grad} and~\ref{appx:hfforce}.  Energy labels were calculated following \appxref{dft-label-val}.  For large orbital integrals in the procedure, proper symmetries in the atomic orbital indices were leveraged to save computation and reduce memory footprint.
Local frame transformation and natural reparameterization (detailed in \appxref{local-frame} and \appxref{nat-reparam}) were applied to the coefficients to obtain the model inputs.

\subsection{Classical OFDFT Methods Implementation}
\label{sec:classic-kedf}

To evaluate the advantages of \ourmethod, we select several classical kinetic energy density functionals (KEDFs) for reference.
The Thomas-Fermi (TF) KEDF~\citep{thomas1927calculation,fermi1928statistische} $T_\TF[\rho] := \frac{3}{10} (3 \pi^2)^\frac23 \int \rho(\bfrr)^{5/3} \ud \bfrr$ is the exact KEDF for uniform electron gas (UEG), \ie, when the electron density $\rho$ is constant.
Expanding KEDF around the UEG limit up to the first-order variation gives the KEDF approximation $T_\TF[\rho] + \frac19 T_\vW[\rho]$ (ref.~\citep{brack1976extended}), where $T_\vW[\rho] := \frac18 \int \frac{\lrVert{\nabla \rho(\bfrr)}^2}{\rho(\bfrr)} \ud \bfrr$ is the von Weizs\"acker KEDF~\citep{weizsacker1935theorie} which corrects the approximation with gradient information of the input density. %
We also include a variant of this correction $T_\TF[\rho] + T_\vW[\rho]$, which is also considered in ref.~\citep{karasiev2012issues}.
The base APBE KEDF~\citep{constantin2011semiclassical} is also tested.

For a fair comparison, these functionals are implemented in the same pipeline and code framework as \ourmethod.
Similar to the way we use the PBE XC functional and the ABPE KEDF as mentioned in \appxref{dft-mat-of}, we also re-implemented the TF, TF+$\frac19$vW, and TF+vW functionals in PySCF using PyTorch.
Their values are evaluated by numerical quadrature on a grid in the real 3-dimensional space.
To solve for the ground-state density, gradient descent is used, for which the gradient of the kinetic energy is directly evaluated by automatic differentiation. The same stopping criterion is used.
For initialization, we use the MINAO method, which we found better than the H\"uckel method.
Since the calculation of these classical KEDFs requires quadrature on grid which is substantially more computationally intensive for large molecules, we only used 50 structures for comparison on chignolin, and omitted the comparison on larger proteins.
For the same reason, we omitted the comparison with other deep learning OFDFT methods~\citep{snyder2012finding,brockherde2017bypassing,meyer2020machine} since they use grid to represent density thus also require the costly grid quadrature, in addition to lack of access to their implementation. %

We also attempted running other more recent KEDFs~\citep{wang1992kinetic,wang1999orbital,huang2010nonlocal} using the implementation in OFDFT software packages, including PROFESS~\citep{Ho2008introducing}, GPAW~\citep{Enkovaara2010electronic}, ATLAS~\citep{Mi2016atlas} and DFTpy~\citep{shao2021dftpy}. However, they are primarily oriented towards periodic material systems and use plane-wave basis.
To avoid the difficulty of expressing sharp density change near nuclei using plane waves, most of the packages use pseudo-potentials (which have to be local ones), but they did not provide pseudo-potentials for elements ($\rmH \rmC \rmN \rmO \rmF$) in our concerned systems, and we encountered difficulties in generating reliable local pseudo-potentials for these elements and in converting the real Coulomb potential into the required pseudo-potential format.
Although GPAW does not require a pseudo-potential, %
we found that its OFDFT calculations on molecular systems are hard to converge, even on small molecules (\eg, ethanol) using the default settings with a looser convergence criterion. %

\subsection{\mlff and \mlffden Implementation} \label{sec:mlff-baseline}

Results~\ref{sec:res-larger-scale} demonstrate that \ourmethod has the ability to extrapolate to large molecular systems, a valuable advantage for quantum chemistry methods. We measure the relevance of this advantage by contrasting \ourmethod with two deep learning-based alternatives, \mlff and \mlffden, that are aimed to predict the total energy (ground-state electronic energy plus inter-nuclear energy) from the molecular structure $\clM$. Both alternatives employ the same Graphormer backbone architecture as \ourmethod, but they do not have the density projection branch (\appxref{den-proj}), and \mlff does not have the density coefficient input branch in its NodeEmbedding module (\appxfigref{model-arch}(c)). The model hyperparameters are also adjusted to be comparable to those of \ourmethod.

\subsection{Curve Fitting Details} \label{sec:curve-fit}
For fitting the curves in \figref{res-outscale}(a) and \figref{res-time}, we restrict the formulas $(\mathtt{a} N_\mathrm{heavy} + \mathtt{b})^\mathtt{c} + \mathtt{d}$ of all curves in each figure to have the same scale, \ie, the same $\mathtt{a}$.
Otherwise, the flexibility of $\mathtt{a}$ would diminish the reflection of the exponent $\mathtt{c}$ on the curvature of the curve on the provided range of $N_\mathrm{heavy}$ (\ie, a smaller $\mathtt{a}$ allows a larger exponent $\mathtt{c}$).
We found that sharing the same $\mathbf{a}$ in different curves hardly hinders the closeness to fitted datapoints. A two-phase fitting strategy is put forward based on this idea: (1) we jointly fit all curves with a shared $\mathtt{a}$, obtaining the pre-optimized $\mathtt{a}'$; (2) taking $\mathtt{a}'$ as an initial guess, we re-fit all curves independently to obtain the post-optimized $\mathtt{a}^*$ using the Trust Region Reflective optimization algorithm, where we restrict the search space of variable $\mathtt{a}$ into $[\mathtt{a}'(1-\xi), \mathtt{a}'(1+\xi)]$, with $\xi=0.5$. The two-phase fitting strategy approximately keeps all post-optimized $\mathtt{a}^*$ in the same scale as well as bring better fitting accuracy. The fitting process is conducted using the SciPy package in Python. %

%% file: appendix.tex
\newcommand{\bfCb}{\bar{\bfC}}
\newcommand{\bfDb}{\bar{\bfD}}
\newcommand{\bfTb}{\bar{\bfT}}
\newcommand{\bfVb}{\bar{\bfV}}
\newcommand{\bfGammab}{\bar{\bfGamma}}
\newcommand{\bfCt}{\tilde{\bfC}}
\newcommand{\bfDt}{\tilde{\bfD}}
\newcommand{\bfLt}{\tilde{\bfL}}
\newcommand{\bfWt}{\tilde{\bfW}}

\newcommand{\ana}{\textnormal{ana}}
\newcommand{\tot}{\textnormal{tot}}
\newcommand{\nuc}{\textnormal{nuc}}
\newcommand{\init}{\textnormal{init}}

\titleformat{\section}{\normalfont\large\bfseries}{Supplementary Section \thesection}{1em}{}
\titleformat{\subsection}{\normalfont\large\bfseries}{Supplementary Section \thesubsection}{1em}{}
\titleformat{\subsubsection}{\normalfont\large\bfseries}{Supplementary Section \thesubsubsection}{1em}{}

\renewcommand{\figurename}{Supplementary Figure}
\renewcommand{\tablename}{Supplementary Table}
\renewcommand{\theequation}{S\arabic{equation}}
\renewcommand{\thefigure}{S\arabic{figure}}
\renewcommand{\thetable}{S\arabic{table}}
\setcounter{equation}{0}
\setcounter{figure}{0}
\setcounter{table}{0}

\appendix
\section*{\LARGE{Supplementary Information}}
\tableofcontents

\newpage

\begin{table}[]
    \centering
    \caption{\textbf{Main notations.} These notations are used consistently throughout the main text and the supplementary text.}
    \begin{tabular}{ll}
        \toprule
        \multicolumn{2}{c}{Basic concepts} \\
        \midrule
        $\bfrr \in \bbR^3$ & Electron coordinate \\ %
        $N$ & Number of electrons in a molecular system \\
        $\psi(\bfrr^{(1)}, \cdots, \bfrr^{(N)})$ & $N$-electron wavefunction \\
        $\braket{f}{g} := \int f(\bfrr) g(\bfrr) \dd \bfrr$ & The standard inner product in function space \\
        $\obraket*{f}{\Oh}{g} := \int f(\bfrr) (\Oh g)(\bfrr) \dd \bfrr$ & Function-space inner product with operator \\
        $(f|g) := \int \frac{f(\bfrr) g(\bfrr')}{\lrVert{\bfrr - \bfrr'}} \dd \bfrr \ud \bfrr'$ & Coulomb integral \\
        \midrule
        \multicolumn{2}{c}{Molecular system} \\
        \midrule
        $\bfxx \in \bbR^3$ & Atom coordinates \\
        $a, b \in \{1,\,2,\, \cdots, A\}$ & Indices for atoms in a molecule \\
        $\bfX := \{\bfxx^{(a)}\}_{a=1}^A$ & Molecular conformation/geometry \\
        $\bfZ := \{Z^{(a)}\}_{a=1}^A$ & Molecular composition \\
        $\clM := \{\bfX, \bfZ\}$ & Molecular structure \\
        $\{\clM^{(d)}\}_{d=1}^D$ & Molecular structures in a dataset \\
        \midrule
        \multicolumn{2}{c}{Density functional theory} \\
        \midrule
        $\Phi := \{\phi_i(\bfrr)\}_{i=1}^N$ & Orbitals \\
        $i, j \in \{1,\,2,\, \cdots, N\}$ & Indices for orbitals or electrons \\
        $\psi_{[\Phi]}(\bfrr^{(1)}, \cdots, \bfrr^{(N)}) := \det[\phi_i(\bfrr^{(j)})]_{ij}$ & Slater determinant from orbitals $\Phi = \{\phi_i(\bfrr)\}_{i=1}^N$ \\
        $\{\eta_\alpha(\bfrr)\}_{\alpha=1}^B$ & Orbital basis \\
        Subscripts $\alpha,\, \beta,\, \gamma,\, \delta \in \{1,\,2,\, \cdots, B\}$ & Indices for orbital basis \\
        $\bfC_{\alpha i}$ & Orbital coefficients \\
        $\bfGamma_{\alpha\beta} := \sum_{i=1}^N \bfC_{\alpha i} \bfC_{\beta i}$ & Density matrix \\
        $\bfS_{\alpha\beta} := \braket{\eta_\alpha}{\eta_\beta}$ & Overlap matrix of orbital basis \\
        $\bfD_{\alpha\beta,\gamma\delta} := \braket{\eta_\alpha \eta_\beta}{\eta_\gamma \eta_\delta}$ & Overlap matrix of paired orbital basis \\
        $\bfDt_{\alpha\beta,\gamma\delta} := (\eta_\alpha \eta_\beta | \eta_\gamma \eta_\delta)$ & 4-center-2-electron Coulomb integral of orbital basis \\
        $\rho(\bfrr)$ & (One-electron reduced) density (function) \\
        $\{\omega_\mu(\bfrr)\}_{\mu=1}^M$ & Density basis \\
        Subscripts $\mu, \nu \in \{1,\,2,\, \cdots, M\}$ & Indices for density basis \\
        Subscript $\mu = (a,\tau)$ & Atom assignment decomposition of basis index \\
        $\bfpp_\mu$ & Density coefficient \\
        $\bfww_\mu := \int \omega_\mu(\bfrr) \dd \bfrr$ & Density basis normalization vector \\
        $\bfW_{\mu\nu} := \braket{\omega_\mu}{\omega_\nu}$ & Overlap matrix of density basis \\
        $\bfWt_{\mu\nu} := (\omega_\mu | \omega_\nu)$ & 2-center-2-electron Coulomb integral of density basis \\
        $\bfL_{\mu,\alpha\beta} := \braket{\omega_\mu}{\eta_\alpha \eta_\beta}$ & Overlap matrix between density basis and orbital basis \\
        $\bfF$ & Fock matrix in KSDFT \\
        $k$ & Step index for SCF iteration or density optimization process \\
        & ($\star$ for the converged step) \\
        $(\bfV_\eff)_{\alpha\beta} := \obraket*{\eta_\alpha}{V_\eff}{\eta_\beta}$ & Effective potential matrix under orbital basis \\
        $(\bfvv_\eff)_\mu := \braket*{\omega_\mu}{V_\eff}$ & Effective potential vector under density basis \\
        $\mu \in \bbR$ & Chemical potential \\
        $\bfveps := \Diag[\veps_1, \cdots, \veps_N]$ & The diagonal $N \times N$ matrix of orbital energies \\
        $U[\rho]$ & Universal functional \\
        $E_\XC[\rho]$ & Exchange-correlation (XC) functional \\
        $T_\tnS[\rho]$ & Kinetic energy density functional (KEDF) \\
        $T_\tnS(\bfpp,\clM)$ & KEDF under atomic basis of density \\
        $T_{\tnS,\theta}(\bfpp,\clM)$ & KEDF model/approximation \\
        $T_\APBE$ & The APBE kinetic functional as the base functional $T_\base$ \\
        $T_\res$ & Residual KEDF on top of the base functional $T_\base$ \\
        $E_\TXC$ & Kinetic and XC functional \\
        $\bfff$ & Hellmann-Feynman force \\
        \bottomrule
    \end{tabular}
    \label{tbl:notations}
\end{table}

\section{Mechanism of Density Functional Theory} \label{appx:dft}

In this section, we introduce details in relevant theory on DFT, including the formulation of OFDFT under atomic basis, and the mechanism and details to use KSDFT to generate value and gradient data for learning KEDF.
Atomic units are used through out the paper.
Notations used in the main text and the supplementary text are listed in \appxtblref{notations}.

\subsection{Basic Formulation of Density Functional Theory} \label{appx:dft-functional}

For brevity, the following formulation is for spinless fermions therefore only consider spacial states.
For the restricted Kohn-Sham calculation we adopt for data generation, a pair of electrons of opposite spins share a common spacial orbital, which amounts to duplicate the orbitals in the following formulation.

The mechanism of DFT may be more intuitively introduced under Levy's constrained search formulation~\citep{levy1979universal}.
The $N$-electron Schr\"odinger equation for ground state is equivalent to the following optimization problem on $N$-electron wavefunctions $\psi(\bfrr^{(1)}, \cdots, \bfrr^{(N)})$ under the variational principle:\footnote{
  In this paper we only consider real-valued wavefunctions (and subsequently real-valued orbitals),
  since we only need to solve for the ground state of stationary Schr\"odinger equation without spin-orbit interaction, for which the Hamiltonian operator $\Hh := \Th + \Vh_\ee + \Vh_\ext$ is Hermitian and real.
}
\begin{align}
  E^\star = \min_{\psi: \, \text{antisym}, \braket{\psi}{\psi} = N} \obraket*{\psi}{\Th + \Vh_\ee + \Vh_\ext}{\psi},
  \label{eqn:engopt-wavefn}
\end{align}
where $\Th := -\frac{1}{2} \sum_{i=1}^N {\nabla^{(i)}}^2$ is the kinetic operator,
$\Vh_\ee := \sum_{1 \le i < j \le N} \frac{1}{\lrVert*{\bfrr^{(i)} - \bfrr^{(j)}}}$ is the electron-electron Coulomb interaction (internal potential),
and $\Vh_\ext := \sum_{i=1}^N V_\ext(\bfrr^{(i)})$ comes from a one-body external potential $V_\ext(\bfrr)$ that commonly arises from the electrostatic field of the nuclei specified by the given molecular structure $\clM = \{\bfX, \bfZ\}$ where $\bfX := (\bfxx^{(1)}, \cdots, \bfxx^{(A)})$ and $\bfZ := (Z^{(1)}, \cdots, Z^{(A)})$:
\begin{align}
  V_\ext(\bfrr) = -\sum_{a=1}^A \frac{Z^{(a)}}{\lrVert{\bfrr - \bfxx^{(a)}}}.
  \label{eqn:Vext-def}
\end{align}

Although the optimization problem is exactly defined, directly optimizing the $N$-electron wavefunction is very challenging computationally.
Specifying the wavefunction $\psi$ and evaluating the energy already require an exponential cost in principle, as $\psi$ is a function on $\bbR^{3N}$ whose dimension increases with $N$.
To make an easier optimization problem, it is then desired to optimize a functional of the one-electron reduced density,
\begin{align}
  \rho_{[\psi]}(\bfrr) := N \int \lrvert*{\psi(\bfrr, \bfrr^{(2)}, \cdots, \bfrr^{(N)})}^2 \dd \bfrr^{(2)} \cdots \ud \bfrr^{(N)},
  \label{eqn:den-def}
\end{align}
which has an intuitive physical interpretation of charge density even under a classical view,
and more importantly, the cost to specify a density is constant (with respect to $N$) in principle, as the density is a function on $\bbR^3$ whose dimension is constant.
This is the starting point of density functional theory (DFT)~\citep{thomas1927calculation,fermi1928statistische,slater1951simplification}, and is first formally verified by \citet{hohenberg1964inhomogeneous}.

In terms of the density, the external potential energy, \ie, the last term in \eqnref{engopt-wavefn}, is already an explicit density functional, since the external potential is one-body:
\begin{align}
  \obraket*{\psi}{\Vh_\ext}{\psi} = \int \rho_{[\psi]}(\bfrr) V_\ext(\bfrr) \dd \bfrr = E_\ext[\rho_{[\psi]}], \quad
  \text{where  }
  E_\ext[\rho] := \int \rho(\bfrr) V_\ext(\bfrr) \dd \bfrr.
  \label{eqn:ext-def}
\end{align}
For the other energy terms, using the density as an intermediate, the optimization problem in \eqnref{engopt-wavefn} can be equivalently%
\footnote{
  The correspondence between the optimization space of $\rho$ and the optimization space of $\psi$ to allow this equivalence is analyzed by \citet{lieb1983density}. %
}
carried out in two levels:
\begin{align}
  E^\star &= \min_{\rho: \ge 0, \int \rho(\bfrr) \dd \bfrr = N} \lrparen*[\Big]{ \min_{\psi: \, \text{antisym}, \rho_{[\psi]} = \rho} \obraket*{\psi}{\Th + \Vh_\ee}{\psi} } + E_\ext[\rho]
  \label{eqn:engopt-twolevel} \\
  &= \min_{\rho: \ge 0, \int \rho(\bfrr) \dd \bfrr = N}
  \bigg\{ E[\rho] :=
    U[\rho] + E_\ext[\rho]
  \bigg\}.
  \label{eqn:engopt-univ}
\end{align}
Here, the result of the first-level optimization problem carrying out a constrained search in \eqnref{engopt-twolevel} defines a density functional,
\begin{align}
  U[\rho] := \min_{\psi: \, \text{antisym}, \rho_{[\psi]} = \rho} \obraket*{\psi}{\Th + \Vh_\ee}{\psi},
  \label{eqn:univ-def}
\end{align}
called the universal functional, as it is independent of system specification (\ie, $\clM$ or $V_\ext$).
It is composed of the kinetic and internal potential energy of the electrons.
The optimization objective is then formally converted to a density functional $E[\rho]$ as shown in \eqnref{engopt-univ}.

To carry out practical computation, variants of the kinetic and internal potential energy that allow explicit calculation or have known properties are introduced, to cover the major part of the corresponding energies in $U[\rho]$.
The internal potential energy is covered by its classical version, \ie assuming no correlation, called the Hartree energy:
\begin{align}
  E_\tnH[\rho] := \frac{1}{2} \int \frac{\rho(\bfrr) \rho(\bfrr')}{\lrVert{\bfrr - \bfrr'}} \dd \bfrr \ud \bfrr'.
  \label{eqn:hart-def}
\end{align}
The kinetic energy is covered by the kinetic energy density functional (KEDF), which is defined in a similar way as the universal functional:
\begin{align}
  T_\tnS[\rho] := \min_{\psi: \, \text{antisym}, \rho_{[\psi]} = \rho} \obraket*{\psi}{\Th}{\psi}.
  \label{eqn:ts-def}
\end{align}
The remainder in the universal functional is called the \emph{exchange-correlation (XC) functional}:
\begin{align}
  E_\XC[\rho] := U[\rho] - T_\tnS[\rho] - E_\tnH[\rho],
\end{align}
which is by definition also a density functional.
Under this decomposition, the density optimization problem in \eqnref{engopt-univ} becomes:
\begin{align}
  E^\star = \min_{\rho: \ge 0, \int \rho(\bfrr) \dd \bfrr = N}
  \bigg\{ E[\rho] =
    T_\tnS[\rho] + \underbrace{ E_\tnH[\rho] + E_\XC[\rho] + E_\ext[\rho] }_{=: E_\eff[\rho]}
  \bigg\}.
  \label{eqn:engopt-den}
\end{align}
Here $E_\eff[\rho]$ is defined for future convenience and denoted after the effective-potential interpretation of its variation detailed later.
Using carefully designed explicit expressions or machine-learning models to approximate the $T_\tnS[\rho]$ and $E_\XC[\rho]$ functionals, practical computation can be conducted.
This is the formulation of orbital-free density functional theory (OFDFT).
Indeed, the object to be optimized is the electron density, which is one function on the constant-dimensional space of $\bbR^3$, and hence greatly reduces computation complexity over the original variational problem \eqnref{engopt-wavefn}.
Under properly designed $T_\tnS[\rho]$ and $E_\XC[\rho]$ approximations, the complexity is favorably $O(N^2)$ under atomic basis.

Considering the KEDF $T_\tnS[\rho]$ is more challenging to approximate than the XC functional $E_\XC[\rho]$, \citet{kohn1965self} leverage an equivalent formulation of KEDF to allow its accurate calculation, at the cost of increasing the complexity.
The alternative formulation optimizes \emph{determinantal} wavefunctions.
A determinantal wavefunction for $N$ electrons is specified by $N$ one-electron wavefunctions $\Phi := \{\phi_i(\bfrr)\}_{i=1}^N$, called orbitals, following the form:
\begin{align}
  \psi_{[\Phi]}(\bfrr^{(1)}, \cdots, \bfrr^{(N)}) := \frac{1}{\sqrt{N!}} \det[\phi_i(\bfrr^{(j)})]_{ij}, \quad
  \text{given orbitals } \Phi := \{\phi_i(\bfrr)\}_{i=1}^N.
  \label{eqn:slater-det}
\end{align}
The equivalent optimization problem in parallel with \eqnref{ts-def} is:\footnote{
  The equivalence on defining $T_\tnS[\rho]$ by \eqnref{ts-def} and \eqnref{ts-orb} is known to be guaranteed if the density $\rho$ comes from the ground state of a non-interacting system which is non-degenerate~\citep[Thm.~4.6]{lieb1983density}.
  Even there exists a density $\rho$ that makes the determinantally-defined $T_\tnS[\rho]$ by \eqnref{ts-orb} different~\citep[Thm.~4.8]{lieb1983density}, the determinantally-defined $T_\tnS[\rho]$ still gives the right ground-state energy in density optimization (up to a closure)~\citep[Thm.~4.9]{lieb1983density}.
}
\begin{align}
  & T_\tnS[\rho] = \min_{\{\phi_i\}_{i=1}^N: \rho_{[\Phi]} = \rho} \obraket*{\psi_{[\Phi]}}{\Th}{\psi_{[\Phi]}}
  = \min_{\subalign{\{\phi_i\}_{i=1}^N: & \text{orthonormal}, \\ & \rho_{[\Phi]} = \rho}} \sum_{i=1}^N \obraket*{\phi_i}{\Th}{\phi_i},
  \label{eqn:ts-orb} \\
  \text{where} \quad &
  \rho_{[\Phi]}(\bfrr) := \sum_{i=1}^N \lrvert{\phi_i(\bfrr)}^2, \quad
  \text{given orthonormal orbitals } \Phi := \{\phi_i(\bfrr)\}_{i=1}^N.
  \label{eqn:den-orb-orthon}
\end{align}
In \eqnref{ts-orb}, the second equality holds %
since the density normalizes to $N$, and a set of (non-collinear) functions can always be orthogonalized, \eg, using the Gram-Schmidt process (\citep[Sec.~3.1.4]{bobrowicz1977self};~\citep[Sec.~14.3]{levine2009quantum}).
This equivalence can be understood from the interpretation of $T_\tnS[\rho]$ as the \emph{non-interacting} portion of kinetic energy.
Indeed, for a non-interacting system, there are only kinetic energy and external potential energy (\ie, taking $\Vh_\ee = 0$), so the two-level optimization in parallel with \eqnref{engopt-twolevel} becomes:
\begin{align}
  E^\star &= \min_{\psi: \, \text{antisym}, \braket{\psi}{\psi} = 1} \obraket*{\psi}{\Th + \Vh_\ext}{\psi} \\
  &= \min_{\rho: \ge 0, \int \rho(\bfrr) \dd \bfrr = N} \lrparen*[\Big]{ \min_{\psi: \, \text{antisym}, \rho_{[\psi]} = \rho} \obraket*{\psi}{\Th}{\psi} } + E_\ext[\rho] \\
  &= \min_{\rho: \ge 0, \int \rho(\bfrr) \dd \bfrr = N} T_\tnS[\rho] + E_\ext[\rho],
  \label{eqn:engopt-nonint-den}
\end{align}
from which we see that $T_\tnS[\rho]$ is the ground-state kinetic energy of the non-interacting system whose ground-state density is $\rho$.
On the other hand, it is known that the ground-state wavefunction of a non-interacting system is commonly determinantal (at least when the ground state is non-degenerate~\citep[Thm.~4.6]{lieb1983density}). %
Hence the optimization can be rewritten as:
\begin{align}
  E^\star &= \min_{\{\phi_i\}_{i=1}^N} \obraket*{\psi_{[\Phi]}}{\Th + \Vh_\ext}{\psi_{[\Phi]}} \\
  &= \min_{\rho: \ge 0, \int \rho(\bfrr) \dd \bfrr = N} \lrparen*[\Big]{ \min_{\{\phi_i\}_{i=1}^N: \rho_{[\Phi]} = \rho} \obraket*{\psi_{[\Phi]}}{\Th}{\psi_{[\Phi]}} } + E_\ext[\rho] \\
  &= \min_{\rho: \ge 0, \int \rho(\bfrr) \dd \bfrr = N} \lrparen*[\bigg]{ \min_{\subalign{\{\phi_i\}_{i=1}^N: & \text{orthonormal}, \\ & \rho_{[\Phi]} = \rho}} \sum_{i=1}^N \obraket*{\phi_i}{\Th}{\phi_i} } + E_\ext[\rho],
  \label{eqn:engopt-nonint-orb}
\end{align}
which indicates \eqnref{ts-orb}.

Back to the main problem, leveraging this knowledge about $T_\tnS[\rho]$ in the variational problem \eqnref{engopt-den} gives:
\begin{align}
  E^\star &= \min_{\rho: \ge 0, \int \rho(\bfrr) \dd \bfrr = N}
  \lrparen*[\bigg]{ \min_{\subalign{\{\phi_i\}_{i=1}^N: & \text{orthonormal}, \\ & \rho_{[\Phi]} = \rho}} \sum_{i=1}^N \obraket*{\phi_i}{\Th}{\phi_i} }
  + E_\tnH[\rho] + E_\XC[\rho] + E_\ext[\rho],
\end{align}
which can be converted into directly optimizing the orbitals in a single-level optimization:
\begin{align}
  E^\star = \min_{\{\phi_i\}_{i=1}^N: \text{orthonormal}} \bigg\{ E[\Phi] := \sum_{i=1}^N \obraket*{\phi_i}{\Th}{\phi_i} +
  \underbrace{ E_\tnH[\rho_{[\Phi]}] + E_\XC[\rho_{[\Phi]}] + E_\ext[\rho_{[\Phi]}] }_{E_\eff[\rho_{[\Phi]}]} \bigg\}.
  \label{eqn:engopt-orb}
\end{align}
This is the formulation of Kohn-Sham density functional theory (KSDFT).
With decades of development of XC functional approximations, KSDFT has achieved remarkable success and becomes among the most popular quantum chemistry method.
In its formulation, the object to be optimized is a set of orbitals $\{\phi_i(\bfrr)\}_{i=1}^N$, which are $N$ functions on $\bbR^3$.
This is still substantially cheaper than optimizing an $N$-electron wavefunction on $\bbR^{3N}$, but has a complexity at least $O(N)$ times more than OFDFT, due to $N$ times more $\bbR^3$ functions to optimize.
Under atomic basis, KSDFT has a complexity at least $O(N^3)$ (using density fitting) without further approximations.
In this triumphant era of deep machine learning, approximating a complicated functional is not as challenging as before.
Powerful deep learning models create the opportunity to approximate KEDF accurately enough to match successful XC functional approximations.
This would enable accurate and practical OFDFT calculation, unleashing its power of lower complexity to push the accuracy-efficiency trade-off in quantum chemistry.

\subsection{KSDFT Calculation Produces Labels of KEDF} \label{appx:dft-variation}

We now explain why a KSDFT calculation procedure could provide value and gradient labels of KEDF.
Computation details under atomic basis are postponed in \appxref{dft-label}.
We start by describing the typical algorithm to solve the optimization problem in KSDFT.
To determine the optimal solution of orbitals $\Phi := \{\phi_i(\bfrr)\}_{i=1}^N$, the variation
of the energy functional $E[\Phi]$ in \eqnref{engopt-orb} with respect to each orbital $\phi_i$ is required:
\begin{align}
  \fracdelta{E[\Phi]}{\phi_i}(\bfrr)
  ={} & \fracdelta{\sum_{j=1}^N (-1/2) \obraket*{\phi_j}{\nabla^2}{\phi_j}}{\phi_i}(\bfrr)
  + \int \fracdelta{E_\eff[\rho_{[\Phi]}]}{\rho}(\bfrr') \fracdelta{\rho_{[\Phi]}(\bfrr')}{\phi_i(\bfrr)} \dd \bfrr' \\
  ={} & 2 \Th \phi_i(\bfrr) + 2 V_{\eff[\rho_{[\Phi]}]}(\bfrr) \phi_i(\bfrr),
  \label{eqn:eng-orb-variation} \\
  \text{where} \quad
  V_{\eff[\rho]}(\bfrr) :={} & \fracdelta{E_\eff[\rho]}{\rho} (\bfrr)
  = \underbrace{ \int \frac{\rho(\bfrr')}{\lrVert{\bfrr' - \bfrr}} \dd \bfrr' }_{=: V_{\tnH[\rho]}(\bfrr)}
  + \underbrace{ \fracdelta{E_\XC[\rho]}{\rho}(\bfrr) }_{=: V_{\XC[\rho]}(\bfrr)}
  + V_\ext(\bfrr).
  \label{eqn:Veff-fn}
\end{align}
The term $V_{\eff[\rho_{[\Phi]}]}$ arises as a variation with respect to the density $\rho$ since the orbitals affect the energy component $E_\eff$ apart from $T_\tnS$ (defined in \eqnref{engopt-den}) only through the density $\rho_{[\Phi]}$ they define.
By \eqnref{den-orb-orthon} we have $\fracdelta{\rho_{[\Phi]}(\bfrr')}{\phi_i(\bfrr)} = 2 \phi_i(\bfrr) \delta(\bfrr - \bfrr')$, which then gives \eqnref{eng-orb-variation}.
Combining \eqnref{eng-orb-variation} with the variation of the orthonormal constraint yields the optimality equation for the problem \eqnref{engopt-orb}:
\begin{align}
  \Fh_{[\rho_{[\Phi]}]} \phi_i :={} & \Th \phi_i + V_{\eff[\rho_{[\Phi]}]} \phi_i
  = \veps_i \phi_i,
  \quad \forall i = 1, \cdots N.
  \label{eqn:ks-eq}
\end{align}
In the derivation, only the Lagrange multipliers $\veps_i$ for the normalization constraints $\braket{\phi_i}{\phi_i} = 1$ are imposed, since from the resulting equations (\ref{eqn:ks-eq}), $\{\phi_i\}_{i=1}^N$ are eigenstates of an Hermitian operator $\Fh_{[\rho_{[\Phi]}]}$ called the Fock operator, and hence are naturally orthogonal in the general case of non-degeneracy.
These equations resemble the Schr\"odinger equation for $N$ non-interacting fermions, where $V_{\eff[\rho_{[\Phi]}]}(\bfrr)$, as a function on $\bbR^3$, acts as an \emph{effective} one-body external potential, hence the name.

Note that $V_{\eff[\rho_{[\Phi]}]}$ is unknown beforehand, as itself depends on the solution of orbitals.
Hence a fixed-point iteration is employed: starting from a set of initial orbitals $\Phi^{(0)} := \{\phi_i^{(0)}\}_{i=1}^N$, construct the Fock operator using results in previous iterations,
\begin{align}
  \Fh^{(k)} := \Th + V_\eff^{(k)},
  \label{eqn:fock-opr}
\end{align}
where $V_\eff^{(k)}$ is taken as $V_{\eff[\rho_{[\Phi^{(k-1)}]}]}$ following this derivation, \ie, $\Fh^{(k)} = \Fh_{[\rho_{[\Phi^{(k-1)}]}]}$,
and solve the corresponding eigenvalue problem for the orbitals in the current iteration:
\begin{align}
  \Fh^{(k)} \phi_i^{(k)} = \veps_i^{(k)} \phi_i^{(k)},
  \quad \forall i = 1, \cdots, N.
  \label{eqn:fock-eq}
\end{align}
The iteration stops until ``self-consistency'' is achieved, \ie, the eigenstate solution $\Phi^{(k)} := \{\phi_i^{(k)}\}_i$ in the current step coincides (up to an acceptable error) with the orbitals $\Phi^{(k-1)} := \{\phi_i^{(k-1)}\}_{i=1}^N$ in the previous step that define $\Fh^{(k)}$.
This is the self-consistent field (SCF) method~\citep{blinder1965basic}.

An important fact of SCF is that, in each iteration $k$, the solution $\{\phi_i^{(k)}\}_{i=1}^N$ exactly defines the ground-state of a \emph{non-interacting} system of $N$ fermions moving in the effective one-body potential $V_\eff^{(k)}$ as the external potential $V_\ext$.
Indeed, the variational problem \eqnref{engopt-nonint-orb} that describes the non-interacting system can be reformulated into a single-level optimization as:
\begin{align}
  E^\star = \min_{\{\phi_i\}_{i=1}^N: \text{orthonormal}} \sum_{i=1}^N \obraket*{\phi_i}{\Th}{\phi_i} + \int \rho_{[\Phi]}(\bfrr) V_\eff^{(k)}(\bfrr) \dd \bfrr,
  \label{eqn:engopt-nonint-scf}
\end{align}
whose variation coincides with \eqnref{fock-eq} thus solved by $\{\phi_i^{(k)}\}_{i=1}^N$.
This reveals the relation of SCF solution to the KEDF:
this solution of orbitals $\{\phi_i^{(k)}\}_{i=1}^N$ achieves the minimum non-interacting kinetic energy $\sum_{i=1}^N \obraket*{\psi_{[\Phi]}}{\Th}{\psi_{[\Phi]}}$ among all orthonormal orbitals that lead to the same density $\rho_{[\Phi^{(k)}]}$ (otherwise \eqnref{engopt-nonint-scf} can be further minimized; can also be seen from the equivalence to \eqnref{engopt-nonint-orb});
by the alternative form of KEDF \eqnref{ts-orb} as non-interacting ground-state kinetic energy, we thus have:
\begin{align}
  T_\tnS[\rho_{[\Phi^{(k)}]}] = \obraket*{\psi_{[\Phi^{(k)}]}}{\Th}{\psi_{[\Phi^{(k)}]}} = \sum_{i=1}^N \obraket*{\phi_i^{(k)}}{\Th}{\phi_i^{(k)}}.
  \label{eqn:ks-label}
\end{align}
This indicates that \emph{every SCF iteration produces a label for $T_\tnS$}.
Moreover, as the non-interacting variation problem \eqnref{engopt-nonint-scf} is equivalent to its two-level optimization form \eqnref{engopt-nonint-orb},
which is in turn equivalent to the density optimization form using KEDF \eqnref{engopt-nonint-den} (which explains the alternative KEDF form \eqnref{ts-orb}),
the density $\rho_{[\Phi^{(k)}]}$ from the solution $\Phi^{(k)} := \{\phi_i^{(k)}\}_{i=1}^N$ of each SCF iteration minimizes \eqnref{engopt-nonint-den}.
Therefore, it satisfies the variation equation (Euler equation) of \eqnref{engopt-nonint-den} (taking $V_\ext$ as $V_\eff^{(k)}$) subject to the normalization constraint with Lagrange multiplier (chemical potential) $\mu^{(k)}$:
\begin{align}
  \fracdelta{T_\tnS[\rho_{[\Phi^{(k)}]}]}{\rho} + V_\eff^{(k)} = \mu^{(k)}.
  \label{eqn:ts-den-variation}
\end{align}
The variation of KEDF $\fracdelta{T_\tnS}{\rho}$ is related to the gradient with respect to density coefficients when the density $\rho$ is expanded on a basis (see \appxref{dft-label-grad}).
Hence, \emph{every SCF iteration also produces a label for the gradient of $T_\tnS$, up to a projection}.

It is worth noting that these arguments still hold when the effective potential $V_\eff^{(k)}$ in SCF iteration $k$ is \emph{not} $V_{\eff[\rho_{[\Phi^{(k-1)}]}]}$,
since the deductions from \eqnref{engopt-nonint-scf} to \eqnref{ts-den-variation} only require $V_\eff^{(k)}$ to be a one-body potential.
This allows more flexible data generation process since in common DFT calculation settings, $V_\eff^{(k)}$ indeed deviates from $V_{\eff[\rho_{[\Phi^{(k-1)}]}]}$ for more stable and faster convergence, \eg when using the ``direct inversion in the iterative subspace'' (DIIS) method~\citep{pulay1982improved,kudin2002blackbox}.
This also indicates that even when the XC functional used in data generation is not accurate, the generated value and gradient labels for $T_\tnS[\rho]$ are still exact, since the XC functional still gives an effective one-body potential to define the non-interacting system \eqnref{fock-eq} or \eqnref{engopt-nonint-scf}, as long as it is pure (\ie, only depends on density features).
In this sense, data generation for KEDF is easier than that for the XC functional.

\subsection{Formulation under Atomic Basis} \label{appx:dft-mat}

For practical calculation, KSDFT typically uses an atomic basis $\{\eta_\alpha(\bfrr)\}_{\alpha=1}^B$ to expand the orbitals for conducting the SCF iteration in \eqnref{fock-eq} for molecular systems. The expansion gives:
\begin{align}
  \phi_i(\bfrr) = \sum_\alpha \bfC_{\alpha i} \eta_\alpha(\bfrr),
  \label{eqn:orb-expd}
\end{align}
which converts solving for eigenfunctions into the common problem of solving for eigenvectors of a matrix.
On the other hand, as emphasized in Introduction~\ref{sec:intro} and Results~\ref{sec:res-framework}, we also hope to represent the density on an atomic basis $\{\omega_\mu(\bfrr)\}_{\mu=1}^M$ for efficient OFDFT implementation,
\begin{align}
  \rho(\bfrr) = \sum_\mu \bfpp_\mu \omega_\mu(\bfrr).
  \label{eqn:den-expd}
\end{align}
The left-hand-sides of \eqnref{orb-expd} and \eqnref{den-expd} may also be denoted as $\phi_{i,\bfC}$ or $\Phi_\bfC$ and $\rho_\bfpp$ to highlight the dependency on the coefficients.
Note that both the orbital basis $\{\eta_\alpha(\bfrr)\}_{\alpha=1}^B$ and density basis $\{\omega_\mu(\bfrr)\}_{\mu=1}^M$ depend on the molecular structure $\clM$, as the location and type of each basis function is determined by the coordinates $\bfxx^{(a)}$ and atomic number $Z^{(a)}$ of the corresponding atom.
Nevertheless, the development in this subsection is for one given molecular system $\clM$, so we omit its appearance for density or orbital representation.
Typically, the numbers of basis $B$ and $M$ increase linearly with the number of electrons $N$, \ie, $O(B) = O(M) = O(N)$.

\subsubsection{KSDFT under Atomic Basis} \label{appx:dft-mat-ks}

For the SCF iteration in \eqnref{fock-eq}, using the expansion of orbitals in \eqnref{orb-expd}, it becomes:
$\sum_\beta \bfC_{\beta i}^{(k)} \Fh^{(k)} \eta_\beta(\bfrr) = \veps_i^{(k)} \sum_\beta \bfC_{\beta i}^{(k)} \eta_\beta(\bfrr),\;
\forall i = 1, \cdots, N$.
Integrating each function equation with basis function $\eta_\alpha(\bfrr)$ gives:
$\sum_\beta \bfC_{\beta i}^{(k)} \obraket*{\eta_\alpha}{\Fh^{(k)}}{\eta_\beta} = \veps_i^{(k)} \sum_\beta \bfC_{\beta i}^{(k)} \braket*{\eta_\alpha}{\eta_\beta}$,
which can then be formulated as a generalized eigenvalue problem in matrix form:
\begin{align}
  & \bfF^{(k)} \bfC^{(k)} = \bfS \bfC^{(k)} \bfveps^{(k)},
  \label{eqn:roothaan-eq} \\
  \text{where} \quad
  \bfF^{(k)}_{\alpha\beta} :={} & \obraket*{\eta_\alpha}{\Fh^{(k)}}{\eta_\beta}
  \stackrel{\text{\eqnref{fock-opr}}}{=} \underbrace{\obraket*{\eta_\alpha}{\Th}{\eta_\beta}}_{= -\frac12 \int \eta_\alpha(\bfrr) \nabla^2 \eta_\beta(\bfrr) \dd \bfrr =: \bfT_{\alpha\beta}}
  + \underbrace{\obraket*{\eta_\alpha}{V_\eff^{(k)}}{\eta_\beta}}_{=: (\bfV_\eff^{(k)})_{\alpha\beta}},
  \label{eqn:fock-mat} \\
  \bfS_{\alpha\beta} :={} & \braket*{\eta_\alpha}{\eta_\beta}, \quad
  \bfveps^{(k)} := \begin{psmallmatrix}
    \veps_1^{(k)} & & \\
    & \ddots & \\
    & & \veps_N^{(k)}
  \end{psmallmatrix}. \label{eqn:orbeng-mat}
\end{align}
Here, $\bfF^{(k)}$ is called the Fock matrix, and $\bfS$ is the overlap matrix of the orbital basis.

To show the expression of the Fock matrix, we first give the expression of the density defined by the orbital coefficients from \eqnref{den-orb-orthon} and \eqnref{orb-expd}:
\begin{align}
  \rho_\bfC(\bfrr) :={} & \rho_{[\Phi_\bfC]}(\bfrr)
  = \sum_{i=1}^N \lrvert*[\bigg]{\sum_\alpha \bfC_{\alpha i} \eta_\alpha(\bfrr)}^2
  = \sum_{i=1}^N \sum_{\alpha\beta} \bfC_{\alpha i} \bfC_{\beta i} \eta_\alpha(\bfrr) \eta_\beta(\bfrr) \\
  ={} & \sum_{\alpha\beta} \bfGamma_{\alpha\beta} \eta_\alpha(\bfrr) \eta_\beta(\bfrr),
  \label{eqn:den-denmat}
\end{align}
where we have defined the density matrix corresponding to the orbital coefficients:
\begin{align}
  \bfGamma := \bfC \bfC\trs, \quad \bfGamma_{\alpha\beta} := \sum_{i=1}^N \bfC_{\alpha i} \bfC_{\beta i}.
  \label{eqn:denmat-coeff}
\end{align}
Note that \eqnref{den-orb-orthon} requires orthonormal orbitals, and hence \eqnref{den-denmat} requires $\bfC$ to satisfy the corresponding orthonormality constraint shown in \eqnref{coeff-orb-orthon} below.
Orbital coefficient solutions $\bfC^{(k)}$ in SCF iterations satisfy this constraint as explained later.

In the derivation of SCF iteration, $V_\eff^{(k)}$ is taken as $V_{\eff[\rho_{[\Phi^{(k-1)}]}]}$, \ie, $\Fh^{(k)} = \Fh_{[\rho_{[\Phi^{(k-1)}]}]}$.
This allows explicit calculation of $\bfV_\eff^{(k)}$ as ${\bfV_\eff}_{\bfC^{(k-1)}}$ based on \eqnref{Veff-fn} and $\bfF^{(k)}$ as $\bfF_{\bfC^{(k-1)}}$, for which we introduce the following series of definitions:
\begin{align}
  & \bfF_\bfC := [\obraket*{\eta_\alpha}{\Fh_{[\rho_\bfC]}}{\eta_\beta}]_{\alpha\beta} = \bfT + {\bfV_\eff}_\bfC \quad \text{($\bfT$ defined in \eqnref{fock-mat})}, \\
  & {\bfV_\eff}_\bfC := [\obraket*{\eta_\alpha}{V_{\eff[\rho_\bfC]}}{\eta_\beta}]_{\alpha\beta}
  = {\bfV_\tnH}_\bfC + {\bfV_\XC}_\bfC + \bfV_\ext, \label{eqn:Veff-mat} \\
  \text{where} \quad
  & ({\bfV_\tnH}_\bfC)_{\alpha\beta} := \obraket*{\eta_\alpha}{V_{\tnH[\rho_\bfC]}}{\eta_\beta}
  \stackrel{\text{\eqnsref{Veff-fn,den-denmat}}}{=} \iint \eta_\alpha(\bfrr) \eta_\beta(\bfrr) \frac{\sum_{\gamma\delta} \bfGamma_{\gamma\delta} \eta_\gamma(\bfrr') \eta_\delta(\bfrr')}{\lrVert{\bfrr - \bfrr'}} \dd \bfrr' \ud \bfrr \\
  & \hspace{1.5cm}
  = \sum_{\gamma\delta} \bfDt_{\alpha\beta,\gamma\delta} \bfGamma_{\gamma\delta}
  = (\bfDt \bfGammab)_{\alpha\beta}, \label{eqn:VH-mat} \\
  & \hspace{1.6cm}
  \text{where  } \bfDt_{\alpha\beta,\gamma\delta} := (\eta_\alpha \eta_\beta | \eta_\gamma \eta_\delta), \label{eqn:cint-4c2e} %
  \text{ $\bfGammab$ is the vector of flattened $\bfGamma$}, \\
  & ({\bfV_\XC}_\bfC)_{\alpha\beta} := \obraket*{\eta_\alpha}{V_{\XC[\rho_\bfC]}}{\eta_\beta} = \int V_{\XC[\rho_\bfC]}(\bfrr) \eta_\alpha(\bfrr) \eta_\beta(\bfrr) \dd \bfrr, \label{eqn:VXC-mat} \\
  & (\bfV_\ext)_{\alpha\beta} := \obraket*{\eta_\alpha}{V_\ext}{\eta_\beta} = -\sum_{a=1}^A Z^{(a)} \int \frac{\eta_\alpha(\bfrr) \eta_\beta(\bfrr)}{\lrVert{\bfrr - \bfxx^{(a)}}} \dd \bfrr. \label{eqn:Vext-mat}
\end{align}
In defining $\bfDt$, we have used the notation of Coulomb integral for brevity:
\begin{align}
  (f|g) := \int \frac{f(\bfrr) g(\bfrr')}{\lrVert{\bfrr - \bfrr'}} \dd \bfrr \ud \bfrr'.
  \label{eqn:cint}
\end{align}
In practice, integrals in $\bfS$, $\bfT$, $\bfV_\ext$, and the Coulomb integral $\bfDt$ can be calculated analytically under Gaussian-Type Orbitals (GTO) as the basis $\{\eta_\alpha\}_{\alpha=1}^B$.
The integral in ${\bfV_\XC}_\bfC$ is conducted numerically on a quadrature grid, as typically used in a DFT calculation.
After convergence, the electronic energy can be calculated from the orbital coefficients $\bfC$ by:
\begin{align}
  & E(\bfC) := E[\Phi_{\bfC}]
  \stackrel{\text{\eqnref{engopt-orb}}}{=}
  T_\tnS(\bfC) + \underbrace{ E_\tnH(\bfC) + E_\XC(\bfC) + E_\ext(\bfC) }_{=: E_\eff(\bfC)}, \label{eqn:Etot-orb-coeff} \\
  \text{where} \quad
  & T_\tnS(\bfC) := \sum_{i=1}^N \obraket*{\phi_{i,\bfC}}{\Th}{\phi_{i,\bfC}}
  = \sum_{\alpha\beta} \bfGamma_{\alpha\beta} \bfT_{\alpha\beta} = \bfGammab\trs \bfTb, \label{eqn:ts-denmat} \\
  & E_\tnH(\bfC) := E_\tnH[\rho_\bfC]
  = \frac12 \sum_{\alpha\beta\gamma\delta} \bfGamma_{\alpha\beta} \bfDt_{\alpha\beta,\gamma\delta} \bfGamma_{\gamma\delta} = \frac12 \bfGammab\trs \bfDt \bfGammab, \label{eqn:EH-denmat} \sbox0{\ref{eqn:EXC-denmat}} \\
  & E_\XC(\bfC) := E_\XC[\rho_\bfC] = E_\XC \Big[ \sum_{\alpha\beta} \bfGamma_{\alpha\beta} \eta_\alpha \eta_\beta \Big], \label{eqn:EXC-denmat} \\
  & E_\ext(\bfC) := E_\ext[\rho_\bfC]
  = \sum_{\alpha\beta} \bfGamma_{\alpha\beta} (\bfV_\ext)_{\alpha\beta} = \bfGammab\trs \bfVb_\ext, \label{eqn:Eext-denmat}
\end{align}
where $\bfGammab$, $\bfTb$, $\bfVb_\ext$ are the vectors of flattened $\bfGamma$, $\bfT$, $\bfV_\ext$, respectively.
The term $E_\XC[\rho_\bfC]$ is again calculated by numerically integrating the defined density by $\bfC$ on a quadrature grid.

\paragraph{Computational Complexity}
Note that the construction of ${\bfV_\tnH}_\bfC$ from \eqnref{VH-mat} and the evaluation of $E_\tnH(\bfC)$ from \eqnref{EH-denmat} require $O(B^4)= O(N^4)$ complexity.
Even when using density fitting which decreases the complexity to $O(N^2)$, the complexity in each SCF iteration of KSDFT is $O(N^3)$ since the complexity of density fitting itself is $O(N^3)$ (see \appxref{dft-label-dfit}).

\paragraph{Orbital Orthonormality}
Under the atomic basis, orbital orthonormality $\braket*{\phi_i}{\phi_j} = \delta_{ij}$ becomes
$\sum_{\alpha\beta} \bfC_{\alpha i} \bfC_{\beta j} \braket*{\eta_\alpha}{\eta_\beta} = \delta_{ij}$, or in matrix form,
\begin{align}
  \bfC\trs \bfS \bfC = \bfI.
  \label{eqn:coeff-orb-orthon}
\end{align}
As mentioned after \eqnref{ks-eq}, only the normalization constraints need to be taken care of, as the orbitals are eigenstates of an Hermitian operator and hence are already orthogonal if non-degenerate.
This property transmits to the matrix form of the problem:\footnote{
  This can also be directly verified in the matrix form:
  for $i \ne j$, $(\bfC\trs \bfS \bfC)_{ij} = \bfC_{:i}\trs \bfS \bfC_{:j}
  \stackrel{\text{\eqnref{roothaan-eq}}}{=} \bfC_{:i}\trs \frac{1}{\veps_j} \bfF \bfC_{:j}
  \stackrel{\text{$\bfF$ is Hermitian}}{=} \frac{1}{\veps_j} (\bfF \bfC_{:i})\trs \bfC_{:j}
  \stackrel{\text{\eqnref{roothaan-eq}}}{=} \frac{\veps_i}{\veps_j} (\bfS \bfC_{:i})\trs \bfC_{:j}
  \stackrel{\text{$\bfS$ is Hermitian}}{=} \frac{\veps_i}{\veps_j} (\bfC\trs \bfS \bfC)_{ij}$,
  which indicates $\big( 1 - \frac{\veps_i}{\veps_j} \big) (\bfC\trs \bfS \bfC)_{ij} = 0$, thus $(\bfC\trs \bfS \bfC)_{ij} = 0$ assuming non-degeneracy $\veps_i \ne \veps_j$.
}
\begin{align}
  \bfC_{:i}\trs \bfS \bfC_{:i} = 1, \quad \forall i = 1, \cdots, N.
  \label{eqn:coeff-orb-normal}
\end{align}
Fulfilling the orthonormality constraint then only needs to normalize each eigenvector $\bfCt^{(k)}_{:i}$ of the problem in \eqnref{roothaan-eq} to form $\bfC^{(k)}$; explicitly,
$\bfC^{(k)}_{:i} = \bfCt^{(k)}_{:i} / \sqrt{{\bfCt^{(k)}_{:i}}\trs \bfS \bfCt^{(k)}_{:i}}$.

\paragraph{Relation to Direct Gradient Derivation}
The matrix form of the optimality equation under basis hence the SCF iteration problem \eqnref{roothaan-eq} can also be derived directly from \eqnref{engopt-orb} by taking the gradient of the energy function of coefficients:
$E(\bfC) := E[\Phi_{\bfC}] = E\big[ \{\sum_\alpha \bfC_{\alpha i} \eta_\alpha\}_{i=1}^N \big]$.
Its gradient is related to the variation of the functional of orbitals by integral with the basis:
\begin{align}
  \big( \nabla_\bfC E(\bfC) \big)_{\alpha i}
  = \int \fracdelta{E[\Phi_{\bfC}]}{\phi_i}(\bfrr) \big( \nabla_\bfC \phi_{i,\bfC}(\bfrr) \big)_{\alpha i} \dd \bfrr
  = \int \fracdelta{E[\Phi_{\bfC}]}{\phi_i}(\bfrr) \eta_\alpha(\bfrr) \dd \bfrr.
  \label{eqn:eng-orb-grad-variation}
\end{align}
The variation is given by \eqnref{eng-orb-variation}, which is $2 \Fh_{[\rho_{\Phi_\bfC}]} \phi_{i,\bfC}(\bfrr) = 2 \sum_\beta \bfC_{\beta i} \Fh_{[\rho_{\Phi_\bfC}]} \eta_\beta(\bfrr)$, which turns the gradient into matrix form:
\begin{align}
  \nabla_\bfC E(\bfC) = 2 \bfF_\bfC \bfC.
  \label{eqn:grad-and-fock}
\end{align}
For the orbital orthonormality constraint, as mentioned, only the normalization constraints require explicit treatment.
By introducing Lagrange multiplier $\veps_i$ for each constraint in \eqnref{coeff-orb-normal} and taking the gradient for the corresponding Lagrange term gives
  $\nabla_\bfC \sum_{i=1}^N \veps_i \big( \bfC_{:i}\trs \bfS \bfC_{:i} - 1 \big)
  = 2 \bfS \bfC \bfveps$.
This leads to the optimality equation in matrix form:
\begin{align}
  \bfF_\bfC \bfC = \bfS \bfC \bfveps.
  \label{eqn:ksdft-scf-converged}
\end{align}
By constructing the corresponding fixed-point iteration, \eqnref{roothaan-eq} is derived.

\paragraph{Accelerating and Stabilizing SCF Iteration}
As mentioned, $\bfF^{(k)}$ and $\bfV_\eff^{(k)}$ may be taken differently from $\bfF_{\bfC^{(k-1)}}$ and ${\bfV_\eff}_{\bfC^{(k-1)}}$ for more stable and faster convergence.
The direct inversion in the iterative subspace (DIIS) method~\citep{pulay1982improved,kudin2002blackbox} is a popular choice for this.
In DIIS, the Fock matrix $\bfF^{(k)}$ in the eigenvalue problem \eqnref{fock-eq} for each SCF iteration $k$ is taken as a weighted mixing of the vanilla Fock matrices $\bfF_{\bfC^{(k')}}, k' < k$ in previous steps:
\begin{align}
  \bfF^{(k)} := \sum_{k'=0}^{k-1} \pi^{(k)}_{k'} \bfF_{\bfC^{(k')}},
  \label{eqn:fock-mat-diis}
\end{align}
where $\{\pi^{(k)}_{k'}\}_{k'=0}^{k-1}$ are the weights that are positive and normalized $\sum_{k'=0}^{k-1} \pi^{(k)}_{k'} = 1$. %
Due to the normalization, the kinetic part $\bfT$ of the matrix remains the same, so it agrees with the form in \eqnref{fock-mat} (or \eqnref{fock-opr} in operator form), where:
\begin{align}
  \bfV_\eff^{(k)} := \sum_{k'=0}^{k-1} \pi^{(k)}_{k'} {\bfV_\eff}_{\bfC^{(k')}}.
  \label{eqn:Veff-mat-diis}
\end{align}

\subsubsection{OFDFT under Atomic Basis} \label{appx:dft-mat-of}

To solve the optimization problem of OFDFT in \eqnref{engopt-den}, it is unnatural to construct a fixed-point SCF iteration process from its variation in \eqnref{ts-den-variation}.
Hence, direct gradient-based density optimization is conducted.
For this, the energy functional of density function in \eqnref{engopt-den} needs to be converted into a function of density coefficients using the basis expansion of density function in \eqnref{den-expd}:
\begin{align}
  & E(\bfpp) := E[\rho_\bfpp] = E \Big[ \sum_\mu \bfpp_\mu \omega_\mu \Big]
  = T_\tnS(\bfpp) + \underbrace{ E_\tnH(\bfpp) + E_\XC(\bfpp) + E_\ext(\bfpp) }_{= E_\eff(\bfpp) := E_\eff[\rho_\bfpp]}, \label{eqn:Etot-den-coeff} \\
  \text{where} \quad
  & T_\tnS(\bfpp) := T_\tnS[\rho_\bfpp] = T_\tnS \Big[ \sum_\mu \bfpp_\mu \omega_\mu \Big], \label{eqn:ts-den-coeff} \\
  & E_\tnH(\bfpp) := E_\tnH[\rho_\bfpp]
  = \frac12 \iint \frac{\sum_\mu \bfpp_\mu \omega_\mu(\bfrr) \sum_\nu \bfpp_\nu \omega_\nu(\bfrr')}{\lrVert*{\bfrr - \bfrr'}} \dd \bfrr \ud \bfrr'
  = \frac12 \bfpp\trs \bfWt \bfpp, \label{eqn:EH-den-coeff} \\
  & E_\XC(\bfpp) := E_\XC[\rho_\bfpp] = E_\XC \Big[ \sum_\mu \bfpp_\mu \omega_\mu \Big], \label{eqn:EXC-den-coeff} \sbox0{\ref{eqn:EXC-den-coeff}} \\
  & E_\ext(\bfpp) := E_\ext[\rho_\bfpp]
  = \int \sum_\mu \bfpp_\mu \omega_\mu(\bfrr) V_\ext(\bfrr) \dd \bfrr
  = \bfpp\trs \bfvv_\ext, \label{eqn:Eext-den-coeff} \\
  & \text{where} \quad
  \bfWt_{\mu\nu} := (\omega_\mu | \omega_\nu), \label{eqn:cint-2c2e} \quad
  (\bfvv_\ext)_\mu := \braket*{\omega_\mu}{V_\ext}. \label{eqn:vext-vec}
\end{align}
The Coulomb integral notation $(\omega_\mu | \omega_\nu)$ is defined in \eqnref{cint}.
Recall that we have omitted the dependency of density basis $\{\omega_\mu\}_\mu$ hence of the functions \eg $T_\tnS(\bfpp)$ on the molecular structure $\clM$ in this subsection.
Integrals for $\bfWt$ and $\bfvv_\ext$ can be calculated directly~\citep{dupuis1976evaluation,rys1983computation} under Gaussian-Type Orbitals (GTO) as the basis $\{\omega_\mu\}_\mu$, using software libraries \eg \texttt{libcint}~\citep{sun2015libcint} in PySCF~\citep{sun2018pyscf}.
The term $E_\XC[\rho_\bfpp]$ is calculated by numerically integrating the defined density $\rho_\bfpp$ on a quadrature grid as typically used in a DFT calculation.
In our \ourmethod, $T_\tnS(\bfpp)$ is calculated directly from the coefficient $\bfpp$ using the deep learning model $T_{\tnS,\theta}(\bfpp,\clM)$.

To carry out direct optimization using a learned KEDF model $T_{\tnS,\theta}(\bfpp)$, the gradient of the electronic energy in \eqnref{Etot-den-coeff} is required, which is given by:
\begin{align}
  \nabla_\bfpp E_\theta(\bfpp) = \nabla_\bfpp T_{\tnS,\theta}(\bfpp) + \underbrace{ \bfWt \bfpp + \nabla_\bfpp E_\XC(\bfpp) + \bfvv_\ext }_{\nabla_\bfpp E_\eff(\bfpp)}.
  \label{eqn:Etot-grad-den-coeff}
\end{align}
This gradient is then used to update the density coefficient after projected onto the linear subspace of normalized densities, following \eqnref{den-gd}.

\paragraph{Relation to Derivation as Integral of the Variation with Basis}
The gradient $\nabla_\bfpp E(\bfpp)$ can also be derived by the relation between gradient and variation that we have already seen in \eqnref{eng-orb-grad-variation}:
\begin{align}
  \big( \nabla_\bfpp E(\bfpp) \big)_\mu
  = \big( \nabla_\bfpp E[\rho_\bfpp] \big)_\mu
  = \int \fracdelta{E[\rho_\bfpp]}{\rho}(\bfrr) \big( \nabla_\bfpp \rho_\bfpp(\bfrr) \big)_\mu \dd \bfrr
  = \int \fracdelta{E[\rho_\bfpp]}{\rho}(\bfrr) \omega_\mu(\bfrr) \dd \bfrr.
  \label{eqn:eng-den-grad-variation}
\end{align}
Integrating the variations given in \eqnref{Veff-fn} with the basis functions $\{\omega_\mu\}_\mu$ recovers the Hartree energy gradient $\bfWt \bfpp$ and external energy gradient $\bfvv_\ext$ in \eqnref{Etot-grad-den-coeff}.
The formula also applies to the gradient of the kinetic energy $\nabla_\bfpp T_\tnS(\bfpp)$ and the gradient of the XC energy $\nabla_\bfpp E_\XC(\bfpp)$.

\paragraph{Automatic Differentiation Implementation for Calculating the Gradient}
In the implementation of \ourmethod, the gradient of the KEDF model $\nabla_\bfpp T_{\tnS,\theta}(\bfpp,\clM)$ is evaluated directly using automatic differentiation~\citep{baydin2018automatic}, which can be conveniently done if implementing the model using common deep learning programming frameworks, \eg, PyTorch~\citep{paszke2019pytorch}.
To calculate $\nabla_\bfpp E_\XC(\bfpp)$ conveniently, we also re-implemented the PBE XC functional~\citep{perdew1996generalized} in PySCF using PyTorch and evaluate its gradient also by automatic differentiation.
For material systems, automatic differentiation implementation of OFDFT is also developed recently~\citep{tan2023automatic}.
When using the residual version $T_{\res,\theta}$ of KEDF model, which is detailed in \eqnref{tsmodel-res} in \appxref{func-var} later, the base KEDF (taken as the APBE KEDF~\citep{constantin2011semiclassical}) is also implemented in this way.

\paragraph{Computational Complexity}
As will be detailed in \appxref{model}, the Transformer-based~\citep{vaswani2017attention} KEDF model for our \ourmethod has a quadratic complexity $O(A^2) = O(N^2)$.
The PBE functional~\citep{perdew1996generalized} for $E_\XC$ and the APBE functional~\citep{constantin2011semiclassical} for the base KEDF are at the GGA level (generalized gradient approximation), so evaluating the energies amounts to calculating the density features with $O(M)$ cost on each grid point, in total with $O(M N_\grid)$ cost where $N_\grid$ is the number of grid points, and then conducting the quadrature with $O(N_\grid)$ cost.
The complexity for these energies is thus $O(M N_\grid)$ which is also quadratic $O(N^2)$ since $N_\grid = O(N)$ (though with a large prefactor).
Evaluating the gradient using automatic differentiation is in the same order of evaluating the function, hence also has $O(N^2)$ complexity.
Evaluating $E_\tnH(\bfpp)$ and $E_\ext(\bfpp)$ using \eqnref{EH-den-coeff} and \eqnref{Eext-den-coeff} and their gradients using \eqnref{Etot-grad-den-coeff} require $O(M^2) = O(N^2)$ complexity.
Therefore, the complexity in \ourmethod has a quadratic complexity $O(N^2)$, which is indeed lower than that of KSDFT (detailed in \appxref{dft-mat-ks} above).

Besides the advantage in asymptotic complexity, the fact that \ourmethod is implemented in PyTorch(see \appxref{model_config}) enables it to leverage GPUs efficiently.
These factors jointly facilitate the much higher throughput of \ourmethod than KSDFT.

\subsection{Label Calculation under Atomic Basis} \label{appx:dft-label}

This subsection details the calculation of the data tuple $(\bfpp^{(k)}, T_\tnS^{(k)}, \nabla_\bfpp T_\tnS^{(k)})$ for learning a KEDF model from the orbital coefficients $\bfC^{(k)}$ of the orbital solution $\Phi^{(k)} := \{\phi_i^{(k)}\}_{i=1}^N$ in each SCF iteration $k$.
Following the previous subsection, we omit the appearance of $\clM$ for density or orbital representation (\eg, in $T_\tnS(\bfpp, \clM)$) and omit the index $d$ for different molecular systems.
We insist keeping the $k$ index to reflect that the deduction is based on the solution in an SCF iteration but does not apply to arbitrary orbital coefficients $\bfC$.

\subsubsection{Density Fitting} \label{appx:dft-label-dfit}

We start with calculating the density coefficient $\bfpp^{(k)}$ under the density basis $\{\omega_\mu(\bfrr)\}_{\mu=1}^M$ for representing the density defined by the orbital coefficient solution $\bfC^{(k)}$.
This process is called \emph{density fitting}~\citep{dunlap2000robust}, which is also used in KSDFT for acceleration, in which context the atomic basis for the density is also called \emph{auxiliary basis}.
The density coefficient $\bfpp^{(k)}$ needs to fit represented density $\rho_{\bfpp^{(k)}}$ by \eqnref{den-expd} to the density $\rho_{\bfC^{(k)}}$ defined by \eqnref{den-denmat}.
Noting that \eqnref{den-denmat} essentially expands the density onto the paired basis $\{\eta_\alpha(\bfrr) \eta_\beta(\bfrr)\}_{\alpha\beta}$ with coefficient as the vector $\bfGammab^{(k)}$ of flattened $\bfGamma^{(k)}$,
this is the classical coordinate transformation problem from the paired basis to the density basis.
The classical approach is by minimizing the standard L2-norm of the residual density:
\begin{align}
  \int \lrvert{ \rho_{\bfpp^{(k)}}(\bfrr) - \rho_{\bfC^{(k)}}(\bfrr) }^2 \dd \bfrr
  ={} & {\bfpp^{(k)}}\trs \bfW \bfpp^{(k)} - 2 {\bfpp^{(k)}}\trs \bfL \bfGammab^{(k)} + {\bfGammab^{(k)}}\trs \bfD \bfGammab^{(k)},
\end{align}
where $\bfW_{\mu\nu} := \braket*{\omega_\mu}{\omega_\nu}$, $\bfL_{\mu,\alpha\beta} := \braket*{\omega_\mu}{\eta_\alpha \eta_\beta}$, and $\bfD_{\alpha\beta,\gamma\delta} := \braket*{\eta_\alpha \eta_\beta}{\eta_\gamma \eta_\delta}$ are the overlap matrices of the density basis, between the density basis and the paired basis, and of the paired basis, respectively.
Noting that this is an quadratic form of $\bfpp^{(k)}$, we know the solution is
$\bfpp^{(k)} = \bfW^{-1} \bfL \bfGammab^{(k)}$.

However, the standard L2-metric on the density function may not be the most favorable metric for density fitting. Instead, energy is the directly concerned quantity.
The kinetic energy is the most desired metric, since this would minimize the mismatch of the fitted density $\bfpp^{(k)}$ to the kinetic energy label $T_\tnS^{(k)}$. But there is no explicit expression to calculate the kinetic energy from density coefficient.
We hence turn to using the Hartree energy and the external energy as the metric.
(Using the XC energy requires an arbitrary choice of a functional approximation, and the calculation is more costly.)

For the Hartree energy as defined in \eqnref{hart-def}, noting that it is quadratic in density, we fit $\bfpp$ by minimizing the ($2 \times$) Hartree energy arising from the residual density:
\begin{align}
  & 2 E_\tnH[\rho_{\bfpp^{(k)}} - \rho_{\bfC^{(k)}}]
  = \iint \frac{\big( \rho_{\bfpp^{(k)}}(\bfrr) - \rho_{\bfC^{(k)}}(\bfrr) \big) \big( \rho_{\bfpp^{(k)}}(\bfrr') - \rho_{\bfC^{(k)}}(\bfrr') \big)}{\lrVert*{\bfrr - \bfrr'}} \dd \bfrr \ud \bfrr' \\
  ={} & {\bfpp^{(k)}}\trs \bfWt \bfpp^{(k)} - 2 {\bfpp^{(k)}}\trs \bfLt \bfGammab^{(k)} + {\bfGammab^{(k)}}\trs \bfDt \bfGammab^{(k)},
\end{align}
where symbols with tilde are the corresponding overlap matrices under integral kernel $\frac{1}{\lrVert{\bfrr - \bfrr'}}$, which, using the symbol of Coulomb integral defined in \eqnref{cint}, are
$\bfWt_{\mu\nu} := (\omega_\mu | \omega_\nu)$, $\bfLt_{\mu,\alpha\beta} := (\omega_\mu | \eta_\alpha \eta_\beta)$, and $\bfDt_{\alpha\beta,\gamma\delta} := (\eta_\alpha \eta_\beta | \eta_\gamma \eta_\delta)$ as already defined in \eqnref{cint-4c2e}.
As a quadratic form, the solution is
$\bfpp^{(k)} = \bfWt^{-1} \bfLt \bfGammab^{(k)}$.
This result can be understood as if the Hartree energy (Coulomb integral) defines a metric on the space of density functions. %

For the external energy as defined in \eqnref{ext-def}, as it is linear in density, we fit $\bfpp$ by directly minimizing the difference between the defined external energies:
\begin{align}
  \big( E_\ext[\rho_{\bfpp^{(k)}}] - E_\ext[\rho_{\bfC^{(k)}}] \big)^2
  = \big( E_\ext(\bfpp^{(k)}) - E_\ext(\bfC^{(k)}) \big)^2
  \stackrel{\text{\eqnsref{Eext-den-coeff,Eext-denmat}}}{=}{} & ( {\bfpp^{(k)}}\trs \bfvv_\ext - {\bfGammab^{(k)}}\trs \bfVb_\ext )^2.
\end{align}

To combine the two metrics, the final optimization problem is a combined least squares problem:
\begin{align}
  \bfpp^{(k)} = \argmin_\bfpp \bfpp\trs \bfWt \bfpp - 2 \bfpp\trs \bfLt \bfGammab^{(k)} + {\bfGammab^{(k)}}\trs \bfDt \bfGammab^{(k)}
    + ( \bfpp\trs \bfvv_\ext - {\bfGammab^{(k)}}\trs \bfVb_\ext )^2,
\end{align}
which corresponds to the over-determined linear equations in matrix form:
\begin{align}
  \sfbW \bfpp^{(k)} = \sfbbb^{(k)},
  \quad \text{where  }
  \sfbW := \begin{pmatrix}
    \bfWt \\ \bfvv_\ext\trs
  \end{pmatrix},
  \sfbbb^{(k)} := \begin{pmatrix}
    \bfLt \bfGammab^{(k)} \\ {\bfGammab^{(k)}}\trs \bfVb_\ext
  \end{pmatrix}.
\end{align}
This is directly solved using least-squares solvers.
In this conversion, we did not explicitly consider the normalization constraint, ${\bfpp^{(k)}}\trs \bfww = N$, since it is already satisfied with a high accuracy, due to the close fit to the original density.

\subsubsection{Value Label Calculation} \label{appx:dft-label-val}

The label for the value of KEDF can be calculated from \eqnref{ks-label} by leveraging the expression \eqnref{ts-denmat} under atomic basis:
\begin{align}
  T_\tnS(\bfC^{(k)}) = \sum_{\alpha\beta} \bfGamma^{(k)}_{\alpha\beta} \bfT_{\alpha\beta} = {\bfGammab^{(k)}}\trs \bfTb,
\end{align}
where $\bfGamma^{(k)} = \bfC^{(k)} {\bfC^{(k)}}\trs$ from \eqnref{denmat-coeff}, and $\bfGammab^{(k)}$ is the vector by flattening.
The corresponding density coefficient $\bfpp^{(k)}$ is calculated from $\bfC^{(k)}$ using density fitting as detailed above.
A subtlety arises since the fitted density $\bfpp^{(k)}$ may differ a little from the original density defined by $\bfC^{(k)}$ due to finite-basis incompleteness, so $T_\tnS(\bfC^{(k)})$ may not be the best kinetic energy label for $\bfpp^{(k)}$.
Indeed, as mentioned, in density fitting, we do not have a way to directly find the density coefficient $\bfpp^{(k)}$ that achieves the kinetic energy closest to $T_\tnS(\bfC^{(k)})$.
Instead, $\bfpp^{(k)}$ is fitted to match the Hartree and external energy.
We hence assume that the electronic energy is less affected by density-fitting error than the kinetic energy, and instead of taking $T_\tnS(\bfpp^{(k)}) \approx T_\tnS(\bfC^{(k)})$, we take
$E(\bfpp^{(k)}) \approx E(\bfC^{(k)})$, which means $T_\tnS(\bfpp^{(k)}) + E_\eff(\bfpp^{(k)}) \approx T_\tnS(\bfC^{(k)}) + E_\eff(\bfC^{(k)})$.
Hence the KEDF value label for $\bfpp^{(k)}$ is taken as:
\begin{align}
  T_\tnS^{(k)} :={} & T_\tnS(\bfC^{(k)}) + E_\eff(\bfC^{(k)}) - E_\eff(\bfpp^{(k)}), \\
  \text{where} \quad
  E_\eff(\bfC^{(k)}) ={} & \underbrace{\frac12 {\bfGammab^{(k)}}\trs \bfDt \bfGammab^{(k)}}_{E_\tnH(\bfC^{(k)})} + E_\XC(\bfC^{(k)}) + \underbrace{{\bfGammab^{(k)}}\trs \bfVb_\ext}_{E_\ext(\bfC^{(k)})}, \\
  E_\eff(\bfpp^{(k)}) ={} & \underbrace{\frac12 {\bfpp^{(k)}}\trs \bfWt \bfpp^{(k)}}_{E_\tnH(\bfpp^{(k)})} + E_\XC(\bfpp^{(k)}) + \underbrace{{\bfpp^{(k)}}\trs \bfvv_\ext}_{E_\ext(\bfpp^{(k)})},
\end{align}
following definitions and expressions of Eqs.~(\ref{eqn:Etot-orb-coeff}-\ref{eqn:Eext-denmat}) and Eqs.~(\ref{eqn:Etot-den-coeff}-\ref{eqn:Eext-den-coeff}).
Labels for the two variants of functional models detailed in \appxref{func-var} can be calculated accordingly.
For the residual version of KEDF $T_\res$, the label is correspondingly modified using values at $\bfpp^{(k)}$:
\begin{align}
  T_\res^{(k)} := T_\tnS^{(k)} - T_\APBE(\bfpp^{(k)}).
\end{align}
For the version $E_\TXC$ that learns the sum of KEDF and the XC energy, the corresponding label is:
$T_\tnS^{(k)} + E_\XC(\bfpp^{(k)})
= T_\tnS(\bfC^{(k)}) + \big( E_\tnH(\bfC^{(k)}) + E_\XC(\bfC^{(k)}) + E_\ext(\bfC^{(k)}) \big) - \big( E_\tnH(\bfpp^{(k)}) + E_\XC(\bfpp^{(k)}) + E_\ext(\bfpp^{(k)}) \big) + E_\XC(\bfpp^{(k)})
= T_\tnS(\bfC^{(k)}) + E_\XC(\bfC^{(k)}) + \big( E_\tnH(\bfC^{(k)}) - E_\tnH(\bfpp^{(k)}) + E_\ext(\bfC^{(k)}) - E_\ext(\bfpp^{(k)}) \big)$,
where the last term can be omitted since the Hartree energy difference and the external energy difference are minimized by $\bfpp^{(k)}$ in density fitting.
We therefore take the label as:
\begin{align}
  E_\TXC^{(k)} := T_\tnS(\bfC^{(k)}) + E_\XC(\bfC^{(k)}).
\end{align}

\subsubsection{Gradient Label Calculation} \label{appx:dft-label-grad}

Under an atomic basis, the kinetic energy functional of density is converted into a function of density coefficient $T_\tnS(\bfpp) := T_\tnS[\rho_\bfpp]$ following \eqnref{ts-den-coeff}.
For its gradient $\nabla_\bfpp T_\tnS(\bfpp)$, following the fact in \eqnref{eng-den-grad-variation}, it is related to the variation of the functional $T_\tnS[\rho]$ by integral with the basis:
\begin{align}
  \big( \nabla_\bfpp T_\tnS(\bfpp) \big)_\mu
  = \int \fracdelta{T_\tnS[\rho_\bfpp]}{\rho}(\bfrr) \big( \nabla_\bfpp \rho_\bfpp(\bfrr) \big)_\mu \dd \bfrr
  = \int \fracdelta{T_\tnS[\rho_\bfpp]}{\rho}(\bfrr) \omega_\mu(\bfrr) \dd \bfrr.
  \label{eqn:ts-den-grad-variation}
\end{align}
The variation corresponding to a known density is given by \eqnref{ts-den-variation} above, which comes from the solution of orbitals in a KSDFT SCF iteration.
If omitting the error in density fitting and approximating $\fracdelta{}{\rho} T_\tnS[\rho_{\bfpp^{(k)}}]$ with $\fracdelta{}{\rho} T_\tnS[\rho_{\bfC^{(k)}}] = \fracdelta{}{\rho} T_\tnS[\rho_{[\Phi^{(k)}]}]$, then the gradient can be accessed by integrating both sides of \eqnref{ts-den-variation} with the density basis:
\begin{align}
  & \nabla_\bfpp T_\tnS(\bfpp^{(k)}) + \bfvv_\eff^{(k)} = \mu^{(k)} \bfww,
  \label{eqn:ts-grad} \\
  \text{where} \quad
  & (\bfvv_\eff^{(k)})_\mu := \braket*{\omega_\mu}{V_\eff^{(k)}} = \int V_\eff^{(k)}(\bfrr) \omega_\mu(\bfrr) \dd \bfrr, \quad
  \label{eqn:veff-vec}
  \bfww_\mu := \int \omega_\mu(\bfrr) \dd \bfrr.
\end{align}
In practice, the chemical potential $\mu^{(k)}$ is not needed, since the gradient matters in its projection on the tangent space of normalized densities in order to keep the density normalized in density optimization (see \eqnref{den-gd}).
The space of normalized densities is a linear space $\{\bfpp \mid \bfpp\trs \bfww = N\}$ since $\int \rho_\bfpp(\bfrr) \dd \bfrr = \sum_\mu \bfpp_\mu \int \omega_\mu(\bfrr) \dd \bfrr = N$, so it coincides with its tangent space.
The projection onto the tangent space is achieved by applying $\bfI - \frac{\bfww \bfww\trs}{\bfww\trs \bfww}$, which gives:
\begin{align}
  \Big( \bfI - \frac{\bfww \bfww\trs}{\bfww\trs \bfww} \Big) \nabla_\bfpp T_\tnS(\bfpp^{(k)}) = -\Big( \bfI - \frac{\bfww \bfww\trs}{\bfww\trs \bfww} \Big) \bfvv_\eff^{(k)}.
  \label{eqn:ts-grad-proj}
\end{align}
Due to the same reason, the gradient loss function \eqnref{loss-grad} also only matches the projected gradient of the model to the projected gradient label.

The remaining task is to evaluate $\bfvv_\eff^{(k)}$ in \eqnref{veff-vec}.
Considering the complication that $V_\eff^{(k)}(\bfrr)$ may not be taken as the explicit-form effective potential $V_{\eff[\rho_{[\Phi^{(k-1)}]}]}(\bfrr) = V_{\eff[\rho_{\bfC^{(k-1)}}]}$ (\eqnref{Veff-fn} and \eqnref{den-denmat}), such as when DIIS (\eqnref{Veff-mat-diis}) is used in SCF iteration,
evaluating $\bfvv_\eff^{(k)}$ can be done by leveraging the orbital-basis representation $\bfV_\eff^{(k)}$ already available in the SCF problem (\eqnref{roothaan-eq}), which is defined in \eqnref{fock-mat} as the integral with paired orbital basis:
$(\bfV_\eff^{(k)})_{\alpha\beta} := \int V_\eff^{(k)}(\bfrr) \eta_\alpha(\bfrr) \eta_\beta(\bfrr) \dd \bfrr$.
Using the expansion coefficients $\bfK$ of the density basis onto the paired orbital basis, that is $\omega_\mu(\bfrr) = \sum_{\alpha\beta} \bfK_{\mu,\alpha\beta} \eta_\alpha(\bfrr) \eta_\beta(\bfrr)$,
we can conduct the conversion by $(\bfvv_\eff^{(k)})_\mu = \sum_{\alpha\beta} \bfK_{\mu,\alpha\beta} (\bfV_\eff^{(k)})_{\alpha\beta}$.

However, solving for $\bfK$ is unaffordable:
using least squares, this amounts to solving $\bfD \bfK = \bfL$ (or solving $\bfDt \bfK = \bfLt$). Since $\bfD$ (or $\bfDt$) has shape $B^2 \times B^2$, the complexity is $O(M B^4) = O(N^5)$.
Even it is only called once for one molecular structure, the cost is still intractable even for medium-sized molecules.
Moreover, the approximation $\fracdelta{}{\rho} T_\tnS[\rho_{\bfpp^{(k)}}] \approx \fracdelta{}{\rho} T_\tnS[\rho_{\bfC^{(k)}}]$ does not guarantee the optimality of density optimization using the learned KEDF model on the same molecular structure as explained later.

We hence turn to another approximation, and use a more direct way to calculate the gradient.
The approximation is that \eqnref{ts-den-variation} also holds for the fitted density $\rho_{\bfpp^{(k)}}$:
\begin{align}
  \fracdelta{T_\tnS[\rho_{\bfpp^{(k)}}]}{\rho} + V_{\eff\{\bfpp^{(k')}\}_{k'<k}} = {\mu'}^{(k)},
  \label{eqn:ts-den-variation-den-coeff}
\end{align}
where $V_{\eff\{\bfpp^{(k')}\}_{k'<k}}$ is the $V_\eff^{(k)}$ constructed from fitted density coefficients in previous SCF iterations, instead of orbital coefficient solutions in previous SCF iterations that the $V_\eff^{(k)}$ in \eqnref{ts-den-variation} uses.
The chemical potential ${\mu'}^{(k)}$ may be different, but due to the above argument for \eqnref{ts-grad-proj}, it is not used.
The approximation holds if density fitting error can be omitted, \eg when using a large basis set.
Following the procedure above, the corresponding kinetic-energy gradient after projection is given by:
\begin{align}
  & \Big( \bfI - \frac{\bfww \bfww\trs}{\bfww\trs \bfww} \Big) \nabla_\bfpp T_\tnS(\bfpp^{(k)}) = -\Big( \bfI - \frac{\bfww \bfww\trs}{\bfww\trs \bfww} \Big) \bfvv_{\eff\{\bfpp^{(k')}\}_{k'<k}}, \label{eqn:ts-grad-proj-label} \\
  \text{where} \quad
  & \big( \bfvv_{\eff\{\bfpp^{(k')}\}_{k'<k}} \big)_\mu := \braket*[\big]{\omega_\mu}{V_{\eff\{\bfpp^{(k')}\}_{k'<k}}} = \int V_{\eff\{\bfpp^{(k')}\}_{k'<k}}(\bfrr) \omega_\mu(\bfrr) \dd \bfrr,
  \label{eqn:veff-vec-den-coeff}
\end{align} correspondingly.
Calculating $\bfvv_{\eff\{\bfpp^{(k')}\}_{k'<k}}$ requires a direct approach.
This can be done following the relation between the known functions $V_{\eff[\rho_{[\Phi^{(k')}]}]}$ and the constructed $V_\eff^{(k)}$ in the SCF iteration.
In DIIS, this relation can be drawn from the construction in \eqnref{Veff-mat-diis} by noting the definitions \eqnref{fock-mat} and \eqnref{Veff-mat} of the matrices, which is a weighted average:
  $V_\eff^{(k)} = \sum_{k'=0}^{k-1} \pi^{(k)}_{k'} V_{\eff[\rho_{\bfC^{(k')}}]}$.
Following this pattern, the required effective potential in \eqnref{ts-den-variation-den-coeff} is constructed as:
$V_{\eff\{\bfpp^{(k')}\}_{k'<k}} = \sum_{k'=0}^{k-1} \pi^{(k)}_{k'} V_{\eff[\rho_{\bfpp^{(k')}}]}$.
The weights $\{\pi^{(k)}_{k'}\}_{k'=0}^{k-1}$ are taken as the same as those computed in the SCF iteration.
Its vector form under the density basis is given by:
\begin{align}
  \bfvv_{\eff\{\bfpp^{(k')}\}_{k'<k}} = \sum_{k'=0}^{k-1} \pi^{(k)}_{k'} {\bfvv_\eff}_{\bfpp^{(k')}}, \label{eqn:veff-vec-den-coeff-diis} \quad
  \text{where  }
  \big( {\bfvv_\eff}_\bfpp \big)_\mu := \braket*{\omega_\mu}{V_{\eff[\rho_\bfpp]}} = \int V_{\eff[\rho_\bfpp]}(\bfrr) \omega_\mu(\bfrr) \dd \bfrr.
\end{align}
Calculation of each ${\bfvv_\eff}_{\bfpp^{(k)}}$ can be carried out directly following \eqnref{Veff-fn} that gives $V_{\eff[\rho_\bfpp]}$ explicitly.
In our implementation, each ${\bfvv_\eff}_{\bfpp^{(k)}}$ is conveniently calculated using our automatic differentiation implementation mentioned in \appxref{dft-mat-of}, since we notice the fact that:
\begin{align}
  {\bfvv_\eff}_\bfpp = \nabla_\bfpp E_\eff(\bfpp),
  \label{eqn:veff-vec-den-coeff-as-grad}
\end{align}
where $E_\eff(\bfpp) := E_\eff[\rho_\bfpp]$ is defined in \eqnref{Etot-den-coeff} and its gradient $\nabla_\bfpp E_\eff(\bfpp)$ is given by \eqnref{Etot-grad-den-coeff}.
This fact is again due to the relation between gradient and variation revealed in \eqnref{eng-den-grad-variation} and noting that $V_{\eff[\rho]}$ is defined as the variation $\fracdelta{E_\eff[\rho]}{\rho}$ of the effective energy functional in \eqnref{Veff-fn}.
As analyzed at the end of \appxref{dft-mat-of}, evaluating the gradient has the same complexity as evaluating the energy $E_\eff(\bfpp)$, which is $O(M^2) + O(M N_\grid) = O(N^2)$, which is much lower than the $O(N^5)$ complexity above.

To sum up, $\{{\bfvv_\eff}_{\bfpp^{(k')}}\}_{k'}$ are first calculated using automatic differentiation following \eqnref{veff-vec-den-coeff-as-grad}, which are used to construct $\bfvv_{\eff\{\bfpp^{(k')}\}_{k'<k}}$ following \eqnref{veff-vec-den-coeff-diis}, then the gradient $\nabla_\bfpp T_\tnS(\bfpp^{(k)})$ is given by \eqnref{ts-grad-proj-label} up to a projection.
Since the loss function \eqnref{loss-grad} for gradient supervision explicitly projects the gradient error, the gradient label itself does not have to be projected before evaluating the loss (\ie, the loss is the same whether the gradient label is projected; since projection is idempotent).
We hence take the gradient label $\nabla_\bfpp T_\tnS^{(k)}$ directly as the density-constructed DIIS effective potential vector:
\begin{align}
  \nabla_\bfpp T_\tnS^{(k)} := -\bfvv_{\eff\{\bfpp^{(k')}\}_{k'<k}}.
  \label{eqn:ts-grad-label-unproj}
\end{align}
For the residual version of KEDF $T_\res$ and the version $E_\TXC$ that also includes XC energy as detailed in \appxref{func-var}, the labels are produced accordingly:
\begin{align}
  \nabla_\bfpp T_\res^{(k)} := \nabla_\bfpp T_\tnS^{(k)} - \nabla_\bfpp T_\APBE(\bfpp^{(k)}), \quad
  \nabla_\bfpp E_\TXC^{(k)} := \nabla_\bfpp T_\tnS^{(k)} + \nabla_\bfpp E_\XC(\bfpp^{(k)}),
\end{align}
where $\nabla_\bfpp T_\APBE(\bfpp^{(k)})$ and $\nabla_\bfpp E_\XC(\bfpp^{(k)})$ are also calculated using automatic differentiation.

Apart from the convenient and efficient calculation using automatic differentiation, this choice of gradient label could also train a KEDF model that leads to the correct optimal density, since the labeling approach is compatible with the density optimization procedure of \ourmethod shown in \eqnref{den-gd} and \eqnref{Etot-grad-den-coeff}.
More specifically, upon the convergence of SCF iteration for which we mark quantities with ``$\star$'', the DIIS effective potential $V_\eff^\star$ is converged to the single-step, explicit-form effective potential $V_{\eff[\rho_{\bfC^\star}]}$ (\eqnref{Veff-fn} and \eqnref{den-denmat}) by design.
Correspondingly, the density-constructed DIIS effective potential vector $\bfvv_{\eff\{\bfpp^{(k')}\}_{k'<\star}}$ (\eqnref{veff-vec-den-coeff}) at convergence coincides with ${\bfvv_\eff}_{\bfpp^\star}$ (\eqnref{veff-vec-den-coeff-diis}), which is $\nabla_\bfpp E_\eff(\bfpp^\star)$ by \eqnref{veff-vec-den-coeff-as-grad}.
This gives a gradient label to the KEDF model through \eqnref{ts-grad-proj-label}, which enforces the model to satisfy:
\begin{align}
  \Big( \bfI - \frac{\bfww \bfww\trs}{\bfww\trs \bfww} \Big) \big( \nabla_\bfpp T_{\tnS,\theta}(\bfpp^\star) + \nabla_\bfpp E_\eff(\bfpp^\star) \big)
  \stackrel{\text{\eqnref{Etot-grad-den-coeff}}}{=} \Big( \bfI - \frac{\bfww \bfww\trs}{\bfww\trs \bfww} \Big) \nabla_\bfpp E_\theta(\bfpp^\star)
  = 0,
  \label{eqn:ts-grad-label-optimal}
\end{align}
which in turn enforces the density optimization process \eqnref{den-gd} to converge to $\bfpp^\star$, the true ground-state density coefficient.
The optimality of density optimization using the learned KEDF model can then be expected.

\subsection{Force Calculation} \label{appx:hfforce}

The force experienced by atoms in a molecular structure is an important quantity as it is directly required for geometry optimization and molecular dynamics simulation.
It is also used as a metric to evaluate the results of \ourmethod (Results~\ref{sec:res-in-scale}, \appxref{res-out-scale-main}).
There are different ways to calculate the force in both KSDFT and OFDFT, and the results may differ. %
Here we describe two common ways for force calculation, which are the Hellmann-Feynman (HF) force~\citep{hellman1937einfuhrung,feynman1939forces} and the analytical force~\citep{pulay1969ab}.
Evaluation protocol using force for \ourmethod and \mlff/\mlffden against KSDFT is detailed at the end.

\paragraph{Hellmann-Feynman Force}
Force is the negative gradient of the total energy of a molecule as a function $E_\tot(\bfX)$ of atom coordinates $\bfX = \{\bfxx^{(a)}\}_{a=1}^A$ (molecular conformation).
We omit the dependency on the molecular composition $\bfZ$ for brevity.
The total energy $E_\tot(\bfX) := E^\star_\bfX + E_\nuc(\bfX)$ comprises both the electronic energy $E^\star_\bfX$ in electronic ground state (including interaction with the nuclei), and also the energy from inter-nuclear interaction:
\begin{align}
  E_\nuc(\bfX) := \frac12 \sum_{\substack{a,b = 1, \cdots, A, \\ a \ne b}} \frac{Z^{(a)} Z^{(b)}}{\lrVert*{\bfxx^{(a)} - \bfxx^{(b)}}},
  \label{eqn:eng-nuc}
\end{align}
which gives the inter-nuclear part of the force,
\begin{align}
  -\nabla_{\bfxx^{(a)}} E_\nuc(\bfX) = -Z^{(a)} \sum_{b :\ne a} Z^{(b)} \frac{\bfxx^{(b)} - \bfxx^{(a)}}{\lrVert*{\bfxx^{(b)} - \bfxx^{(a)}}^3}.
  \label{eqn:force-nuc}
\end{align}
The electronic energy $E^\star_\bfX$ is the minimum after a variational optimization process for solving the electronic ground state of the molecule in conformation $\bfX$.
In the most fundamental form, $E^\star_\bfX$ is determined by the variational problem on $N$-electron wavefunctions as shown in \eqnref{engopt-wavefn}.
The Hamiltonian operator therein $\Hh_\bfX = \Th + \Vh_\ee + \Vh_{\ext,\bfX}$ depends on the conformation $\bfX$ through $V_{\ext,\bfX}$, which is given in \eqnref{Vext-def}.
The ground-state wavefunction $\psi^\star_\bfX$ and energy $E^\star_\bfX = \obraket*{\psi^\star_\bfX}{\Hh_\bfX}{\psi^\star_\bfX}$ hence also depend on $\bfX$.
The gradient of $E^\star_\bfX$ can then be reformed as: $\nabla_\bfX E^\star_\bfX =$
\begin{align}
  \nabla_\bfX \obraket*{\psi^\star_\bfX}{\Hh_\bfX}{\psi^\star_\bfX}
  &= \obraket*{\nabla_\bfX \psi^\star_\bfX}{\Hh_\bfX}{\psi^\star_\bfX} + \obraket*{\psi^\star_\bfX}{\nabla_\bfX \Hh_\bfX}{\psi^\star_\bfX} + \obraket*{\psi^\star_\bfX}{\Hh_\bfX}{\nabla_\bfX \psi^\star_\bfX} \\
  &\stackrel{\text{(*)}}{=} E^\star_\bfX \braket*{\nabla_\bfX \psi^\star_\bfX}{\psi^\star_\bfX} + \obraket*{\psi^\star_\bfX}{\nabla_\bfX \Hh_\bfX}{\psi^\star_\bfX} + E^\star_\bfX \braket*{\psi^\star_\bfX}{\nabla_\bfX \psi^\star_\bfX} \\
  &= \obraket*{\psi^\star_\bfX}{\nabla_\bfX \Hh_\bfX}{\psi^\star_\bfX} + E^\star_\bfX \nabla_\bfX \braket*{\psi^\star_\bfX}{\psi^\star_\bfX} \\
  &\stackrel{\text{(\#)}}{=} \obraket*{\psi^\star_\bfX}{\nabla_\bfX \Hh_\bfX}{\psi^\star_\bfX},
  \label{eqn:hf-theorem}
\end{align}
where the equality (*) is due to that $\psi^\star_\bfX$ is an eigenstate of the Hermitian operator $\Hh_\bfX$ with real eigenvalue $E^\star_\bfX$, and the equality (\#) is due to that the wavefunction is normalized $\braket*{\psi^\star_\bfX}{\psi^\star_\bfX} = 1$ for all $\bfX$.
\eqnref{hf-theorem} is the Hellmann-Feynman (HF) theorem~\citep{hellman1937einfuhrung,feynman1939forces}.
To continue the calculation, the gradient of the Hamiltonian operator in the expression can be derived as $\nabla_{\bfxx^{(a)}} \Hh_\bfX = \nabla_{\bfxx^{(a)}} \Vh_{\ext,\bfX}$, and by noting that $\Vh_{\ext,\bfX}$ is multiplicative and one-body as shown in \eqnref{Vext-def}, we have
$\nabla_{\bfxx^{(a)}} V_{\ext,\bfX}(\bfrr) = -Z^{(a)} \frac{\bfrr - \bfxx^{(a)}}{\lrVert*{\bfrr - \bfxx^{(a)}}^3}$,
and subsequently, the electronic force can be calculated as:
\begin{align}
  -\nabla_{\bfxx^{(a)}} E^\star_\bfX
  = -\obraket*{\psi^\star_\bfX}{\nabla_{\bfxx^{(a)}} \Hh_\bfX}{\psi^\star_\bfX}
  = Z^{(a)} \int \rho^\star_\bfX(\bfrr) \frac{\bfrr - \bfxx^{(a)}}{\lrVert*{\bfrr - \bfxx^{(a)}}^3} \dd \bfrr
  =: \bfff^{(a)}_\bfX,
  \label{eqn:hfforce}
\end{align}
where $\rho^\star_\bfX(\bfrr) := \rho_{[\psi^\star_\bfX]}(\bfrr)$ defined in \eqnref{den-def}.
This is the \emph{Hellmann-Feynman (HF) force} $\bfff_\bfX$.
This expression coincides with the electrostatic force under a classical view, indicating ``there are no `mysterious quantum-mechanical forces' acting in molecules''~\citep{levine2009quantum}.
From the expression, evaluating the HF force only requires a good approximation to the ground-state electron density $\rho^\star_\bfX(\bfrr)$, which is available in various quantum chemistry methods, including $\rho_{\bfC^\star_\bfX}$ in KSDFT given by \eqnref{den-denmat} and $\rho_{\bfpp^\star_\bfX}$ in \ourmethod given by \eqnref{den-expd}.
The total force on the nuclei is the sum with the inter-nuclear force in \eqnref{force-nuc}.

\paragraph{Analytical Force}
However, when using the atomic basis, there emerges approximation error in the function representation, since the basis is incomplete.
Consequently, the conditions of the HF theorem do not hold exactly.
This makes the HF force in \eqnref{hfforce} only an approximation to the true electronic force $-\nabla_{\bfxx^{(a)}} E^\star_\bfX$, and other approximations are possible.
For example, $-\nabla_{\bfxx^{(a)}} E^\star_\bfX$ can also be estimated by directly taking the analytical gradient of the electronic energy $E^\star_\bfX$ expressed under the atomic basis with the optimal coefficients~\citep{pulay1969ab}.
This way of estimating the electronic force is hence called \emph{analytical force} $\bfff_{\ana,\bfX}$.
For KSDFT, $E^\star_\bfX = E_\bfX(\bfC^\star_\bfX)$, where $E_\bfX(\bfC)$ is given by Eqs.~(\ref{eqn:Etot-orb-coeff}-\ref{eqn:Eext-denmat}) (note that the basis functions $\eta_\alpha$, $\eta_\beta$ and matrices $\bfT$, $\bfDt$ and $\bfV_\ext$ all depend on $\bfX$), and $\bfC^\star_\bfX$ is the optimal orbital coefficients.
The corresponding analytical force on an atom $a$ can be expanded as:
\begin{align}
  \bfff^{(a)}_{\ana,\bfX}
  := -\nabla_{\bfxx^{(a)}} E_\bfX(\bfC^\star_\bfX)
  = -(\nabla_{\bfxx^{(a)}} E_\bfX) (\bfC^\star_\bfX)
  - \tr\Big( (\nabla_{\bfxx^{(a)}} \bfC^\star_\bfX)\trs \nabla_\bfC E_\bfX(\bfC) \Big|_{\bfC = \bfC^\star_\bfX} \Big),
  \label{eqn:anaforce}
\end{align}
where $(\nabla_{\bfxx^{(a)}} E_\bfX)$ takes the gradient with a fixed $\bfC$, and $\nabla_{\bfxx^{(a)}} \bfC^\star_\bfX$ is the Jacobian, $(\nabla_{\bfxx^{(a)}} \bfC^\star_\bfX)_{\alpha i, \xi} := \partial(\bfC^\star_\bfX)_{\alpha i} / \partial \bfxx^{(a)}_\xi$ in which $\xi \in \{1,2,3\}$ indices the three spacial components, and matrix operations, including transpose, matrix multiplication and trace, act on indices $\alpha$ and $i$.
When $\bfC^\star_\bfX$ is indeed accurately optimized, self-consistency in \eqnref{ksdft-scf-converged} and orthonormality in \eqnref{coeff-orb-orthon} are satisfied.
By also noting \eqnref{grad-and-fock}, the second term in \eqnref{anaforce} vanishes:
$\tr\big( (\nabla_{\bfxx^{(a)}} \bfC^\star_\bfX)\trs \nabla_\bfC E_\bfX(\bfC) \big|_{\bfC = \bfC^\star_\bfX} \big)
\stackrel{\text{\eqnref{grad-and-fock}}}{=} 2 \tr\big( (\nabla_{\bfxx^{(a)}} \bfC^\star_\bfX)\trs \bfF_{\bfC^\star_\bfX} \bfC^\star_\bfX \big)
\stackrel{\text{\eqnref{ksdft-scf-converged}}}{=} 2 \tr\big( (\nabla_{\bfxx^{(a)}} \bfC^\star_\bfX)\trs \bfS_\bfX \bfC^\star_\bfX \bfveps^\star_\bfX \big)
= \tr\Big( (\nabla_{\bfxx^{(a)}} \bfC^\star_\bfX)\trs \nabla_{\bfC^\star_\bfX} \tr\big( ( {\bfC^\star_\bfX}\trs \bfS_\bfX \bfC^\star_\bfX - \bfI ) \bfveps^\star_\bfX \big) \Big)
= \nabla_{\bfxx^{(a)}} \tr\big( ( {\bfC^\star_\bfX}\trs \bfS_\bfX \bfC^\star_\bfX - \bfI ) \bfveps^\star_\bfX \big)
\stackrel{\text{\eqnref{coeff-orb-orthon}}}{=} 0$.
If in practice $\bfC^\star_\bfX$ is not exactly optimized, the contribution from $\nabla_{\bfxx^{(a)}} \bfC^\star_\bfX$ should also be considered; see \citet{pulay1969ab} for detailed treatments.

Note that only the $-(\nabla_{\bfxx^{(a)}} \bfVb_{\ext,\bfX})\trs \bfGammab^\star_\bfX$ term ($\bfGammab^\star_\bfX$ is the vector of flattened optimal density matrix $\bfGamma^\star_\bfX := \bfC^\star_\bfX {\bfC^\star_\bfX}\trs$), as a part of $-(\nabla_{\bfxx^{(a)}} E_\bfX) (\bfC^\star_\bfX)$, corresponds to the HF force (see \eqnref{Eext-denmat}).
Other terms in the analytical force $\bfff^{(a)}_{\ana,\bfX}$ in \eqnref{anaforce} are collectively called the Pulay force after~\citep{pulay1969ab}, \ie, $\bfff_{\ana,\bfX} - \bfff_\bfX$.
They come in the form of the gradient of the basis functions with respect to atom coordinates, and hence are non-zero when using atomic basis and differentiate the analytical force from the HF force.

\paragraph{Implementation}
Although the analytical force is regarded as a more accurate estimation when using atomic basis, we still take the HF force to evaluate the results, since the way to calculate the analytical force is different for different methods:
KSDFT and \ourmethod require different types of Coulomb integrals under different basis sets, and \mlff and \mlffden only require back-propagating the gradient through the deep learning model.
Moreover, the HF force in \eqnref{hfforce} only depends on the density from the ground-state solution, so it can also be seen as a relevant metric to evaluate the solved density.

For each molecular system, we take the true value of HF force as that given by the KSDFT solution, where the density is taken after density fitting (see \appxref{dft-label-dfit}).
The HF force by \ourmethod is calculated directly from the optimized density.
For \mlff and \mlffden, as they cannot provide the density, only the corresponding analytical forces, $-\nabla_\bfX E_{\tot,\theta}(\bfX)$ and $-\nabla_\bfX E_{\tot,\theta}(\bfX, \bfpp^\init)$, are available.
Note that since they predict the total energy, the inter-nuclear force $-\nabla_\bfX E_\nuc(\bfX)$ (\eqnref{force-nuc}) is included in the gradients.
To convert them into HF forces, we extract from the gradients with the inter-nuclear force as well as the Pulay force:
$\bfff_{\textnormal{\mlff},\bfX}^{(a)} = -\nabla_{\bfxx^{(a)}} E_{\tot,\theta}(\bfX) - (-\nabla_{\bfxx^{(a)}} E_\nuc(\bfX)) - (\bfff_{\ana,\bfX}^{(a)} - \bfff^{(a)}_\bfX)$,
and similarly for $\bfff_{\textnormal{\mlffden},\bfX}^{(a)}$, where the analytical force $\bfff_{\ana,\bfX}^{(a)}$ and the HF force $\bfff^{(a)}_\bfX$ are calculated from KSDFT.
Following previous works~\citep{chmiela2017machine,schutt2018schnet}, the error in predicted force is measured by the mean absolute error (MAE) over each of the three spacial components of the force on each atom in each molecule in the test set.

\subsection{Scaling Property under Atomic Basis} \label{appx:dft-scaling-property}

The KEDF has an exact scaling property which describes its change after uniformly scaling (stretching or squeezing) a density.
Under a scaling (squeezing) rate $\lambda$, the uniformly scaled density is given by the rule of change of variables:
$\lambdah \rho(\bfrr) := \lambda^3 \rho(\lambda \bfrr)$.
The scaling property is stated as the following~\citep{parr1989density}:
\begin{align}
  T_\tnS[\lambdah \rho] = \lambda^2 T_\tnS[\rho].
  \label{eqn:ts-scaling}
\end{align}
Designing the model to exactly satisfy this condition not only guarantees reasonable results in some physical sense, but would also reduce the functional space where the model needs to search by learning, hence improving accuracy and the generalization and extrapolation ability.
Some prior investigations~\citep{hollingsworth2018can,kalita2021learning} for machine-learning KEDF indeed observed accuracy improvement in some cases (though not as substantial in some other cases).

Now we consider leveraging this property for the KEDF model $T_\tnS(\bfpp, \clM)$ under atomic basis.
As input density is represented under an atomic basis $\{\omega_\mu\}_{\mu=1}^M$ using coefficient $\bfpp$ as $\rho_\bfpp$ given by \eqnref{den-expd}, we need to first represent the scaled density $\lambdah \rho_\bfpp$ under the same basis with coefficient $\lambdah(\bfpp)$:
\begin{align}
  \lambdah \rho_\bfpp := \sum_\mu \bfpp_\mu \lambdah \omega_\mu(\bfrr)
  = \sum_\mu \lambdah(\bfpp)_\mu \omega_\mu(\bfrr).
  \label{eqn:scaled-den-expd}
\end{align}
Solving for $\lambdah(\bfpp)$ using least squares gives
$\lambdah(\bfpp) = \bfW^{-1} \bfW^{(\lambda)} \bfpp$,
where $\bfW^{(\lambda)}_{\mu\nu} := \braket*{\omega_\mu}{\lambdah \omega_\nu}$ here, and $\bfW_{\mu\nu} := \braket*{\omega_\mu}{\omega_\nu}$ as the same as above.
The scaling property \eqnref{ts-scaling} is then transformed as:
\begin{align}
  T_\tnS(\bfW^{-1} \bfW^{(\lambda)} \bfpp, \clM) = \lambda^2 T_\tnS(\bfpp, \clM).
  \label{eqn:ts-scaling-den-coeff}
\end{align}

However, to preserve \eqnref{ts-scaling} exactly, the expansion \eqnref{scaled-den-expd} must hold exactly.
But this is not the case: the finite basis set $\{\omega_\mu\}_{\mu=1}^M$ is incomplete, and the scaled basis functions $\{\lambdah \omega_\mu\}_{\mu=1}^M$ are not linearly dependent on the original ones.
Hence, the transformed scaling property \eqnref{ts-scaling-den-coeff} under the atomic basis is \emph{not exact}, thus may not provide much benefit to the KEDF model $T_{\tnS,\theta}(\bfpp, \clM)$.

There seems to be a possibility when using the even-tempered atomic basis set, which comes in the following form:
\begin{align}
  & \omega_{\mu=(a,\tau,\bfxi)}(\bfrr) = w_{a,\tau,\bfxi}(\bfrr - \bfxx^{(a)}), \\
  \text{where} \quad &
  w_{a,\tau,\bfxi}(\bfrr=(x,y,z)) := x^{\bfxi_1} y^{\bfxi_2} z^{\bfxi_3} \exp(-\alpha_{a,\lrvert{\bfxi}} \beta^\tau \lrVert{\bfrr}^2),
  \label{eqn:even-tempered-basis}
\end{align}
with tempering ratio $\beta > 1$, monomial-exponent parameter $\bfxi = (\bfxi_1, \bfxi_2, \bfxi_3)$, and exponent parameter $\alpha_{a,\lrvert{\bfxi}}$ shared across $\bfxi$ values with the same $\lrvert{\bfxi} := \bfxi_1 + \bfxi_2 + \bfxi_3$.
The index $\tau$ runs from 0 to $\clT_{a,\lrvert{\bfxi}}$.
Therefore, under the scaling with $\beta^{-\frac12}$, we have:
$\widehat{\beta^{-\frac12}} \omega_{\mu=(a,\tau,\bfxi)}(\bfrr)
= \widehat{\beta^{-\frac12}} w_{a,\tau,\bfxi}(\bfrr - \bfxx^{(a)})
= \beta^{-\frac32} w_{a,\tau,\bfxi}(\beta^{-\frac12} \bfrr - \bfxx^{(a)})
= (\beta^{-\frac12})^{3+\lrvert{\bfxi}} w_{a,\tau-1,\bfxi}(\bfrr - \bfxx^{(a)} / \beta^{-\frac12})$,
which is in the same even-tempered basis set but on a scaled molecular structure $\clM / \beta^{-\frac12} := \{\bfZ, \bfX / \beta^{-\frac12}\}$, as long as $\tau > 0$.
For $\tau = 0$, it corresponds to the most flat basis function, and its coefficient $\bfpp_{\mu=(a,\tau=0,\bfxi)}$ is close to zero and hence can be omitted for density representation.
Therefore, under the scaling with $\beta^{-\frac12}$, the scaled density can be expressed as:
\begin{align}
  \widehat{\beta^{-\frac12}} \rho_{\bfpp,\clM}(\bfrr)
  &= \sum_{a,\bfxi} \sum_{\tau=0}^{\clT_{a,\lrvert{\bfxi}}} \bfpp_{a,\tau,\bfxi} \widehat{\beta^{-\frac12}} \omega_{a,\tau,\bfxi}(\bfrr) \\
  &= \sum_{a,\bfxi} \sum_{\tau=1}^{\clT_{a,\lrvert{\bfxi}}} \bfpp_{a,\tau,\bfxi} (\beta^{-\frac12})^{3+\lrvert{\bfxi}} w_{a,\tau-1,\bfxi}(\bfrr - \bfxx^{(a)} / \beta^{-\frac12}) + \sum_{a,\bfxi} \bfpp_{a,0,\bfxi} \widehat{\beta^{-\frac12}} \omega_{a,0,\bfxi}(\bfrr) \\
  &\approx \sum_{a,\bfxi} \sum_{\tau=1}^{\clT_{a,\lrvert{\bfxi}}} \bfpp_{a,\tau,\bfxi} (\beta^{-\frac12})^{3+\lrvert{\bfxi}} w_{a,\tau-1,\bfxi}(\bfrr - \bfxx^{(a)} / \beta^{-\frac12}) \\
  &=: \sum_{a,\bfxi} \sum_{\tau=0}^{\clT_{a,\lrvert{\bfxi}}} \bfpp^{(\beta^{-\frac12})}_{a,\tau,\bfxi} w_{a,\tau,\bfxi}(\bfrr - \bfxx^{(a)} / \beta^{-\frac12})
  = \rho_{\bfpp^{(\beta^{-\frac12})}, \clM / \beta^{-\frac12}}(\bfrr),
\end{align}
where the new coefficients are defined as:
\begin{align}
  \bfpp^{(\beta^{-\frac12})}_{a,\tau,\bfxi} :=
  \begin{cases}
    (\beta^{-\frac12})^{3+\lrvert{\bfxi}} \bfpp_{a,\tau+1,\bfxi}, & \tau < \clT_{a,\lrvert{\bfxi}}, \\
    0, & \tau = \clT_{a,\lrvert{\bfxi}}.
  \end{cases}
\end{align}
The scaling property \eqnref{ts-scaling} can then be written as:
\begin{align}
  T_\tnS(\bfpp^{(\beta^{-\frac12})}, \clM / \beta^{-\frac12}) = (\beta^{-\frac12})^2 T_\tnS(\bfpp, \clM).
\end{align}
However, this holds only for one value of scaling depending on the basis set (also for $(\beta^{-\frac12})^n$ for integer $n$ as long as the contributions from the first $n$ coefficients can be omitted).
Moreover, this constraint connects the original input to a \emph{scaled conformation} $\clM / \beta^{-\frac12}$, which can be far from an equilibrium structure.
The behavior of the $T_{\tnS,\theta}$ model for these scaled conformations may be less relevant for real applications.
Hence still, it does not seem definite to gain benefits from the scaling property.

Another possibility to exactly expressing the scaling property is to replace the molecular structure $\clM$ with the basis overlap matrix $\bfW$ to characterize the atomic basis.
In this way, the rescaling of atomic basis can be described by the change of the overlap matrix:
\begin{align}
  \lambdah \bfW_{\mu\nu} := \int \lambdah \omega_\mu(\bfrr) \, \lambdah \omega_\nu(\bfrr) \dd \bfrr
  = \lambda^3 \int \omega_\mu(\bfrr) \omega_\nu(\bfrr) \dd (\lambda \bfrr)
  = \lambda^3 \bfW_{\mu\nu}.
\end{align}
Hence, we do not need to expand the rescaled basis onto the original one, and the scaling property in \eqnref{ts-scaling} becomes:
\begin{align}
  T_\tnS(\bfpp, \lambda^3 \bfW) = \lambda^2 T_\tnS(\bfpp, \bfW).
\end{align}
It is promising future work to investigate whether the overlap matrix is sufficient to specify the spacial and type configurations of the basis functions, and to design proper model architecture for effectively processing such pairwise feature and ensuring the above scaling property.

\begin{figure}[H]
  \centering
  \includegraphics[width=0.92\textwidth]{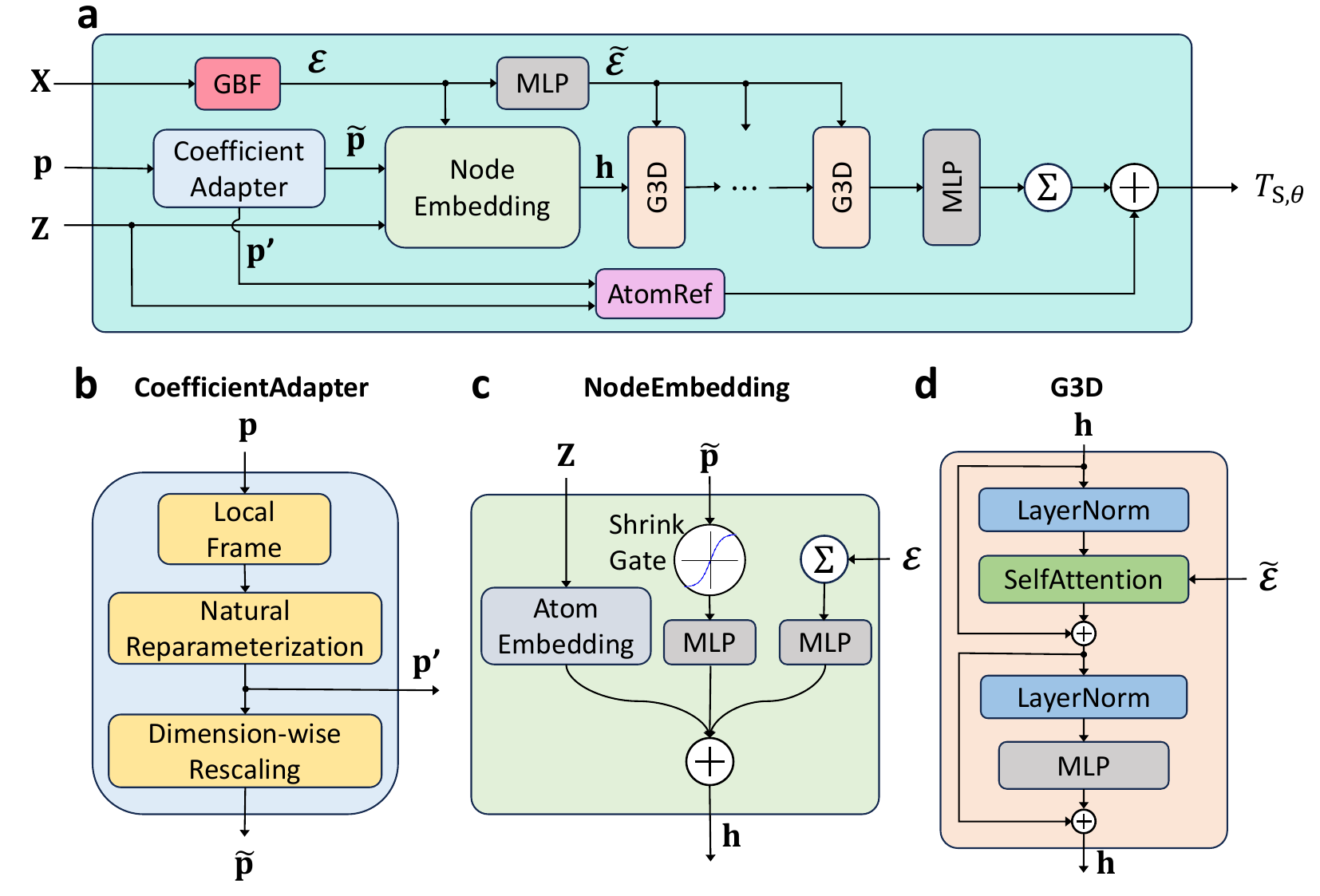}
  \caption{\textbf{The KEDF model architecture.}
  \textbf{(a)}~(See also Alg.~\ref{alg:kedf}.) Overview of the model architecture. The model calculates the non-interacting kinetic energy (or a variant of it) from the given density, specified by the density coefficient vector $\bfpp$ on an atomic basis set (\appxref{coeff-spec}), as well as the atomic numbers $\bfZ$ and coordinates $\bfX$ of all atoms in the target molecule for characterizing the basis functions.
  The coefficient vector $\bfpp$ is first processed by the \emph{CoefficientAdapter} module to make an $\mathrm{SE}(3)$-invariant density features $\bfppt$ and reduce the gradient scale for the rest of the model to fit.
  The $\bfppt$ features are then used to construct initial node features $\bfhh$ by the \emph{NodeEmbedding} module (\appxref{model-node-embed}), for which the types $\bfZ$ and positions $\bfX$ of the atoms are also incorporated to provide information to interpret the coefficient features.
  The positional input $\bfX$ is perceived by the model only in terms of pairwise distance features $\bm{\clE}$ produced by the Gaussian basis function (\emph{GBF}) module (\appxref{model-gbf}).
  The node features $\bfhh$ are subsequently updated by several Graphormer-3D (\emph{G3D}) modules (\appxref{model-g3d}). They calculate the interaction of features on every pair of nodes hence cover the non-local effect, in which relative position features $\tilde{\bm{\clE}}$ between each pair of nodes are considered, which are processed from $\bm{\clE}$ by a multi-layer perceptron (\emph{MLP}) module (\eqnref{mlp}).
  The final node features are aggregated by another MLP module and summed over the nodes to produce a scalar, and the final output is its addition with the output of the \emph{AtomRef} enhancement module (Methods~\ref{sec:learn-grad}, \appxref{atom-ref}) which shares the burden to model a large-scale gradient.
  \textbf{(b)}~(See also Alg.~\ref{alg:coeffada}.) Structure of the \emph{CoefficientAdapter} module. It consists of the \emph{LocalFrame} module (Methods~\ref{sec:local-frame}; \appxref{local-frame}) to convert $\mathrm{SE}(3)$-equivariant features into invariant features, followed by two enhancement modules \emph{NaturalReparameterization} and \emph{DimensionwiseRescaling} (Methods~\ref{sec:learn-grad}; \appxref{nat-reparam} and~\ref{appx:dim-rescale}) which reduce the gradient scale for subsequent modules to fit.
  \textbf{(c)}~Structure of the \emph{NodeEmbedding} module (\appxref{model-node-embed}). It integrates information from three sources for each node: atom-type (for basis-set type) $\bfZ$ which is mapped to a feature vector by atom embedding, positional feature in terms of pairwise distance feature $\bm{\clE}$ which is aggregated over nodes and processed by an MLP module, and density features $\bfppt$ which is mapped to a mild numerical range by a shrink gate and processed by another MLP module.
  \textbf{(d)}~Structure of the \emph{G3D} module (\appxref{model-g3d}). The \emph{SelfAttention} module updates node features by the non-local cross-node interaction of features $\bfhh$ based on spacial relation features $\tilde{\bm{\clE}}$ between nodes. An MLP module further processes the updated node features. The \emph{LayerNorm} module is applied before each of the two modules for numerical convenience.
  Note that the data on each streamline is the concatenated features across the nodes. Only the \emph{SelfAttention} module models the interaction across nodes, while other modules are applied to the features of each node independently.
  }
  \label{fig:model-arch}
\end{figure}

\section{\ourmethod Technical Details} \label{appx:model}

In this section, we provide more technical details of the proposed \ourmethod framework.
We first provide an overall description of the non-local neural network model for KEDF and detail the architecture of neural network components in \appxref{model-spec}.
We then elaborate on other components in the model that handle the unconventional challenges for learning a neural network model, including the use of local frame to guarantee geometric invariance of the model in \appxref{local-frame}, and enhancement modules for expressing a large gradient range in \appxref{tech-effi-train}. %
Furthermore, we discuss explorations on learning different density functional variants in \appxref{func-var}.
Finally, we detail the density optimization techniques in \appxref{den-opt}, which is the usage of \ourmethod to solve a given molecular structure.

\subsection{Model Specification} \label{appx:model-spec}

We start with more explanations on the design idea of the model formulation.
As motivated in the main paper, \ourmethod utilizes the expansion coefficients $\bfpp$ of the density under an atomic basis set as the direct density feature input into the functional model.
Compared to the grid-based representation, a common alternative for density representation, using the atomic-basis representation saves thousands of times the representation dimension.
This is critical for implementing a non-local calculation in the model, which is well-known vital for approximating KEDF~\citep{garcia1996nonlocal,wang1999orbital,wang2000orbital,mi2018nonlocal,teale2022dft}, %
but would otherwise be prohibited by the millions times more cost.
The aggregation to atoms further reduces the required number of interacting nodes (from grids to atoms) in the non-local calculation.
Noting that atomic basis are also used in material systems, so this way of density representation hence \ourmethod is not restricted to non-periodic molecular systems in formulation.
We note that there are also works in the context of learning the XC functional that develop machine-learning models also using features under atomic basis~\citep{dick2020machine,chen2021deepks}, but the models do not take the molecular structure in input and focus on processing the features hosted on each atom individually, hence are essentially not non-local models. %

Since the atomic basis depends on the molecular structure $\clM := \{\bfZ, \bfX\}$, the model also needs to include $\clM$ into its input for a complete specification of the input density, where the atom types $\bfZ := \{Z^{(a)}\}_{a=1}^A$ are used to specify the types of basis functions (since atoms of different elements are assigned with different types (\ie, parameters) of basis functions) and the conformation $\bfX := \{\bfxx^{(a)}\}_{a=1}^A$ to specify the centers of the basis functions.
Although including $\clM$ into the model input sacrifices formal universality, explicit dependency on $\clM$ is arguably inevitable for an efficient density representation, which requires the structure of the problem, \ie, the pattern of the density (``inductive bias'' in machine learning), to reduce the representation dimension.
Even the irregular grid representation inherits the pattern of molecular structure, hence a non-local model using grid input feature still requires generalization across molecular systems.
On the other hand, the non-local Graphormer model architecture, based on which our neural network model is designed, has shown attractive capability to generalize across conformations and chemicals for a range of molecular tasks in previous studies~\citep{ying2021transformers,shi2022benchmarking,zheng2023towards}.
Our extrapolation study in Results~\ref{sec:res-larger-scale} directly validates the superior generalization capability for OFDFT. %
Particularly, \figref{res-outscale}(c) shows that \ourmethod outperforms classical KEDFs which only use raw electron density input by a large margin even in an extrapolation setup, indicating the benefit of accuracy of non-local calculation enabled by the concise density representation based on the molecular structure $\clM$ outweighs the sacrifice of formal universality.
As for the generalization to molecules with unseen elements, since the atom type $Z$ here as perceived by the model only represents the type of basis functions that is used to hold the electron density near the atom but does not represent the physics of the actual nucleus or its interaction with electrons, we can assign the atom of unseen element with a seen atom type, and use the basis functions of the seen element to hold the electron density around that atom.
Nevertheless, %
due to a different electron structure, the model has not seen the pattern of density coefficients for the unseen element, so there exists a generalization challenge. This could potentially be mitigated by using a common basis set for all elements or including more elements in training, which will be investigated in future work.

The KEDF model takes atomic numbers $\bfZ$, positions $\bfX$, and density coefficients $\bfpp$ of all atoms in the molecule as input.
Note that given $\bfpp$ and $\clM$, there is no need of explicit density gradient or Laplacian features since they are transmitted to the features of the basis functions and thus already embodied in $\clM$.
The input variables are first used to construct node features encoding information about the electron density surrounding each atom. These features are further transformed and employed to predict the non-interacting kinetic energy. The gradient of the KEDF with respect to density coefficients for density optimization is obtained through auto-differentiation~\citep{paszke2019pytorch}.
Considering the fact that non-local calculation is indispensable for KEDF, we build the non-local model based on the Graphormer architecture~\citep{ying2021transformers,shi2022benchmarking}. Notably, Graphormer can handle varying-length input feature (as needed since different molecules have different numbers of atoms) in a permutation-invariant manner, which is not straightforward using other popular architectures such as multi-layer perceptions alone. The architecture has shown attractive performance in processing molecular structure to predict various properties, \eg, ground-state energy and HOMO-LUMO gap of molecular systems~\citep{ying2021transformers,shi2022benchmarking}, and also structure sampling from a thermodynamical ensemble~\citep{zheng2023towards}. %
The non-local architecture has an $O(N^2)$ complexity, which does not increase the complexity of OFDFT. Our empirical results in \appxref{ablat-nonlocal} indicate the non-local formulation is crucial; restricting atomic interactions with distance cutoffs leads to a performance decline on considered molecular systems.

\begin{algorithm}[t]
\caption{Evaluation of the KEDF model $T_{\tnS,\theta}(\bfpp,\clM)$ (or the kinetic residual model $T_{\res,\theta}(\bfpp,\clM)$ or the TXC model $E_{\TXC,\theta}(\bfpp,\clM)$; see also \appxfigref{model-arch})} \label{alg:kedf}
\begin{algorithmic}[1]
\Require Input molecular structure $\clM = \{\bfX, \bfZ\}$ comprising positions $\bfX:= \{\bfxx^{(a)}\}_{a=1}^A$ and atomic numbers $\bfZ := \{Z^{(a)}\}_{a=1}^A$ of all atoms in the molecule, input density coefficients $\bfpp$ (see \appxref{coeff-spec}).
\State Construct pairwise distance features $\bm{\clE} \leftarrow \texttt{GBF}(\bfX)$ and %
$\tilde{\bm{\clE}} \leftarrow \texttt{MLP}(\bm{\clE})$ (\eqnref{gbf}, \eqnref{mlp-pw-dist}, \eqnref{mlp});
\State Process coefficient features: $(\bfppt, \bfpp') \leftarrow \texttt{CoefficientAdapter}(\bfpp)$ (Alg.~\ref{alg:coeffada});
\State Construct initial atomic representations: $\bfhh \leftarrow \texttt{NodeEmbedding}(\bfZ, \bm{\clE}, \bfppt)$ (\eqnref{nodeembed});
\For {$i$ in $1 \cdots L$}
\State Update atomic representations using the $i$-th G3D module: $\bfhh \leftarrow \texttt{G3D}^{(i)}(\bfhh, \tilde{\bm{\clE}})$ (\eqnref{g3d}, \eqnref{layernorm}, \eqnref{selfatt});
\EndFor
\State Compute the output of the atomic reference module: $T' \leftarrow T_\mathrm{AtomRef}(\bfpp',\clM)$ (\eqnref{atom-ref});
\State Compute the kinetic energy: $T_{\tnS} \leftarrow \sum_{a=1}^A \texttt{MLP}(\bfhh_a) + T'$ (\eqnref{model-output});
\State \Return $T_{\tnS}$
\end{algorithmic}
\end{algorithm}

An overview of the KEDF model is sketched in \appxfigref{model-arch}(a) and summarized in Alg.~\ref{alg:kedf}.
The process can be narrated in four stages.

\itemi To adapt the Graphormer architecture for learning a physical functional, a few modifications are needed.
To address the additional learning challenges mentioned in Methods~\ref{sec:local-frame} and~\ref{sec:learn-grad}, the input density coefficients, formulated following the details in \appxref{coeff-spec}, are first processed by the CoefficientAdapter module (\appxfigref{model-arch}(b)), in which the local frame module (Methods~\ref{sec:local-frame}; \appxref{local-frame}) first converts the rotational equivariant coefficients to rotational invariant coefficients, and two enhancement modules NaturalReparameterization and DimensionwiseRescaling (Methods~\ref{sec:learn-grad}; \appxref{nat-reparam} and~\ref{appx:dim-rescale}) follow in order so as to reduce the gradient scale.

\itemii Before leveraging the Graphormer architecture, initial node (\ie, atom) features need to be prepared, which, in addition to the node type ($\bfZ$) and geometry ($\bm{\clE}$, processed from $\bfX$) information in the original version of Graphormer, density coefficients ($\bfppt$, processed from $\bfpp$) should also be blended in so that the resulted node features hold the information of electron density around the respective atoms. This process is handled by the NodeEmbedding module (\appxfigref{model-arch}(c); \appxref{model-node-embed}).
Note that the conformation input $\bfX$ is perceived by the model only in the form of pairwise distance features $\bm{\clE}$, which is produced by the Gaussian Basis Function (GBF) module (\appxref{model-gbf}) from pairwise distances of all atom pairs. These features are consumed by the NodeEmbedding module and also by the G3D module next. In this way, the model is naturally invariant with respect to the translation and rotation of atom coordinates $\bfX$.

\itemiii Several concatenated Graphormer modules then process the node features. Since in processing the interaction between node features, the spacial relation between the two nodes needs to be considered, we use the Graphormer-3D (G3D) version~\citep{shi2022benchmarking}, where the conformation information is input as pairwise distance features processed by the GBF module and a multi-layer perceptron (MLP) module. In the G3D module (\appxfigref{model-arch}(d); \appxref{model-g3d}), the SelfAttention module carries out calculation on any pair of node features, which covers non-local interaction or correlation of density features at distance.
The LayerNorm module in the G3D module is adopted following successful experiences of such models, which shifts and scales the feature distribution to make the following layer easier to process numerically.

\itemiv Finally, the last-layer node features are processed by MLP and then got aggregated into one scalar, which is added with the output of the AtomRef enhancement module $T_{\rm AtomRef}(\bfpp', \clM)$ (Methods~\ref{sec:learn-grad}; \appxref{atom-ref}) to construct the final energy output (\appxref{model-output}). The AtomRef module shares the burden to express large gradient thus reduces the difficulty to fit large gradient for the neural-network G3D branch.

We next detail the neural network components in the model, including GBF, NodeEmbedding, and G3D modules. Other rule-based or pre-arranged modules fall in standalone topics, so we provide their details in subsequent subsections, including local frame in \appxref{local-frame}, and enhancement modules of DimensionwiseRescaling, NaturalReparameterization and AtomRef in \appxref{tech-effi-train}.
We start with handling the formatting of density coefficient features.

\subsubsection{Density Basis and Coefficient Specification} \label{appx:coeff-spec}

As mentioned in Methods~\ref{sec:method}, \ourmethod adopts atomic basis as an efficient density representation for molecules.
Each basis function $\omega_\mu$ is specified by the position of the center atom and the basis function index, and we hence use $\mu = (a, \tau)$ to index the $\tau$-th basis function centered at atom $a$. The basis coefficients therefore can be correspondingly attributed to each atom, which naturally serve as node-wise density features for the atom point cloud.

Specifically, we choose an even-tempered basis set~\citep{bardo1974even} (\eqnref{even-tempered-basis}) as the density basis set, with its $\beta$ parameter taken as $2.5$.
Each atom type $Z$ (\ie, atomic number) has its own set of basis functions and the size of each set $\clT_Z$ varies from different atom types. The detailed composition of each atom type is summarized in \appxtblref{basis-orb-compose}.

\begin{figure}[ht]
  \centering
  \includegraphics[width=0.8\textwidth]{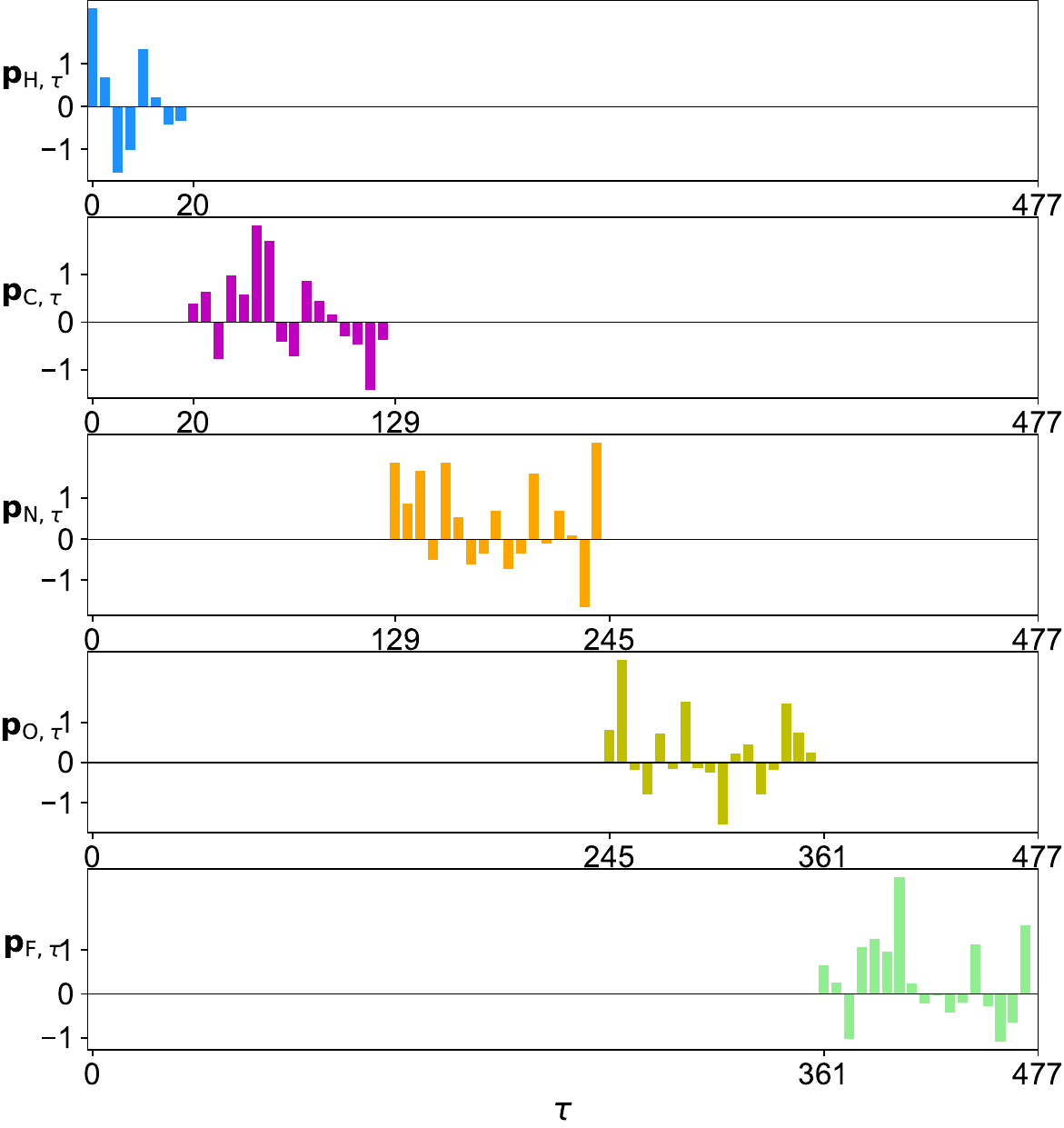}
  \caption{\textbf{Concatenation of atomic basis for different atom types.} $\tau$ is the index of density coefficient dimension and $\bfpp_{Z, \tau}$ denotes the coefficient vector for atom type $Z$. Note that this is a schematic illustration, so the coefficient dimensions may not correspond to the actual dimensions precisely.}
  \label{fig:aux-basis}
\end{figure}

The difference of basis functions on different types of atoms makes the coefficient features hold different meanings and even come with different dimensions. This is unconventional and challenging for machine learning models to process.
To make the coefficient vector homogeneous over all the atoms, the basis function sets on different atom types are joined together, and this united basis set is broadcast to all atoms, making a unified $\clT$-dimensional density coefficient vector $\bfpp_a$ on any atom $a$, where $\clT := \sum_Z \clT_Z$ is the sum of the number of basis functions over all considered atom types.
The final density coefficient vector for the entire system is thus the concatenation of these $\clT$-dimensional vectors: $\bfpp := \mathrm{concat}(\{\bfpp_a\}_{a=1}^A)$.
In more detail, in the QM9 dataset, the 477-dimensional concatenated coefficient vector consists of 20, 109, 116, 116 and 116 basis functions which correspond to $\rmH$, $\rmC$, $\rmN$, $\rmO$ and $\rmF$, respectively. For example, given the coefficient of a hydrogen ($\rmH$) atom, we place them at the first 20 dimensions and use zero-valued vectors to mask other positions.
A graphic illustration is shown in \appxfigref{aux-basis}.
The zero-valued mask is also employed to mask the predicted gradient during density optimization, avoiding introducing irrelevant gradient information from masked positions.

\begin{table}[h]
  \centering
  \caption{\textbf{Orbital composition of the basis set associated with each atom type.} Note that even though the compositions for N, O and F are the same, their basis sets can still be different from each other because they can have different exponents and contraction coefficients.}
  \begin{tabular}{cc}
    \toprule
    Atom Type & Orbitals \\
    \midrule
    $\rmH$ & 6s3p1d \\
    $\rmC$ & 11s8p7d3f2g \\
    $\rmN$ & 11s8p7d4f2g \\
    $\rmO$ & 11s8p7d4f2g \\
    $\rmF$ & 11s8p7d4f2g \\
    \bottomrule
  \end{tabular}
  \label{tbl:basis-orb-compose}
\end{table}

\subsubsection{Gaussian Basis Function (GBF) Module} \label{appx:model-gbf}

As shown in \appxfigref{model-arch}(a), the model converts the conformation (atom coordinates) input $\bfX$ into pairwise distance features before sent to the rest part of the model, so that geometric invariance with respect to $\bfX$ is guaranteed.
These features are produced by the GBF module. It first converts atom coordinates $\bfX = \{\bfxx^{(a)}\}_{a=1}^A$ into pairwise distances, and then expand each distance value $\lrVert{\bfxx_a - \bfxx_b}$ into a feature vector by evaluation under a series of Gaussian basis functions:
\begin{align}
  \text{define } \bm{\clE} := \texttt{GBF}(\bfX): \quad
  \bm{\clE}_{ab}^k := \frac{1}{\sqrt{2\pi} \sigma_{k}} e^{-\frac{(\lrVert{\bfxx_a - \bfxx_b} - \mu_k)^2}{2 \sigma_k^2}},
  \label{eqn:gbf}
\end{align}
where $\mu_k$ and $\sigma_k$ are learnable scalar parameters representing the center and scale of the $k$-th Gaussian basis function.

The pairwise distance features are used by the SelfAttention module in the G3D module (\appxfigref{model-arch}(d); \appxref{model-g3d}), which help identify the strength and ways of interactions between node features.
They are also used to construct the initial node features in the NodeEmbedding module introduced next (\appxfigref{model-arch}(c); \appxref{model-node-embed}).

\subsubsection{NodeEmbedding Module} \label{appx:model-node-embed}
As shown in \appxfigref{model-arch}(c), we design a NodeEmbedding module to integrate various input features into initial node features $\bfhh$, which, for each atom $a$, embody the atom type $Z^{(a)}$ (for specifying the type of basis functions centered at that location), the spacial position relation with other atoms (basis centered at other locations) specified by $\bfX$, and the density coefficient vector $\bfpp_a$ on that atom representing the electron density around the atom (see \appxref{coeff-spec} above). %
Specifically, \itemi for the atom type $Z^{(a)}$, an AtomEmbedding module assigns a learnable feature vector $\bfcc^{Z^{(a)}}$ to the atom according to its type $Z^{(a)}$.
\itemii For positional features encoding the spacial relation of the atom $a$ with respect to other atoms, distance features with respect to all other atoms are summed: $\sum_{b=1}^A \bm{\clE}_{ab}$.
\itemiii For the density coefficient $\bfpp_a$ on the atom $a$, it is first processed by the CoefficientAdapter module (\appxfigref{model-arch}(b)) detailed later in \appxref{local-frame} and \appxref{tech-effi-train}. The processed coefficient vector $\bfppt_a$ becomes translation and rotation (\ie, $\mathrm{SE}(3)$) invariant, and is properly transformed to exhibit a mild scale range of itself and of the corresponding gradient.
Due to the trade-off between the scale range of $\bfppt_a$ and the corresponding gradient, the scale of $\bfppt_a$ is still large for a neural network to process according to our trials. Therefore, we introduce a ShrinkGate module:
\begin{align}
  \texttt{ShrinkGate}(\bfppt_a) := \lambda_{\rm co} \, \tanh(\lambda_{\rm mul} \, \bfppt_a),
\end{align}
where $\lambda_{\rm co}$ and $\lambda_{\rm mul}$ are learnable scalar parameters. The $\tanh$ function is applied element-wise, which maps to a bounded space hence suppresses extreme values to avoid numerical challenges. Due to the $\mathrm{SE}(3)$-invariant property of $\bfppt_a$, applying such a nonlinear operation on it still preserves this geometry invariance.

Before aggregating features from the three sources, positional features and density coefficient features are processed by multi-layer perceptron (MLP) modules. MLP (also called feed-forward network in some context) are the most classical neural network architecture, which has been proven to be able to approximate any continuous function under certain limit~\citep{hornik1989multilayer}, and are indispensable components in modern neural network architectures.
Specifically, all the MLP modules in our model shown in \appxfigref{model-arch} follow the general expression: %
\begin{align}
  \texttt{MLP}(\bfxx) := \bfU^{(2)} \texttt{gelu}(\bfU^{(1)} \bfxx + \bfbb^{(1)}) + \bfbb^{(2)},
  \label{eqn:mlp}
\end{align}
where $\bfxx$ here represents a general feature vector of a unit (node or atom pair), $\bfU^{(1)}$ and $\bfU^{(2)}$ are learnable weight matrix parameters, $\bfbb^{(1)}$ and $\bfbb^{(2)}$ are learnable bias vector parameters, and $\texttt{gelu}(x) := x \Phi(x)$ for any scalar input $x \in \bbR$ is the Gaussian error linear unit activation function~\citep{hendrycks2016gaussian} that introduces nonlinearity to the module, where $\Phi(x)$ is the cumulative distribution function of the standard Gaussian distribution.
When applied to a feature vector, $\texttt{gelu}$ operates element-wise: $\texttt{gelu}(\bfxx)_k := \texttt{gelu}(\bfxx_k)$.

To sum up, the whole NodeEmbedding module can be formulated as:
\begin{align}
  \text{define } \bfhh :={} & \texttt{NodeEmbedding}(\bfZ, \bm{\clE}, \bfppt): \\
  \bfhh_a :={} & \bfcc^{Z^{(a)}} + \texttt{MLP}^{(\bm{\clE})} \Big( \sum_{b=1}^A \bm{\clE}_{ab} \Big) + \texttt{MLP}^{(\bfpp)} \big( \texttt{ShrinkGate}(\bfppt_a) \big),
  \label{eqn:nodeembed}
\end{align}
where $\texttt{MLP}^{(\bm{\clE})}$ and $\texttt{MLP}^{(\bfpp)}$ are two MLP instances with independent parameters.

\subsubsection{Graphormer-3D (G3D) Module} \label{appx:model-g3d}

The Graphormer-3D (G3D) module (\appxfigref{model-arch}(d)) is the main neural network module in the KEDF model (\appxfigref{model-arch}(a)). It is a Transformer-based~\citep{vaswani2017attention} graph neural network, which performs non-local calculation between every pair of node features even located at distance, while is tailored to taking into account the spacial relations between the two locations.
Each G3D module contains two layer normalization (LayerNorm) modules, a SelfAttention module, and an MLP module, assembled in the way shown in \appxfigref{model-arch}(d).

\paragraph{LayerNorm}
The Layer Normalization module~\citep{ba2016layer} is a widely adopted technique in Transformer-based architectures. It shifts and scales the running feature vector in a neural network so that the values over the vector components distribute with zero mean and unit variance, which is the numerical range over which neural network modules are the most sensitive, thus facilitates stable and faster training and a better fit to data.
This module is applied before sending node features to the SelfAttention module and the MLP module.
It normalizes the node feature vector on each node independently:
\begin{align}
  & \texttt{LayerNorm}(\bfhh_a) := \bfss \odot \frac{\bfhh_a - \mu_a}{\sigma_a} + \bfbb, \\
  \text{where } &
  \mu_a := \frac1{D_{\rm hid}} \sum_{k=1}^{D_{\rm hid}} \bfhh_a^k, \quad
  \sigma_a := \sqrt{\frac1{D_{\rm hid}} \sum_{k=1}^{D_{\rm hid}}(\bfhh_a^k - \mu_a)^2},
  \label{eqn:layernorm}
\end{align}
$D_{\rm hid}$ is the dimension of the input feature vector $\bfhh_a$ for node $a$, $\bfss \in \bbR^{D_{\rm hid}}$ and $\bfbb \in \bbR^{D_{\rm hid}}$ are learnable vector parameters, and $\odot$ denotes element-wise multiplication.

\paragraph{SelfAttention}
The SelfAttention module is the processor responsible for non-local calculation that produces the result of interaction between any two node feature vectors.
For updating feature vector on node (\ie, atom) $a$, the vanilla SelfAttention module~\citep{vaswani2017attention} considers interaction of the current feature vector $\bfhh_a$ of node $a$ with the feature vector $\bfhh_b$ of each of the other nodes as well as itself.
This is done by constructing a ``query'' feature vector $\bfQ_a := \bfU^\mathrm{(query)} \bfhh_a$ for node $a$ to interact with other nodes (including itself), and each of the other nodes, say, node $b$ (could be $a$), constructs a ``key'' feature vector $\bfK_b := \bfU^\mathrm{(key)} \bfhh_b$ to respond to the ``query'', and a ``value'' feature vector $\bfV_b := \bfU^\mathrm{(value)} \bfhh_b$ to convey its contribution.
Here, $\bfU^\mathrm{(query)}$, $\bfU^\mathrm{(key)}$ and $\bfU^\mathrm{(value)}$ are learnable weight matrix parameters in $\bbR^{D'_{\rm hid} \times D_{\rm hid}}$, where $D_{\rm hid}$ is the dimension of each node feature vector $\bfhh_a$ or $\bfhh_b$, and $D'_{\rm hid}$ is another hyperparameter determining the dimension of the key, query and value feature vectors.
The respond of the ``key'' to the ``query'' is modeled by $\frac{\bfQ_a\trs \bfK_b}{\sqrt{D'_{\rm hid}}}$ which is treated as the unnormalized log-probability, or logit, to determine the portion that node $b$ contributes to the new node feature vector of node $a$. The probability, or the portion of contribution, is recovered by applying the softmax function to the logits:
\begin{align}
  \texttt{softmax}(\bm{\ell})_b := \frac{e^{\bm{\ell}_b}}{\sum_{b'} e^{\bm{\ell}_{b'}}}.
  \label{eqn:softmax}
\end{align}
In matrix form, the updated node feature vectors $\bfhh' := [\bfhh'_1, \cdots, \bfhh'_A] \in \bbR^{D'_{\rm hid} \times A}$ as a stacked array over all the nodes is:
\begin{align}
  & \bfhh' := \bfV \, \texttt{softmax}\Big( \frac{\bfQ\trs \bfK}{\sqrt{D'_{\rm hid}}} \Big)\trs \in \bbR^{D'_{\rm hid} \times A}, \\
  \text{where } &
  \bfQ := \bfU^\mathrm{(query)} \bfhh, \,
  \bfK := \bfU^\mathrm{(key)} \bfhh, \,
  \bfV := \bfU^\mathrm{(value)} \bfhh, \,
  \bfhh := [\bfhh_1, \cdots, \bfhh_A] \in \bbR^{D_{\rm hid} \times A}.
\end{align}
To enlarge expressiveness, it is common practice to introduce ``multi-head attention'', where the above self attention calculation is repeated $D_{\rm head}$ times, and the $D_{\rm head}$-fold results are concatenated. More explicitly, the updated node feature vector array is:
\begin{align}
  & \bfhh' := [\bfhh'_1, \cdots, \bfhh'_A] \in \bbR^{(D_{\rm head} D'_{\rm hid}) \times A}, \\
  \text{where }
  & \bfhh'_a := \texttt{concatenate}(\{{\bfhh'}^{(1)}_a, \cdots, {\bfhh'}^{(D_{\rm head})}_a\}) \in \bbR^{D_{\rm head} D'_{\rm hid}}, \forall a = 1 \cdots A,
\end{align}
are reshaped from:
\begin{align}
  & [{\bfhh'}^{(d)}_1, \cdots, {\bfhh'}^{(d)}_A] := \bfV^{(d)} \, \texttt{softmax}\Big( \frac{{\bfQ^{(d)}}\trs \bfK^{(d)}}{\sqrt{D'_{\rm hid}}} \Big)\trs \in \bbR^{D'_{\rm hid} \times A}, \forall d = 1 \cdots D_{\rm head}, \\
  \text{where } &
  \bfQ^{(d)} := \bfU^{(\mathrm{query},d)} \bfhh, \,
  \bfK^{(d)} := \bfU^{(\mathrm{key},d)} \bfhh, \,
  \bfV^{(d)} := \bfU^{(\mathrm{value},d)} \bfhh, \,
  \bfhh := [\bfhh_1, \cdots, \bfhh_A] \in \bbR^{D_{\rm hid} \times A}.
\end{align}
To let the updated feature vector array $\bfhh'$ have the same shape as the original $\bfhh$, hyperparameters $D_{\rm head}$ and $D'_{\rm hid}$ are chosen such that $D_{\rm head} D'_{\rm hid} = D_{\rm hid}$.

The vanilla SelfAttention module handles general featured point (node) cloud input, but for a set of featured atoms, there is a spacial or positional relation between a pair of atoms. Distance is a natural way to describe such relation. For example, a shorter distance generally indicates a stronger interaction between the two node feature vectors. To inform the attention mechanism of this characteristic, Graphormer-3D (G3D)~\citep{shi2022benchmarking} introduce pairwise distance features into the attention mechanism.
To accommodate for the different usage of pairwise distance features from that in the NodeEmbedding module, a learnable MLP layer is applied to the original pairwise distance features, pair by pair, which maps each distance feature vector to dimension $D_{\rm head}$:
\begin{align}
  \tilde{\bm{\clE}} := \texttt{MLP}(\bm{\clE}) \in \bbR^{A \times A \times D_{\rm head}}: \quad
  \tilde{\bm{\clE}}_{ab} := \texttt{MLP}(\bm{\clE}_{ab}) \in \bbR^{D_{\rm head}},
  \label{eqn:mlp-pw-dist}
\end{align}
where the $\texttt{MLP}$ in the latter expression follows the general formulation in \eqnref{mlp}.
This new pairwise distance feature array is incorporated into the vanilla self attention as a bias to the contribution logits.
For an explicit expression, let $\tilde{\bm{\clE}}^{(d)} := [\tilde{\bm{\clE}}^{(d)}_{ab}]_{ab}$ denote the $A \times A$ matrix combining the $d$-th distance feature for all the pairs. The expression for the SelfAttention module is:
\begin{align}
  \begin{split} \label{eqn:selfatt}
    \text{define } & \bfhh' = [\bfhh'_1, \cdots, \bfhh'_A] := \texttt{SelfAttention}(\bfhh, \tilde{\bm{\clE}}) \in \bbR^{D_{\rm hid} \times A}
    \text{ for } \bfhh := [\bfhh_1, \cdots, \bfhh_A] \in \bbR^{D_{\rm hid} \times A}: \\
    & \bfhh'_a := \texttt{concatenate}(\{{\bfhh'}^{(1)}_a, \cdots, {\bfhh'}^{(D_{\rm head})}_a\}) \in \bbR^{D_{\rm head} D'_{\rm hid} = D_{\rm hid}}, \forall a = 1 \cdots A, \\
    \text{where } & [{\bfhh'}^{(d)}_1, \cdots, {\bfhh'}^{(d)}_A] := \bfV^{(d)} \, \texttt{softmax}\Big( \frac{{\bfQ^{(d)}}\trs \bfK^{(d)}}{\sqrt{D'_{\rm hid}}} + \tilde{\bm{\clE}}^{(d)} \Big)\trs \in \bbR^{D'_{\rm hid} \times A}, \forall d = 1 \cdots D_{\rm head}, \\
    &
    \bfQ^{(d)} := \bfU^{(\mathrm{query},d)} \bfhh, \,
    \bfK^{(d)} := \bfU^{(\mathrm{key},d)} \bfhh, \,
    \bfV^{(d)} := \bfU^{(\mathrm{value},d)} \bfhh.
  \end{split}
\end{align}

\paragraph{Assembly}
The third component in the G3D module is an MLP module, which follows the same form as given in \eqnref{mlp}, and is applied to the node feature vector of each node independently (\ie, the nodes share the same MLP module to process their feature vectors).
These modules are combined to make the G3D module following the illustration in \appxfigref{model-arch}(d). Explicitly in equation,
\begin{align}
  \begin{split} \label{eqn:g3d}
    \text{define } & \bfhh' := \texttt{G3D}(\bfhh, \tilde{\bm{\clE}}): \\
    & \bfhh' := \texttt{MLP}(\texttt{LayerNorm}(\bfhh'')) + \bfhh'', \\
    & \bfhh'' := \texttt{SelfAttention}(\texttt{LayerNorm}(\bfhh), \tilde{\bm{\clE}}) + \bfhh.
  \end{split}
\end{align}

\subsubsection{Output Process} \label{appx:model-output}
To produce a scalar output, each node feature vector $\bfhh_a$ is processed to produce a scalar by an MLP module following the form of \eqnref{mlp}, whose output dimension (\ie, number of rows of $\bfU^{(2)}$) is 1. The scalars from all the nodes are summed up, and the summed value is added with the output of the AtomRef module (Methods~\ref{sec:learn-grad}; \appxref{atom-ref}) to produce the final energy output:
\begin{align}
  T_{\tnS} := \sum_{a=1}^A \texttt{MLP}(\bfhh_a) + T_\mathrm{AtomRef}(\bfpp', \clM).
  \label{eqn:model-output}
\end{align}

\subsubsection{Model Configuration} \label{appx:model_config}
In all experimental settings, we employ the same backbone architecture, Graphormer, specifically utilizing the Graphormer-3D (\emph{G3D}) encoder module. To ensure a fair comparison, most hyperparameters in all models are maintained consistently (\eg, model depth and hidden dimension). A summary of all hyperparameter choices can be found in \appxtblref{model-hyper-param}. It is worth mentioning that in the shrink gate of \emph{NodeEmbedding}, the initial value of learnable parameters $\nu_{\rm co}$ is set to 10, while the initial value of $\nu_{\rm mul}$ is set to 0.02 for the ethanol dataset and 0.05 for all other datasets. We did not conduct an extensive hyperparameter search, and most hyperparameters were chosen following Graphormer~\citep{ying2021transformers}. All models are implemented using the PyTorch deep learning framework~\citep{paszke2019pytorch}. 

\begin{table}[h]
  \centering
  \caption{\textbf{Hyperparameters of the Graphormer model.}  These hyperparameters are adopted in all methods in the present work (\ourmethod, \mlff, \mlffden).}
  \begin{tabular}{cc}
    \toprule
    Hyperparameter & \ourmethod \\
    \midrule
    G3D Modules & {12} \\
    Hidden Dimension & {768} \\
    MLP Hidden Dimension & {768} \\
    Number of Heads & {32} \\
    GBF Dimension & {128} \\
    Dropout  & {0.1} \\
    Attention Dropout  & {0.1} \\
    Optimizer & {Adam} \\
    Learning Rate Schedule  & {Linear decay} \\
    Adam ($\beta_1,\, \beta_2$) & {(0.95, 0.99)} \\
    Adam $\epsilon$ & $1\e{-8}$ \\
    \bottomrule
  \end{tabular}
  \label{tbl:model-hyper-param}
\end{table}

\subsubsection{Training Hyperparameters}

Following Graphormer~\citep{ying2021transformers}, all models are trained using the Adam optimizer and a linear decay learning schedule. The peak learning rate is set to $1\e{-4}$ for our functional models and $3\e{-4}$ for baseline models (\ie, \mlff and \mlffden). Other optimizer hyperparameters can be found in \appxtblref{model-hyper-param}. A warmup stage featuring a linearly increasing learning rate is introduced to stabilize training during the initial stage. The number of warmup steps is set to 30k for \ourmethod and 60k for baseline models. The batch size is set to 256 for the ethanol dataset and 128 for all other datasets. All models are trained on Nvidia Tesla V100 GPUs.

For different molecule datasets, the number of training epochs is determined by examining the loss curve. Training is halted once the validation loss fails to decrease for 20 epochs. Specifically, our functional models are trained for approximately 600 and 700 epochs on the ethanol and QM9 datasets, respectively. For the QMugs dataset, our model is trained for approximately 700 epochs, while the \mlff and \mlffden models are trained for 2,300 epochs. Notably, we propose a series of QMugs datasets with increasing molecular size in Results~\ref{sec:res-larger-scale}, where the number of SCF datapoints will slightly increase as the average molecular size increases (larger molecules generally require more SCF iterations to converge). To ensure a fair comparison, we maintain approximately the same number of training epochs for different datasets. In the chignolin experiment, there are four chignolin datasets with increasing peptide length and data size. Our functional models are trained for 1,400, 1,100, 800, and 750 epochs, respectively, while the \mlff and \mlffden models are trained for 3,000, 3,000, 1,000, and 900 epochs, respectively.

In practice, we employ a weighted loss function to optimize the KEDF model:
\begin{align}
L = \xi_{\rm eng} L_{\rm eng} + \xi_{\rm grad} L_{\rm grad} + \xi_{\rm den} L_{\rm den},
\end{align}
where $L_{\rm eng}$, $L_{\rm grad}$, and $L_{\rm den}$ represent the KEDF energy loss (\eqnref{loss-eng}), gradient loss (\eqnref{loss-grad}), and projected density loss (\eqnref{loss-den}), respectively. $\xi_{\rm eng}$, $\xi_{\rm grad}$, and $\xi_{\rm den}$ are the corresponding loss weights. The selection of loss weights is determined through grid search on the validation set for various datasets independently, and the utilized loss weights are provided in \appxtblref{loss-weights}.

\begin{table}[h]
  \centering
  \caption{\textbf{Loss weights for various datasets and learning targets.} The loss weights are tuned on each setting by conducting a grid search on the validation set.}
  \begin{tabular}{lccc}
    \toprule
    Dataset & $\xi_{\rm eng}$ & $\xi_{\rm grad}$ & $\xi_{\rm den}$ \\
    \midrule
    Ethanol-$T_\res$ & 1 & 0.1 & 0.08 \\
    Ethanol-$E_\TXC$ & 1 & 0.03 & 1 \\
    QM9-$T_\res$ & 1 & 0.12 & 0.005 \\
    QM9-$E_\TXC$ & 1 & 0.12 & 0.05 \\
    QMugs & 1 & 0.1 & 1\\
    Chignolin & 1 & 0.1 & 1 \\
    \bottomrule
  \end{tabular}
  \label{tbl:loss-weights}
\end{table}

\subsection{Local Frame Module for Geometric Invariance} \label{appx:local-frame}

\begin{figure}[t]
  \centering
  \includegraphics[width=0.7\textwidth]{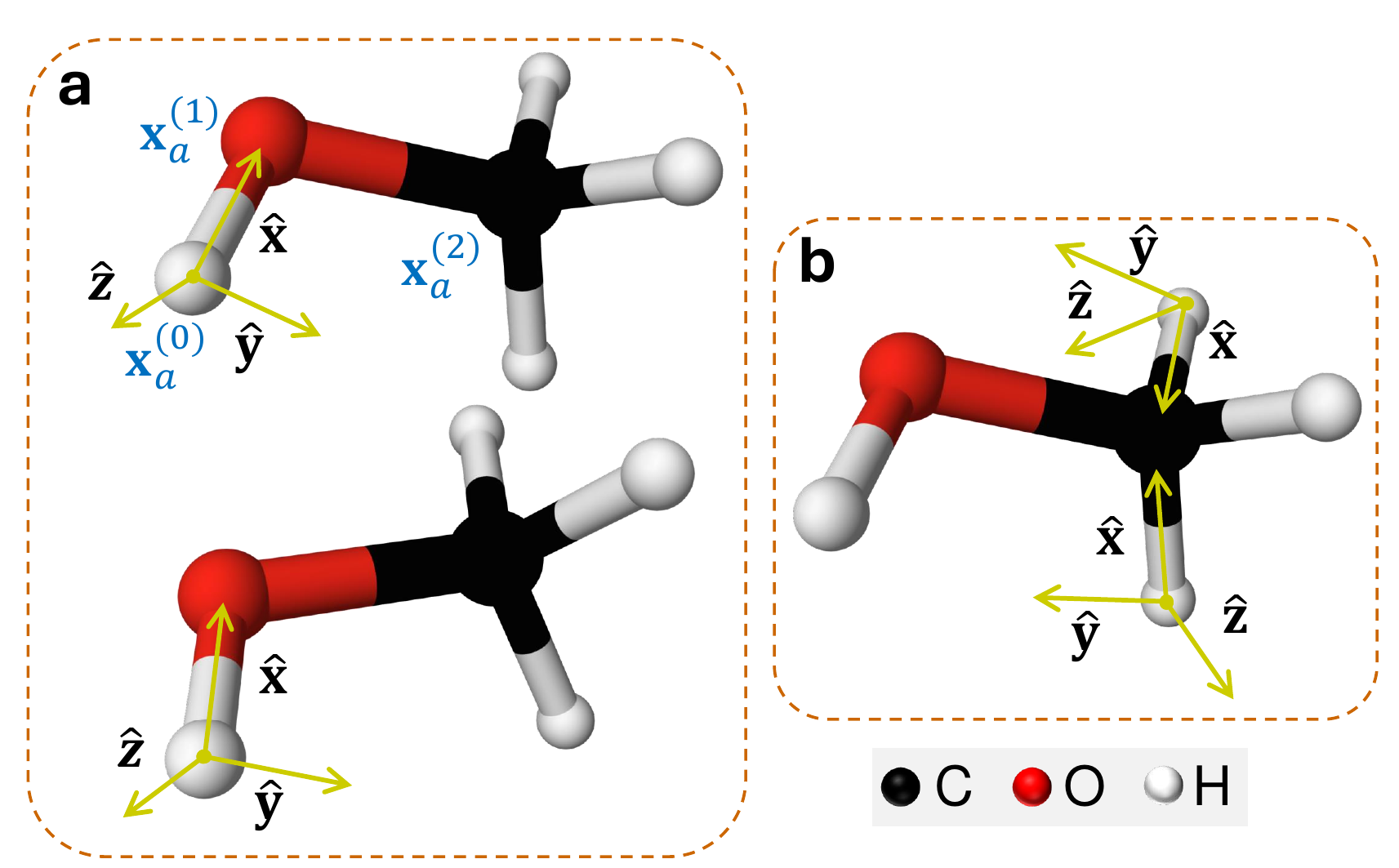}
  \caption{\textbf{Illustration of the local frame}. \textbf{(a)}~The local frame is constructed for each atom according to its local environment, which is equivariant with respect to. the transformation of the global frame. \textbf{(b)}~Local frames of similar substructures (\eg, the three $\rmH$-$\rmC$ bonds in the methyl group) rotate with substructures accordingly, resulting in invariant coefficient features for locally similar density patterns.
  }
  \label{fig:local-frame}
\end{figure}

As mentioned in Methods~\ref{sec:local-frame}, by expanding the electron density on a set of atomic basis, the expansion coefficients $\bfpp$ mathematically comprise geometric tensors of various orders equivariant to rotations and translations. A local frame simultaneously rotating with the structure is adopted to decouple the density feature from the change of the coordinate system and unnecessary geometric variability.
As shown in \appxfigref{local-frame}(a), to construct the local frame at an atom at $\bfxx_a^{(0)}$, we choose $\hat{\bfxx}$ pointing to its nearest atom $\bfxx_a^{(1)}$.
The $\hat{\bfzz}$ axis lies in the line of the cross-product of $\hat{\bfxx}$ with the direction to the second-nearest not-on-$\hat{\bfxx}$ atom $\bfxx_a^{(2)}$ and
the $\hat{\bfyy}$ axis is then given by $\hat{\bfyy} = \hat{\bfzz} \times \hat{\bfxx}$.
Note that we exclude the hydrogen atoms
in the neighborhood of the center atom following \cite{han2018deep}, making the obtained local frame depend more on heavy atoms, whose positions are more stable than those of hydrogen atoms and reflect more reliable local structures. 
Denote the local frame associated with atom $a$ as:
\begin{align}
    \mathcal{R}_a := (\hat{\bfxx}, \hat{\bfyy}, \hat{\bfzz}),
\end{align}
the density coefficient vector and the gradient vector under the local frame are calculated by:
\begin{align}
  \bfpp'^l_a = \bfD^l(\mathcal{R}_a) \bfpp^l_a,
  \quad
  \nabla_{\bfpp'^l_a} T_\tnS = \bfD^l(\mathcal{R}_a) \nabla_{\bfpp^l_a} T_\tnS.
  \label{eqn:locframe}
\end{align}
where $l$ is the degree of tensors or azimuthal quantum number. $\bfpp^l$ corresponds to the coefficient of basis functions of the degree $l$ (or type-$l$ tensors in mathematics). $\bfD^l(\mathcal{R}_a)$ is the Wigner-D matrix of degree $l$. 

\begin{figure}[h]
  \centering
  \begin{subfigure}[b]{0.42\textwidth}
      \centering
      \includegraphics[width=\textwidth]{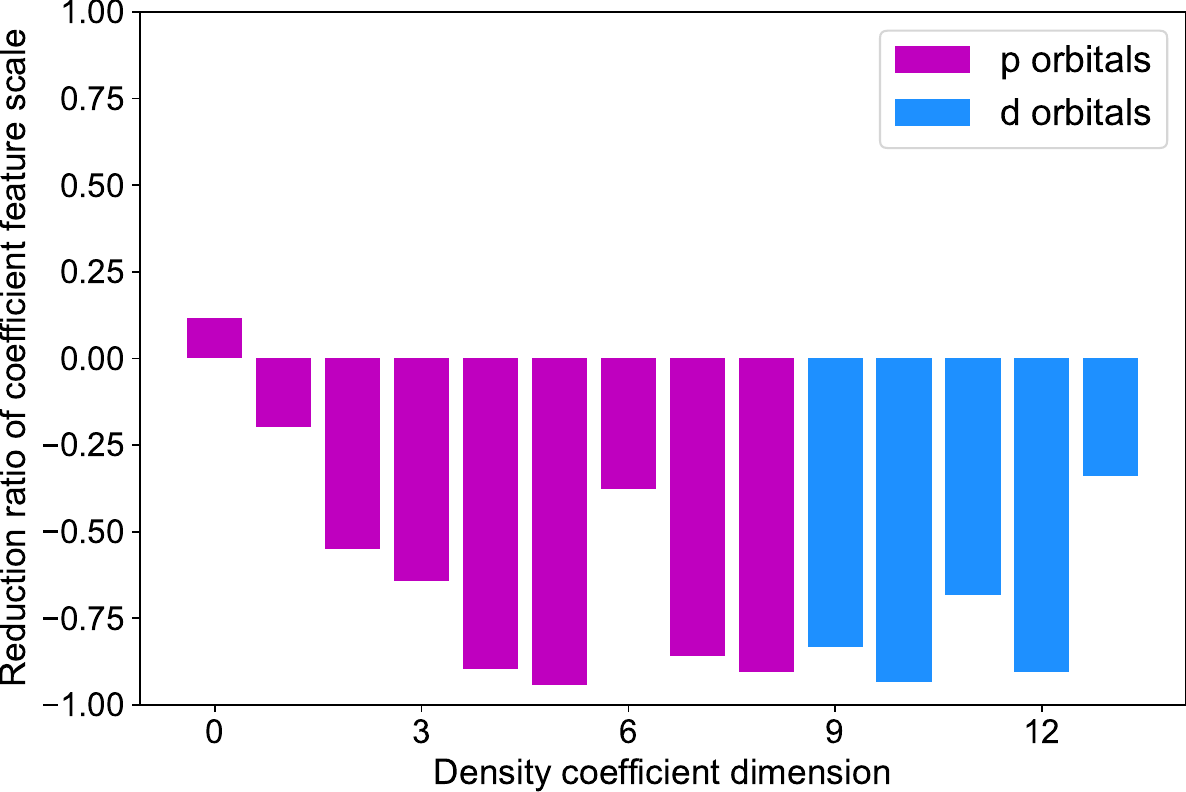}
      \caption{H atom}
      \label{fig:coeff-lofra-H}
  \end{subfigure}
  \vfill
  \begin{subfigure}[b]{0.80\textwidth}
      \centering
      \includegraphics[width=\textwidth]{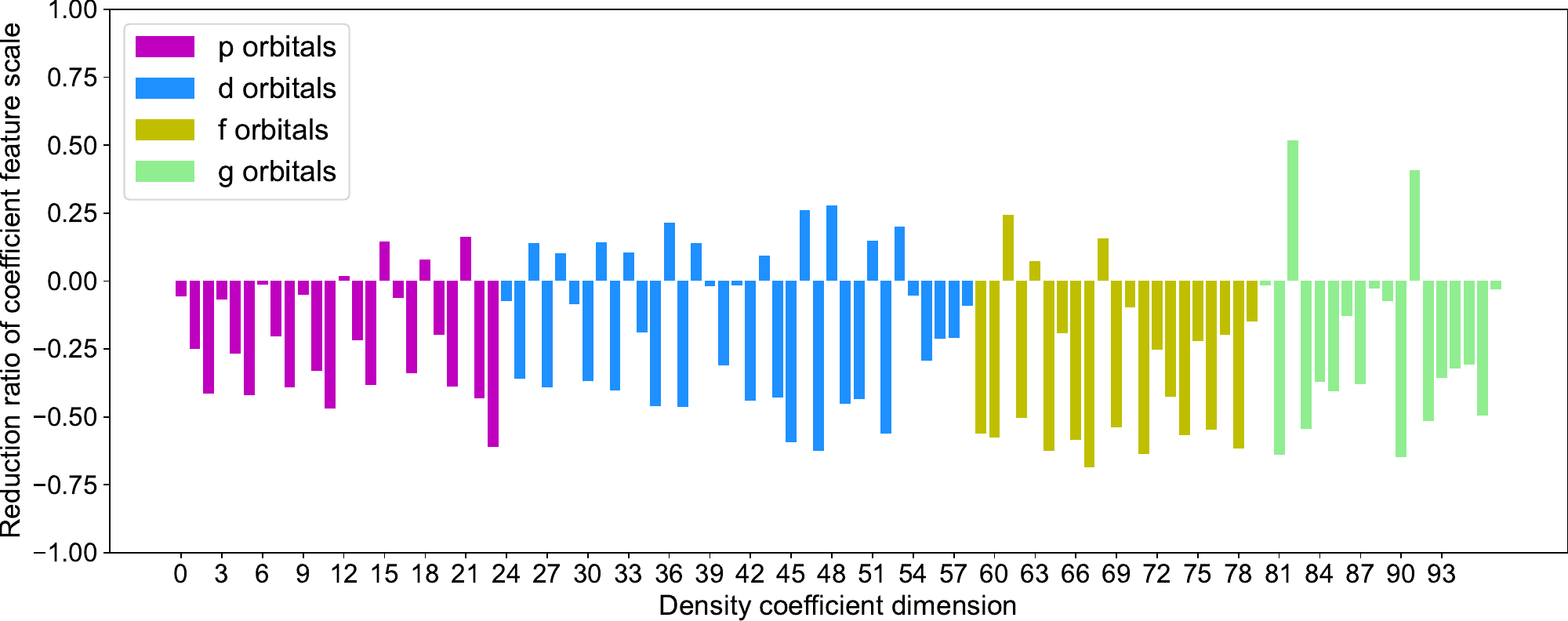}
      \caption{C atom}
      \label{fig:coeff-lofra-C}
  \end{subfigure}
  \vfill
  \begin{subfigure}[b]{0.80\textwidth}
      \centering
      \includegraphics[width=\textwidth]{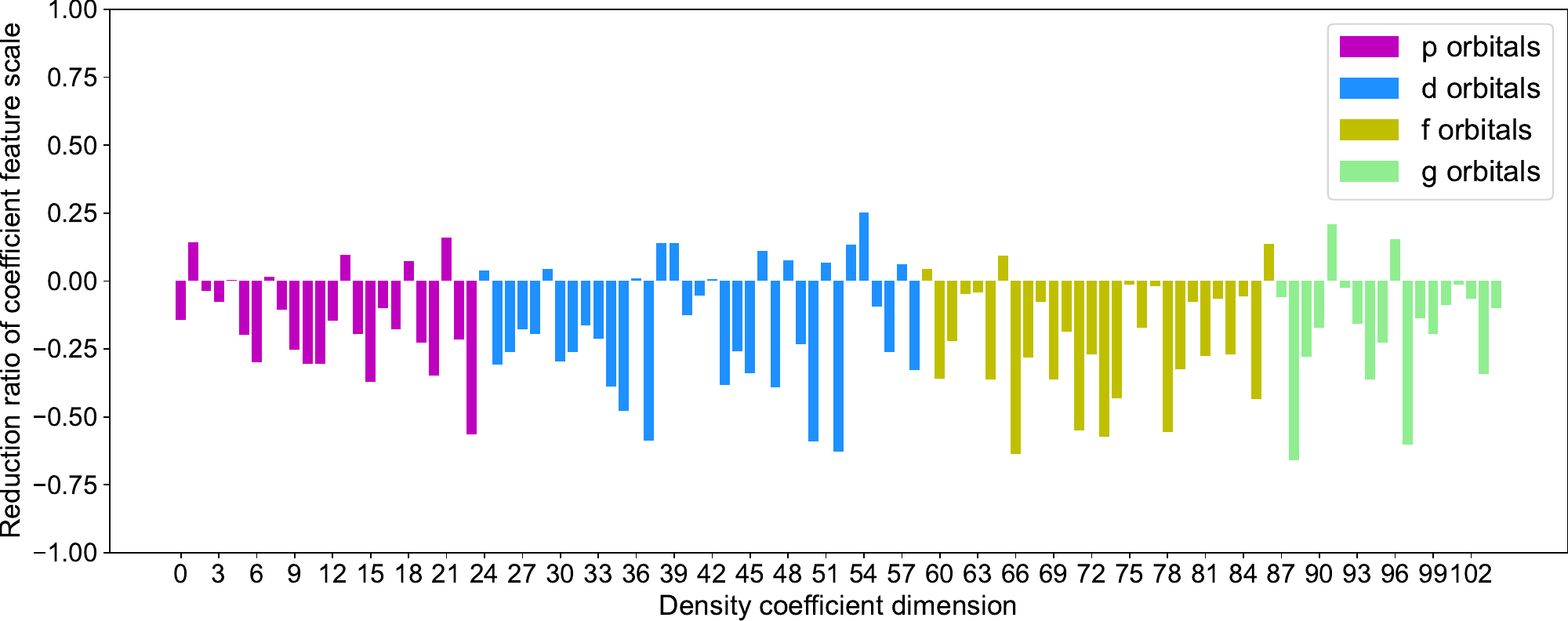}
      \caption{N atom}
      \label{fig:coeff-lofra-N}
  \end{subfigure}
  \vfill
  \begin{subfigure}[b]{0.80\textwidth}
      \centering
      \includegraphics[width=\textwidth]{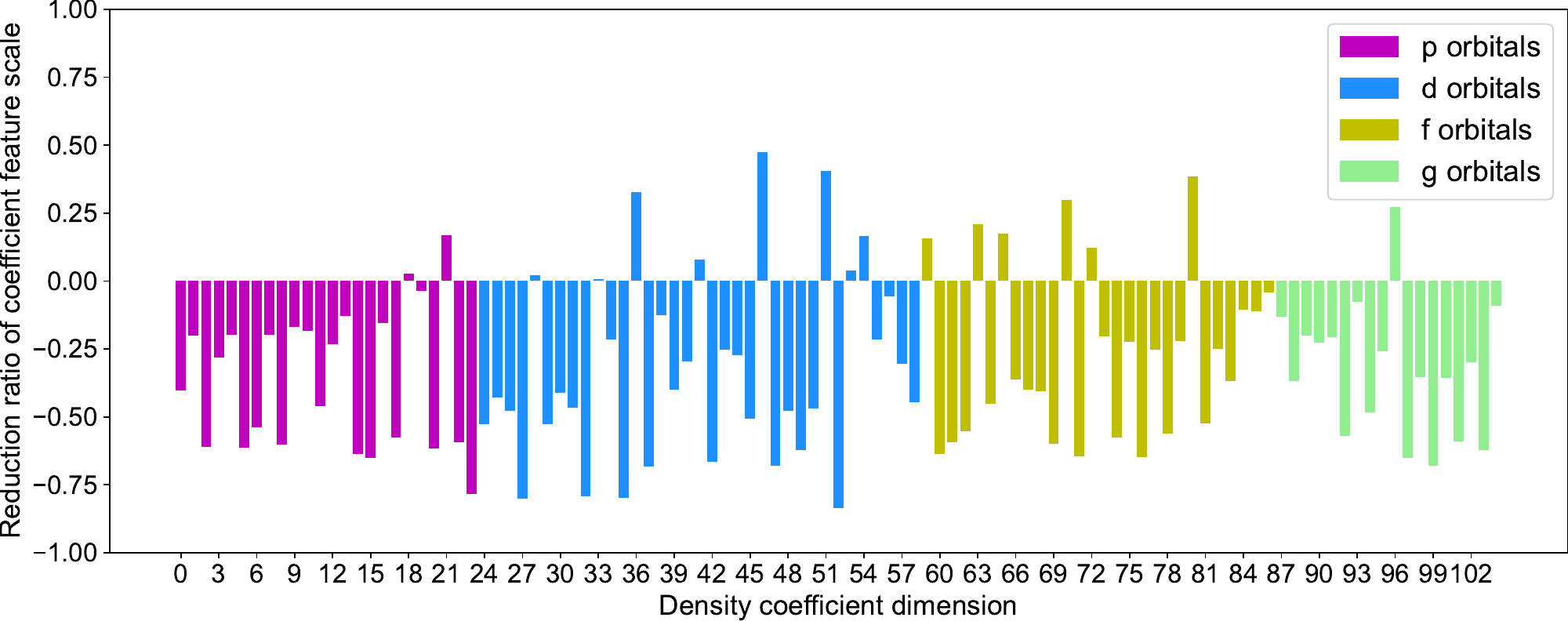}
      \caption{O atom}
      \label{fig:coeff-lofra-O}
  \end{subfigure}
  \caption{\textbf{Coefficient scale changes by the use of the local frame.} For each coefficient dimension $\tau$ of atom type $Z$, the height of the bar represents the relative change in scale, calculated as $\frac{{\rm std\_coeff'}_{Z,\tau} - {\rm std\_coeff}_{Z,\tau}}{{\rm std\_coeff}_{Z,\tau}}$,
  where ${\rm std\_coeff}_{Z,\tau} := \std\{\bfpp_{a,\tau}^{(d,k)}\}_{a: Z^{(a)} = Z, \; k, \, d}$
  and 
  $
  {\rm std\_coeff'}_{Z,\tau} 
  := \std\{{\bfpp'}_{a,\tau}^{(d,k)}\}_{a: Z^{(a)} = Z, \; k, \, d}
  $ measure the coefficient scale before and after being transformed by the local frame transformation, respectively. 
  A negative value indicates a reduction in the scale of the coefficient due to the local frame transformation. Importantly, the local frame substantially decreases the coefficient scale across diverse basis dimensions and atom types.}
  \label{fig:coeff-local-frame}
\end{figure}

\begin{figure}[h]
  \centering
  \begin{subfigure}[b]{0.42\textwidth}
      \centering
      \includegraphics[width=\textwidth]{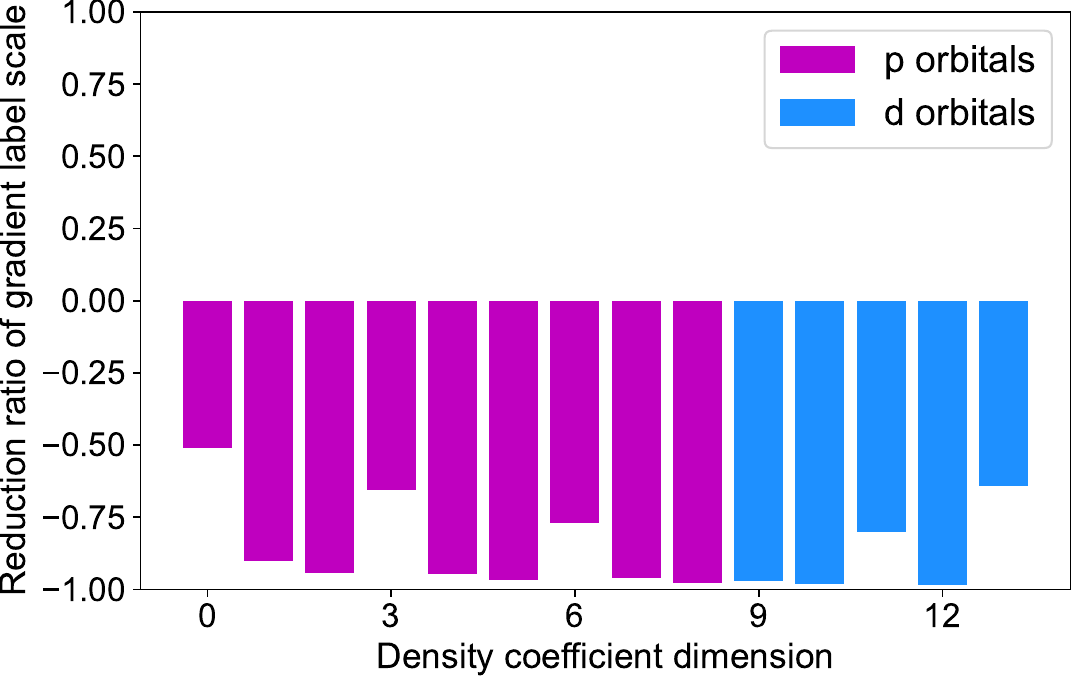}
      \caption{H atom}
      \label{fig:grad-lofra-H}
  \end{subfigure}
  \vfill
  \begin{subfigure}[b]{0.84\textwidth}
      \centering
      \includegraphics[width=\textwidth]{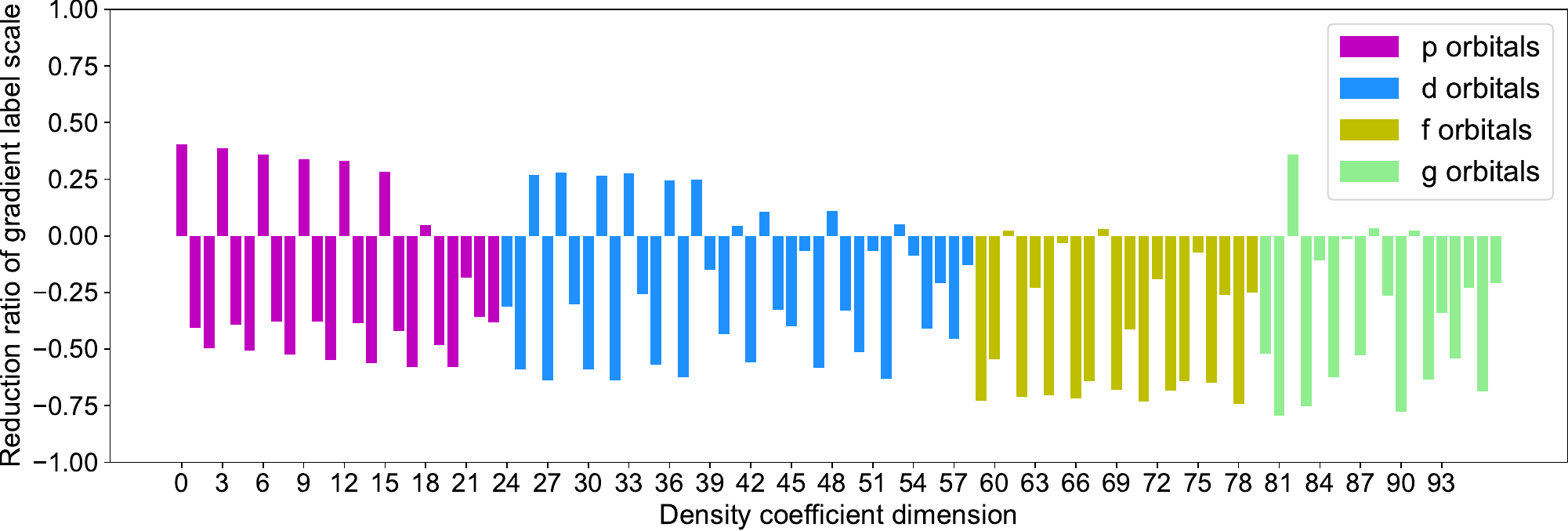}
      \caption{C atom}
      \label{fig:grad-lofra-C}
  \end{subfigure}
  \vfill
  \begin{subfigure}[b]{0.84\textwidth}
      \centering
      \includegraphics[width=\textwidth]{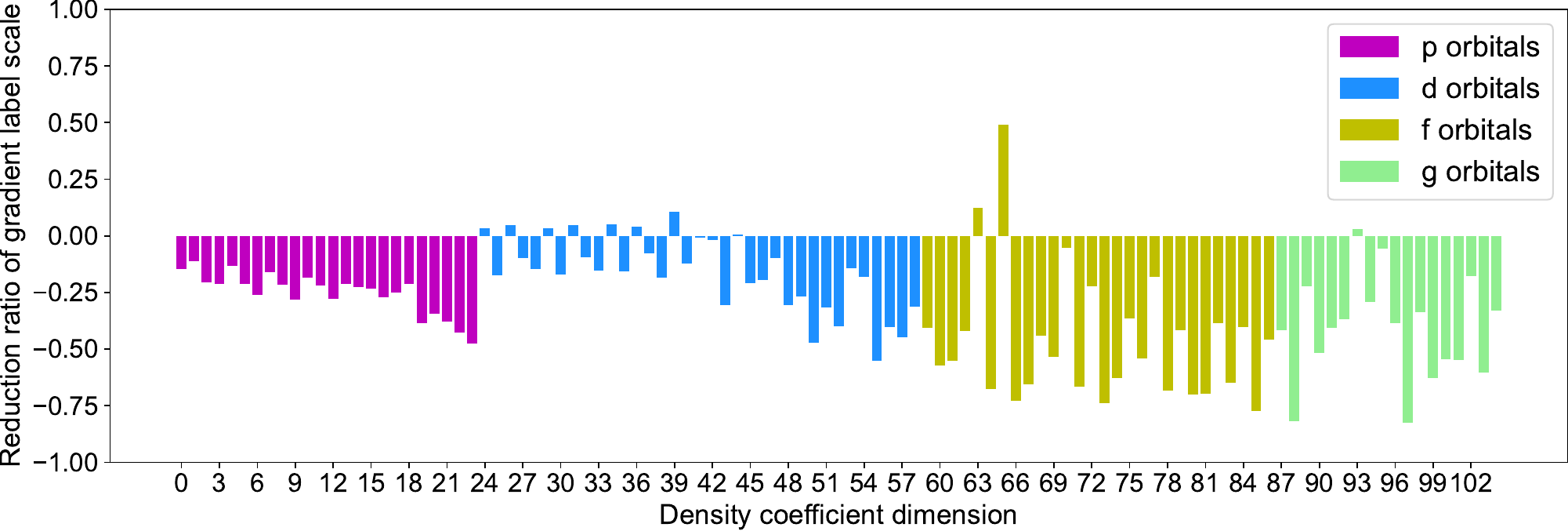}
      \caption{N atom}
      \label{fig:grad-lofra-N}
  \end{subfigure}
  \vfill
  \begin{subfigure}[b]{0.84\textwidth}
      \centering
      \includegraphics[width=\textwidth]{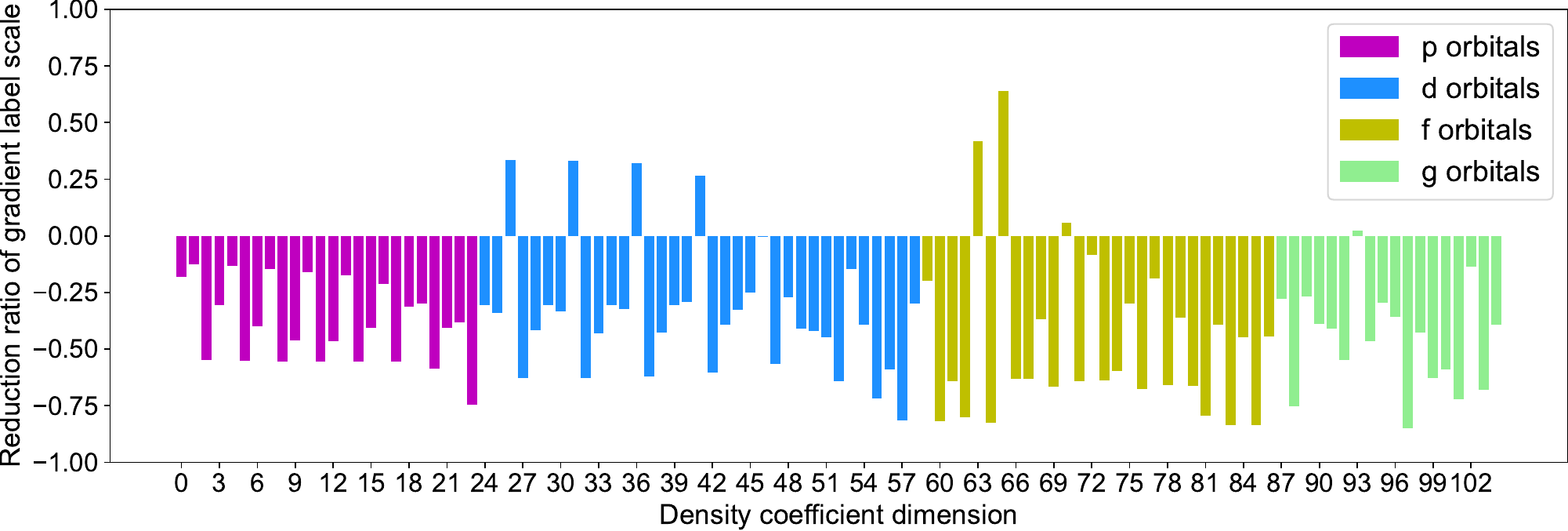}
      \caption{O atom}
      \label{fig:grad-lofra-O}
  \end{subfigure}
  \caption{\textbf{Gradient scale changes by the use of the local frame.} For each coefficient dimension $\tau$ for atom type $Z$, the height of the bar represents the relative change in the gradient scale, calculated by 
  $\frac{\mathrm{max\_abs\_grad'}_{Z,\tau} - \mathrm{max\_abs\_grad}_{Z,\tau}}{\mathrm{max\_abs\_grad}_{Z,\tau}}$,
  where 
  $\mathrm{max\_abs\_grad}_{Z,\tau} := \max \big\{\big\lvert \nabla_{\bfpp_{a,\tau}} T_\tnS^{(d,k)} \big\rvert\big\}_{a: Z^{(a)} = Z, \; k, \, d}$ 
  and 
  $\mathrm{max\_abs\_grad'}_{Z,\tau} := \max \big\{\big\lvert \nabla_{\bfpp_{a,\tau}}{T'}_\tnS^{(d,k)} \big\rvert\big\}_{a: Z^{(a)} = Z, \; k, \, d}$ 
  measure the gradient scale before and after being transformed by the local frame transformation, respectively. The local frame transformation notably reduces the gradient scale across a wide range of basis dimensions and atom types.
  }
  \label{fig:grad-local-frame}
\end{figure}

Notably, an additional benefit of the local frame is that it makes a stable feature for locally similar density patterns. For example, consider the coefficients corresponding to the atomic basis on the three hydrogen atoms in a methyl group (\appxfigref{local-frame}(b)). As the three $\rmH$-$\rmC$ bonds are highly indistinguishable, the densities on the bonds are very close 
and contribute almost identically to the kinetic energy. But as the three bonds have different orientations relative to a common global coordinate system, the coefficients on the hydrogen atoms are vastly different if the basis functions on different hydrogen atoms are aligned to the same directions.
In contrast, basis functions under local frames are oriented equivariantly with the orientation of the local substructures, leaving the coefficients largely invariant, which makes more physical meanings to predict the energy.
For example, coefficients on basis functions aligning with the $\hat{\bfxx}$ direction always represent the density on the shortest bond with the atom.
As a result, the local frame makes almost the same density coefficients on the three hydrogen atoms in the methyl group \emph{even} though the $\rmH$-$\rmC$ bonds have different orientations. %
We also provide a numerical visualization of the ethanol dataset, where the target functional is chosen as the residual energy of the non-interacting kinetic energy after deducted by the value given by the base kinetic energy functional APBE~\citep{constantin2011semiclassical}.
As shown in \appxfigref{coeff-local-frame}, the local frame always attains a noticeable scale (standard deviation) reduction across various atom types, demonstrating its capability to eliminate unnecessary geometric variability caused by various orientations of the chemical bonds. More importantly, as shown in \appxfigref{grad-local-frame}, this approach also brings a substantial scale reduction for gradient labels across various atom types, especially for $\rmH$ atoms, where almost all coefficient dimensions exhibit a $>$ 60\% gradient scale reduction.
This is crucial to alleviate the large gradient range and make the subsequent dimension-wise rescaling module easier (See more discussions in \appxref{dim-rescale}).
These two advantages of the local frame are beneficial for the efficient optimization of the KEDF model. The empirical results in \appxref{ablat-den-prep} suggest that the KEDF model achieves a 2-fold lower training and test error by using this technique.

\subsection{Enhancement Modules for Expressing Vast Gradient Range}
\label{appx:tech-effi-train}

As mentioned in Methods~\ref{sec:learn-grad}, the raw gradient range of the input is vast, and conventional data normalization techniques are not applicable in our task since minimizing the gradient scale is conflicting with minimizing the coefficient scale. For the same reason we neither can take the logarithm of the density feature, as a number of works have adopted~\citep{changyong2014log}, since the gradient would be correspondingly scaled up. %
Another way to decrease the gradient scale is to downscale the energy value (\ie, using a larger energy unit). However, this does not bring further improvement in our trials.
As mentioned in the main context, this is due to the trade-off between easy learning and model resolution: an error in the normalized (small) scale will be enlarged in the original scale.
It neither works to use a separate model to learn the gradient directly, as we have empirically observed that the energy value model easily overfits the training data, and during density optimization the gradient model unnecessarily decreases the energy constructed from the value model and even appears non-conservative as the optimization diverges.

Having both large coefficients and gradients implies a function with a large Lipschitz coefficient, which leads to a great challenge for the optimization of conventional neural networks.
According to these observations, we turn to elaborating on a series of enhancement modules to express the vast gradient range and enable effective training.

\subsubsection{Dimension-wise Rescaling} \label{appx:dim-rescale}

Dimension-wise rescaling of coefficients and gradients is applied after the process of local frame and natural reparameterization.
The two modules made the rescaling easier, but there is still a scale trade-off between coefficients and gradients.
We hence choose moderate values for the scale parameters in \eqnref{dim-rescale-def}:
the target gradient scale $s_\grad$ is set to $0.05$ and the maximal coefficient scale $s_\coeff$ is set to $50$. 
Compared to the gradient scale, handling input coefficients with a larger range is more manageable. For example, incorporating a ShrinkGate module to further compress the coefficients into a bounded space, as discussed in \appxref{model-node-embed}. We found this technique yields an admirable performance empirically. Consequently, we prefer allowing a larger coefficient scale in the trade-off.
In the context of deep learning, the Lipschitz coefficient (\ie, the maximum absolute gradient that the neural network model could express) is usually used to indicate the capability of a model fitting gradient label. Following this convention, we take the maximum absolute gradient value to measure the label scale.

\begin{figure}[h]
  \centering
  \begin{subfigure}[b]{0.8\textwidth}
      \centering
      \includegraphics[width=\textwidth]{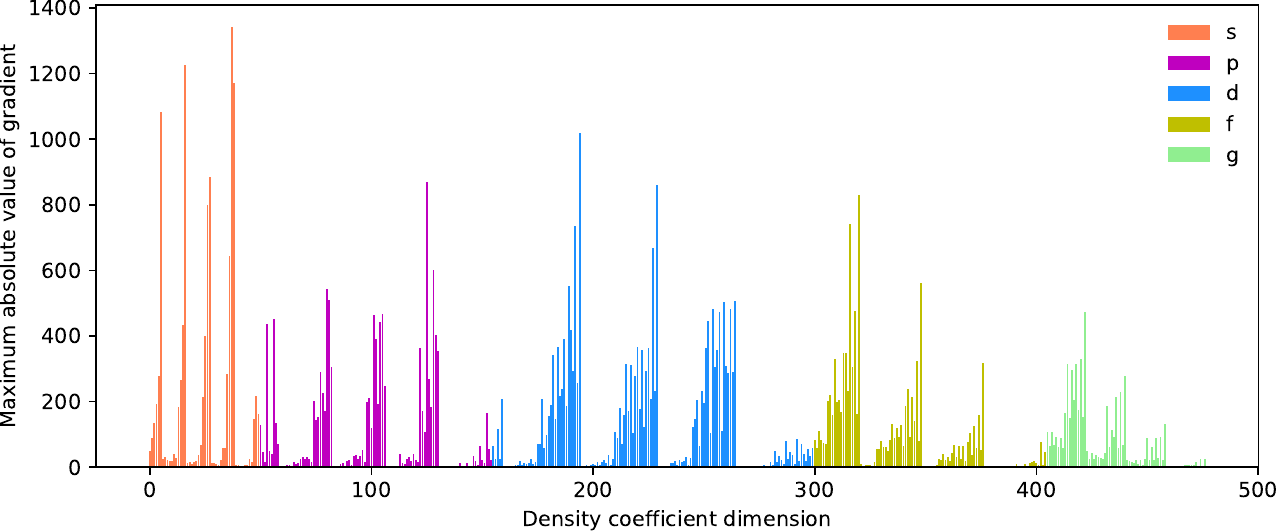}
      \caption{Gradient scales after processed by local frame, natural reparameterization, and atomic reference model (before dimension-wise rescaling).}
      \label{fig:raw-grad-scale}
  \end{subfigure}
  \vfill
  \begin{subfigure}[b]{0.8\textwidth}
      \centering
      \includegraphics[width=\textwidth]{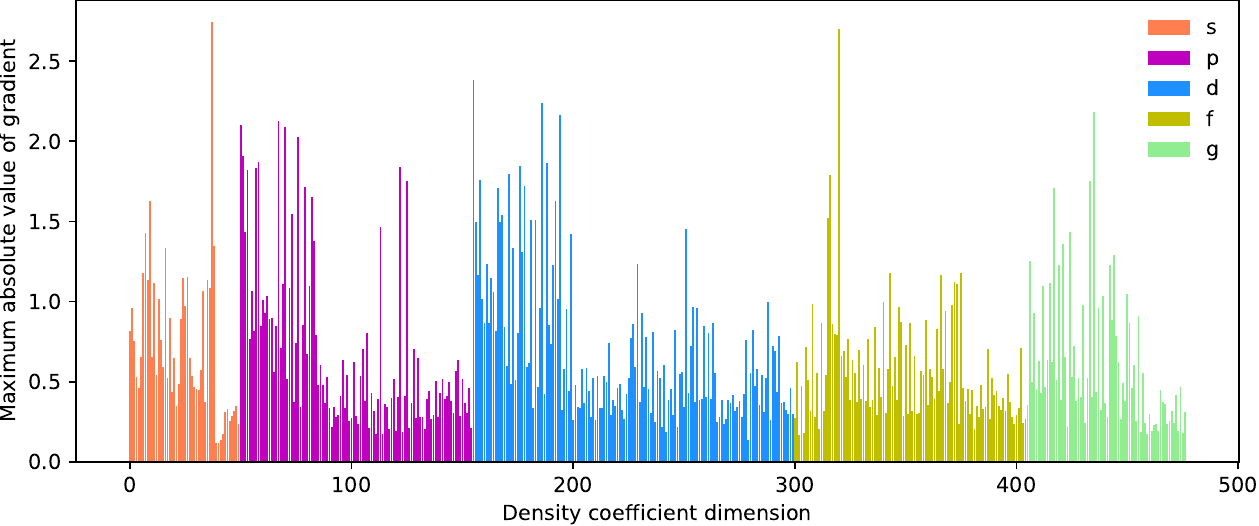}
      \caption{Gradient scales further after dimension-wise rescaling.}
      \label{fig:grad-scale}
  \end{subfigure}
  \vfill
  \begin{subfigure}[b]{0.8\textwidth}
      \centering
      \includegraphics[width=\textwidth]{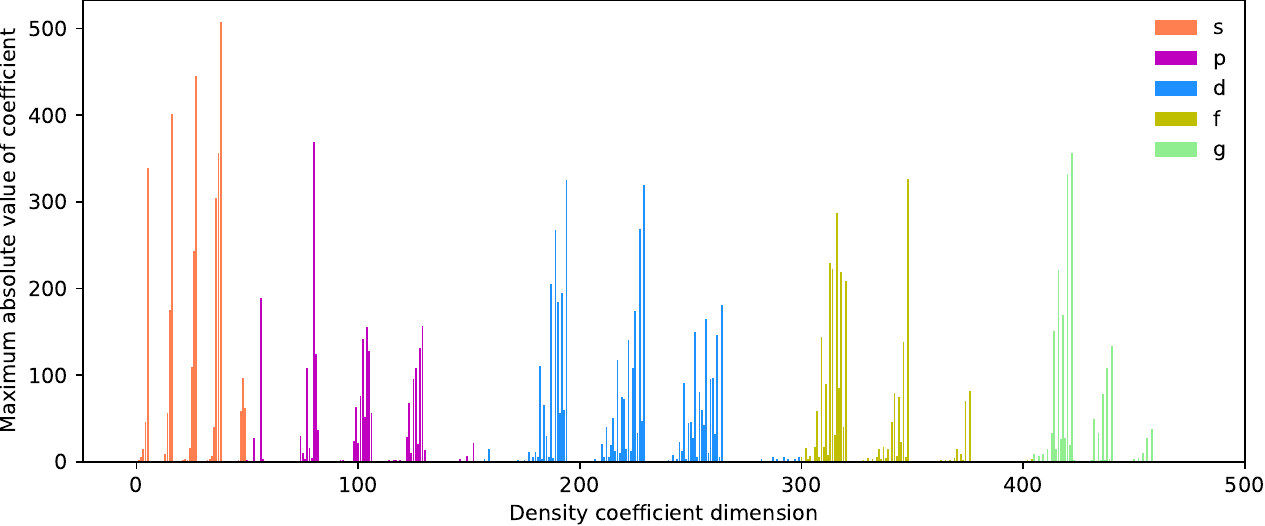}
      \caption{Density coefficient scales further after dimension-wise rescaling.}
      \label{fig:coeff-scale}
  \end{subfigure}
  \caption{\textbf{Gradient and density coefficient scales over dimensions on the QM9 dataset in the setting of learning residual KEDF with APBE base KEDF.} The maximum absolute value is used to measure the maximum scale of data. \textbf{(a)-(b)}~present the gradient scale before and after the dimension-wise rescaling transformation. This technique properly balances the scales of gradient and density coefficient in each dimension, hence reduces the gradient scale for easier learning, while maintaining a reasonable density coefficient scale. \textbf{(c)}~shows the corresponding coefficient scale after the dimension-wise rescaling transformation.}
  \label{fig:dim-rescale}
\end{figure}

To illustrate the importance of dimension-wise rescaling, 
we visualize in \appxfigref{dim-rescale} the scales of gradients and density coefficients over the dimensions before and after the processing of dimension-wise rescaling in the setting of learning residual KEDF with APBE base KEDF on the QM9 dataset.
The shown gradient scales are estimated after centralizing the gradient values by subtracting the gradient mean of each dimension with the atomic reference module. %
As plotted in \appxfigref{dim-rescale}(a), many gradient dimensions have an extremely large scale. We found this leads to great difficulty in the practical optimization of deep learning models. After dimension-wise rescaling, most dimensions are rescaled to have the desired gradient scale %
(\appxfigref{dim-rescale}(b)).
As illustrated in \appxfigref{dim-rescale}(c), the density coefficients are rescaled to a larger scale, and a ShrinkGate module is introduced to normalize the coefficients into a friendly space (\appxref{model-node-embed}).
As a result, we found that the neural network model without the dimension-wise rescaling technique hardly converges and the loss curve is particularly volatile, due to the smoothness of common neural network architectures which restricts the range of the output gradient of the model. Applying the dimension-wise rescaling technique effectively mitigates this dilemma and enables efficient optimization.

\subsubsection{Natural Reparameterization} \label{appx:nat-reparam}

Although dimension-wise rescaling allows the trade-off between numerical scales of coefficients and gradients, the trade-off often has a difficult frontier that the two sides cannot be simultaneously made in a mild scale.
A way to improve the trade-off is to balance the difficulties over the dimensions.
The dimensions exhibit different scales since they have different sensitivity to the density hence the energy.
Different basis functions spread over different regions in space, so a same amount of coefficient change on different basis functions influences the density function differently. %
To understand and balance the sensitivities, we derive how the change of coefficients affect the density function.
Consider a small change $\Delta \bfpp$ to the coefficients $\bfpp$, which leads to a change in the density function $\Delta \rho(\bfrr) := \rho_{\bfpp + \Delta \bfpp}(\bfrr) - \rho_\bfpp(\bfrr)$.
This difference is typically measured by the L2-metric in the function space:
$\int \lrvert{\rho_{\bfpp + \Delta \bfpp}(\bfrr) - \rho_\bfpp(\bfrr)}^2 \dd \bfrr
= \int \lrvert{\Delta \rho(\bfrr)}^2 \dd \bfrr
= \int \lrvert{\rho_{\Delta \bfpp}(\bfrr)}^2 \dd \bfrr
= \int \sum_\mu \Delta \bfpp_\mu \omega_\mu(\bfrr) \sum_\nu \Delta \bfpp_\nu \omega_\nu(\bfrr) \dd \bfrr
= \Delta \bfpp\trs \bfW \Delta \bfpp$,
where $\bfW_{\mu\nu} := \int \omega_\mu(\bfrr) \omega_\nu(\bfrr) \dd \bfrr$ is the overlap matrix of the atomic basis for density.
As the atomic basis is not orthonormal, $\bfW$ is not isotropic (\ie, cannot be turned proportional to the identity matrix by orthogonal transformations), so the same amount of coefficient change in different dimensions has different effects on the density function.

The proposed natural parameterization $\bfppt := \bfM\trs \bfpp$, where $\bfM$ is a square matrix satisfying $\bfM \bfM\trs = \bfW$, fulfills the desired balance.
To see this, noting that $\bfW$ is non-singular hence is $\bfM$, we have $\bfpp = \bfM^{-\top} \bfppt$, so the density change is $\int \lrvert{\Delta \rho(\bfrr)}^2 \dd \bfrr
= \Delta \bfpp\trs \bfW \Delta \bfpp = \Delta \bfppt\trs \bfM^{-1} \bfW \bfM^{-\top} \Delta \bfppt
= \Delta \bfppt\trs \bfM^{-1} \bfM \bfM\trs \bfM^{-\top} \Delta \bfppt
= \Delta \bfppt\trs \Delta \bfppt$,
hence the change of coefficient in any dimension contributes equally to the density change.
The gradient is reparameterized accordingly, following $\nabla_{\bfppt} T_\tnS = (\nabla_{\bfppt} \bfpp)\trs \nabla_\bfpp T_\tnS = \bfM^{-1} \nabla_\bfpp T_\tnS$.

Choosing the matrix $\bfM$ still faces a degree of freedom of an orthogonal transformation.
We choose:
\begin{align}
  \bfM = \bfQ \sqrt{\bfLambda},
  \label{eqn:sqrtW}
\end{align}
where $\sqrt{\bfLambda}$ denotes element-wise square-root operation, and the diagonal matrix $\bfLambda$ and orthogonal matrix $\bfQ$ come from the eigenvalue decomposition of $\bfW = \bfQ \bfLambda \bfQ\trs$ (note $\bfW$ is symmetric positive definite so such decomposition exists).
We found it achieves more balanced sensitivity than using Cholesky decomposition in terms of the resulted largest gradient scale after dimension-wise rescaling.
Although the eigenvalue decomposition requires a cubic complexity, it is only needed once per molecular structure, hence does not introduce a large overhead compared to the cost in density optimization.
Empirically, natural reparameterization considerably stabilizes the optimization of the functional model and brings lower training and validation loss, as presented in \appxref{ablat-den-prep}, underscoring the necessity of balancing the density coefficient with a physical metric.

\subsubsection{Atomic Reference Module} \label{appx:atom-ref}
As mentioned in Methods~\ref{sec:learn-grad}, the per-type bias statistics $\{\Tb_Z\}_Z$ and the global bias $\Tb_\mathrm{global}$ on the dataset are solved from the following over-determined linear equations using least squares method:
\begin{align}
  \Tb_{\clM^{(d)}} := \sum_Z A_Z^{(d)} \Tb_Z + \Tb_{\rm global}
  = \mathrm{mean} \{ T_\tnS^{(d,k)} %
  - {\bar{\bfgg}_{\clM^{(d)}}}\trs \bfpp^{(d,k)}
  \}_k,
  \quad \forall d,
  \label{eqn:atom-ref-bias}
\end{align}
where $A_Z^{(d)} := \#\{a \in \clM^{(d)}: Z^{(a)} = Z\}$ is the number of atoms of type $Z$ in molecule $\clM^{(d)}$.
Note that on the ethanol and chignolin datasets where all structures share the same constitution, we only use the global bias, \ie, $\Tb_{\clM} := \Tb_{\rm global}$.
A similar treatment for energy bias is also adopted in~\citep{musaelian2023learning}.

\begin{figure}[h]
  \centering    
  \includegraphics[width=0.7\textwidth]{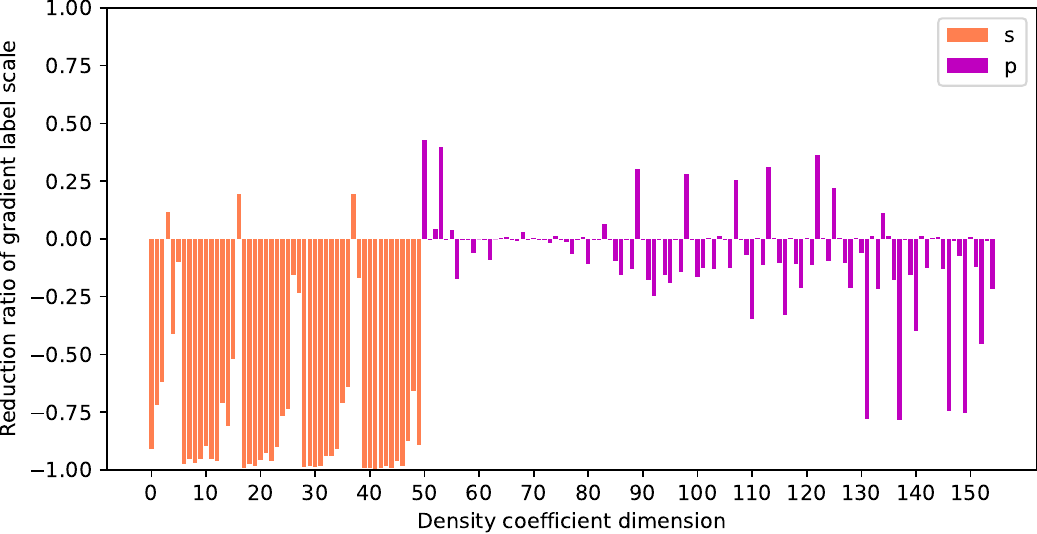}
  \caption{\textbf{Gradient scale changes by the use of the atomic reference module.} For coefficient dimension $\tau$ of the atom type $Z$, the height of the bar is also calculated by $\frac{\mathrm{max\_abs\_grad'}_{Z,\tau} - \mathrm{max\_abs\_grad}_{Z,\tau}}{\mathrm{max\_abs\_grad}_{Z,\tau}}$,
  where 
  $\mathrm{max\_abs\_grad}_{Z,\tau} := \max \big\{ \big\lvert \nabla_{\bfpp_{a,\tau}} T_\tnS^{(d,k)} \big\rvert\big\}_{a: Z^{(a)} = Z, \; k, \, d}$
  and 
  $\mathrm{max\_abs\_grad'}_{Z,\tau} := \max \big\{\big\lvert \nabla_{\bfpp_{a,\tau}}{T'}_\tnS^{(d,k)} \big\rvert\big\}_{a: Z^{(a)} = Z, \; k, \, d}$ 
  measure the gradient scale before and after being centralized by the atomic reference module, respectively. The atomic reference module considerably covers the vast gradient scale.}
  \label{fig:dim-rescale-woatomref}
\end{figure}

To demonstrate the benefit of the atomic reference module, we visualize the gradient scale reduction by the module on the QM9 dataset in \appxfigref{dim-rescale-woatomref}. We find that the atomic reference module substantially reduces the scale (variance) of gradient labels, especially for dimensions corresponding to `s' atomic orbital functions. Since `s' orbitals usually have large gradient mean values but small variance values, subtracting the gradient mean from these dimensions greatly simplifies the learning of gradient labels.

\paragraph{CoefficientAdapter Module}
These enhancement modules and the neural network model are combined in the way shown in the model architecture \appxfigref{model-arch}(a).
In the architecture, the natural reparameterization module and the dimension-wise rescaling module are combined with the local frame technique into the CoefficientAdapter module (\appxfigref{model-arch}(b)).
It transforms the vanilla density coefficient $\bfpp$ to a neural-network-friendly density feature $\bfppt$ as the direct input to the neural network model.
It not only guarantees geometric invariance of density features, but also facilitates efficient learning with vast gradient labels. We outline the implementation in Alg.~\ref{alg:coeffada}.

\begin{algorithm}[t]
\caption{Evaluation of the CoefficientAdapter module} \label{alg:coeffada}
\begin{algorithmic}[1]
\Require Input density coefficients $\bfpp$.
\Require Pre-computed dimension-wise rescaling factors $\{\lambda_{Z,\tau}\}_{Z,\tau}$ (see \eqnref{dim-rescale-def} and \appxref{dim-rescale}); pre-computed quantities for the target molecular structure $\clM$:
Wigner-D matrices $\{\{\bfD_a^l := \bfD^l(\clR_a)\}_{l=0}^{l_{\rm max}}\}_{a=1}^A$ (see \eqnref{locframe}) for transforming coefficients on each atom onto the local frame of the atom, and the square-root matrix $\bfM$ of the density-basis overlap matrix $\bfW$ (see \eqnref{sqrtW}).
\For {$a$ in $1 \cdots A$}
\For {$l$ in $0 \cdots l_{\rm max}$}
\State Transform density coefficients using the Wigner-D matrix: $\bfpp_a^l \leftarrow \bfD_a^l \bfpp_a^l$ (\eqnref{locframe});
\EndFor
\EndFor
\State Conduct natural reparameterization: $\bfpp' \leftarrow \bfM\trs \bfpp$ (\eqnref{nat-reparam});
\For {$a$ in $1 \cdots A$}
\For {$\tau$ in $1 \cdots \clT$}
\State Rescale density coefficients dimension-wise: $\bfppt_{a,\tau} \leftarrow \lambda_{Z^{(a)},\tau} \, \bfpp'_{a,\tau}$ (\eqnref{dim-rescale});
\EndFor
\EndFor
\State \Return $(\bfppt$, $\bfpp')$
\end{algorithmic}
\end{algorithm}

\subsection{Functional Variants}
\label{appx:func-var}

In principle, besides the unknown KEDF, any energy density functionals in the decomposition formulation of ground-state energy (\eqnref{engopt-den}) can be taken as the training objective of the proposed \ourmethod framework. Here we first describe the two versions we mainly employed in the implementation of \ourmethod, and then describe our exploration in other functional variants.

\subsubsection{Residual KEDF} \label{appx:func-var-tsres}
In the implementation of \ourmethod, the first functional version we introduced aims to reduce the difficulty of modeling KEDF directly by employing an existing KEDF as the base functional and learning the residual:
\begin{align}
  T_{\tnS,\theta}(\bfpp,\clM) := T_{\res,\theta}(\bfpp) + T_\base(\bfpp,\clM).
  \label{eqn:tsmodel-res}
\end{align}
Specifically, we chose the APBE KEDF~\citep{constantin2011semiclassical} as the base KEDF since it best fits the training data on the QM9 dataset among four functionals mentioned in Methods~\ref{sec:classic-kedf}.

\begin{figure}[h]
  \centering
  \begin{subfigure}[b]{0.8\textwidth}
      \centering
      \includegraphics[width=\textwidth]{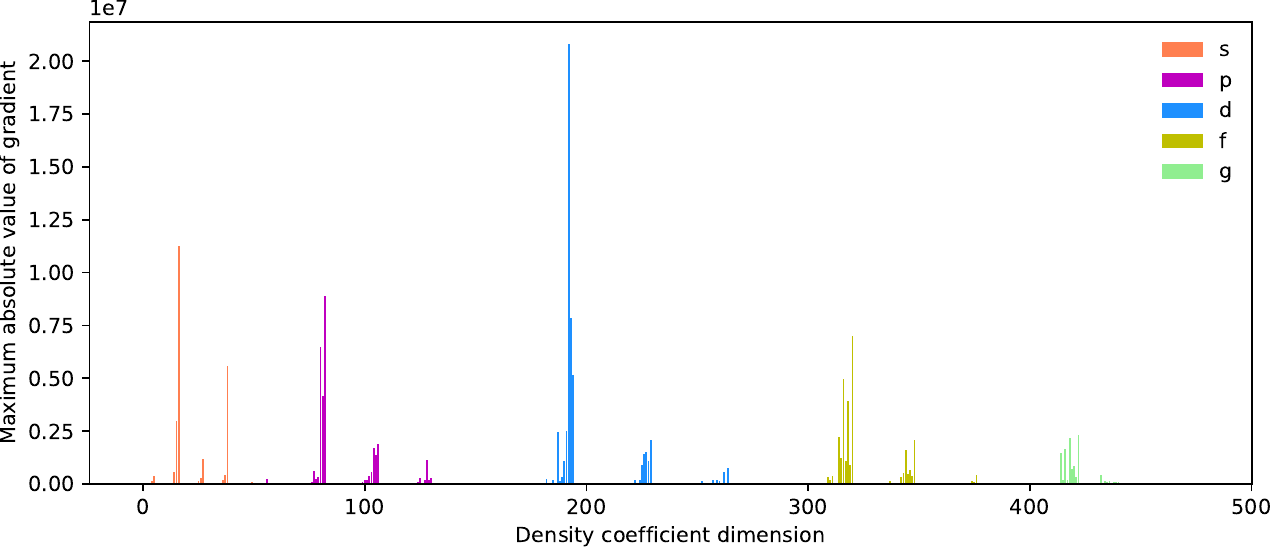}
      \caption{Gradient scales after processed by local frame, natural reparameterization, atomic reference model, and dimension-wise rescaling, in parallel with \appxfigref{dim-rescale}(b).}
      \label{fig:vw-grad-scale}
  \end{subfigure}
  \vfill
  \begin{subfigure}[b]{0.8\textwidth}
      \centering
      \includegraphics[width=\textwidth]{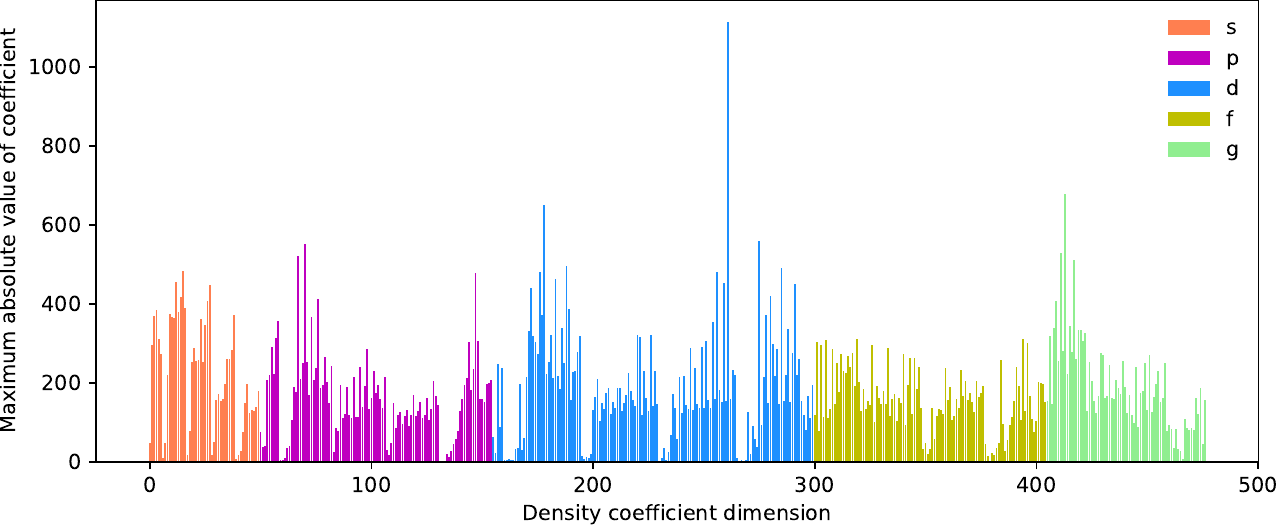}
      \caption{Density coefficient scales after processed by local frame, natural reparameterization, atomic reference model, and dimension-wise rescaling, in parallel with \appxfigref{dim-rescale}(c).}
      \label{fig:vw-coeff-scale}
  \end{subfigure}
  \caption{\textbf{Gradient and density coefficient scales over dimensions on the QM9 dataset in the setting of learning residual KEDF with the vW base KEDF.} %
  The maximum absolute value is used to measure the maximum scale of data. \textbf{(a)} and \textbf{(b)}~respectively present the gradient scales and density coefficient scales after processed by local frame and all enhancement modules (natural reparameterization, atomic reference model, and dimension-wise rescaling), in parallel with \appxfigref{dim-rescale}(b) and (c) for APBE base KEDF respectively.
  Although these techniques have reduced the original gradient scale under a reasonable density scale, the processed gradient scale is still exceedingly large $\sim 10^{7}$ for a neural network to learn.}
  \label{fig:vw-rescale}
\end{figure}

The residual KEDF formulation also allows leveraging a lower bound of KEDF which introduces non-negativity to the residual model, in hope for better extrapolation performance by encoding this analytical property into the model.
Nevertheless, we find that it is not straightforward to gain benefits from this property, as it introduces additional training challenge which outweighs its merit.
Specifically, we take the von Weizs\"acker (vW) KEDF~\citep{weizsacker1935theorie} as the base KEDF, which is a lower bound of the true KEDF~\citep[Thm.~1.1]{lieb1983density}. %
The training challenge is that it renders the gradient for the residual model to learn in vast range.
The visualization of processed gradient and coefficient scales presented in \appxfigref{vw-rescale}.
Even after the processing of local frame, natural reparameterization, atomic reference model, and dimension-wise rescaling, the processed gradient scale, in terms of the maximum absolute value across all dimensions and all datapoints, is $2.08 \times 10^{7}$ on the QM9 dataset (\appxfigref{vw-rescale}(a)), which is orders larger than the scale 2.74 when using the APBE as the base KEDF (\appxfigref{dim-rescale}(b)), and the scale 4.82 for the TXC version below (\ie, learning the sum of the KEDF and XC functional), even allowing a larger processed density coefficient scale of 1113.87 (\appxfigref{vw-rescale}(b)) vs. 507.44 for APBE base KEDF (\appxfigref{dim-rescale}(c)) and 451.59 for the TXC version.
This large gradient scale impedes any effective training of the neural network model.
We even tried dropping out datapoints with particularly large gradient labels that exceed a chosen threshold for training, but observed a performance degradation, due to reduced information on a broader range of densities.
We suspect that this difficulty may be due to the divergence between learning an easier rule and learning a numerically more friendly target. The vW functional leaves the neural network model to learn a non-negative residual, which can be regarded as an easier rule.
But since the vW functional is a lower bound, it may not approximate the KEDF closely, hence could leave the residual and its gradient in a large scale.

Nevertheless, in \appxref{res-in-scale} we empirically verified the learned KEDF model satisfies this lower bound.

\subsubsection{TXC Functional} \label{appx:func-var-tsexc}
The residual KEDF version has achieved a simpler learning target with a tractable gradient range, but it has a computational bottleneck of evaluating the value and gradient of the APBE base KEDF as well as the PBE XC functional from the density coefficient, which is conducted on a grid.
As discussed in \appxref{dft-mat-of}, the time complexity of calculating the value is $O(M N_\grid)$ (recall $M$ is the number of basis functions). %
Evaluating the gradient by automatic differentiation requires the same time complexity as evaluating the value, and moreover, it also requires $O(M N_\grid)$ memory occupation to store intermediate values for back-propagation.
Considering the large prefactor of $N_\grid$ (commonly $\sim 10^3 N$), the computational cost for residual KEDF to conduct density optimization becomes unaffordable for large-scale systems. Such cost was observed on the QMugs dataset, for which \appxref{scala-exp} presents more detailed results.

To get rid of calculation on grid, the second version of learning target for the deep learning model is introduced as the sum of the KEDF and the XC functional, which we call the TXC functional, denoted as $E_{\TXC,\theta}(\bfpp,\clM)$.

\subsubsection{Learning Other Functionals} \label{appx:func-var-others}
Besides the two versions above, we also experimented other direct targets for the deep learning model to learn, including directly learning the KEDF $T_\tnS[\rho]$ (\ie, without a base KEDF), the universal functional $U[\rho]$ (\eqnref{univ-def}), and the total energy functional $E_{\tot,\clM}[\rho]$, \ie, electronic energy $E[\rho]$ in \eqnref{engopt-den} plus inter-nuclear energy $E_\nuc(\clM)$ in \eqnref{eng-nuc} later. We found that both directly learning the KEDF and the universal functional are hard to optimize due to their large gradient range, which is even larger than that for learning the residual KEDF model and the TXC model, and cannot be effectively handled even using the proposed techniques.
Interestingly, %
learning the total energy functional $E_{\tot,\clM}[\rho]$, \ie, adding the external energy $E_\ext$ and inter-nuclear energy $E_\nuc(\clM)$ to the universal functional as the target, substantially reduces the gradient range and enables stable optimization, especially on protein systems we considered.
Specifically, on the 50 test chignolin structures considered in Results~\ref{sec:res-larger-scale}, learning the total energy model $E_{\tot,\theta}(\bfpp,\clM)$ using data of all peptides gives a better density optimization result (per-atom energy MAE 0.071$\,\mathrm{kcal/mol}$ as in \figref{res-outscale}(e) `\ourmethod/Pretrain', per-atom \emph{relative} energy MAE 0.097$\,\mathrm{kcal/mol}$ as in \figref{res-outscale}(c) `\ourmethod') than learning the TXC model $E_{\TXC,\theta}$ (per-atom energy MAE 0.102$\,\mathrm{kcal/mol}$, per-atom \emph{relative} energy MAE 0.148$\,\mathrm{kcal/mol}$).
Nevertheless, these results by the $E_{\TXC,\theta}$ model are still reasonable and effective, and are substantially better than the classical OFDFT using the APBE KEDF (per-atom \emph{relative} energy MAE 0.684$\,\mathrm{kcal/mol}$ as in \figref{res-outscale}(c) `APBE').

\begin{algorithm}[t]
\caption{Usage of \ourmethod to solve a given molecular system (density optimization process)} \label{alg:denopt}
\begin{algorithmic}[1]
\Require A trained kinetic residual model $T_{\res,\theta}(\bfpp, \clM)$ or TXC model $E_{\TXC,\theta}(\bfpp, \clM)$, molecular structure $\clM = \{\bfX, \bfZ\}$ of the given system comprising atomic numbers $\bfZ$ and positions $\bfX$ of all atoms, gradient descent step size $\veps$, initialization method \texttt{init\_method}, a trained density coefficient projection branch $\Delta \bfpp_\theta$ if \texttt{init\_method == `ProjMINAO'} (\appxref{den-proj}).
\State Compute constants for this molecular structure $\clM$:
density basis normalization vector $\bfww$, overlap matrix $\bfW$ and its square-root matrix $\bfM$ for natural reparameterization,
2-center-2-electron Coulomb integral $\bfWt$ for evaluating $E_\tnH$ (\eqnref{EH-den-coeff}),
external potential vector $\bfvv_\ext$ for evaluating $E_\ext$ (\eqnref{Eext-den-coeff}),
rotation matrices and Wigner-D matrices for local frame,
build grid and calculate density basis values on grid points for evaluating $E_\XC$ and $T_\base$ if using $T_{\res,\theta}$ model;
\If{\texttt{init\_method == `H\"uckel'}}
\State $\bfpp^{(1)} \leftarrow \texttt{H\"uckel\_init}(\clM)$;
\ElsIf{\texttt{init\_method == `ProjMINAO'}}
\State $\bfpp \leftarrow \texttt{MINAO\_init}(\clM)$;
\State $\bfpp^{(1)} \leftarrow \bfpp - \Delta \bfpp_\theta (\bfpp, \clM)$;
\EndIf
\State $k \leftarrow 1$
\While{stopping criterion (described in \appxref{stop-cri}) is not met}
\State Compute electronic energy using the PyTorch implementation (see Alg.~\ref{alg:kedf} for evaluating $T_{\res,\theta}$ or $E_{\TXC,\theta}$):
\If{using the kinetic residual model}
\State $E^{(k)} \!\leftarrow\! T_{\res,\theta}(\bfpp^{(k)}\!, \clM) + T_\base(\bfpp^{(k)}\!, \clM) + E_\tnH(\bfpp^{(k)}\!, \bfWt) + E_\XC(\bfpp^{(k)}\!, \clM) + E_\ext(\bfpp^{(k)}\!, \bfvv_\ext)$;
\ElsIf{using the TXC model}
\State $E^{(k)} \leftarrow E_{\TXC,\theta}(\bfpp^{(k)}, \clM) + E_\tnH(\bfpp^{(k)}, \bfWt) + E_\ext(\bfpp^{(k)}, \bfvv_\ext)$;
\EndIf
\State Calculate the gradient of the electronic energy with respect to the density coefficients:
\Statex \quad \, $\nabla_\bfpp E^{(k)} \leftarrow \texttt{auto\_grad} (E^{(k)}, \bfpp^{(k)})$;
\State Update the density coefficients using projected gradient: 
\Statex \quad \, $\bfpp^{(k+1)} \leftarrow \bfpp^{(k)} - \veps \lrparen{\bfI - \frac{\bfww \bfww\trs}{\bfww\trs \bfww}} \nabla_\bfpp E^{(k)}$;
\State $k \leftarrow k + 1$
\EndWhile
\State \Return $(E, \bfpp)$ at the iteration determined by the stopping criterion.
\end{algorithmic}
\end{algorithm}

\subsection{Density Optimization Details} \label{appx:den-opt}

After a functional model is learned, \ourmethod solves a given molecular structure by density optimization, as shown in Alg.~\ref{alg:denopt}.
Here we provide additional results and details for density optimization.

\begin{figure}[h]
  \centering
  \includegraphics[width=0.7\textwidth]{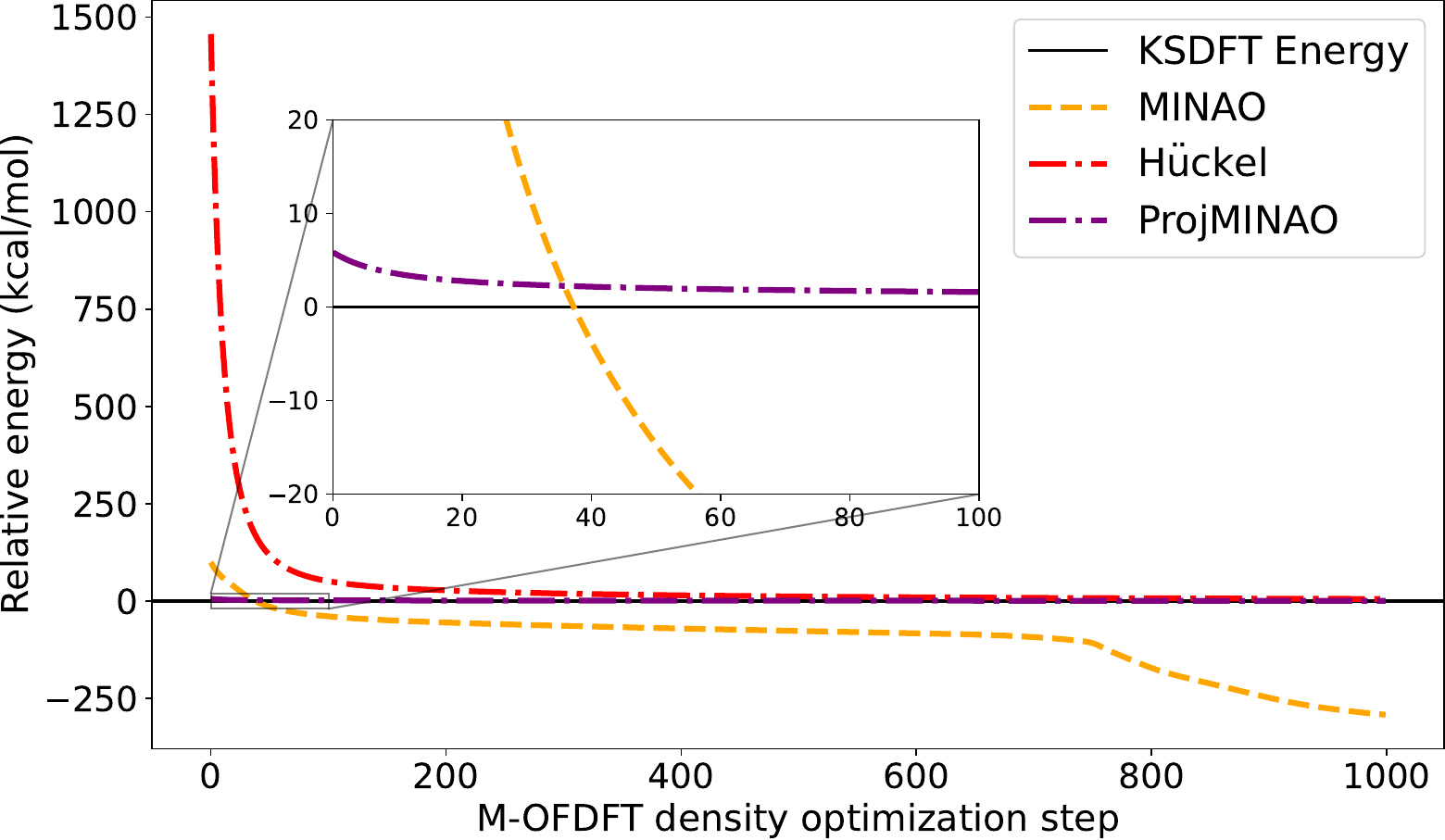}
  \caption{\textbf{Typical density optimization curves of \ourmethod for a QM9 molecule with different initialization methods.}
  MINAO, the common KSDFT initialization, leads the optimization to a large gap from the target energy, since it is not from the eigensolution to an effective Hamiltonian matrix, hence lies off the training-data manifold (out of distribution). %
  H\"uckel initialization solves an eigenvalue problem, which indeed converges the curve with a much smaller energy error of 5.34 $\mathrm{kcal/mol}$.
  ProjMINAO initialization uses a deep learning model to project the MINAO density onto the training-data manifold, which also converges the curve and achieves the best result of 0.60 $\mathrm{kcal/mol}$ energy error.
  The inset figure highlights the role of density optimization even though the ProjMINAO density is close to the ground-state density.
  }
  \label{fig:stage2-init}
\end{figure}

\subsubsection{Additional Density Optimization Results} \label{appx:den-res}

As mentioned in Methods~\ref{sec:stage2}, \ourmethod can produce reasonable optimization curves for a given molecular structure from either of the two initialization methods H\"uckel and ProjMINAO, and gives accurate energy and HF force results.
Here we illustrate the density optimization behaviors of various initialization strategies using a QM9 molecule. As shown \figref{stage2-init}, the optimization process starting from the MINAO density leads to an obvious gap from the target KSDFT energy. In contrast, the optimization curve starting from H\"uckel density converges closely to the target energy, although it shows a larger energy error than MINAO density at initialization. 
Furthermore, ProjMINAO initialization also converges closely to the target energy, and finally achieves a better performance than H\"uckel initialization. These findings demonstrate that both H\"uckel and ProjMINAO initializations have a better alignment with the training-density manifold, and thus lead to better generalization performance during density optimization.
More attractively, \figref{stage2-init} implies that \ourmethod only requires an on-manifold initialization but does not need projection in each density optimization step, implying its better robustness than previous methods.

\begin{figure}[t]
  \centering
  \includegraphics[width=0.95\textwidth]{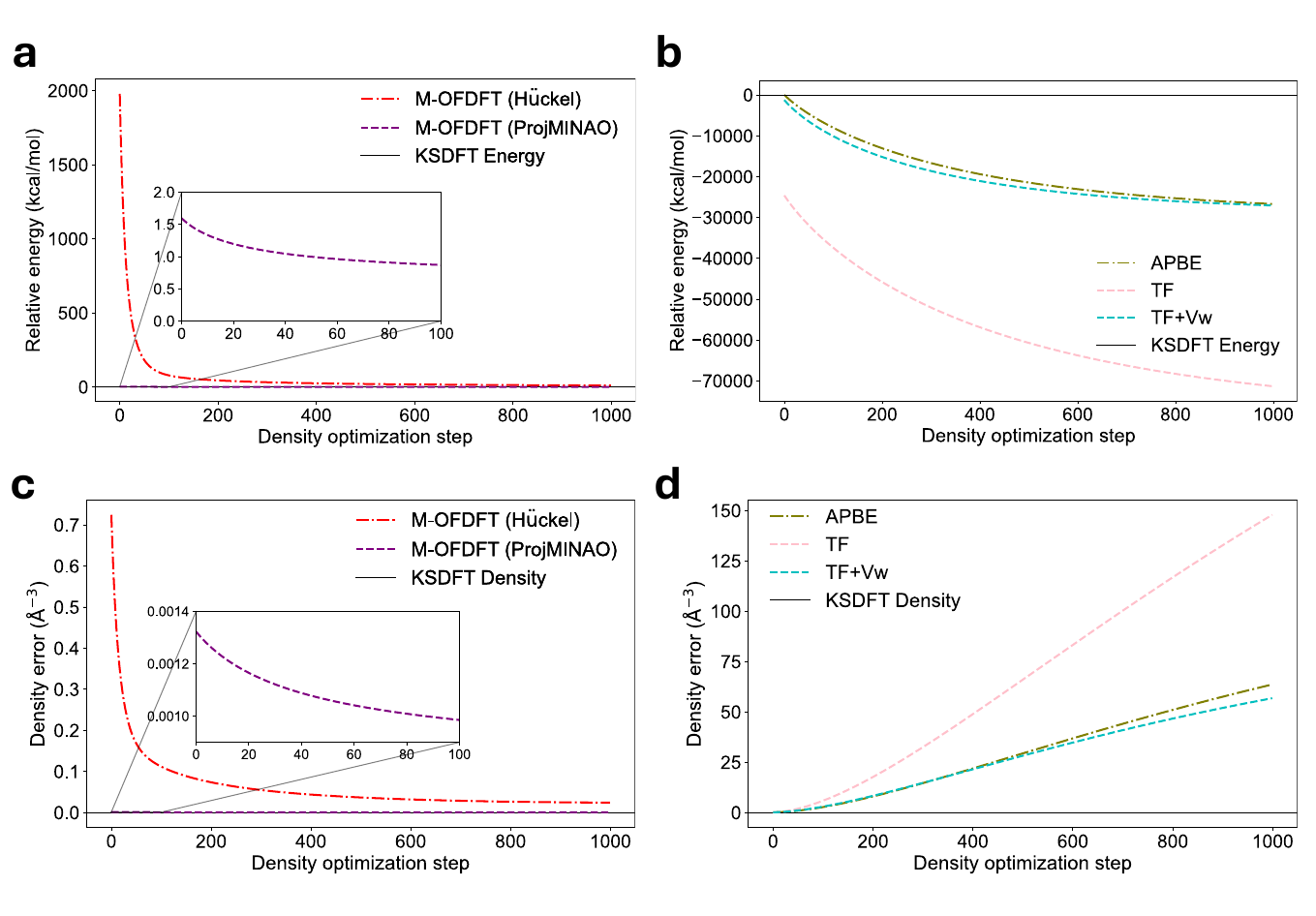}
  \caption{\textbf{Density optimization curves by \ourmethod and OFDFT with classical KEDFs.}
  The figures come in parallel with \figref{stage2-init} but on a different QM9 molecule.
  \textbf{(a)}~Curves in relative energy along the density optimization process of \ourmethod with different initialization methods. \ourmethod converges closely to the true ground-state energy using either initialization method.
  \textbf{(b)}~Curves in relative energy along the density optimization process of OFDFT using classical KEDFs. The curves run below the true ground-state energy with a large gap.
  \textbf{(c)}~Curves in density error along the density optimization process of \ourmethod with different initialization methods. \ourmethod converges closely to the true ground-state density using either initialization method, even though the optimization is not driven by minimizing the density error.
  \textbf{(d)}~Curves in density error along the density optimization process of OFDFT using classical KEDFs. The curves diverge quickly.
  Curves for TF+$\frac19$vW are omitted in \textbf{(b)} and \textbf{(d)} since they soon run out of the shown range.
  }
  \label{fig:den-res}
\end{figure}

To further show the advantage of the density optimization results, we compare \ourmethod with classical OFDFT using well-established KEDFs, following the same setting as \figref{stage2-init} except on a different QM9 molecule.
First, we compare the optimization curves in energy by \ourmethod and OFDFT with classical KEDFs including TF~\citep{thomas1927calculation,fermi1928statistische}, TF+$\frac19$vW~\citep{brack1976extended}, TF+vW~\citep{karasiev2012issues}, and APBE~\citep{constantin2011semiclassical}, from the conventional MINAO~\citep{sun2018pyscf} initialization.
As illustrated in \appxfigref{den-res}(a-b), \ourmethod converges closely to the true ground-state energy using both initialization methods, while all classical KEDFs lead to optimization curves falling below the true ground-state energy. %
We did not plot the curve for TF+$\frac19$vW since it diverges so vastly that the curve soon runs out of the shown range. 

We further plot the density error along with the optimization process, measured in the L2-metric $\int \lrvert{\rho(\bfrr) - \rho^\star(\bfrr)}^2 \dd \bfrr$ from the KSDFT ground-state density $\rho^\star(\bfrr)$ (see \appxref{nat-reparam} for calculation details). %
As shown in \appxfigref{den-res}(c-d), \ourmethod from either initialization continuously drives the density towards the true ground-state density and results in a small density error, even though the optimization process is not driven by minimizing the density error (but by minimizing electronic energy).
This indicates that the energy objective of \ourmethod constructed from the learned functional model is not artificial in the sense of only producing the correct energy after optimization, but it also leads to the correct electron density, so the functional holds the desired physical meaning. %
In contrast, density error curves by classical KEDFs diverge quickly and present an ascending trend, revealing a problem for applying these functionals to molecular systems. %
Again, we did not plot the curve for TF+$\frac19$vW since it soon runs out of the shown range.

A subtlety in the density optimization process is the non-negativity of the density value everywhere in space. As the basis functions follow the form of the multiplication of a Gaussian radial function which is always non-negative, and a spherical harmonic function or a monomial (as is the case of the even-tempered basis in \eqnref{even-tempered-basis}) that accounts for the angular anisotropicity which can take negative values. The coefficients hence need to be within a certain region to guarantee that the represented density function is non-negative everywhere.
Due to the complexity of the basis functions, an explicit expression for such a constraint is not obvious.
We hence implemented a numerical guarantee that enforces the non-negativity of density value on each grid point, %
by adding the following artificial energy penalty term to the minimization objective \eqnref{Etot-den-coeff} of density optimization:
\begin{align+}
  E_\mathrm{nonneg}(\bfpp, \clM) := \sum_{g=1}^{N_\grid} \max\{-\rho_\bfpp(\bfrr^{(g)}), 0\},
\end{align+}
where $\rho_\bfpp(\bfrr) := \sum_\mu \bfpp_\mu \omega_\mu(\bfrr)$ is the electron density function represented by coefficient vector $\bfpp$ (see \eqnref{den-expd}), and $\{\bfrr^{(g)}\}_{g=1}^{N_\grid}$ is a set of grid points for the molecular structure $\clM$.
Nevertheless, in our empirical trials, we found that this additional term is seldom activated during density optimization, and density optimization without this term already leads to an electron density that is non-negative on almost all grid points.
The number of exceptional grid points is even smaller than that due to density fitting error.
Considering the cost of evaluating density values on grid points, we hence omitted this step.

For ensuring density non-negativity, it is possible to represent the density as the square of linear combination of the atomic orbital basis functions, as adopted in many existing OFDFT implementations (\eg, refs.~\citep{wang1999orbital,fujinami2020orbital,imoto2021order,remme2023kineticnet}). But this would revert the computational complexity to quartic ($O(N^4)$) due to the Hartree term, and sacrifice the advantage over KSDFT.
Future explorations could be expanding the density onto a set of non-negative-valued basis functions.

\subsubsection{Density Initialization Details} \label{appx:den-proj}

Recall that we introduced two initialization methods in Methods~\ref{sec:stage2}, H\"uckel and ProjMINAO, for stable density optimization in \ourmethod.
Initialization is worthy of the attention since it should stay in the training-data manifold of the functional model for reliable prediction.
Training data come from eigensolutions of effective one-electron Hamiltonian matrix, which motivates the first choice of the H\"uckel initialization~\citep{hoffman1963an,lehtola2019assessment} which follows the same mechanism.
The second choice is to project the initial density onto the manifold, inspired by previous methods, \eg, using local PCA~\citep{snyder2012finding,brockherde2017bypassing} or kernel PCA~\citep{snyder2013orbital}.
But these existing projection techniques is not easily applicable for \ourmethod since the density optimization domain is the coefficient space generated by the given molecular structure, which differs for different molecular structures, so it is not straightforward to determine the training-data manifold within the coefficient space for an unseen molecular structure.
We hence use an additional deep learning model to conduct the projection of the MINAO initialization, which we call ProjMINAO.

To implement the ProjMINAO initialization, we construct an additional output branch $\Delta \bfpp_{\theta}(\bfpp, \clM)$ to predict the required correction to project the given density coefficient $\bfpp$ towards the ground-state density coefficient $\bfpp^\star$ for the molecular structure $\clM$.
Due to the formulation of KSDFT, the ground-state density corresponds to the eigensolution of an effective Hamiltonian (\appxref{dft-variation}), hence is on the training-data manifold for sure. Projecting towards the ground state could also accelerate convergence in density optimization.
The branch is built on top of the last \emph{G3D} module shown in \appxfigref{model-arch}(a) of the KEDF model.
It processes the hidden representations of each atom through an MLP module to predict the corresponding coefficient difference.
The branch is trained to project the density coefficient data along the SCF iteration of KSDFT onto the corresponding ground-state coefficient:
\begin{align}
  \sum_d \sum_k \lrVert{ \Delta \bfpp_{\theta}(\bfpp^{(d, k)}, \clM^{(d)}) - (\bfpp^{(d, k)} - \bfpp^{(d, \star)})}.
  \label{eqn:loss-den}
\end{align}
Note that even though the projection is aimed at the ground state, we do not rely on it for accuracy, as the functional model could continue optimizing the density (\figref{stage2-init}).

Since the MINAO or H\"uckel initialization primarily produces orbital coefficients, density fitting (\appxref{dft-label-dfit}) is required to convert them to density coefficients to complete the density initialization.
Note that this procedure is performed only \emph{once} for each molecular structure, so the cubic computational cost of density fitting (and generating H\"uckel initialized orbitals) does not dominate over the quadratic scaling of density optimization on regular workloads. %

For very large molecules, we also propose an alternative initialization to directly generate the density coefficients by superposition of isolated-atom densities, in the same spirit of MINAO which uses the superposition of isolated-atom orbitals.
This amounts to concatenating the fitted MINAO density coefficients of each atom as if in isolation, %
which only has a linear computational cost.
This constructed density coefficients are then fed to the projection branch for the correction, which is the final initialized density.
In practice, this technique drastically reduces the time needed to construct the initial coefficients, further reducing the overall time consumption of \ourmethod, and renders only a minor increase of error (approximately 0.02$\,\mathrm{kcal/mol}$ in per-atom energy error on the interested protein systems in Results~\ref{sec:res-time}).

The effectiveness of this approach to handle off-the-training-data-manifold issue in density optimization is demonstrated qualitatively by the convergent curves of energy and density in \figref{stage2-init} and \appxfigref{den-res}, and quantitatively by the results in Results.~\ref{sec:res-in-scale} and Results.~\ref{sec:res-larger-scale}.
In contrast to some other deep learning OFDFT methods~\citep{snyder2012finding,snyder2013orbital,brockherde2017bypassing} which require the on-manifold projection in each density optimization step, \ourmethod only requires an on-manifold initialization, meaning at most one projection.
This makes the implementation much more efficient especially for large systems, and also indicates that the learned functional generalizes better since it is more robust to unseen densities. %

\subsubsection{Stopping Criterion} \label{appx:stop-cri}

Another required specification for density optimization is stopping criterion.
Ideally, the solution from \ourmethod is expected at the stationary point with zero projected gradient of the electronic energy. But in practice, the exact zero-gradient point may be missed due to discretization error. Therefore, we propose a set of stopping criteria to determine the practical stationary point over a chosen number of optimization steps.
\begin{itemize}
  \item For molecules in a similar scale as those in training, the stopping criterion is chosen as the step with minimal single-step energy update, \ie, at the global (over the chosen number of optimization steps) minimum of single-step energy update.
  This is the closest step to the stationary point in practice.
  \item For larger-scale molecules than those in training, however, the above criterion is not as effective as on in-scale systems. The reason is that the reliable coefficient region around the ground-state density for the model is likely smaller and anisotropic due to the extrapolation risk, so the coefficients may find a mistaken direction to "sneak off" the ground state in an extensive optimization. Note that it is hard to project the confronted density during optimization onto the training-data manifold precisely, since the manifold of coefficient is unknown for a molecular structure not present in training.
  Therefore, we mitigate this problem by controlling the extent of density optimization to balance optimization effectiveness and reliability.
  Specifically, we consider two additional criteria: \itemone the step where the projected (onto the tangent space of normalized coefficients; not the training-data manifold) gradient norm \emph{first} stops decreasing, \ie, the first local minimum of the projected gradient norm. \itemtwo the step where the single-step energy \emph{first} stops decreasing, \ie, the first local minimum of single-step energy update.
  These two criteria characterize a local minimum on the optimization process, hence represent a certain level of optimization effectiveness. Meanwhile, they are chosen to be the first encountered point with such characterization, where is likely still in the reliable region for the model. %
  Based on empirical results on the validation set, we prefer first invoking criterion \itemone. If this criterion is not met within the chosen optimization steps, we seek for criterion \itemtwo. If this is still not met, we resort to the original criterion of the global minimum of single-step energy update, which always exists.
\end{itemize}

\subsubsection{Density Optimization Hyperparameters}
During the deployment stage, we use the ProjMINAO as the initialization choice for all settings in Results~\ref{sec:res}.
The adopted gradient descent steps and step size $\veps$ are chosen according to the performance of evaluation datasets. The gradient descent steps and step size are set to 1000 and $1\e{-3}$ for all molecular datasets and functional variants (including traditional KEDF baselines), with the exception of implementing the learned KEDF models $T_{\res, \theta}$ and $E_{\TXC, \theta}$ on the ethanol dataset, where the step size $\veps$ is set to $5\e{-4}$.
The vanilla Stochastic Gradient Descent (SGD) optimizer is adopted in the density optimization process.

\section{Related Work} \label{appx:relw}

In this section, we give a more detailed discussion on prior works that leverage machine learning to approximate the kinetic energy density functional (KEDF) for OFDFT.
Kernel ridge regression is employed in pioneering works~\citep{snyder2012finding,snyder2013orbital,li2016pure,brockherde2017bypassing,hollingsworth2018can,kalita2021learning}, including the extension that leverages kernel gradients~\citep{meyer2020machine}. These works have proven the success of the idea of machine-learning OFDFT. Such models can be seen as non-local, but the costly calculation on grid restricts the applications to 1-dimensional systems. %
Some other works fit a linear combination of classical KEDF approximations with explicit expression~\citep{alghadeer2021highly}, including non-local ones~\citep{kumar2022accurate}, but the demonstrated systems are still effectively 1-dimensional.
Deep neural networks have also been explored recently, including multi-layer perceptrons (also called feed-forward neural networks)~\citep{seino2018semi,seino2019semi,golub2019kinetic,fujinami2020orbital,imoto2021order,del2023variational} and convolutional neural networks~\citep{yao2016kinetic,meyer2020machine,ryczko2021orbital,remme2023kineticnet}. %
Many of them learn the kinetic energy density at each grid point from semi-local density features on that point~\citep{imoto2021order}, and to compensate for non-local effects, third-order~\citep{seino2018semi,seino2019semi,fujinami2020orbital} and fourth-order~\citep{golub2019kinetic} density derivative features are leveraged. Others consider the interaction of density features at different locations hence are non-local~\citep{yao2016kinetic,meyer2020machine,ryczko2021orbital,del2023variational,remme2023kineticnet}.
Many of such works enable the calculation on 3-dimensional systems~\citep{yao2016kinetic,seino2018semi,seino2019semi,golub2019kinetic,imoto2021order,ryczko2021orbital,remme2023kineticnet}, and the lower computational complexity than KSDFT has been shown empirically~\citep{imoto2021order}. %
Nevertheless, the demonstrated systems are still limited to tiny molecules of dozen atoms, with exceptions of ref.~\citep{yao2016kinetic} with about 30 atoms but restricted to alkanes, and ref.~\citep{imoto2021order} with a few thousands atoms but on material systems and without accuracy evaluation.
The contribution of our work \ourmethod is the applicability to diverse and larger molecules (\eg, the chignolin polypeptide containing 168 atoms) with extrapolation capability while maintaining the low complexity of OFDFT.
Technical contributions are summarized in the beginning of Methods~\ref{sec:method}.

\begin{figure}[h]
  \centering
  \includegraphics[width=0.8\textwidth]{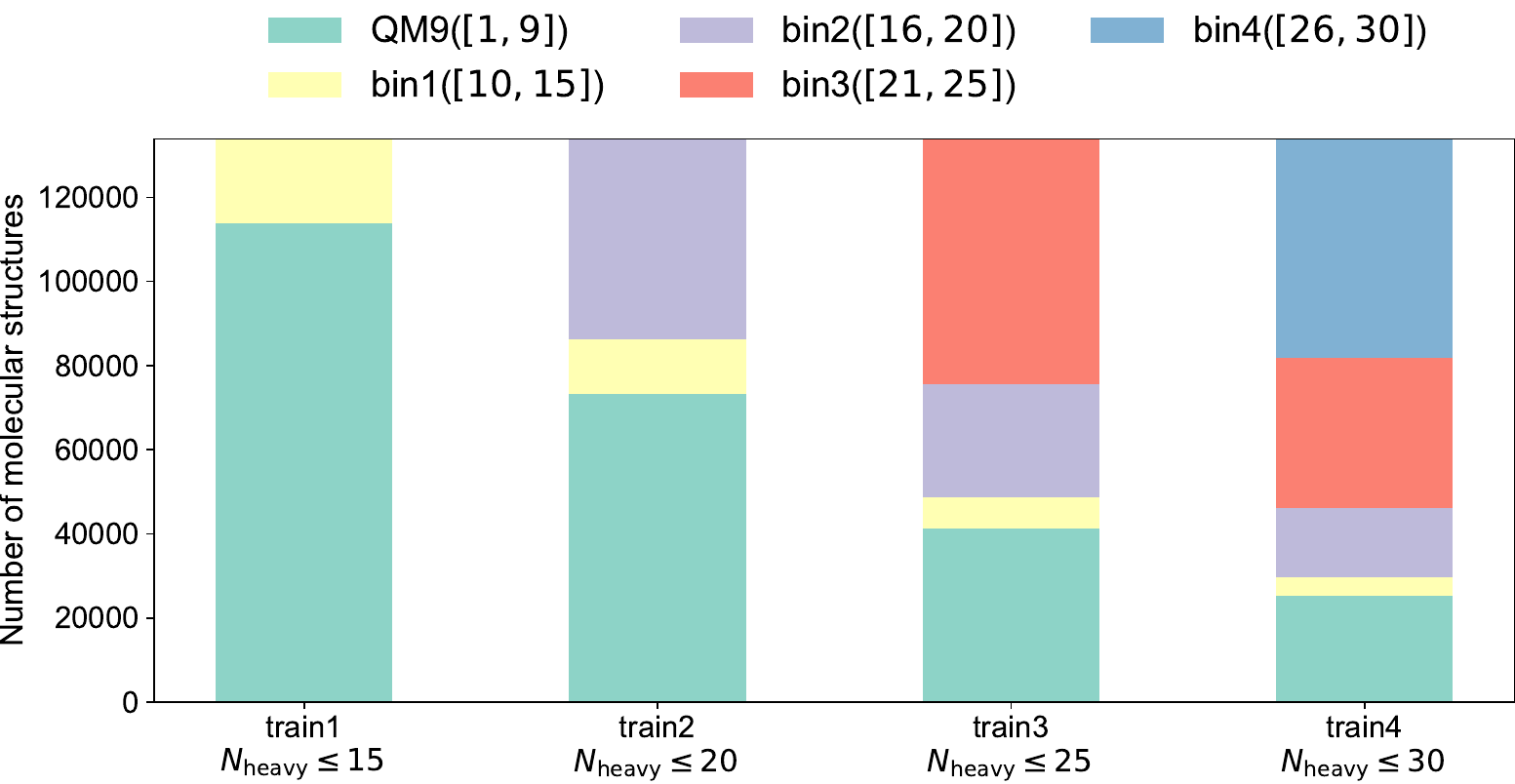}
  \caption{\textbf{Compositions of training datasets from QM9 and QMugs for the extrapolation setting in \figref{res-outscale}(b).}
  The training datasets contain the same number of datapoints (molecular structures), while involve increasingly larger molecules. The proportion of molecules in each scale range (\ie, QM9, bin1-bin4) respects the proportion in the joint dataset. The range of the number of heavy atoms in each data source is listed in the figure (\ie, $[1, 9]$ for QM9).
  }
  \label{fig:qmugs-comp-ratio}
\end{figure}

\section{Additional Empirical Results}

\subsection{In-Scale Results}

\subsubsection{Quantitative Results} \label{appx:res-in-scale}

\begin{table}[h]
  \centering
  \caption{\textbf{Performance of \ourmethod with different learning targets and initialization configurations, as an extension to the results narrated in Results~\ref{sec:res-in-scale}.}
    The mean absolute errors (MAEs) in energy and Hellmann-Feynman (HF) force are listed in $\mathrm{kcal/mol}$ and $\mathrm{kcal/mol/\angstrom}$, respectively. The results are calculated on ethanol test structures ($n$=10,000) and QM9 test molecules ($n$=13,388).
  }
  \begin{tabular}{lcccc}
    \toprule
    \multirow{2}{*}{Method} & \multicolumn{2}{c}{Ethanol} & \multicolumn{2}{c}{QM9} \\
    \cmidrule(lr){2-3} \cmidrule(lr){4-5}
    & Energy & HF force & Energy & HF force \\
    \midrule
    $T_\res$-H\"uckel & 5.27 & 27.73 & 9.88 & 12.73 \\
    $T_\res$-ProjMINAO & \textbf{0.18} & \textbf{1.18} & 0.93 & \textbf{2.91} \\
    $E_\TXC$-H\"uckel & 3.95 & 126.65 & 9.69 & 123.51 \\
    $E_\TXC$-ProjMINAO & \textbf{0.18} & 2.07 & \textbf{0.73} & 4.73 \\
    \midrule
    \mlff & 0.10 & 25.24 & 0.25 & 7.72 \\
    \bottomrule
  \end{tabular}
  \label{tbl:res-in-scale-abs}
\end{table}

\paragraph{Functional variants and initialization}
As mentioned in Results~\ref{sec:res-in-scale}, \ourmethod achieves chemical accuracy on unseen conformations (ethanol) and chemicals (QM9) in a similar scale as seen during training, using a deep learning model targeting the residual KEDF on top of a base KEDF.
Here we provide results under more technical choices in \appxtblref{res-in-scale-abs}.
Firstly, we let the model learn the sum of the kinetic and XC energy $E_\TXC$ to get rid of the large prefactor of grid computation for calculation on large-scale molecules (\appxref{func-var-tsexc}).
This is adopted in time complexity (Results~\ref{sec:res-time}) and extrapolation (Results~\ref{sec:res-larger-scale}) studies, and here we list its performance in this in-scale case.
Secondly, as presented in Methods~\ref{sec:stage2}, two initialization strategies are proposed to address the out-of-distribution issue of initialized density. %
The results reported in Results~\ref{sec:res-in-scale} are produced by ProjMINAO initialization, and here we also list the results produced by the H\"uckel initialization method, which is one of the standard initialization methods in common DFT calculation~\citep{sun2018pyscf,lehtola2019assessment}, and does not require an additional deep learning model (or module).
We also provide the results of \mlff, \ie, a deep learning model with the same architecture that directly predicts the ground-state energy of the given molecular structure in an end-to-end way.

From \appxtblref{res-in-scale-abs}, we can see that although the $E_\TXC$ model combines with an additional unknown density functional, it can still be effectively trained, and does not have obvious negative effects on the accuracy. The energy accuracy is even improved on both datasets given the same initialization method.
For the initialization methods, even without the deep learning shortcut, \ourmethod still achieves a reasonable accuracy using the H\"uckel initialization. With an additional deep learning module for predicting a correction, the ProjMINAO initialized density stays close to the ground-state density, so it can yield better results. %
Finally, \ourmethod achieves a comparable energy accuracy with the end-to-end method \mlff, even though the process to give the energy by \ourmethod (density optimization using the KEDF model) is not directly supervised.
Notably, on tasks that do not have a direct supervision, \ourmethod is superior over \mlff.
As the table shows, \ourmethod achieves a substantially more accurate HF force estimation than \mlff, even though neither has seen HF force label in training.
Abundant evidence in Results~\ref{sec:res-larger-scale} also shows the dominating advantage of \ourmethod in extrapolation to molecules larger than seen during training.

\paragraph{Details on the potential energy surface (PES) study}
As shown in \figref{res-in-scale}(c) of Results~\ref{sec:res-in-scale}, \ourmethod can accurately reproduce the PESs of ethanol. Here we provide more implementation specifics of the evaluation process. To benchmark the PES, we generate a series of ethanol structures by varying either the \ch{H-C-C-O} torsion angle or the \ch{O-H} bond length, starting from the equilibrium ethanol 
conformation (optimized by classical molecular dynamics simulation). The torsion angles are taken uniformly on $[-180^\circ, 180^\circ)$ with 15$^\circ$ increment, where the 0$^\circ$ angle is defined when the four atoms are on the same plane and \ch{H} and \ch{O} are on the same side.
The bond lengths are taken uniformly on $[0.856\,\angstrom, 0.144\,\angstrom]$ with 0.015$\,\angstrom$ increment. The interval is taken as the range of the \ch{O-H} bond length in the training dataset.
The energy curves in \figref{res-in-scale}(c) are produced by the residual KEDF $T_{\res,\theta}$ version (\appxref{func-var-tsres}) of \ourmethod with ProjMINAO density initialization. This setting is consistent with other results shown in Results~\ref{sec:res-in-scale}.
The curves are %
\emph{not} evaluated on densities solved by KSDFT~\citep{yao2016kinetic,seino2019semi}.

\paragraph{Geometry optimization study}
To investigate the utility of \ourmethod for geometry optimization, we integrate the \ourmethod implementation with the geometry optimization framework in PySCF~\citep{sun2018pyscf}, wherein the HF force (\appxref{hfforce}) by \ourmethod is used.
We generate a set of initial ethanol structures by varying the \ch{H-C-C-O} torsion angle from the equilibrium ethanol conformation.
The torsion angles are taken uniformly on $[-180^\circ, 60^\circ]$ with 30$^\circ$ increment.
For each initial structure, we relax the structure using both KSDFT and \ourmethod for at most 100 steps.
For \ourmethod, the residual KEDF $T_{\res,\theta}$ version (\appxref{func-var-tsres}) with ProjMINAO density initialization is used, which is consistent with other results shown in Results~\ref{sec:res-in-scale}.

\begin{figure}
  \centering
  \includegraphics[width=\textwidth]{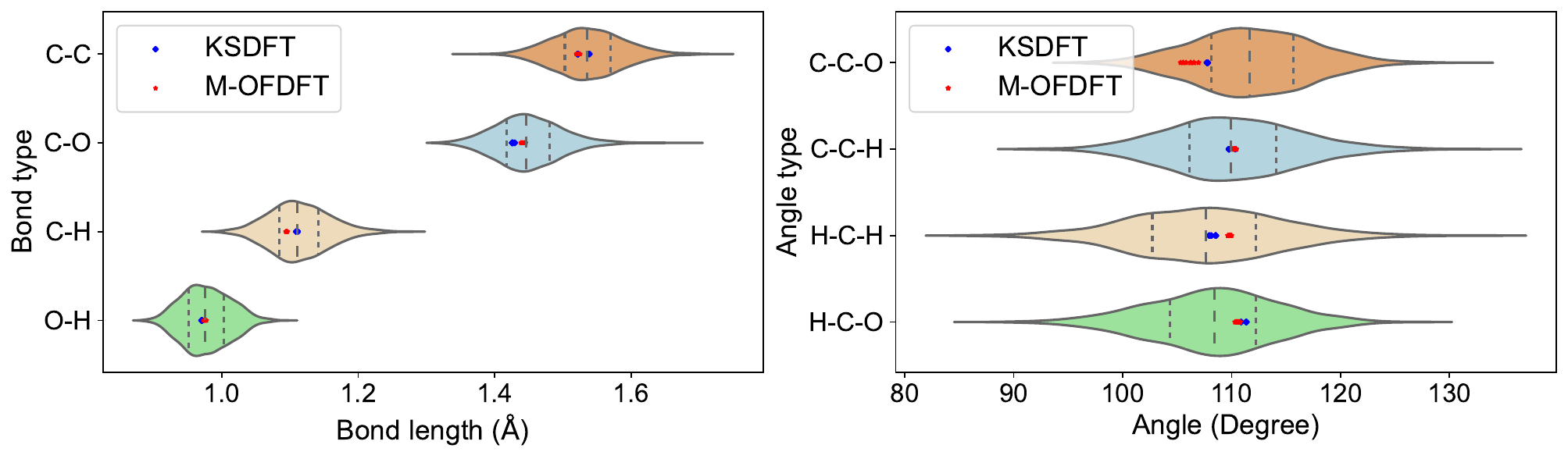}
  \caption{\textbf{Bond lengths and bond angles of ethanol structures by geometry optimization with \ourmethod and KSDFT.} Different points represent the optimizes results from different initial structures. The violin plots in the background depict the distributions of the corresponding bond lengths or angles under thermodynamic equilibrium, which are calculated on ethanol structures ($n$=20,000). The vertical dashed lines represent the quantiles.
  The mean RMSD between the optimized structure by \ourmethod and by KSDFT over the initial structures is 0.07$\angstrom$. %
  }
  \label{fig:torsion_go}
\end{figure}

To evaluate the optimized structures by \ourmethod, we first calculate the rooted mean square deviation (RMSD) between the optimized structure by \ourmethod and the optimized structure by KSDFT for each initial ethanol structure. The mean RMSD value across all the initial structures is 0.07$\,\angstrom$, indicating a good consistency of \ourmethod with KSDFT.
To further evaluate the optimized structures by \ourmethod, we compare the bond lengths and angles against those of optimized structures by KSDFT. For a reference to assess the error, for each bond or angle type, we plot the distribution in the form of violin plot of the bond length or angle value in the ethanol training dataset. As the dataset is from the MD17 dataset~\citep{chmiela2017machine, chmiela2019sgdml}, the plot represents the distribution in thermodynamic equilibrium.
As depicted in \appxfigref{torsion_go}, the majority of the bond lengths and angles of the optimized structures by \ourmethod exhibit good agreement with the results of KSDFT, and align closely with the high-density region of the corresponding thermodynamic equilibrium distributions. The difference from KSDFT results is also substantially smaller than the span of the corresponding thermodynamic equilibrium distribution.
This result underscores the practical efficacy of \ourmethod for geometry optimization.

\begin{figure}
  \centering
  \includegraphics[width=\textwidth]{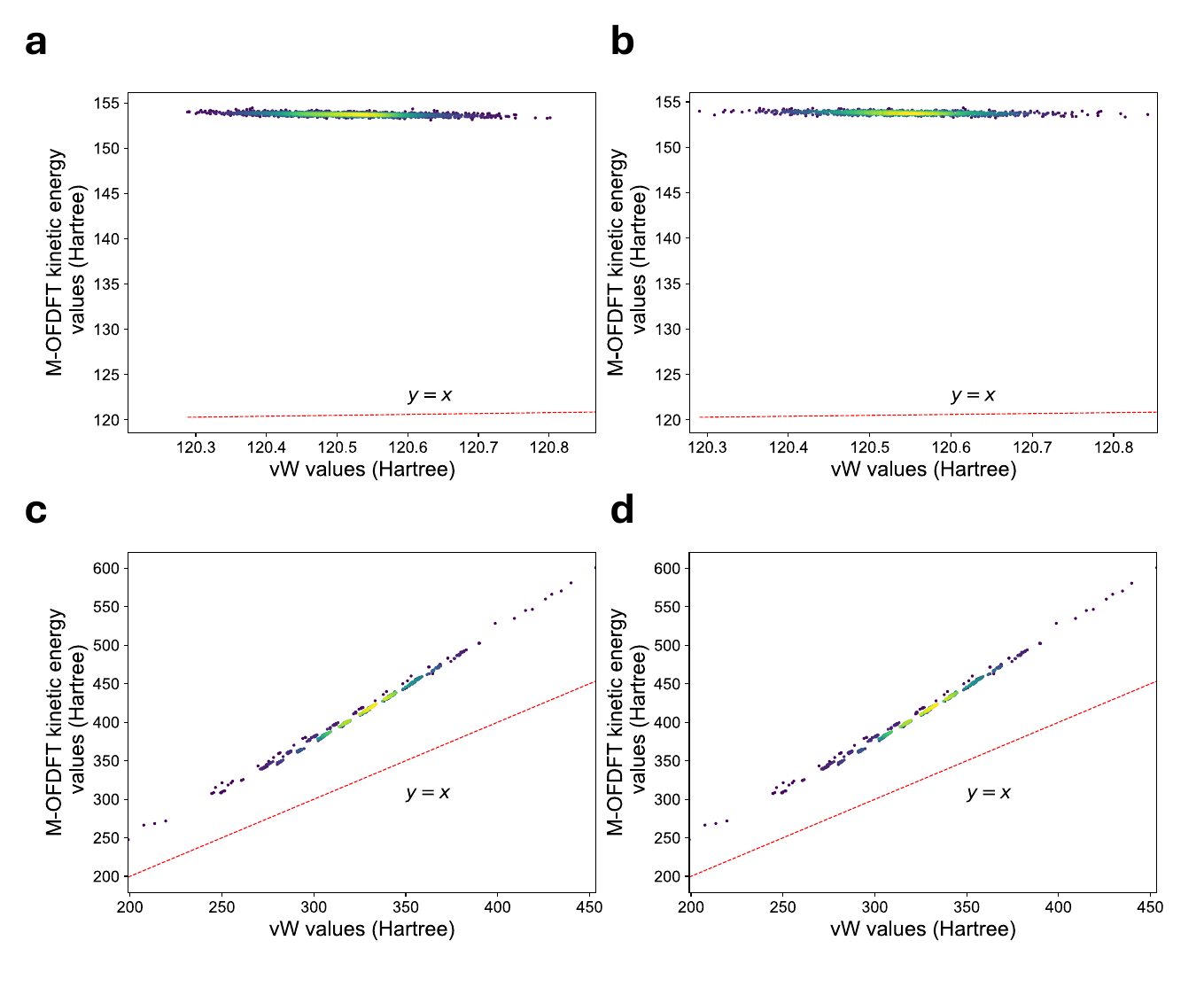}
  \caption{\textbf{Kinetic energy value comparison between \ourmethod and the vW KEDF.} Each point represents the vW KEDF value (x-axis coordinate) and the kinetic energy value by our learned KEDF model (y-axis coordinate) of each electron density.
  Two versions of our learned KEDF model are considered, including the residual KEDF version \textbf{(a,c)} and the TXC functional version \textbf{(b,d)}.
  The densities for evaluation come from ground-state densities optimized by our KEDF model on 10,000 unseen ethanol test structures \textbf{(a-b)} or 13,388 unseen QM9 test structures \textbf{(c-d)}.
  }
  \label{fig:kinetic_eng}
\end{figure}

\paragraph{Comparison with vW as a lower bound}
As mentioned in \appxref{func-var-tsres}, since the von Weizs\"acker (vW) KEDF~\citep{weizsacker1935theorie} is a lower bound of the true KEDF~\citep[Thm.~1.1]{lieb1983density}, taking it as the base KEDF could inform the $T_\res$ model to be non-negative, but unfortunately introduces more training challenges. Hence it remains to be explored to leverage the lower-bound property of the vW KEDF.
Nevertheless, we can empirically verify that our learned KEDF models already satisfy this lower bound.
For this, we present a scatter plot in \appxfigref{kinetic_eng}, where each point represents the vW KEDF value (x-axis coordinate) and the kinetic energy value by our learned KEDF model (y-axis coordinate) of each electron density.
The densities for this evaluation are taken as the ground-state densities optimized by our KEDF model on unseen ethanol test structures (\appxfigref{kinetic_eng}(a-b)) or unseen QM9 test structures (\appxfigref{kinetic_eng}(c-d)), following the setting in Results~\ref{sec:res-in-scale}.
Our KEDF model takes either the residual KEDF version (\appxref{func-var-tsres}) with APBE base KEDF (\appxfigref{kinetic_eng}(a,c)), or the TXC functional version (\appxref{func-var-tsexc}) which gives the kinetic energy value by subtracting the $E_\XC$ value from the model-predicted value $E_\TXC$ (\appxfigref{kinetic_eng}(b,d)).
The figure clearly shows that in all cases, the kinetic energy values by our KEDF model are larger than the corresponding vW values, hence our learned KEDF models satisfy this lower bound property.

\subsubsection{Visualization of Optimized Densities} \label{appx:visual-den}

\begin{figure}[h]
  \centering
  \begin{subfigure}[b]{0.49\textwidth}
      \centering
      \includegraphics[width=\textwidth]{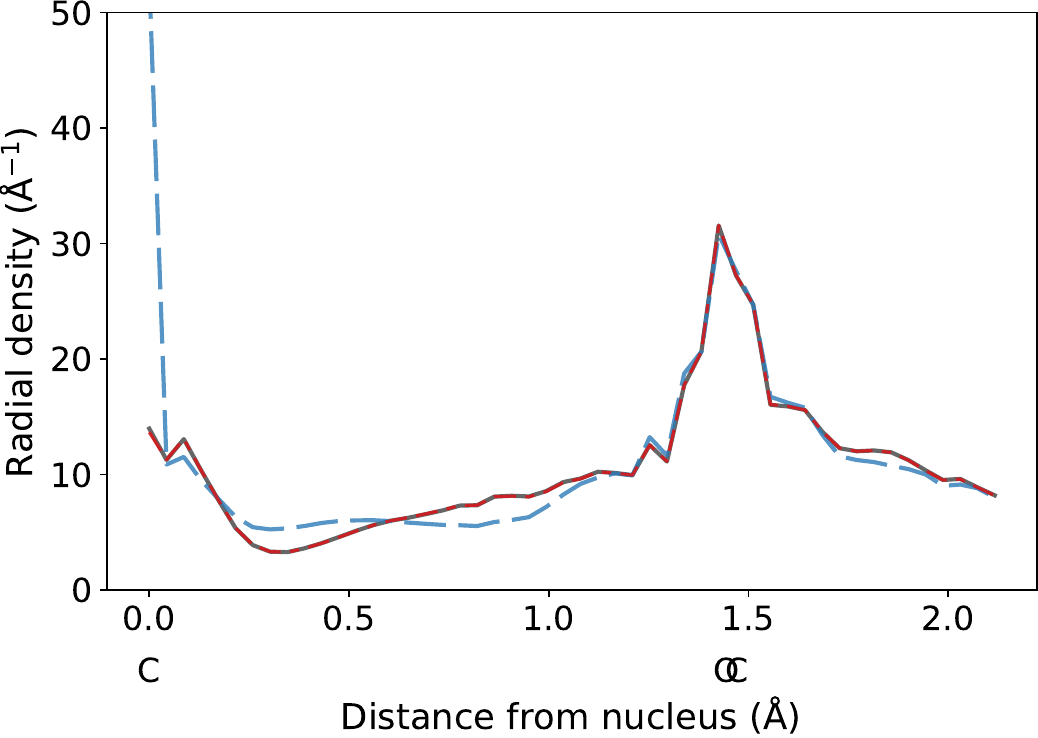}
      \caption{$\alpha$-carbon}
      \label{fig:optimized-density-center-C}
  \end{subfigure}
  \hfill
  \begin{subfigure}[b]{0.49\textwidth}
      \centering
      \includegraphics[width=\textwidth]{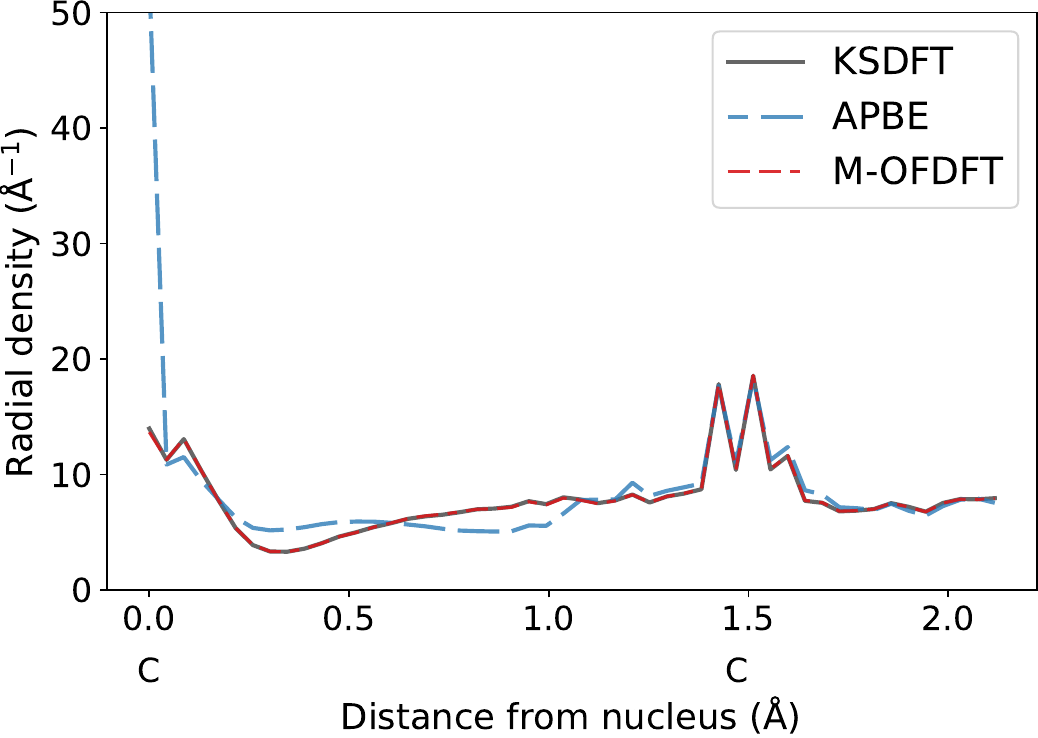}
      \caption{$\beta$-carbon}
      \label{fig:optimized-density-side-c}
  \end{subfigure}
  \caption{\textbf{Additional visualization of the density optimized by various methods.} Integrated density on spheres of varying radii around each of the two carbon atoms in an ethanol structure is plotted in each figure, in parallel with \figref{res-in-scale}(b).
  }
  \label{fig:extra-optimized-density}
\end{figure}

\paragraph{Radial density plot}
For more qualitative investigation of \ourmethod, we provide more visualization results of the optimized density by \ourmethod, in addition to \figref{res-in-scale}(b).
\appxfigref{extra-optimized-density} shows the radial density by spherical integral around the other two heavy atoms in the ethanol molecule, \ie, the $\alpha$-carbon atom (the one connected to the oxygen atom) and the $\beta$-carbon atom.
The results demonstrate that the density curves of \ourmethod again align almost identically with the KSDFT density curves.
In contrast, the classical KEDF using APBE for KEDF again leads to a substantial deviation around bonding regions between peaks of core electrons.

\begin{figure}[h]
    \centering
    \includegraphics[width=0.6\textwidth]{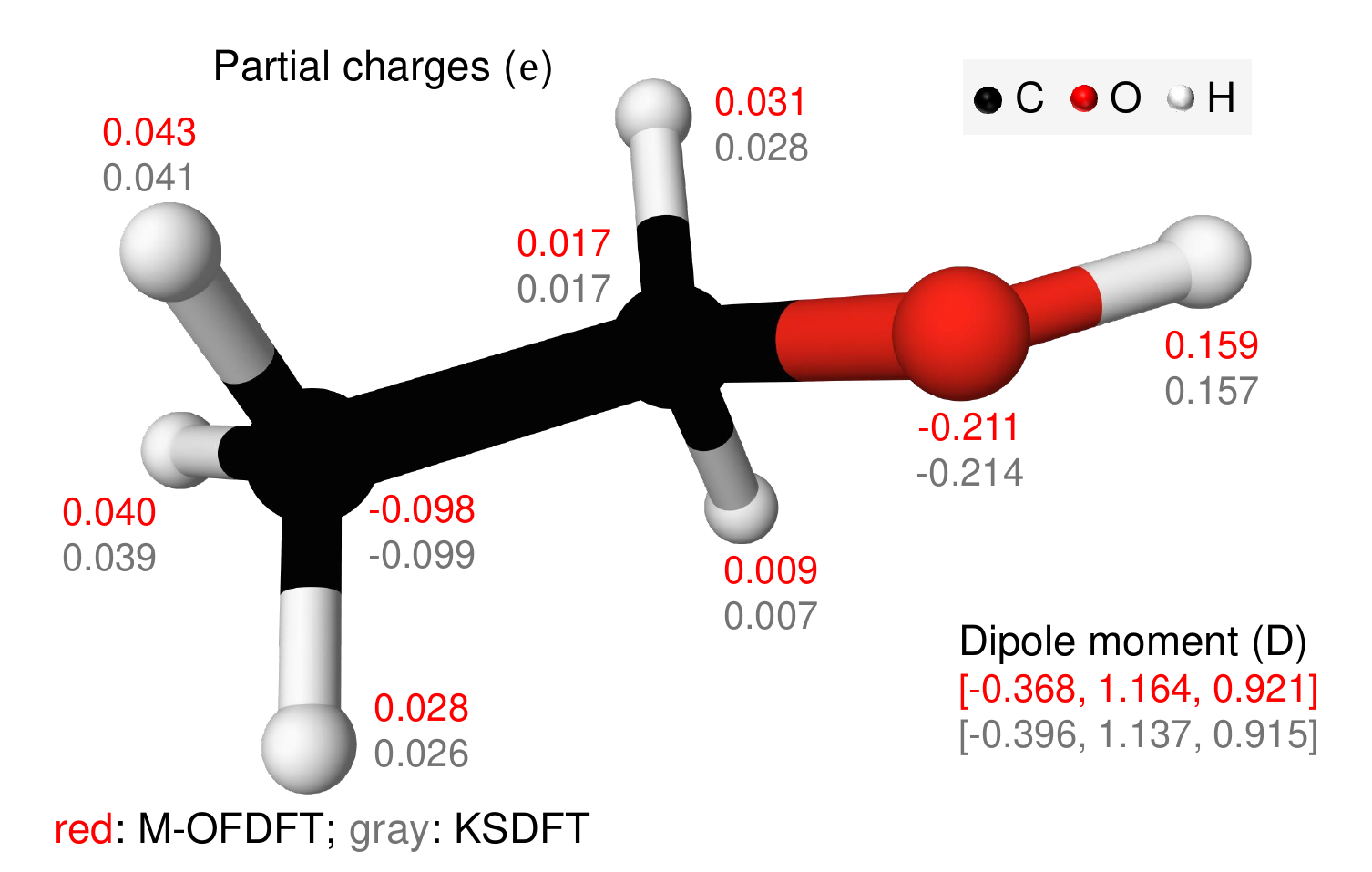}
    \caption{\textbf{Visualization of Hirshfeld atomic partial charge on each atom in an ethanol structure as well as the dipole moment of the structure.} Both the atomic partial charges and the dipole moment derived from the solved electron density by \ourmethod align closely with those solved by KSDFT.}
    \label{fig:par_charge}
\end{figure}

\paragraph{Partial charge and dipole moment visualization}
As reported in Results~\ref{sec:res-in-scale}, the optimized density of \ourmethod can accurately reproduce Hirshfeld partial charges~\citep{hirshfeld1977} and dipole moments of molecules. To further illustrate the results, we provide a representative example in \appxfigref{par_charge} of atomic partial charge on each atom in an unseen test ethanol structure as well as the dipole moment of the structure, based on the optimized density by \ourmethod. The results show that both the Hirshfeld partial charges for each atom and the dipole moment from \ourmethod are in close agreement with those obtained from KSDFT in the ethanol molecule.

\subsection{Extrapolation Results} \label{appx:res-out-scale}

\subsubsection{Results on QMugs and Chignolin} \label{appx:res-out-scale-main}

\paragraph{Validation error in the extrapolation experiments}
As shown in Results~\ref{sec:res-larger-scale}, \ourmethod has shown a superior extrapolation performance in energy than end-to-end deep learning counterparts \mlff and \mlffden that predict the total energy directly.
We emphasize here that this is not due to that \mlff and \mlffden are not well trained or overfit to the training dataset.
\appxtblref{e2e-out-scale} shows that \mlff and \mlffden achieve a better validation error than \ourmethod, but their extrapolation errors increase much more vastly than that of \ourmethod.

\begin{table}[h]
  \centering
  \caption{\textbf{Training, validation and extrapolation error (per-atom energy MAE in $\mathrm{kcal/mol}$) of our \ourmethod and reference methods \mlff and \mlffden on the QMugs and chignolin extrapolation setups.} For the QMugs case, all the methods are trained and validated on two disjoint splits of QM9 molecules and QMugs molecules with no more than 15 heavy atoms, and are tested for extrapolation on QMugs molecules ($n$=50) with 56-60 heavy atoms.
  For the chignolin case, all the methods are trained and validated on two disjoint splits of fragment structures of all peptide lengths (2-5), and are tested for extrapolation on chignolin structures ($n$=1,000).
  Note that for \ourmethod, the training and validation energy errors are from end-to-end model output, while the extrapolation error is from the result after density optimization using the model, consistent with \figref{res-outscale}.}
  \begin{tabular}{lcccccc}
    \toprule
    \multirow{2}{*}{Method} & \multicolumn{3}{c}{QMugs} & \multicolumn{3}{c}{Chignolin} \\
    \cmidrule(lr){2-4} \cmidrule(lr){5-7}
    & Train & Validation & Extrapolation & Train & Validation & Extrapolation \\
    \midrule
    \mlff    & 0.008 & \textbf{0.017} & 1.768 & 0.003 & 0.003 & 0.084 \\
    \mlffden & \textbf{0.004} & 0.019 & 1.824 & \textbf{0.002} & \textbf{0.002} & 0.079 \\
    \ourmethod & 0.044 & 0.055 & \textbf{0.112} & 0.015 & 0.016 & \textbf{0.071} \\
    \bottomrule
  \end{tabular}
  \label{tbl:e2e-out-scale}
\end{table}

\begin{figure}[h]
  \centering
  \begin{subfigure}[b]{0.49\textwidth}
      \centering
      \includegraphics[width=\textwidth]{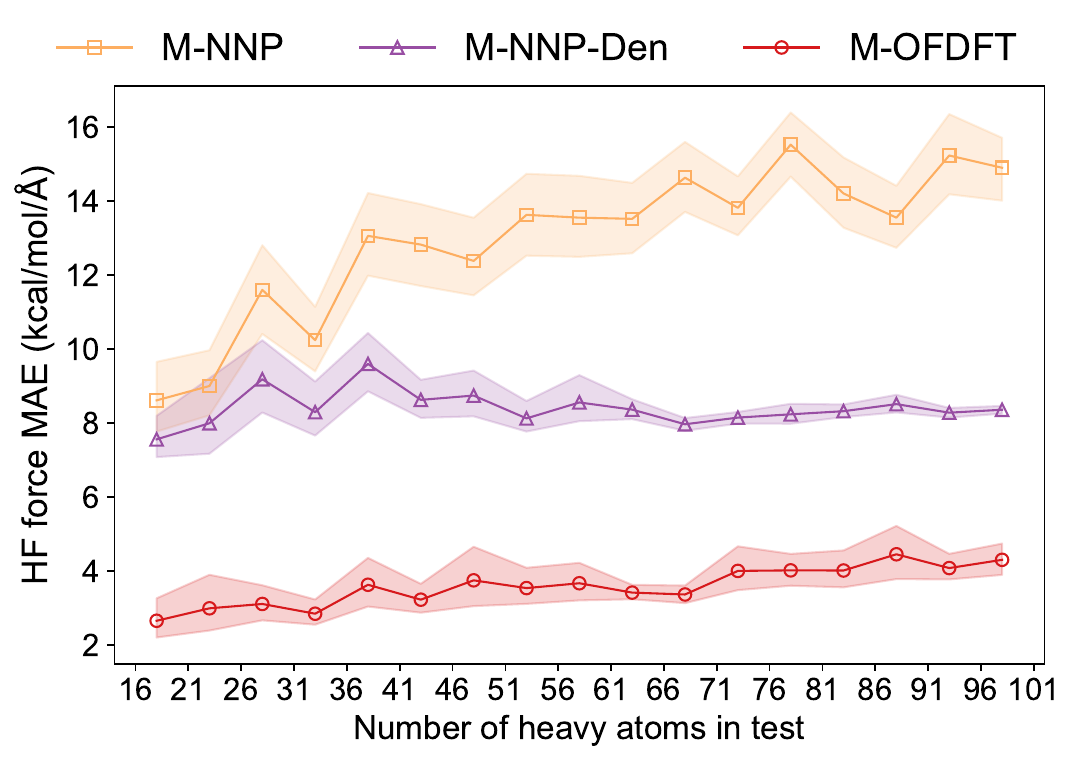}
      \caption{QMugs (fixed training dataset, in parallel with \figref{res-outscale}(a))}
      \label{fig:appx-force-qmugs}
  \end{subfigure}
  \hfill
  \begin{subfigure}[b]{0.49\textwidth}
      \centering
      \includegraphics[width=\textwidth]{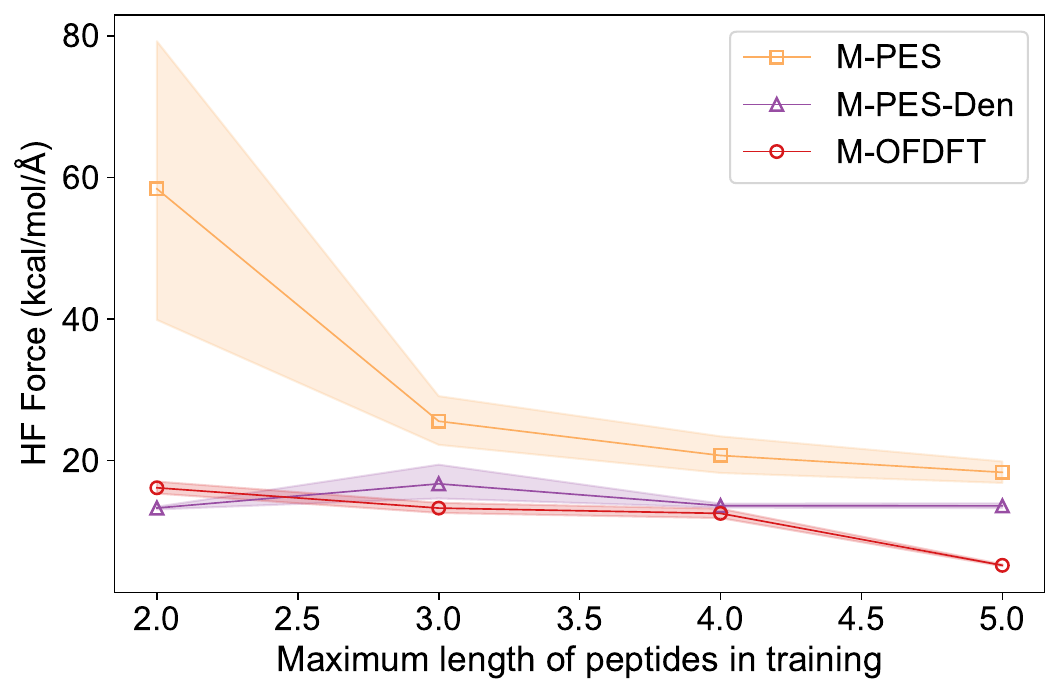}
      \caption{Chignolin (fixed test dataset, in parallel with \figref{res-outscale}(d))}
      \label{fig:appx-force-chig}
  \end{subfigure}
  \vfill
  \begin{subfigure}[b]{0.98\textwidth}
      \centering
      \includegraphics[width=\textwidth]{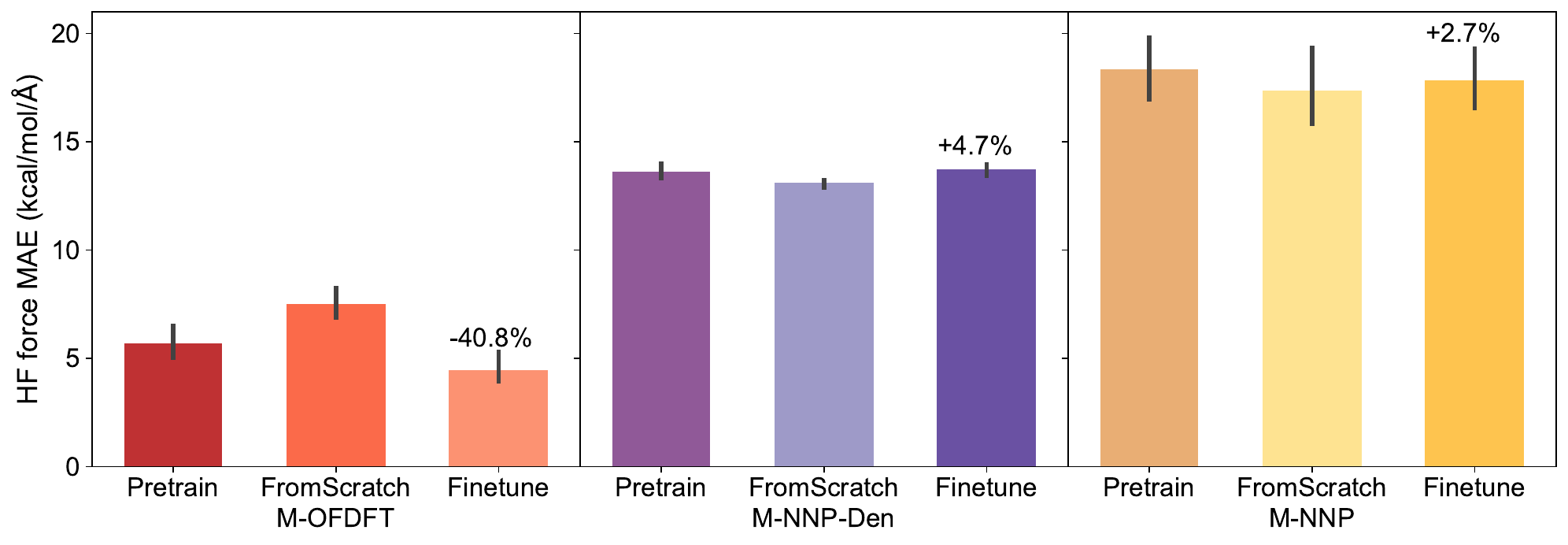}
      \caption{Chignolin (finetuning setting, in parallel with \figref{res-outscale}(e))}
      \label{fig:appx-force-chig-ft}
  \end{subfigure}
  \caption{\textbf{Extrapolation performance in HF force of M-OFDFT compared with \mlff and \mlffden on the two extrapolation settings.}
  The data are mean and the shades as well as error bars in all figures show~95\% confidence intervals.
  \textbf{(a)}~HF force MAE on increasingly larger molecules from the QMugs dataset, using models trained on molecules with no more than 15 heavy atoms from QM9 and QMugs datasets, in parallel with \figref{res-outscale}(a). Each value is calculated on 50 QMugs molecules.
  \textbf{(b)}~HF force MAE on chignolin structures ($n$=1,000), using models trained on a series of datasets including increasingly longer peptides, in parallel with \figref{res-outscale}(d).
  \textbf{(c)}~HF force MAE on chignolin structures ($n$=50), using models trained on all peptides without (`Pretrain') and with (`Finetune') fine-tuning on 500 chignolin structures, in parallel with \figref{res-outscale}(e). Also marked are error reduction ratios by the finetuned models over models trained from scratch (‘FromScratch’) on the 500 chignolin structures only.
  }
  \label{fig:appx-force}
\end{figure}

\paragraph{Extrapolation performance in HF force}
We then show the better extrapolation performance of \ourmethod in HF force. As shown in \appxfigref{appx-force}(a), \ourmethod with functional model trained on molecules with fewer than 15 heavy atoms consistently maintains a substantially lower HF force MAE than \mlff and \mlffden on larger-scale molecules.
\ourmethod also outperforms the two end-to-end counterparts across various fragment datasets in the chignolin experiment (\appxfigref{appx-force}(b)).

In the finetuning setting of chignolin, \appxfigref{appx-force}(c) shows that \ourmethod always achieves much better HF force results than \mlff and \mlffden in each of the three setups.
\ourmethod attains a substantial 40.8\% HF force error reduction in finetuning over training from scratch, showing the ability of capturing transferable knowledge from accessible-scale data, and more efficiently leveraging limited large-scale data.
In contrast, the end-to-end deep learning methods exhibit even worse HF force results when incorporating accessible-scale data, manifesting an extrapolation difficulty to a larger scale.

\paragraph{Extrapolation comparison to NNP with other architectures}
Results~\ref{sec:res-larger-scale} have demonstrated the qualitatively better extrapolation performance of \ourmethod than direct energy prediction, \ie, the neural network potential (NNP) formulation, using the same model architecture as \ourmethod, denoted as \mlff.
We further verify that this conclusion still holds even using more advanced and recent architectures for NNP. We consider Equivariant Transformer (ET)~\citep{tholke2021equivariant} and Equiformer~\citep{liao2023equiformer}, which are recently proposed NNP architectures that have shown remarkable performance and competitiveness in the field. Notably, ET is one of state-of-the-art equivariant NNP architectures that use Cartesian vector features to maintain $\mathrm{SE}(3)$-equivariance, and Equiformer is a cutting-edge approach amongst NNP architectures that leverage high-order spherical harmonics tensors to encode molecular features.

\begin{figure}
  \centering
  \includegraphics[width=0.8\textwidth]{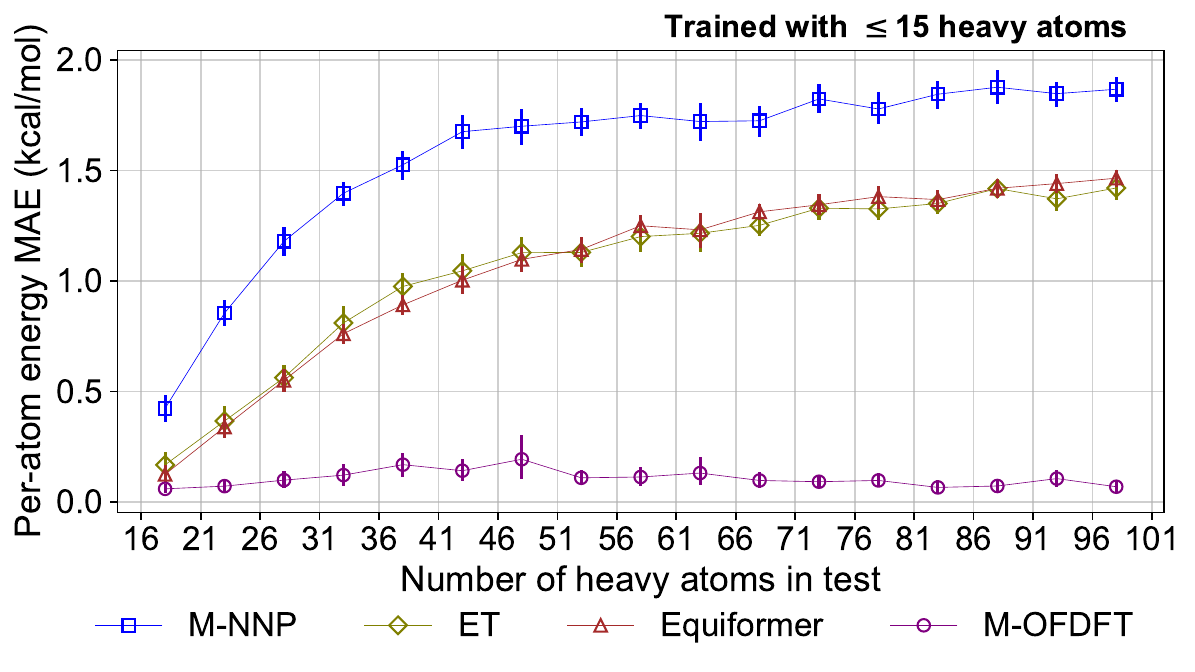}
  \caption{\textbf{Extrapolation performance of \ourmethod compared to other advanced deep learning architectures.} Each line denotes the mean absolute error (MAE) in per-atom energy on increasingly larger molecules from the QMugs dataset, using a model trained on molecules with no more than 15 heavy atoms from QM9 and QMugs datasets. Each value is calculated on 50 QMugs molecules and the bars show 95\% confidence intervals.
  The setting is in parallel with \figref{res-outscale}(a). Beyond \mlff, \ie, the NNP using the same model architecture as \ourmethod and already presented in \figref{res-outscale}(a), two more architectures for NNP are investigated: Equivariant Transformer (ET)~\citep{tholke2021equivariant} and Equiformer~\citep{liao2023equiformer}.}
  \label{fig:extra_mlff}
\end{figure}

Following the setting of~\figref{res-outscale}(a) introduced in Results~\ref{sec:res-larger-scale}, we train the NNP models on QM9 and QMugs molecules with no more than 15 heavy atoms, and test them on increasingly larger molecules from the QMugs dataset.
Results in \appxfigref{extra_mlff} demonstrate that although using the more advanced architectures improves the performance over \mlff, the error in the extrapolation cases is still larger than that of \ourmethod, and the error still increases with molecule size, while the error of \ourmethod keeps constant or even decreasing.

\paragraph{Evaluation of electron density in the chignolin experiment}
For a thorough assessment of the extrapolation capability of \ourmethod, we evaluate the electron density solved by \ourmethod on peptide structures in various lengths.
As peptides are relatively large molecules, it is inconvenient to directly visualize the densities. We hence calculate the Hirshfeld partial charge~\citep{hirshfeld1977} and dipole moment from the solved density.

This evaluation is conducted in parallel with the setting in Results~\ref{sec:res-larger-scale}. To construct the evaluation benchmark, we prepare a test set encompassing a range of short-peptide structures, from dipeptides to pentapeptides, as well as chignolin structures (of length 10). We sample 50 structures for each category of peptides. More details about the peptide structures are described in Methods~\ref{sec:data-chig}.
For solving test peptide structures of lengths 2 to 5, we apply the total energy functional model $E_{\tot,\theta}$ trained on all training peptide structures (of lengths 2-5), following the same setting as \figref{res-outscale}(c). %
For solving test chignolin structures, we further finetune the model on 800 training chignolin structures.
We take KSDFT results to evaluate error, and compare the results with the classical OFDFT using the TF+$\frac19$vW KEDF.
Results in \appxtblref{pep_dipole} demonstrate that \ourmethod consistently outperforms the classical OFDFT over peptides of all lengths, in terms of the accuracy on these density-related quantities. The partial charge MAE of \ourmethod is substantially lower, by two orders of magnitude. This substantial improvement further underscores the power of \ourmethod.

\begin{table}[h]
  \centering
  \caption{\textbf{Hirshfeld partial charge and dipole moment results in mean absolute error (MAE) from KSDFT.} The units for partial charge and dipole moment are $\mathrm{e}$ and $\mathrm{D}$, respectively. Each dataset contains 50 structures.}
  \begin{tabular}{lccc}
    \toprule
    Test dataset & Quantity & \ourmethod & TF+$\frac19$vW \\
    \midrule
    \multirow{2}{*}{Dipeptide}   & Partial charges & \textbf{2.62$\times 10^{-3}$} & 0.147 \\
        & Dipole moment & \textbf{0.217} & 2.970 \\
    \midrule
    \multirow{2}{*}{Tripeptide} & Partial charges & \textbf{2.68$\times 10^{-3}$} & 0.153 \\
    & Dipole moment & \textbf{0.283} & 3.556 \\
    \midrule
    \multirow{2}{*}{Tetrapeptide} & Partial charges & \textbf{2.68$\times 10^{-3}$} &  0.139 \\
    & Dipole moment & \textbf{0.390} & 3.420 \\
    \midrule
    \multirow{2}{*}{Pentapeptide} & Partial charges & \textbf{2.85$\times 10^{-3}$} & 0.141 \\
    & Dipole moment & \textbf{0.543} & 3.474 \\
    \midrule
    \multirow{2}{*}{Chignolin} & Partial charges & \textbf{3.32$\times 10^{-3}$} & 0.132 \\
    & Dipole moment & \textbf{1.077} & 12.049 \\
    \bottomrule
  \end{tabular}
  \label{tbl:pep_dipole}
\end{table}

\subsubsection{Additional Extrapolation Study from QM9} \label{appx:res-out-scale-qm9}

\paragraph{Results on charged molecules}
As clarified in \appxref{model-spec}, the input atom types $\bfZ$ and positions $\bfX$ are only used to inform the KEDF model of the types and centers of the atomic basis functions under which the expansion coefficient input is defined, but not for the physics of the actual atoms. This formulation makes \ourmethod inherently general for handling input densities from either neutral or charged molecular systems, just by feeding the KEDF model with expansion coefficients of the corresponding electron density onto the atomic basis functions whose types and positions are specified by $\bfZ$ and $\bfX$, even if the KEDF model is trained on data only from neutral molecules.

Here, we demonstrate the efficacy of \ourmethod in handling charged molecules.
To construct an evaluation benchmark, we randomly select five unseen carboxylic acid molecules from the QM9 test set, and deprotonate the hydrogen cation from the carboxyl group (\ch{-C(O)OH}) of each molecule, thereby generating five carboxylate anions (\ch{-C(O)O-}).
We employ the TXC functional model $E_{\TXC,\theta}$ trained on the QM9 training set, which comprises neutral molecules only, to solve these charged systems.
We initialize the electron density using ProjMINAO (Methods~\ref{sec:stage2}), which projects the electron density from the MINAO initialization onto the training-data manifold by a deep learning model. Since MINAO gives the initial electron density by treating the system as neutral, we rescale the density coefficients produced by ProjMINAO to normalize to the correct number of electrons of the charged system. Note that in the density optimization process (\eqnref{den-gd}), the number of electrons is kept throughout, so the ground-state density solution also respects the correct charge of the system.
We evaluate the performance of \ourmethod in terms of the mean absolute error (MAE) from KSDFT results over the five systems in the energy difference between the neutral and the corresponding charged system.

The result is that \ourmethod achieves a 3.80$\,\mathrm{kcal/mol}$ MAE of energy difference. In comparison, the result of the classical OFDFT using the TF+$\frac19$vW KEDF is 30692.16$\,\mathrm{kcal/mol}$, an error of five orders larger.
This result indicates that \ourmethod trained on neutral systems is still effective in handling charged molecules, an extrapolation capability of a new kind.
The capability for charged molecules can be further improved if data from charged molecules are included.

\paragraph{Results on unseen chemical environments}
Extrapolating a trained machine-learning model to unseen chemical environments, \eg, bond types not present in training data, is another challenging extrapolation evaluation.
To investigate this type of extrapolation capability of \ourmethod, we apply \ourmethod to the carbon monoxide molecule $\mathrm{CO}$, which contains a triple bond \ch{C+O} that is not encountered in training the KEDF model.
The initial $\mathrm{CO}$ structure is generated using the RDKit software~\citep{rdkit} which gives the bond length of 1.118$\,\angstrom$. We then augment four additional $\mathrm{CO}$ structures by adjusting bond lengths to 1.102$\,\angstrom$, 1.112$\,\angstrom$, 1.122$\,\angstrom$ and 1.132$\,\angstrom$, containing both squeezed and stretched bond lengths.
The residual KEDF $T_{\tnS, \theta}$ model trained on QM9 training set is used to solve these systems.
The H\"uckel method is chosen for density initialization, which exhibits better robustness to various bond lengths than the ProjMINAO initialization in our trials. %

We evaluate the results in mean absolute error (MAE) with respect to KSDFT results.
The Hirshfeld partial charge MAE and dipole moment MAE of optimized densities by \ourmethod over the five $\mathrm{CO}$ structures are 0.102$\,\mathrm{e}$ and 0.150$\,\mathrm{D}$, respectively.
As a reference, these MAE numbers are 0.296$\,\mathrm{e}$ and 0.496$\,\mathrm{D}$ when using the classical OFDFT with the TF+$\frac19$vW KEDF.
Hence \ourmethod still achieves a substantial improvement over classical OFDFT even in this extrapolation scenario.

It should be noted that neither charged molecular systems nor the triple bond \ch{C+O} has been encountered in our training data, thus they are indeed challenging extrapolation tasks.
While the formulation of \ourmethod is designed to be universally applicable to all densities and molecular systems, its performance as a neural network model is hard to completely avoid %
extrapolation error. This is reflected in the (absolute) energy MAE of \ourmethod on charged QM9 molecules, which stands at 3.79$\,\mathrm{kcal/mol}$, higher than 0.73$\,\mathrm{kcal/mol}$ on neutral molecules from the same dataset. Despite this, the extrapolation performance of \ourmethod is still reasonable, and is still substantially better than classical OFDFT methods, showcasing the potential of \ourmethod in these more challenging scenarios. The performance of \ourmethod on these systems can be further improved by enriching the train data with charged molecules and new bond patterns such as triple bond \ch{C+O}, which will be investigated in future work.

\subsection{Empirical Time Cost Results}
\label{appx:scala-exp}

\paragraph{Settings}
As mentioned in Results~\ref{sec:res-time}, we aim to demonstrate the lower time complexity of \ourmethod compared to KSDFT by empirically benchmarking the two methods on two types of large molecular systems. For KSDFT calculations on all systems, we %
adopt the same settings as introduced in Methods~\ref{sec:basis-setup}. \ourmethod calculations are benchmarked on a 32-core CPU server with 216 GiB memory, which is equipped with one Nvidia A100 GPU with 80 GiB memory.
\figref{res-time} is produced by running \ourmethod and KSDFT on QMugs molecules.
For QMugs molecules, since they are generally much larger than ethanols and QM9 molecules to the extent that the residual KEDF $T_{\res,\theta}$ formulation becomes substantially costly due to grid-based computation for $T_\base$ and $E_\XC$, we apply the learned TXC functional model $E_{\TXC,\theta}$ in the same setting as is used to produce \figref{res-outscale}(a).

\begin{figure}
  \centering
  \includegraphics[width=1.0\textwidth]{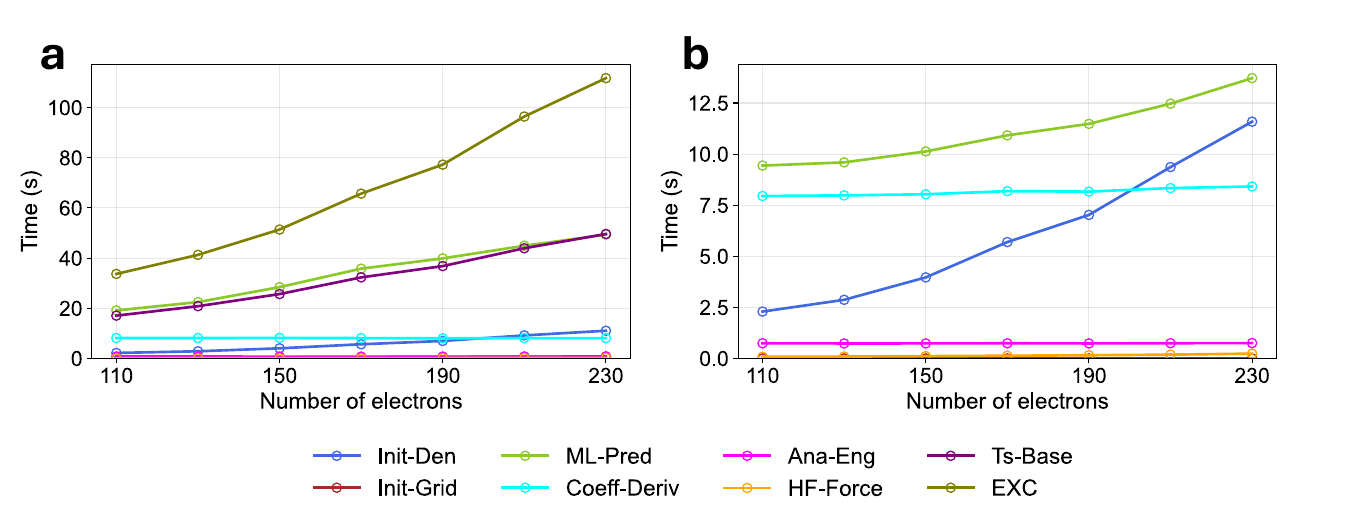}
  \caption{Empirical time cost of various computational components in the density optimization process of \ourmethod, under \textbf{(a)}~the residual KEDF $T_{\res,\theta}$ formulation and \textbf{(b)}~the TXC functional $E_{\TXC,\theta}$ formulation.
  Computational components are defined in the following.
  ``Init-Den'': initialization of density (including density fitting);
  ``Init-Grid'' (only for the $T_{\res,\theta}$ formulation): generation of grid points and evaluation of basis function values on them;
  ``ML-Pred'': evaluation of the deep learning model, $T_{\res,\theta}$ or $E_{\TXC,\theta}$, including the local frame module (\appxref{local-frame}) and enhancement modules (\appxref{tech-effi-train});
  ``Coeff-Deriv'': automatic differentiation to compute the gradient of the deep learning model with respect to input density coefficients;
  ``Ana-Eng'': computation of the values and gradients with respect to density coefficients of energy terms that have analytical expressions, \ie, the Hartree energy $E_\tnH$ (\eqnref{EH-den-coeff}) and the external potential energy $E_\ext$ (\eqnref{Eext-den-coeff});
  ``HF-Force'': Hellmann-Feynman force computation (\appxref{hfforce}) conducted after density optimization;
  ``Ts-Base'' (only for the $T_{\res,\theta}$ formulation): evaluation of the value and gradients with respect to density coefficients of the base KEDF on the grid;
  ``EXC'' (only for the $T_{\res,\theta}$ formulation): evaluation of the value and gradients with respect to density coefficients of the XC functional on the grid.
  }
  \label{fig:tsres_tsexc_time_compare}
\end{figure}

\paragraph{Time cost of each computational component}
To better understand the structure of the time cost in the density optimization process (\ie, the process to use \ourmethod to solve a queried molecular system), we split the time cost into various computational components in \ourmethod.
Both the residual KEDF $T_{\res,\theta}$ formulation (\appxref{func-var-tsres}) and the TXC functional $E_{\TXC,\theta}$ formulation (\appxref{func-var-tsexc}) of \ourmethod are considered.
As shown in \appxfigref{tsres_tsexc_time_compare}(a), in the $T_{\res,\theta}$ formulation, the three major parts of the time cost are the evaluation of the XC functional (denoted as ``EXC''), the evaluation of the base KEDF (denoted as ``Ts-Base''), and the evaluation of the $T_{\res,\theta}$ model (denoted as ``ML-Pred'').
Noting that the first two components are evaluated on grid, we conclude that grid-based computation is the main restriction to running \ourmethod on large molecules, conforming to \appxref{func-var-tsexc}, hence the TXC functional $E_{\TXC,\theta}$ formulation is motivated, which does not require any grid-based computation. %
We also note that grid-based computation also occupies a substantial amount of GPU memory (\appxref{func-var-tsexc}). Using the hardware specified above, we can only afford systems of up to 230 electrons under the $T_{\res,\theta}$ formulation, which is where the plot ends.
To compare the component-wise time cost of the $T_{\res,\theta}$ formulation and the $E_{\TXC,\theta}$ formulation, we conduct the same analysis with the $E_{\TXC,\theta}$ KEDF model.
As shown in \appxfigref{tsres_tsexc_time_compare}(b), due to the removal of grid-based computations, the total running time is substantially reduced. %
We note that in this case, the evaluation of the $E_{\TXC,\theta}$ model takes the largest computational cost.
This part has an $O(N^2)$ computational complexity due to the need for non-local calculation.
As the molecular size increases, this could lead to considerable computational demands. %
Despite the importance of the non-local calculation (\appxref{ablat-nonlocal}), its influence presumably does not extend infinitely, thus allowing us to reduce the complexity by using a distance cutoff for large molecular systems. 
Specifically, with a distance cutoff $r_{c}$, the Transformer-based model, Graphormer, can be modified to capture non-local interactions between one atom and its neighboring atoms within the cutoff. The complexity is then $O(A A_\mathrm{rc})$, where $A_\mathrm{rc}$ is the average number of neighboring atoms within the distance cutoff $r_{c}$. As $r_{c}$ is taken as a constant, $A_\mathrm{rc}$ is a constant. Hence the modification reduces the complexity of the model to linear: $O(A A_\mathrm{rc}) = O(A) = O(N)$. 
We also note that analogous approaches to trim the neighborhood based on distance cutoffs have been utilized to achieve linear cost scaling %
using the Transformer architecture in the realm of machine learning~\citep{beltagy2020longformer,yang2021focal},
which pave the way for further improvement of \ourmethod.

\paragraph{Details on the time evaluation on proteins}
To further highlight the scaling advantage of \ourmethod, we evaluate it on two large-scale protein molecular systems containing 709 and 738 atoms. The all-atom conformations of these systems are obtained from \citet{lindorff2011fast} and neutralized following the same pipeline for processing chignolin structures (detailed in Methods~\ref{sec:data-chig}). Such a scale is already seldom encountered in the context of KSDFT calculations.
Considering the substantial scale difference between the protein systems evaluated and the molecules used during training, we employ the total energy functional model $E_{\tot,\theta}(\bfpp,\clM)$ trained on all chignolin fragments and finetuned on 800 chignolin structures. This model is expected to provide the best extrapolation capacity within our available resources.

The results indicate that \ourmethod achieves a 25.6-fold and 27.4-fold speedup over KSDFT on the two protein systems, respectively.
The per-atom energy MAE of \ourmethod on the two test systems is 0.23$\,\mathrm{kcal/mol}$ and 0.31$\,\mathrm{kcal/mol}$, respectively. While higher than the error on the chignolin case, the result still demonstrates a substantial advantage over \mlff, which gives per-atom energy MAE of 0.36$\,\mathrm{kcal/mol}$ and 0.63$\,\mathrm{kcal/mol}$, respectively.

\subsection{Ablation Study} \label{appx:exp-ablation}

\subsubsection{Multi-Step Data and Gradient Label} \label{appx:abl-mult-grad}
To capture the energy landscape in the density coefficient space, we generate multiple $\bfpp$ samples for each molecular structure $\clM$ and compute the gradient label $\nabla_{\bfpp}T_\tnS$ for additional supervision information. An ablation experiment is conducted on the ethanol dataset to investigate the importance of each type of supervision by excluding the supervision label during the training process. The ablation results are presented in \appxtblref{res-ablation-grad}. We observe that removing the gradient label $\nabla_{\bfpp}T_\tnS$ leads to a considerable decrease in energy and HF force accuracy for both density initialization strategies, emphasizing the importance of maintaining the optimization of \ourmethod on a physical track. Incorporating multi-step $\bfpp$ samples is also crucial for enhancing the performance of \ourmethod, particularly for the ProjMINAO initialization, as the accuracy of the density corrector model depends on the size of training densities.

\begin{table}[h]
  \centering
  \caption{\textbf{Ablation study for various data augmentation strategies.} All results are evaluated on test ethanol molecules with the learned $T_{\res, \theta}$ model. The MAEs in energy and HF force are listed in $\mathrm{kcal/mol}$ and $\mathrm{kcal/mol/\angstrom}$, respectively.The results are calculated on ethanol test structures ($n$=10,000).}
  \begin{tabular}{ccccc}
    \toprule
    Multi-step $\bfpp$            &  Gradient $\nabla_{\bfpp}T_\tnS$      & Density Initialization & Energy & HF Force\\
    \midrule
    \multirow{2}{*}{\checkmark} & \multirow{2}{*}{\checkmark} & H\"uckel  & 5.27   & 27.73 \\
                                &                             & ProjMINAO & \textbf{0.18}   & \textbf{1.18} \\
    \midrule
    \multirow{2}{*}{\checkmark}           & \multirow{2}{*}{} & H\"uckel  & 166.84 & 144.78\\
                                &                             & ProjMINAO & 170.83 & 143.67 \\
    \midrule
    \multirow{2}{*}{} & \multirow{2}{*}{\checkmark}           & H\"uckel  & 6.43   & 32.83 \\
                                &                             & ProjMINAO & 20.61  & 38.56 \\
    \bottomrule
  \end{tabular}
  \label{tbl:res-ablation-grad}
\end{table}

\subsubsection{Non-locality} \label{appx:ablat-nonlocal}
\begin{figure}[h]
  \centering
  \begin{subfigure}[b]{0.49\textwidth}
      \centering
      \includegraphics[width=\textwidth]{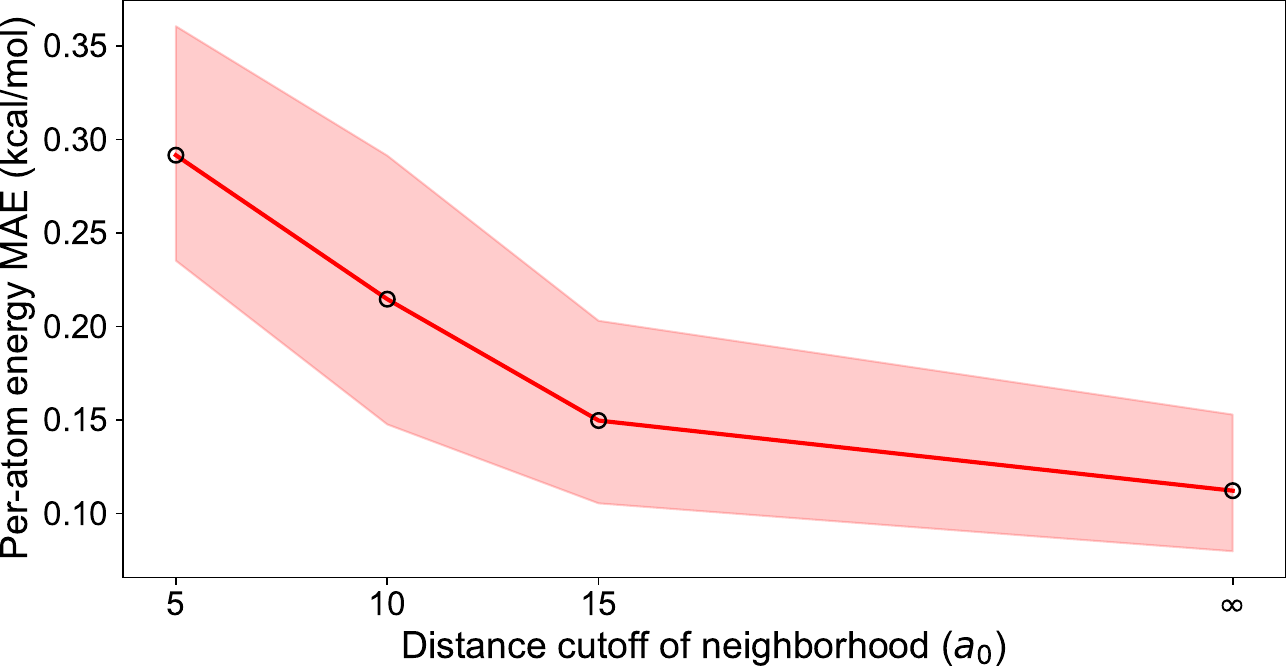}
      \caption{Energy results}
      \label{fig:qmugs-nonlocal-eng}
  \end{subfigure}
  \hfill
  \begin{subfigure}[b]{0.49\textwidth}
      \centering
      \includegraphics[width=\textwidth]{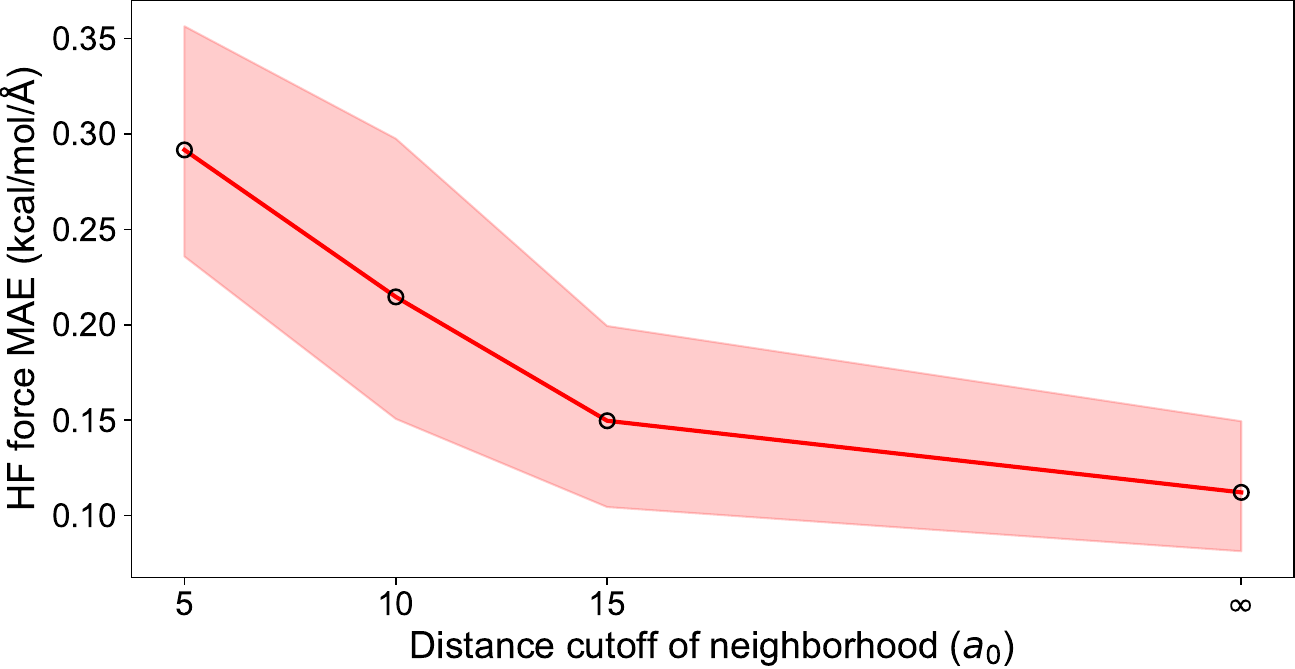}
      \caption{HF force results}
      \label{fig:qmugs-nonlocal-force}
  \end{subfigure}
  \caption{\textbf{Ablation study results for the non-locality of the functional model architecture.} The shades show~95\% confidence intervals.
  \textbf{(a)} and \textbf{(b)} show the energy error and HF force error of \ourmethod on QMugs molecules ($n$=50) with 56-60 heavy atoms, for which the model is trained on QM9 and QMugs molecules of no more than 15 heavy atoms.
  The distance cutoff $\infty$ represents a fully-connected graph (non-local model). The experiment is conducted on the extrapolation setting in parallel with \appxtblref{e2e-out-scale}.
  }
  \label{fig:res-qmugs-nonlocal}
\end{figure}

Considering the non-local dependency of KEDF on electron density, it is essential to incorporate non-local calculations in OFDFT, which motivates us to use Graphormer as our backbone architecture. To demonstrate the importance of non-locality, we conduct an ablation study by gradually decreasing the receptive field (\ie, the distance cutoff of atom neighborhood) of the functional model (specifically, the TXC model (\appxref{func-var-tsexc})) of \ourmethod. The experiment follows the extrapolation setting in parallel with \appxtblref{e2e-out-scale}, \ie, the functional model is trained on molecules with no more than 15 heavy atoms and the resulting \ourmethod is tested on 50 QMugs molecular structures with 56-60 heavy atoms. As illustrated in \appxfigref{res-qmugs-nonlocal}, both the energy and HF force prediction error tend to worsen as the distance cutoff decreases, providing empirical evidence in our setting that a non-local model is indeed indispensable for well approximating a functional involving the kinetic energy density functional (KEDF).

\subsubsection{Local Frame and Enhancement Modules} \label{appx:ablat-den-prep}

As discussed in Methods~\ref{sec:local-frame}, the input coefficients are tensors equivariant to rotation, but the output energy is a scalar hence invariant, which requires the model to have such invariance built-in. In addition, the vast range of gradient data makes optimizing the KEDF model challenging.
A CoefficientAdapter module is proposed to guarantee the geometric invariance of the model and express the vast gradient range. We conducted an ablation experiment to study the importance of each component in this module. As shown in \appxtblref{res-ablation-den}, utilizing the local frame results in a considerable performance improvement compared to the baseline model using the standard global frame (\eg, derived by PCA on the atom coordinates), demonstrating the effectiveness of the local frame in reducing the geometric variability of input data.

Furthermore, since different basis functions have varying importance in representing a density, some coefficient dimensions can substantially influence the density function and the electronic energy. This is difficult for \ourmethod to handle, as the model treats each dimension equally. During density optimization, this influence is amplified when the input density is far from the ground state. We address this issue by introducing the natural reparameterization technique to balance the impact of each coefficient dimension. The results in \appxtblref{res-ablation-den} show that implementing natural reparameterization can substantially improve the performance of the model, especially for the H\"uckel initialization, which is far from the ground-state density. Here, the energy MAE is reduced by one or two orders of magnitude, highlighting the powerful capability of natural reparameterization in balancing sensitivities across coefficient dimensions.

We also attempted to examine the importance of the other two enhancement modules, the atomic reference module and the dimension-wise rescaling module, but found it difficult to optimize the learning objective when removing either module from the model. Consequently, the unreasonable evaluation results are omitted in \appxtblref{res-ablation-den}. Moreover, the substantial impact on expressing a vast gradient range and the quantitative results for reducing geometric variability, as clarified in \appxref{dim-rescale}, also strongly support the necessity of these two modules.

\begin{table}[h]
  \centering
  \caption{\textbf{Ablation study for components in the CoefficientAdapter module.} All results are evaluated on test ethanol molecules with the learned $T_{\res, \theta}$ model. The MAEs in energy and HF force are listed in $\mathrm{kcal/mol}$ and $\mathrm{kcal/mol/\angstrom}$, respectively.The results are calculated on ethanol test structures ($n$=10,000).}
  \begin{tabular}{ccccc}
    \toprule
    Local Frame            &  Nat. Reparam.     & Density Initialization & Energy & HF Force\\
    \midrule
    \multirow{2}{*}{}           & \multirow{2}{*}{}           & H\"uckel  & 567.69  & 73.17 \\
                                &                             & ProjMINAO & 0.57    & 2.02 \\
    \midrule
    \multirow{2}{*}{\checkmark} & \multirow{2}{*}{}           & H\"uckel  & 514.56  & 36.55 \\
                                &                             & ProjMINAO & 0.30    & 1.29 \\
    \midrule
    \multirow{2}{*}{}           & \multirow{2}{*}{\checkmark} & H\"uckel  & 25.41 & 37.18\\
                                &                             & ProjMINAO & \textbf{0.18} & 1.81 \\
    \midrule
    \multirow{2}{*}{\checkmark} & \multirow{2}{*}{\checkmark} & H\"uckel  & 5.77    & 27.73 \\
                                &                             & ProjMINAO & \textbf{0.18}    & \textbf{1.18} \\
    \bottomrule
  \end{tabular}
  \label{tbl:res-ablation-den}
\end{table}

\subsubsection{Results Using Other Training Strategies} \label{appx:other-train-strategy}

As stressed in the main paper, one of the additional challenges for learning a functional model beyond conventional machine learning tasks is that the model is used to construct an objective function for optimizing the input, hence in addition to accurate end-to-end prediction on a discrete set of input queries, the model also needs to capture the output landscape (tendency of change in each locality) to properly guide the optimization.
To address this challenge, we introduced techniques to generate multiple density datapoints each also with a gradient label for each molecular structure (Methods~\ref{sec:stage1}, \appxref{dft-label}), which provides a more holistic depiction of the output landscape. %
While there are alternative methods to regularize the optimization behavior in the context of learning the XC functional model~\citep{kirkpatrick2021pushing,nagai2020completing,li2021kohn,chen2021deepks}, our experiments indicate that they are not as effective for our task. %

\citet{kirkpatrick2021pushing} design an SCF loss based on second-order perturbation theory to regularize the functional stationary at provided self-consistent solutions. %
This effectively acts as a gradient supervision at the ground state, \ie, the given self-consistent orbitals, for each molecular structure.
The SCF loss itself is not directly suitable for learning the KEDF model as it is for supervising the gradient with respect to orbitals but not density (also explained by \citet{del2023variational}).
Approaches under the same spirit to supervise the gradient at the electronic ground state of each molecular structure are also explored in learning a KEDF model~\citep{meyer2020machine,fujinami2020orbital,imoto2021order,del2023variational}.
In contrast, \ourmethod introduces gradient labels on \emph{multiple} electron density states, instead of only the ground-state density, for each molecular structure, which provides richer and broader landscape information, leading to a substantial performance improvement as shown in our ablation study results in \appxtblref{res-ablation-grad} (row 1 and row 3).

Other studies shape the output landscape and regularize the optimization behavior by directly supervising the model-optimized energy and density (or orbitals). This approach supervises the end goal, but unfortunately complicates the learning process of the model, since the dependency of the model-optimized results on model parameters is nested iteratively following an optimization iteration.
Some works resort to gradient-free optimization methods (\eg, evolutionary-style algorithms) to optimize the model parameters~\citep{nagai2020completing}, which are not as efficient as gradient-based optimization.
\citet{li2021kohn} choose to directly invoke automatic differentiation on the loss with respect to model parameters through the optimization process of orbitals, which is, in the context of learning the XC functional for KSDFT, typically conducted through the self-consistency field (SCF) iteration that solves the Kohn-Sham equations, and hence called the Kohn-Sham regularizer. %
We tried its counterpart for learning the KEDF for OFDFT, by supervising the model-optimized energy and density after the density optimization process using gradient descent (note SCF iteration is unnatural here), and then applying automatic differentiation through the gradient-descent iteration. In our experiments on QM9, we found this incurs substantially more computational cost for computing the gradient for model parameters. Even only using the $E_\TXC$ version of functional model (\appxref{func-var-tsexc}) to get rid of the costly grid-based computation, only a few number of density optimization steps up to 8 is affordable. %
With this maximally affordable cost, this strategy could work when applied to an end-to-end pretrained model on the energy and gradient labels, as we observed improved energy accuracy on held-out test molecules using 8 steps of density optimization. However, an undesired observation indicates that the model learned in this way still does not meet the goal: the density optimization process often does not converge, and even when it does, the accuracy is much worse than that at the 8-th step. This makes the method, instead of learning a density functional that holds a physical meaning, more like an end-to-end ground-state energy predictor constructed by unrolling the machine-learning model 8 times through the gradient descent process. %

Under the same idea to supervise model-optimized energy, \citet{chen2021deepks} find it is possible to avoid automatic differentiation through the iterative orbital optimization process if the resulting energy $E_\theta(\bfC_\theta^\star)$ is indeed optimal in the view of the current functional model, since
$\nabla_\theta E_\theta(\bfC_\theta^\star) = (\nabla_\theta E_\theta) \big\vert_{\bfC_\theta^\star} + (\nabla_\theta {\bfCb}_\theta^\star)\trs (\nabla_{\bfCb} E_\theta) \big\vert_{{\bfCb}_\theta^\star} = (\nabla_\theta E_\theta) \big\vert_{\bfC_\theta^\star}$
($\bfCb$ denotes the vector of flattened $\bfC$ matrix, and $\nabla_\theta {\bfCb}_\theta^\star$ is the Jacobian matrix) as the optimality condition indicates $(\nabla_{\bfCb} E_\theta) \big\vert_{{\bfCb}_\theta^\star} = 0$ on the admissible space of orbital coefficients, so $\nabla_\theta {\bfCb}_\theta^\star$ is avoided.
The same argument also applies to the optimization process of density coefficients.
Since finding the optimal orbitals after every model update is expensive, the method opts to optimize the model and density alternately. However, this modification biases the orbital/density optimality thus the optimization process may also deviate from being effective.
In our experiment on chignolin, we iterated the alternate optimization for four steps (and even carefully optimized the density in each step with a large cost), but observed only marginal performance improvement. Specifically, the performance gains from each step are 0.5\%, -0.06\%, 1.7\%, and -0.8\%, respectively, where the negative signs indicate that some steps even worsen the results.